\documentclass[pdflatex,sn-mathphys-num,iicol]{sn-jnl}

\usepackage{graphicx}%
\usepackage{multirow}%
\usepackage{amsmath,amssymb,amsfonts}%
\usepackage{amsthm}%
\usepackage{mathrsfs}%
\usepackage[title]{appendix}%
\usepackage[table]{xcolor}%
\usepackage{textcomp}%
\usepackage{manyfoot}%
\usepackage{booktabs}%
\usepackage{algorithm}%
\usepackage{algorithmicx}%
\usepackage{algpseudocode}%
\usepackage{listings}%
\usepackage[acronym,nonumberlist,toc=false]{glossaries}
\glsdisablehyper
\usepackage{makecell} % for multi-line headers
\usepackage{pifont}
\usepackage{array}
\usepackage[table]{xcolor}
\usepackage{longtable}
\usepackage{subcaption}

\theoremstyle{thmstyleone}%

\theoremstyle{thmstyletwo}%

\theoremstyle{thmstylethree}%

\newacronym{rfm}{RFM}{Robotic Foundation Model}
\newacronym{fm}{FM}{Foundation Model}
\newacronym{cobot}{cobot}{Collaborative Robot}
\newacronym{scara}{SCARA}{Selective Compliance Assembly Robot Arm}
\newacronym{sme}{SME}{Small- and Medium-sized Enterprise}
\newacronym{hrc}{HRC}{Human-Robot Collaboration}
\newacronym{hri}{HRI}{Human-Robot Interaction}
\newacronym{iot}{IoT}{Internet of Things}
\newacronym{gpu}{GPU}{Graphics Processing Unit}
\newacronym{ai}{AI}{Artificial Intelligence}
\newacronym{ml}{ML}{Machine Learning}
\newacronym{amr}{AMR}{Autonomous Mobile Robot}
\newacronym{va}{VA}{Vision-Action Model}
\newacronym{vla}{VLA}{Vision-Language-Action Model}
\newacronym{nl}{NL}{Natural Language}
\newacronym{nlp}{NLP}{Natural Language Processing}
\newacronym{ood}{OOD}{Out-Of-Distribution}
\newacronym{llm}{LLM}{Large Language Model}
\newacronym{vlm}{VLM}{Vision Language Model}
\newacronym{vfm}{VFM}{Vision Foundation Model}
\newacronym
[plural=DoFs,
firstplural=Degrees of Freedom (DoFs)]
{dof}{DOF}{Degree of Freedom}
\newacronym{tcp}{TCP}{Tool Center Point}
\newacronym{ccd}{CCD}{Charge-Coupled Device}
\newacronym{lidar}{LiDAR}{Light Detection and Ranging}
\newacronym{radar}{RADAR}{Radio Detection and Ranging}
\newacronym{sonar}{Sonar}{Sound Navigation and Ranging}
\newacronym{tof}{TOF}{Time-of-flight}
\newacronym{pmd}{PMD}{Photonic Mixing Device}
\newacronym{imu}{IMU}{Inertial Measurement Unit}
\newacronym{rfid}{RFID}{Radio Frequency Identification}
\newacronym{gps}{GPS}{Global Positioning System}
\newacronym{asqs}{ASQS}{Academic Search Query Syntax}
\newacronym{db}{DB}{Database}
\newacronym{vpa}{v/P\&A}{velocity/ precision \& accuracy}
\newacronym{acc}{ACC}{Average Criteria Coverage}
\newcommand{\runinheading}[1]{%
	\par\addvspace{0.7\baselineskip}%
	\noindent\textbf{\textit{#1}}\\[0.2\baselineskip]%
	\noindent
}

\definecolor{medgray}{gray}{0.75}    % Medium gray
\definecolor{lightgray}{gray}{0.92}  % Light gray
\definecolor{gr}{RGB}{210,235,205}  % Green
\definecolor{ye}{RGB}{255,255,170}  % Yellow
\definecolor{or}{RGB}{255,220,180}  % Orange
\definecolor{re}{RGB}{252,210,210}  % Red
\definecolor{LightRed}{rgb}{1.00,0.58,0.69}    % FF94B0 = (255,148,176)
\definecolor{Lavender}{rgb}{0.75,0.51,1.00}    % C082FF = (192,130,255)
\definecolor{MintGreen}{rgb}{0.60,0.94,0.74}   % 98EFBD = (152,239,189)
\definecolor{SkyBlue}{rgb}{0.45,0.87,0.97}     % 72DDF7 = (114,221,247)
\definecolor{Beige}{rgb}{0.95,0.88,0.67}       % F2E0AA = (242,224,170)
\definecolor{PastelPink}{rgb}{0.99,0.77,0.96}  % FDC5F5 = (253,197,245)
\definecolor{Platinum}{rgb}{0.89,0.89,0.89}    % E4E4E4 = (228,228,228)
\definecolor{DesertSand}{rgb}{0.93,0.78,0.67}  % EEC7AC = (238,199,172)
\definecolor{CoolGray}{rgb}{0.58,0.56,0.72}    % 938FB8 = (147,143,184)
\definecolor{LightCoral}{rgb}{0.93,0.45,0.45}  % EE7272 = (238,114,114)
\lstdefinestyle{searchquery}{
	basicstyle=\ttfamily\footnotesize,
	columns=fullflexible,
	breaklines=true,
	breakatwhitespace=true,
	keepspaces=true,
	frame=single,
	xleftmargin=0pt,
	xrightmargin=0pt
}
\begin{document}

\title[Robotic Foundation Models for Industrial Control: A Comprehensive Survey and Readiness Assessment Framework]{Robotic Foundation Models for Industrial Control: A Comprehensive Survey and Readiness Assessment Framework}  % [short title]{full title}, for very long titles 

%%=============================================================%%
%% GivenName	-> \fnm{Joergen W.}
%% Particle	-> \spfx{van der} -> surname prefix
%% FamilyName	-> \sur{Ploeg}
%% Suffix	-> \sfx{IV}
%% \author*[1,2]{\fnm{Joergen W.} \spfx{van der} \sur{Ploeg} 
%%  \sfx{IV}}\email{iauthor@gmail.com}
%%=============================================================%%

\author*[1,2]{\fnm{David} \sur{Kube}}\email{david.kube@siemens.com}

\author[1,2]{\fnm{Simon} \sur{Hadwiger}}\email{simon.hadwiger@siemens.com}

\author[2]{\fnm{Tobias} \sur{Meisen}}\email{meisen@uni-wuppertal.de}

\affil*[1]{\orgdiv{Factory Automation}, \orgname{Siemens AG}, \orgaddress{\street{Gleiwitzer Stra\ss e 555}, \city{Nuremberg}, \postcode{90475}, \country{Germany}}}

\affil[2]{\orgdiv{Institute for Technologies and Management of Digital Transformation}, \orgname{Bergische Universit\"at Wuppertal}, \orgaddress{\street{Gau\ss stra\ss e 20}, \city{Wuppertal}, \postcode{42119}, \country{Germany}}}

\abstract{Robotic foundation models (RFMs) are emerging as a promising route towards flexible, instruction- and demonstration-driven robot control, however, a critical investigation of their industrial applicability is still lacking.
This survey gives an extensive overview over the RFM-landscape and analyses, driven by concrete implications, how industrial domains and use cases shape the requirements of RFMs, with particular focus on collaborative robot platforms, heterogeneous sensing and actuation, edge-computing constraints, and safety-critical operation.
We synthesise industrial deployment perspectives into eleven interdependent implications and operationalise them into an assessment framework comprising a catalogue of 149 concrete criteria, spanning both model capabilities and ecosystem requirements.
Using this framework, we evaluate 324 manipulation-capable RFMs via 48,276 criterion-level decisions obtained via a conservative LLM-assisted evaluation pipeline, validated against expert judgements.
The results indicate that industrial maturity is limited and uneven: even the highest-rated models satisfy only a fraction of criteria and typically exhibit narrow implication-specific peaks rather than integrated coverage.
We conclude that progress towards industry-grade RFMs depends less on isolated benchmark successes than on systematic incorporation of safety, real-time feasibility, robust perception, interaction, and cost-effective system integration into auditable deployment stacks.}


\maketitle

\captionsetup[table]{%
	font=small,
	labelfont=bf,
	justification=raggedright,
	singlelinecheck=false
}

\section{Introduction}\label{sec:intro}
Over the past decades, industrial robotics is undergoing a transition from isolated, highly optimised automation towards flexible and increasingly collaborative deployment paradigms.
In particular, the growing adoption of \glspl{cobot} reflects a shift towards automation solutions that can be commissioned rapidly, redeployed across frequently changing processes, and operated in closer proximity to human workers.
While these platforms offer high mechanical precision and repeatability alongside comparatively restrained power consumption, their industrial utility is increasingly limited by the availability of robust, adaptive and intuitive control methods rather than by actuation performance alone.
This motivates a renewed focus on learning-based robot intelligence that can cope with variability in tasks, environments, actuation and sensing, while respecting industrial constraints such as safety, reliability, low-latency operation and cost-effective integration.
Recent progress in large-scale learning has led to the emergence of \glspl{fm}, which promise to reduce manual engineering effort by transferring knowledge across tasks -- a capability particularly attractive in high-mix, low-volume industrial settings.
Within this development, \glspl{rfm} aim to supply generalist competence for embodied decision-making by providing a flexible platform for transfers across tasks, interfaces and embodiments.
In this survey, we focus on models capable of generating low-level actions for direct robot actuation -- \glspl{rfm} for control or integrated \glspl{rfm} as per our definition in Section~\ref{sec:rfms} -- because this interface is central for real-world deployment and exposes the strongest coupling to industrial constraints (e.g. latency, safety gating, and hardware-specific action spaces).
While paradigms for building such models vary widely -- including diffusion-based approaches~\cite{chiDiffusionPolicyVisuomotor2023, liangDiscreteDiffusionVLA2025}, dreaming-based methods~\cite{zhangDreamVLAVisionLanguageActionModel2025, liangDreamitateRealWorldVisuomotor2024}, and dual-system designs~\cite{buSynergisticGeneralizedEfficient2025, linOneTwoVLAUnifiedVisionLanguageAction2025, chenFastinSlowDualSystemFoundation2025} -- we deliberately chose not to review architectural taxonomy, as related works already cover those topics; for architectural-centred surveys, we thus refer to Section~\ref{sec:related-works}.
Instead, we use the industrial lens to connect the rapid \gls{rfm}s' methodological progress to the practical constraints that determine whether such systems can be deployed in industrial environments.
To structure this discussion and to position \glspl{rfm} within the broader evolution of robot control, we propose a comprehensible categorisation of robotic control methods, summarised by the hierarchy in Figure~\ref{fig:hierarchy-levels}:
\begin{figure}[hbt]
	\centering
	\includegraphics[width=0.5\textwidth]{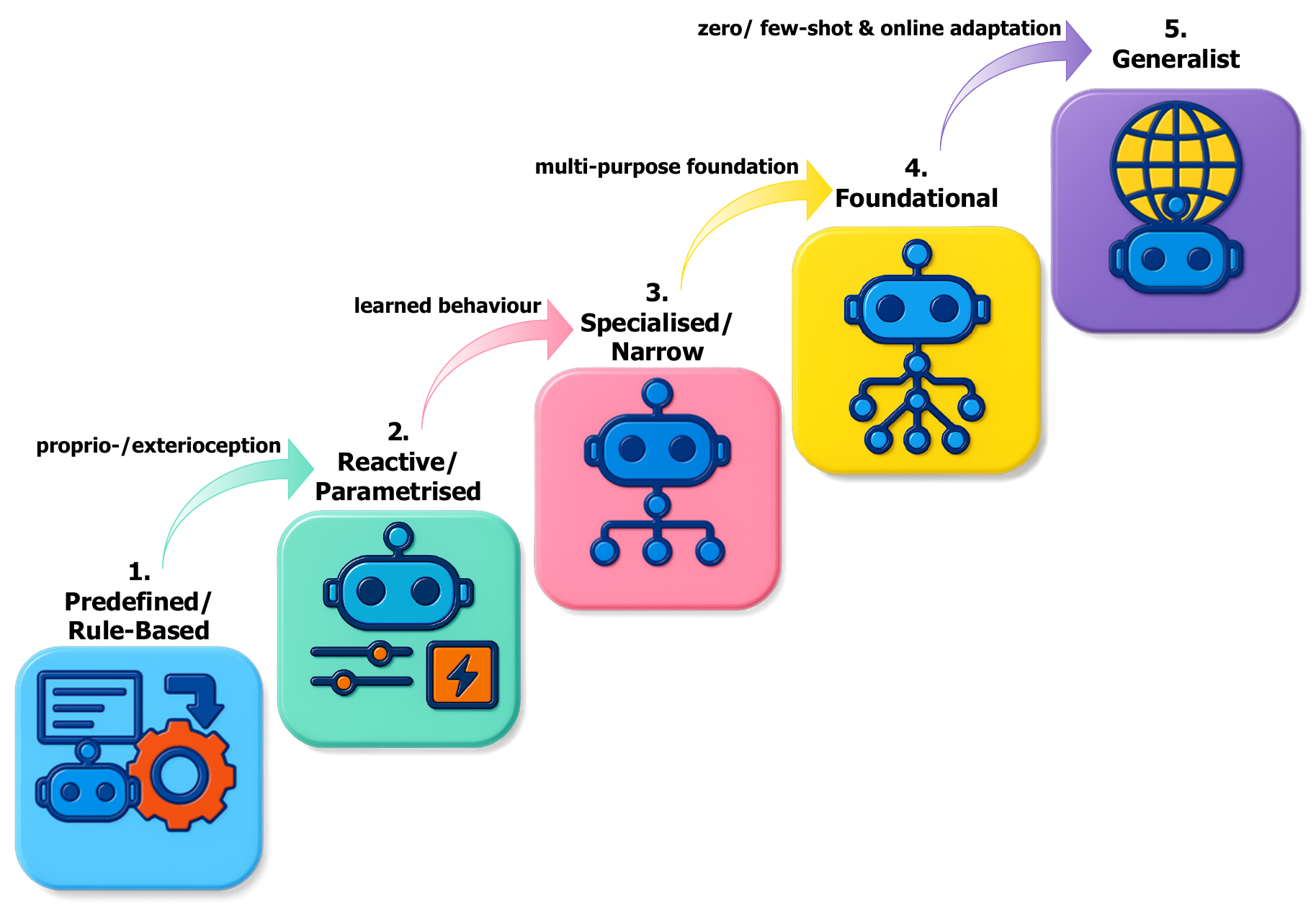}
	\caption{Robotic control-method hierarchy: 
		\textit{Each level increases flexibility, intelligence and generalisation capabilities, while reducing manual engineering effort and required expertise for operation and adaptation.}}
	\label{fig:hierarchy-levels}
\end{figure}
\begin{enumerate}
	\item \textbf{Predefined/Rule-based}: Employ fixed trajectories or simple triggers, lacking dynamic adaptation.\
	\textit{Transition}: Upgraded by integrating sensor-driven, dynamic input.
	\item \textbf{Reactive/Parametrised}: Enable limited adaptation via sensors and parameter adjustment.\
	\textit{Transition}: Move beyond predefined behaviours by introducing learning capabilities.
	\item \textbf{Specialised/Narrow}: Solutions tailored for a specific problem on a given hardware setup.\
	\textit{Transition}: Achieve generic, multi-purpose competence.
	\item \textbf{Foundational}: Broadly pretrained systems capable of efficient fine-tuning with minimal dedicated engineering.\
	\textit{Transition}: Eliminate need for further manual tuning and enable online adaptation.
	\item \textbf{Generalist}: Deliver zero-shot or in-context generalisation to new tasks and configurations, requiring only high-level instructions or online demonstration, akin to human operators in industry.
\end{enumerate}
This hierarchy highlights the core developmental trajectory relevant to industrial deployment: moving from rigid, task-specific automation towards adaptable, instruction- and demonstration-driven operation with decreasing commissioning effort.
The introduction of \glspl{fm} in robotics expressively marked the progression from specialised control systems (3.) to foundational robotics (4.), producing cross-embodiment and multi-task control systems.
Yet, as our subsequent analysis shows, this step does not automatically yield industry-grade autonomy: reaching the next level towards truly generalist systems (5.) requires capabilities such as robust online adaptation, dependable instruction following, and safe operation under uncertainty -- capabilities that are particularly important in dynamic, cost-driven industrial environments.
The pace at which the field is moving makes a dedicated, industrially grounded review a necessity.
\begin{figure}[hbt]
	\centering
	\includegraphics[width=0.5\textwidth]{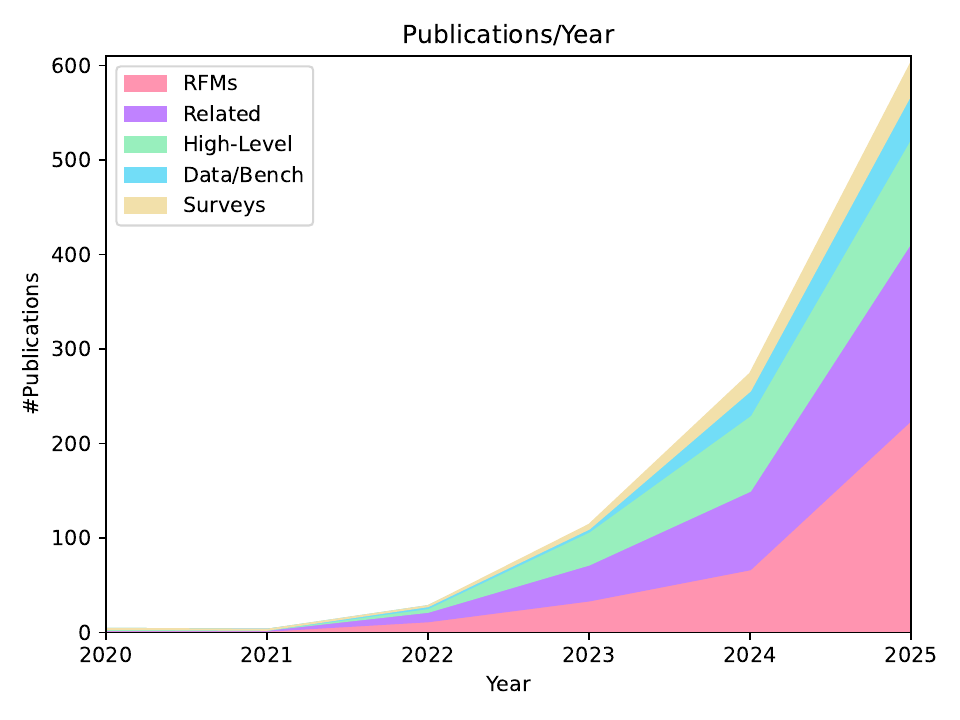}
	\caption{Publication increase.
		\textit{Visualising the amount of publications gathered per category.}}
	\label{fig:num-pub}
\end{figure}
As shown in Figure~\ref{fig:num-pub}, publications relevant to \glspl{rfm} have increased sharply in recent years, with the most pronounced rise occurring in 2025.
This rapid growth risks fragmentation and inflated claims without structured industrial evaluation.
A similar growth trend is visible in related literature addressing enabling components, applications, and evaluation.
While several surveys already exist (see Section~\ref{sec:related-works}), none provides a systematic, implication-grounded assessment of industrial deployment readiness of \glspl{rfm}.
At the same time, the rapid expansion of the literature implies that infrequent snapshots quickly become outdated, motivating a review that both consolidates the current \gls{rfm} landscape and provides an assessment structure that remains useful as new models appear.
\begin{figure}[hbt]
	\centering
	\includegraphics[width=0.5\textwidth]{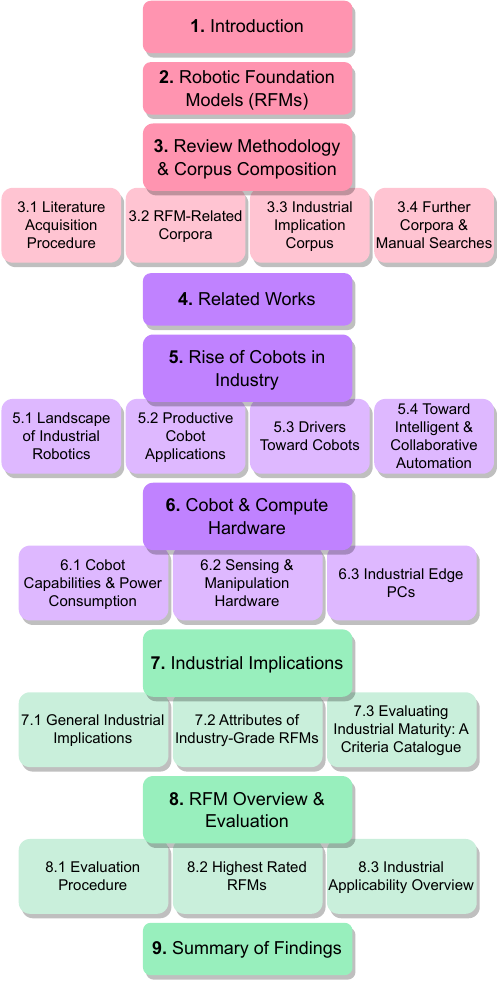}
	\caption{Table of Contents.
	\textit{Highlighting preliminary sections in red, background in purple and the main review in green.}}
	\label{fig:survey_toc}
\end{figure}
\\ Accordingly, this survey is structured to progressively connect foundational concepts to industrial requirements and, subsequently, to a systematic assessment of current \glspl{rfm} (cf. Figure~\ref{fig:survey_toc}):
We first define \glspl{rfm} and position them within the robotics control landscape (Section~\ref{sec:rfms}).
We then describe our review methodology and corpora composition (Section~\ref{sec:methods}), followed by related works (Section~\ref{sec:related-works}) and further background to motivate industrial relevance: the rise of \glspl{cobot} and their deployment domains (Section~\ref{sec:towards_cobots}) as well as the practical constraints of hardware, sensing, and edge computing that shape feasible \gls{rfm} deployment (Section~\ref{sec:cobot_capabilities_hardware}). \\
Building on these foundations, we distil industrial deployment viewpoints into general implications and operationalise them into an explicit assessment framework (Section~\ref{sec:industrial-implications}).
Finally, we apply this framework to the manipulation-capable \gls{rfm} landscape, providing both an overview and a criteria-based industrial maturity evaluation (Section~\ref{sec:survey}), before concluding with a synthesis and discussion of our findings (Section~\ref{sec:summary}).
A key preview of our results is that, despite impressive benchmark-level progress and increasing breadth of contributions, industrial maturity remains limited and uneven across the field: current models tend to address a small number of enabling dimensions at a time, rather than offering integrated coverage of the deployment-critical requirements that dominate real industrial settings.
This motivates our primary goal: to move the discussion from isolated benchmark successes towards a traceable, implication-grounded view of what it would mean for \glspl{rfm} to be credible candidates for industrial deployment.
Our main contributions are:
\begin{enumerate}
	\item A clear categorisation of robotic control approaches that situates \glspl{rfm} within the broader progression from rule-based automation to generalist robotic intelligence (Figure~\ref{fig:hierarchy-levels}).
	\item A concrete definition and clustering of \glspl{rfm} (Section~\ref{sec:rfms}).
	\item A comprehensive overview of the current \gls{rfm} literature and state of the art, with a focus on manipulation-capable systems (Table~\ref{tab:rfms-all}).
	\item An industrially motivated perspective on the rise of \glspl{cobot}, clarifying why their adoption amplifies the need for more general, instructable and adaptable control methods (Sections~\ref{sec:towards_cobots} and~\ref{sec:cobot_capabilities_hardware}).
	\item A synthesis of domain-agnostic industrial deployment implications distilled from industrial literature into eleven interdependent viewpoints (Section~\ref{ssec:ii-gi}).
	\item An extensive, implication-grounded criteria catalogue comprising $149$ items to assess industrial maturity of \glspl{rfm} (including ecosystem requirements), which we believe has not previously been developed in this form for \gls{rfm}-centred industrial readiness assessment (Section~\ref{ssec:ii-cc}).
	\item A large-scale evaluation of the current \gls{rfm} landscape against this catalogue, elucidating strengths, weaknesses, and research gaps that must be addressed to advance towards industry-grade \glspl{rfm} (Sections~\ref{sec:survey} and~\ref{sec:summary}).
\end{enumerate}
\section{Robotic Foundation Models}\label{sec:rfms}
\glspl{fm} represent a significant milestone in artificial intelligence, forming the basis for highly generalisable systems that can be efficiently adapted to diverse downstream tasks. 
These models are trained on extensive, heterogeneous datasets, often employing self-supervised learning methods to extract patterns and relationships well beyond their original training objectives \cite{bommasaniOpportunitiesRisksFoundation2022, firooziFoundationModelsRobotics2023, yuanPowerFoundationModels2022}.
This breadth of data enables \glspl{fm} to exhibit emergent behaviour -- capabilities and skills that substantially exceed the sum of their training data and initial design intentions. 
As Marsh~\cite{marshDEMYSTIFICATIONEMERGENTBEHAVIOR2013} describes, emergent behaviour is “greater than the sum of the parts”.
In practice, this capability is shown in \glspl{llm}, trained simply to predict the next word in a sentence, which demonstrate abilities in arithmetic, programming, and reasoning that were never explicitly targeted during training, but arise from their scale and dataset diversity \cite{bommasaniOpportunitiesRisksFoundation2022}.
Thus, as seen in domains like \gls{nlp} where input-output formats and modalities can often be standardised, these models deliver robust, adaptable performance across a variety of tasks that previous methods were unable to solve \cite{bommasaniOpportunitiesRisksFoundation2022, firooziFoundationModelsRobotics2023, yuanPowerFoundationModels2022}. \\
A fundamental interpretability challenge in \glspl{fm} centres around their dualistic nature: on the one hand, they may function as single, general models whose mechanisms are widely applicable; on the other hand, their ability to adapt to profoundly different tasks may stem from an inherent collection of specialised expert models, each tailored to a specific context~\cite{bommasaniOpportunitiesRisksFoundation2022}.
To efficiently interpret the behaviour of \gls{fm}s and enable both their adaptation and the creation of new \gls{fm}s for novel domains, it is vital to understand the mechanisms by which a \gls{fm} acquires, represents, and processes knowledge.
However, when extending the \gls{fm} paradigm to robotics, additional domain-specific complexities emerge:
The robotic context introduces high-dimensional, closed-loop decision processes, in which a robot’s actions directly and recurrently influence the environment's state, thus its sensory feedback and subsequent behaviour -- a dynamic fundamentally distinct from static, offline data scenarios prevalent in language and vision tasks~\cite{bommasaniOpportunitiesRisksFoundation2022}.
Moreover, robotic platforms display vast heterogeneity in physical embodiments, configurations, and operational contexts.
For this reason, classical control solutions are typically tailored to specific robots, environments, and tasks. 
These systems are characterised by determinism, yet inflexibility and have historically lacked generalisability across tasks or hardware, highlighting the need for control methods that are inherently task-agnostic, cross-embodiment, and open-ended.~\cite{firooziFoundationModelsRobotics2023}
Foundational methods -- capable of leveraging highly diverse robotic datasets -- can fill this gap, when satisfying criteria implicated by industrial use cases and needs.
We argue that, analogous to how \glspl{fm} in \gls{nlp} seamlessly handle varied languages or dialects, emergence in robotic-specific \glspl{fm} could result in the capability to operate across different robot embodiments, modalities, and environments.
In effect, such \glspl{rfm} could be modelled to interpret every robot platform and configuration simply as yet another dialect or language, enabling generalisable control, reasoning, and planning capacity irrespective of the current task or embodiment. \\
Definitions of \glspl{rfm} vary among researchers:
Hu~et~al.~\cite{huGeneralPurposeRobotsFoundation2024} distinguish between "single-purpose" \glspl{rfm} (focused on one modular capability, e.g. perception, planning, or control) and "general-purpose" \glspl{rfm} -- models that can span perception, planning, control, and even non-robotic tasks. 
However, we argue that only the latter, models capable of integrating perception with actionable outputs (such as control, trajectory or plan generation), qualify as true \glspl{rfm}.
Methods restricted to perception or control alone lack the necessary generalism, functioning more as encoders, decoders or specialised solutions rather than \textit{holistic} foundational models.
Furthermore, recent perspectives sometimes conflate \glspl{vla} and \glspl{rfm}, suggesting those terms are equivalent \cite{maSurveyVisionLanguageActionModels2025}. 
However, similar to Xiang~et~al.~\cite{xiangParallelsVLAModel2025} defining \glspl{vla} as a class of \glspl{rfm}, we also consider \glspl{vla} a distinct \gls{rfm} sub-category.
In particular, \glspl{vla} employ multimodal inputs, incorporating language and vision, to provide human-like instructability and generalisation, generating actions for robot control. 
Zhou~et~al.~\cite{zhouBridgingLanguageAction2025} note that \glspl{vla} unify vision, language, and action by aligning all modalities into a common format, such as tokens, thus allowing end-to-end training and prediction.
Xu~et~al.~\cite{xuAnatomyVisionLanguageActionModels2025} describe \glspl{vla} as typically built from three modules: perception, reasoning/planning, and action. 
While this categorisation may be true for certain candidates, we think that models not explicitly modelling or defining a strict separation of those concerns also qualify as \glspl{vla}.
Beyond modalities, Zhong~et~al.~\cite{zhongSurveyVisionLanguageActionModels2025} highlight that \glspl{vla} belong to both digital- and embodied-\gls{ai}, thus need to bridge digital intelligence with physical interaction, demanding robust adaptability to real-world complexity and robot hardware constraints. \\
Synthesising these interpretations, we define \glsreset{rfm} \textbf{\glspl{rfm}} as highly adaptable, data-driven models specialised for addressing robotic tasks, distinguished by a generalist core:
The ability to efficiently adapt to varied tasks, settings, embodiments and hardware configurations with low dedicated engineering or retraining effort -- e.g. through high-level instructions, flexibility in action representation/ interpretation or providing a generalist foundation for task-specific tuning.
A model that does not exhibit such a generalist core, e.g. in the sense of cross-task or cross-embodiment adaptability, does not qualify as an \gls{rfm} under our definition.
To satisfy their multi-purpose nature, \glspl{rfm} must encompass:
\begin{itemize}
	\item \textbf{Multi-Modality}: Capability to ingest diverse proprio- and exterioceptive signals, such as sensor data, images and \gls{nl}.
	\item \textbf{Output flexibility}: Producing actions, plans, trajectories or similar outputs suitable for a spectrum of robots and task domains.
	\item \textbf{Adaptability}: Ability to adapt or generalise to tasks, environments or embodiments with little manual intervention.
\end{itemize}
Given the above definition, it becomes clear that \glspl{vla} form a prominent and distinct sub-category within the broader class of \glspl{rfm}~\cite{xiangParallelsVLAModel2025}, as \glspl{vla} fulfil \gls{rfm}s' key criteria: they harness multimodal inputs in the form of natural language instructions and visual perception, which grants them the flexibility to be intuitively prompted by human operators across a wide array of tasks, while producing low-level actions for direct robot actuation as output.
However, \glspl{rfm} are not constrained to producing low-level actions, hence they can be categorised according to the principal functions they perform within robotic systems, such as (but not limited to):
\begin{itemize}
	\item \textbf{\glspl{rfm} for Control}: These models directly generate low-level action outputs for robot actuation, such as motor commands, joint velocities or force/torque signals, across diverse embodiments and environments. 
	\item \textbf{\glspl{rfm} for Planning}: These models focus on high-level reasoning, task decomposition, (re)planning or explainability generation.
	They may operate by integrating multimodal perception and context, producing sequential plans or sub-goals that downstream controllers execute.
	\item \textbf{Integrated \glspl{rfm}}: Advanced models jointly combining planning and control capabilities.
	These models most closely align with the vision of general-purpose robotic intelligence, capable of internal reasoning, failure-explanation or -recovery and long-horizon (re)planning combined with direct robot control.
\end{itemize}
A similar taxonomy (control \& planning) has been proposed by Ma~et~al.~\cite{maSurveyVisionLanguageActionModels2025} for \glspl{vla}, however, we contend that particularly planners align more closely with the overarching \gls{rfm} paradigm, as their outputs are predominantly of a high-level nature.
Other than through \gls{nl} leveraged in \glspl{vla}, adaptability can equally arise through alternative modalities, methodologies or instructioning mechanisms -- such as using goal images, state representations, or exhibiting a general high adaptability to novel tasks, e.g. through demonstration.
Therefore, subcategories like \glspl{va}, such as DP~\cite{chiDiffusionPolicyVisuomotor2023} or Dreamitate~\cite{liangDreamitateRealWorldVisuomotor2024}, devoid of \gls{nl} understanding, still qualify as \glspl{rfm} if they possess a multi-purpose nature, such as robust foundations for adaptation to various tasks/ embodiments or multi-task/ domain integration. \\
\glspl{rfm} also encompass a myriad of specialised subclasses, targeting specific application areas or domains such as manipulation~\cite{kimOpenVLAOpenSourceVisionLanguageAction2024, liVisionLanguageFoundationModels2024, aminP06VLAThat2025}, locomotion/navigation~\cite{xiaoAnyCarAnywhereLearning2024, zhangEmbodiedNavigationFoundation2025, chengNaVILALeggedRobot2025} or aerial~\cite{fuentesVLHVisionLanguageHapticsFoundation2025} and underwater~\cite{guUSIMU0VisionLanguageAction2025} robotics. Others target combinations thereof, such as mobile-manipulation~\cite{black$p_0$VisionLanguageActionFlow2024, black$p_05$VisionLanguageActionModel2025, brohanRT1RoboticsTransformer2023, brohanRT2VisionLanguageActionModels2023}.
Such specialisation to embodiments or task-categories often stems from predefined action spaces or fine-tuning data composition.
In the context of industrial deployment, manipulation stands out as the most pertinent capability, underpinning essential functions like (dis)assembly, pick-and-place operations, sorting, and dexterous material handling.
Arguably, manipulation constitutes one of the central challenges to resolve in general robotics, laying the groundwork for mastering other domains such as locomotion or navigation, which themselves can be interpreted as a form of self-manipulation. 
Ultimately, however, \glspl{rfm} should pursue the goal to be \textit{both} application- and embodiment-independent, underscoring their foundational nature.\\
In summary, within the scope of this survey, a model qualifies as an \gls{rfm} only if it exhibits a generalist core in the sense of cross-task or cross-embodiment adaptability, while providing multimodal input integration and flexible output generation.
Models lacking these properties are considered specialised or narrow approaches rather than \glspl{rfm}.
\section{Review Methodology \& Corpora Composition}\label{sec:methods}
\begin{figure}[hbt]
	\centering
	\includegraphics[width=0.5\textwidth, trim=90pt 10pt 15pt 10pt, clip]{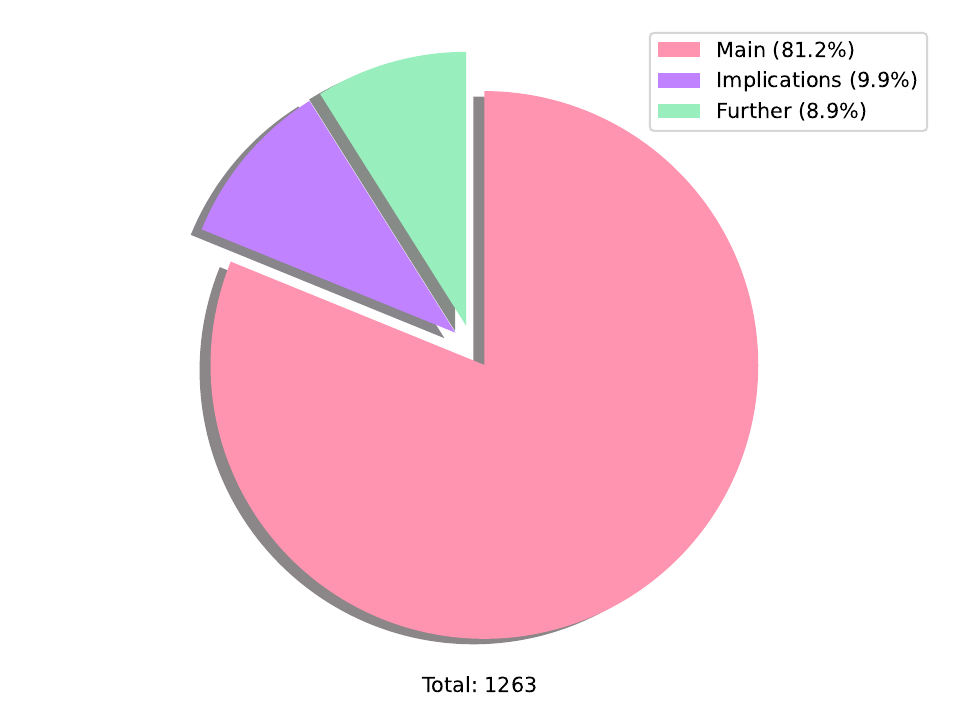}  % plotted with 0.08 explosion
	\caption{Overview of all utilised corpora and number of considered publications, as presented within this section.%: 
		%\textit{}
	}
	\label{fig:corpus_overview}
\end{figure}
This section provides a detailed description of the procedures used to assemble, filter, and categorise the literature corpora that underpin this survey.
Our methodology is designed to ensure systematic coverage, transparency, volume management and reproducibility throughout all phases of literature acquisition and analysis.
Figure~\ref{fig:corpus_overview} presents an overview of all main corpora -- comprising the principal RFM-related corpora (main), the industrial implication corpus, and further supplementary sets -- together with the proportion of all considered publications.
In the following subsections, we describe the automated acquisition pipeline, filtering strategies, and post-processing steps for each corpus, as well as the criteria guiding their composition and specific role within the wider context of this review.
\subsection{Literature Acquisition Procedure}\label{ssec:methods-acq}
\begin{figure}[hbt]
	\centering
	\includegraphics[width=0.5\textwidth, trim=2pt 1pt 7pt 0pt, clip]{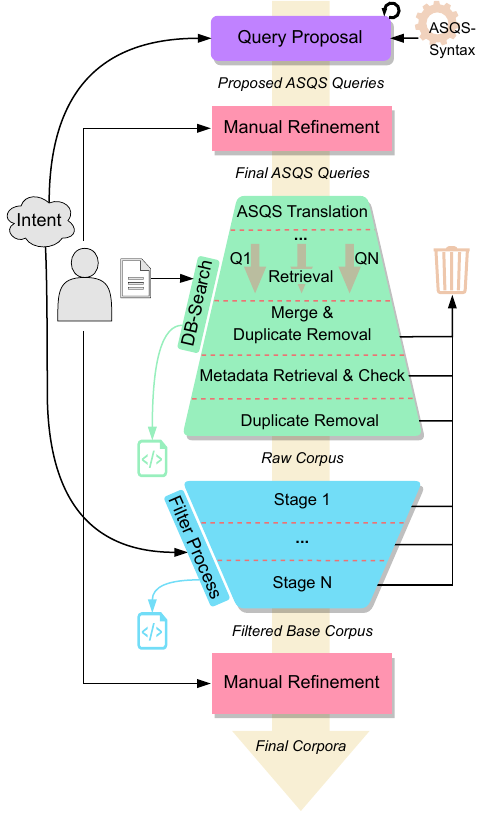}
	\caption{General Literature Acquisition Pipeline Procedure: \textit{Manual user involvement is required only for the Manual Refinement steps (highlighted in pink); all other processes execute automatically based on user-input. Only the Query Proposal and Filter Process modules leverage LLMs, while the DB-Search is fully deterministic.}}
	\label{fig:literature-pipeline}
\end{figure}
To ensure both comprehensive coverage and reproducibility of our literature acquisition, we developed a modular Python pipeline for automated article retrieval and filtering as depicted in Figure~\ref{fig:literature-pipeline}.
This approach was designed to maximise retrieval breadth across relevant literature while minimising blind spots, standardising the literature research procedure throughout all stages of corpora generation, and reducing manual intervention. \\
The main tool in our pipeline (DB-Search) starts with following user-provided information:
\begin{itemize}
	\item One or multiple search queries formulated in a novel, structured \gls{asqs}, tailored to thoroughly capture all relevant literature domains.
	\item The publication date range to be applied for article retrieval (e.g. from 2020 to the present).
	\item Selection of target \glspl{db} used for querying, from the list of implemented \glspl{db} (arXiv, GoogleScholar, OpenAlex, Scopus and SemanticScholar).
	\item Specification of retrieval preference methodology (latest vs. most relevant results).
	\item Upper bounds for the number of articles to retrieve per \gls{db}-query combination, serving as a cut-off to manage volume (e.g. 100 results per \gls{db}-query pair).
\end{itemize}
The pipeline then automatically translates each ASQS query into the respective syntax of each target \gls{db}, relaxing unsupported features where necessary (such as proximity operators or fuzzy matching).
All selected \glspl{db} are then automatically scanned for each specified query -- resulting in the usage of all possible \gls{db}-query permutations. 
During the search process, all metadata returned by the used APIs for the resulting articles are merged.
Thereafter, identified duplicates, based on unique identifiers (such as DOIs), are initially removed for all articles containing sufficient information to permit duplicate removal.
Subsequently, the system retrieves missing metadata from auxiliary APIs (e.g. SemanticScholar, arXiv, Crossref) to ensure completeness.
Metadata validity is programmatically checked -- e.g. by verifying consistency between titles and abstracts.
Articles lacking essential information (not uniquely identifiable) or presenting errors (e.g. title-abstract mismatch) are excluded from the results.
Another round of duplicate removal is performed after these checks. \\
The resulting collection of validated articles is stored in serialised files (JSON or YAML), and, if desired, also exported to PDF format with structured details including, but not limited to: title, URL, authors, abstract, DOI, originating query, and unique identifier assigned by the retrieval-pipeline.
Comprehensive reports documenting the entire retrieval process and its results are automatically generated.
These include:
\begin{itemize}
	\item \textbf{Overall Retrieval Report:}
	\begin{itemize}
		\item Actual (\gls{asqs}-derived) queries employed per database.
		\item Time span and database selection.
		\item Maximum and actual number of retrieved results.
		\item Pipeline execution timestamps.
	\end{itemize}
	\item \textbf{Per-\gls{db}-Query Combination:}
	\begin{itemize}
		\item Unique query ID, timestamp, ASQS/base and translated query.
		\item Applied time and sorting restrictions.
		\item Actual and maximum possible number of results returned.
	\end{itemize}
\end{itemize}
After the main retrieval, an \gls{llm}-based filtering tool (Filter Process) using GPT-4o is leveraged on-demand for semantic relevance scoring. This tool splits the corpus for parallel processing and individually passes each article’s title and abstract, alongside the user-intended relevance criteria, through the \gls{llm}.
It outputs a binary relevance label, a floating-point relevance score in the interval $[0, 1]$ and a concise \gls{nl} justification of its decision.
Sequential or layered filtering can be adopted, using progressively stricter relevance criteria to optimise classification accuracy within the resulting corpora. \\
The pipeline is further supplemented by another component (Query Proposal) for initial query recommendation generation, utilising two further \gls{llm}-instances of GPT-4o.
Given the \gls{asqs} syntax-definition and user intent, they jointly craft and propose a suite of search queries.
This assists in establishing a robust baseline for exploration and iterative query refinement. \\
Combining these tools, we developed all principal literature corpora used throughout this work and presented within the following subsections. 
The pipeline's results were then complemented by manual post-processing, including enrichment (with articles from references or iterative hand-searches), correction, and the application of additional manual filters to ensure quality, completeness, and relevance.
\subsection{RFM-Related Corpora}\label{ssec:methods-crfm}
\begin{figure}[hbt]
	\centering
	\includegraphics[width=0.5\textwidth, trim=80pt 10pt 15pt 10pt, clip]{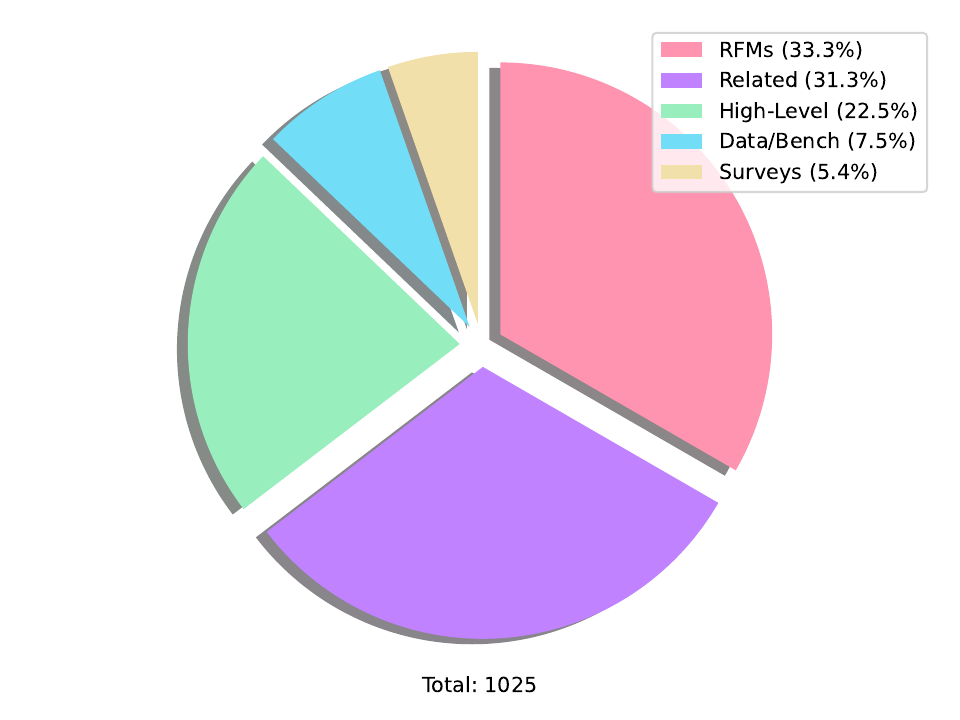}  % plotted with 0.08 explosion
	\caption{Main RFM-Related Corpus Composition%: 
		%\textit{}
	}
	\label{fig:main_corpus}
\end{figure}

The \gls{rfm}-related corpus constitutes the principal literature base for this survey, aiming to target all major research directions regarding \glspl{rfm} and their associated domains.
\begin{figure}[hbt]
	\centering
	\includegraphics[width=0.45\textwidth, trim=18pt 0pt 0pt 0pt, clip]{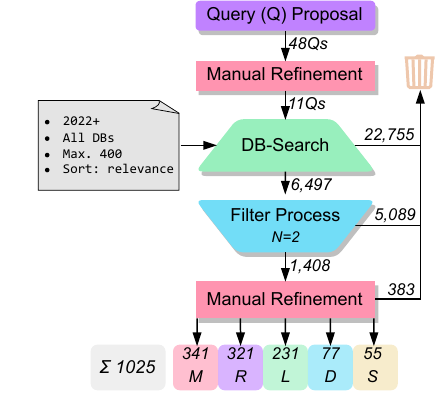}
	\caption{Literature Acquisition Pipeline:
		\textit{Process of gathering the RFM-related corpora (see Figure~\ref{fig:main_corpus}). Results yielded by manual cross-search are included for simplicity.}}
	\label{fig:literature-pipeline-rfm}
\end{figure}
This includes not only \gls{rfm} themselves, but also adjacent contributions such as training protocols, universal extensions, concrete applications and evaluation studies as well as broader considerations on \glspl{rfm} (e.g. safety, robustness, or adversarial challenges).
As we focus on \glspl{rfm} capable of direct low-level control -- as per our definition from Section~\ref{sec:rfms}: \glspl{rfm} for control \& integrated \glspl{rfm} -- we separate those approaches from \gls{llm}-based, high-level or narrow (too specialised) approaches, denoted as "High-Level".
In addition, we cluster literature concerning datasets, benchmarks and simulators, especially those highly relevant to \gls{rfm} research.
To give an overview of related works (see Section~\ref{sec:related-works}), survey and review publications relating to \glspl{rfm} or general embodied \gls{ai} form the last category to be distinguished within the main corpus.\\
To get an initial overview of the literature-landscape and obtain both a comprehensive and up-to-date corpus, we performed \textit{two} iterations of large-scale literature retrievals using our proposed standardised Python pipeline (Section~\ref{ssec:methods-acq}), each configured with identical parameters and subsequently merged to maximise article coverage.
Figure~\ref{fig:literature-pipeline-rfm} shows the concrete retrieval process combining both searches, for which the following settings were applied: \\
\begin{itemize}
	\item \textbf{Queries:} Eleven extensive \gls{asqs}-queries, defined to ensure broad coverage across the complete scope of the main corpus (as shown in Table~\ref{tab:rfm-queries} of the Appendix).
	\item \textbf{Publication Year Range:} From $2022$ to date of search (see below).
	\item \textbf{Databases:} arXiv, GoogleScholar, OpenAlex, Scopus and SemanticScholar.
	\item \textbf{Results Per Query:} A retrieval limit of $400$ articles per database ($x5$) per query ($x11$), leading to a maximal possible result set of $22,000$ articles.
\end{itemize}
We intentionally combined a large set of non-restrictive and targeted queries with a high cut-off value, to ensure a comprehensive result list, while managing volume with targeted filters afterwards. \\
\textbf{Search 1:} Executed on the $4th$ of July $2025$, returning $10,728$ valid article entries, of which $5,035$ were determined unique after deduplication.
\textit{Purpose: Initial overview.}\\
\textbf{Search 2:} Conducted on the $30th$ of October $2025$, returned $12,027$ valid articles, yielding $4,999$ unique entries after improved duplicate removal.
\textit{Purpose: Currentness.}\\
Notably, despite identical parameters, the initial corpus increased significantly in this short period, illustrating the rapid pace of the field and increased database coverage.
Additionally, the duplicate removal algorithm was improved between search $1$ \& $2$ leading to fewer unique articles in the latter. \\
These two result sets were merged, followed by a cautious duplicate removal protocol that erred on the side of retaining ambiguous entries, resulting in a base corpus of $6,497$ as unique classified articles.
This base was then filtered and refined through a multi-stage process: \\
\begin{enumerate}
	\item \textbf{Robotics-Relevant Filtering:} A coarse LLM-based filter was first applied to remove articles not relevant to robotics.
	Of the $6,497$ publications, $4,834$ were retained as relevant, while $1,663$ were disregarded.
	\item \textbf{RFM-Relevant Filtering:} The remaining $4,834$ articles were subjected to a stricter LLM filter, targeting relevance for RFMs, VLAs, and additional foundational approaches or relevant publications as per our targeted specification for this corpus.
	This yielded $1,408$ relevant articles.
\end{enumerate}
For both filter stages, the not-relevant entries were manually sampled and cross-checked.
Within this process no false negative cases were identified. \\
The resulting set of $1,408$ records then became the subject of \textit{manual} in-depth review, further filtering, enrichment (e.g. inclusion of key references and works found through manual chain-searching), and categorisation. Duplicates not resolved algorithmically (e.g. due to identifier inconsistencies) were discarded manually.
The procedure resulted in the following final, categorised main corpus (see also Figure~\ref{fig:main_corpus}):
\begin{itemize}
	\item \textbf{Total articles:} $1,025$ unique and highly relevant works.
	\item \textbf{Models:} $341$ control or integrated \glspl{rfm}, of which 324 focus on manipulation, mobile manipulation, or are generally applicable to multiple robotic domains.
	The remaining $17$ target other specific embodiments or domains (drones, surgical robots, locomotion, navigation or underwater).
	\item \textbf{Related Works:} $321$ articles, e.g. training protocols, extensions (not further categorised for this review).
	\item \textbf{High-Level:} $231$ articles, covering LLM-based (multi-agent) systems, as well as foundational but more narrowly scoped models or \glspl{rfm} for planning.
	\item \textbf{Data and Benchmarks:} $77$ articles, including dataset, simulators, benchmark suites, and data generation procedures.
	\item \textbf{Surveys:} $55$ literature survey and review papers, comprising $24$ on general or embodied AI, and $31$ that include RFMs (of which $11$ focus specifically on RFMs).
	Across the $55$, only $2$ focus on industrially relevant topics.
	For more detail, see Section~\ref{sec:related-works}.
\end{itemize}
%
%%%%%%%%%%%%%%%%%%%%%%%%%%%%%%%
This categorised corpus provides the reference set for all subsequent, review-related analyses, ensuring broad and systematic coverage of \gls{rfm} and closely associated fields.
\subsection{Industrial Implication Corpus}\label{ssec:methods-cinim}
To systematically capture the requirements, challenges, and context-specific attributes of real-world robotic applications in industrial domains, we gathered an industrial implication corpus, utilising the same modular literature research pipeline described in Section~\ref{ssec:methods-acq}.
\begin{figure}[hbt]
	\centering
	\includegraphics[width=0.45\textwidth, trim=18pt 0pt 0pt 0pt, clip]{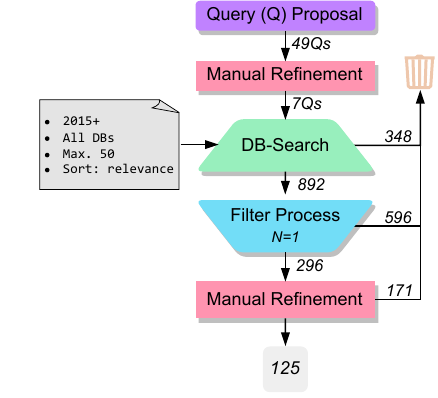}
	\caption{Literature Acquisition Pipeline:
		\textit{Process of gathering the industrial implication-related corpus.}}
	\label{fig:literature-pipeline-impl}
\end{figure}
The intention underlying this corpus was dual: to gather (1) studies of concrete industrial robotics applications and use cases that \textit{implicitly} reveal relevant industrial requirements, and (2) works providing \textit{explicit} discussions or analyses of robotics requirements, constraints, and challenges unique to industrial environments. \\
For this purpose, a comprehensive search depicted in Figure~\ref{fig:literature-pipeline-impl} was conducted according to the following parameters:
\begin{itemize}
	\item \textbf{Queries:} Seven broad ASQS queries (as shown in Table~\ref{tab:impl-queries} of the Appendix), to ensure coverage across a variety of industrial fields, including applications, case studies, analyses, and review papers.
	\item \textbf{Publication Year Range:} From $2015$ up to July $2025$.
	\item \textbf{Databases:} arXiv, GoogleScholar, OpenAlex, Scopus and SemanticScholar.
	\item \textbf{Results Per Query:} A retrieval limit of $50$ articles, yielding a maximum possible result set of $1,750$ publications.
\end{itemize}
The pipeline yielded $1,240$ valid entries, of which $892$ were determined to be unique after deduplication.
To reduce the volume and focus on relevant material, a coarse \gls{llm}-based relevance filter, configured with the same intent as the original search, was applied.
As a result, the corpus was narrowed to $296$ relevant articles and 596 classified as unrelated.
Thereafter, a manual in-depth review was performed on $125$ of the relevant articles that were assigned the highest \gls{llm}-based relevance score (relevance $\geq 0.9$).
We chose to focus on this subset of most relevant articles, since the volume seemed sufficient for the process of gathering industrial implications.
Lesser-scored entries (relevance $< 0.9$) were not included in the detailed review.
The insights distilled from this core set have been iteratively synthesised, clustered, and mapped to industrial implications for robotic systems.
An overview and in-depth discussion of the extracted implications and their interpretation within the \gls{rfm} paradigm are provided in Section~\ref{sec:industrial-implications}.
\subsection{Further Corpora \& Manual Searches}\label{ssec:methods-further}
In addition to the two main corpora presented in the previous sections, additional automated and manual searches about industrial robots, \glspl{cobot}, compute hardware, sensors, manipulators, efficiency considerations, and further related fields as well as norms/ standards have been performed, resulting in a total of $113$ auxiliary publications.
Those have been considered and used throughout the process of crafting and writing of this work.
\section{Related Works}\label{sec:related-works}
Within our extensive literature review, as described in Section~\ref{ssec:methods-crfm}, we identified a total of $55$ relevant survey and review articles.
\begin{figure}[hbt]
	\centering
	\includegraphics[width=0.5\textwidth, trim=100pt 10pt 15pt 10pt, clip]{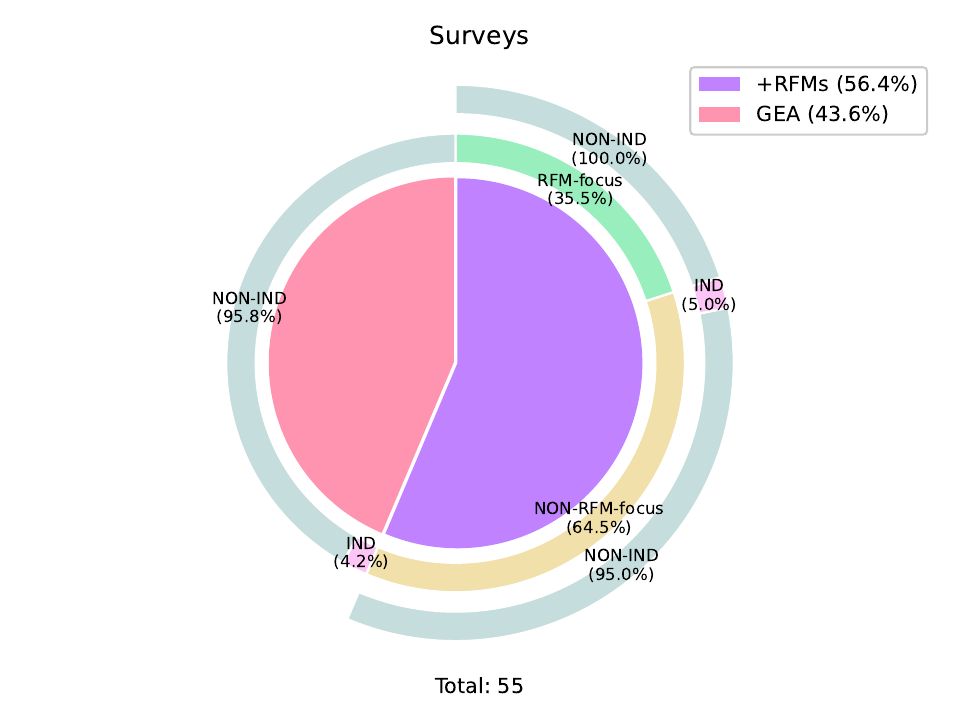}
	\caption{Identified Related Works (Survey corpus) composition: 
		\textit{GEA=general embodied AI, +RFMs=including RFMs, IND=industrial-focus.}}
	\label{fig:surveys_overview}
\end{figure}
Figure~\ref{fig:surveys_overview} shows the distribution of this survey corpus across general embodied AI (GEA), those including \glspl{rfm} (+RFMs), and with explicit industrial focus (IND).
\begin{table*}[ht]
	\centering
	\renewcommand{\arraystretch}{1.35}
	\caption{\textbf{RFM:} Overview of most relevant survey/ review papers with RFM focus.
	\textit{Reporting the focus area, the review's main considerations (Scope), the types of targeted model families (Models), the timespan of reviewed papers (Span), whether the survey is considered concise or extensive (Broad), and the survey's publication year.}}
	\label{tab:survey-overview_rfm}
	\footnotesize  
	\begin{tabular}{@{}
			>{\centering\arraybackslash}m{0.4cm}
			>{\centering\arraybackslash}m{1.7cm}
			>{\centering\arraybackslash}m{7.15cm}
			>{\centering\arraybackslash}m{1.5cm}
			>{\centering\arraybackslash}m{0.7cm}
			>{\centering\arraybackslash}m{0.8cm}
			>{\centering\arraybackslash}m{0.6cm}
			@{}}
		%\toprule
		\rowcolor{CoolGray!50}
	\cellcolor{Platinum!60}
	\textbf{\rule{0pt}{2.5ex}Ref}&
	\textbf{\rule{0pt}{2.5ex}Focus Area}&
	\textbf{\rule{0pt}{2.5ex}Scope}&
	\textbf{\rule{0pt}{2.5ex}Models}&
	\textbf{\rule{0pt}{2.5ex}Span}&
	\textbf{\rule{0pt}{2.5ex}Broad}&
	\textbf{\rule{0pt}{2.5ex}Year}\\
		%\midrule
		\cellcolor{CoolGray!50}\cite{maSurveyVisionLanguageActionModels2025}&%Ma-SurVis
		VLAs for embodied AI& 
		Architectures, challenges, components, control, datasets, objectives, resources, simulators, tasks, task planning, taxonomy&
		Encoder, LLM, VLA, VLM&
		14-24&
		\ding{51}&
		2025\\
		%\midrule
		\cellcolor{CoolGray!50}\cite{zhongSurveyVisionLanguageActionModels2025}&%Zhong-SurVis
		Action tokenisation in VLAs& 
		Action token categorisation: (affordance, code, goal state, language, latent, raw action, reasoning, trajectory), datasets, FM evolution&
		LFM, VFM, VLA, VLM& 
		17-25& 
		\ding{51}&
		2025\\
		%\midrule
		\cellcolor{CoolGray!50}\cite{xuAnatomyVisionLanguageActionModels2025}&%Xu-AnaVis 
		Challenges and roadmap in VLAs& 
		Applications, challenges: (dataset and evaluation, execution, generalization, representation, safety), evolution, roadmap, milestones, modules, opportunities&
		VLA&
		22-25& 
		\ding{51}&
		2025\\
		%\midrule
		\cellcolor{CoolGray!50}\cite{zhouBridgingLanguageAction2025}&%Zhou-BriLan 
		Language-conditioned robot manipulation& 
		Auxiliary tasks, challenges, datasets, evaluation, environments, learning paradigms, limitations, neuro-symbolic, planning, reasoning, reward shaping, semantics extraction, task representation&
		FM, LLM, VLA, VLM&
		18-24& 
		\ding{51}&
		2025\\
		%\midrule
		\cellcolor{CoolGray!50}\cite{shaoLargeVLMbasedVisionLanguageAction2025}&%Shao-LarVlm 
		Large VLM-based VLAs&
		Architectures (hierarchical, monolithic), benchmarks, characteristics, datasets, task planning, learning paradigms, memory mechanisms, multi-agents, operational strengths, taxonomy&
		VLA&
		23-25&
		\ding{51}&
		2025\\
		%\midrule
		\cellcolor{CoolGray!50}\cite{cuiOpenHelixShortSurvey2025}&%Cui-OpeSho 
		Dual-System VLAs& 
		Architectures, definition, design-elements, empirical evaluation, training strategies&
		VLA&
		24-25&
		\ding{55}&
		2025\\
		%\midrule
		\cellcolor{CoolGray!50}\cite{xiangParallelsVLAModel2025}&%Xiang-ParVla 
		Post-training strategies for VLAs& 
		Challenges, embodiment awareness, model development, multi-component integration, perception, post-training methods, task comprehension, taxonomy& 
		VLA&
		23-25&
		\ding{51}&
		2025\\
		%\midrule
		\cellcolor{CoolGray!50}\cite{zhangPureVisionLanguage2025}&%Zhang-PurVis 
		VLA overview and taxonomy& 
		Applications, architectures, benchmarks, challenges, classification, datasets, hardware, paradigms, simulators, taxonomy &
		VLA&
		21-25& 
		\ding{51}&
		2025\\
		%\midrule
		\cellcolor{CoolGray!50}\cite{liuGeneralistRobotPolicies2025}&%Liu-TowGen 
		VLA design and guidelines& 
		Design choices (architectures, backbones, cross-embodiment), guidelines, performance comparison, RoboVLMs&
		VLA& 
		22-24&
		\ding{55}&
		2025\\
		%\midrule
		\cellcolor{CoolGray!50}\cite{sapkotaVisionLanguageActionModelsConcepts2025}&%Sapkota-VisLan 
		VLA advancement synthesis& 
		Agentic AI, architectures, challenges, concepts, control, cross-embodiment, domains, evolution, foundations, planning, real-time, training strategies&
		VLA&
		22-25&
		\ding{51}&
		2025\\
		%\midrule
		\cellcolor{CoolGray!50}\cite{huGeneralPurposeRobotsFoundation2024}&%Hu-TowGen 
		FMs for general-purpose robotics& 
		Benchmarks, challenges (generalisation, data scarcity, safety, requirements, task specification), control, datasets, FM application, manipulation, perception, RFM characteristics, task planning, taxonomy& 
		LLM, RFM, VFM, VLM&
		22-24&
		\ding{51}&
		2024\\
		%\bottomrule
	\end{tabular}
\end{table*}
\begin{table*}[ht]
	\centering
	\caption{\textbf{IND:} Overview of most relevant survey/ review papers with industrial focus.}
	\label{tab:survey-overview_ind}
	\footnotesize  
	\begin{tabular}{@{}
			>{\centering\arraybackslash}m{0.4cm}
			>{\centering\arraybackslash}m{2.6cm}
			>{\centering\arraybackslash}m{6.25cm}
			>{\centering\arraybackslash}m{1.5cm}
			>{\centering\arraybackslash}m{0.7cm}
			>{\centering\arraybackslash}m{0.8cm}
			>{\centering\arraybackslash}m{0.6cm}
			@{}}
		%\toprule
		\rowcolor{CoolGray!50}
		\cellcolor{Platinum!60}
		\textbf{\rule{0pt}{2.5ex}Ref}&
		\textbf{\rule{0pt}{2.5ex}Focus Area}&
		\textbf{\rule{0pt}{2.5ex}Scope}&
		\textbf{\rule{0pt}{2.5ex}Models}&
		\textbf{\rule{0pt}{2.5ex}Span}&
		\textbf{\rule{0pt}{2.5ex}Broad}&
		\textbf{\rule{0pt}{2.5ex}Year}\\
		%\midrule
		\cellcolor{CoolGray!50}\cite{fanVisionlanguageModelbasedHumanrobot2025}&%Fan-VisLan
		VLMs in HRC for smart manufacturing&
		Architectures, human-robot skill transfer, limitations, manipulation, navigation, pretraining, task planning&
		LLM, VLM&
		20-24&
		\ding{51}&
		2025\\
		%\midrule
		\cellcolor{CoolGray!50}\cite{renEmbodiedIntelligenceFuture2024}&%Ren-EmbInt
		Embodied intelligence in smart manufacturing&
		Applications, architecture, challenges, characteristics, capabilities, limitations&
		N/A&
		N/A&
		\ding{55}&
		2024\\
		%\bottomrule
	\end{tabular}
\end{table*}
A comprehensive list of all gathered surveys is provided in the Appendix (Table~\ref{tab:app-rl}).
Here, however, we concentrate on works that are most relevant to our objectives: surveys focused on \glspl{rfm}, as well as those dedicated to industrial scenarios -- even if their main scope does not extend to \glspl{rfm}.
Most surveys to date have concentrated on \glspl{vla} (as distinct sub-category of \glspl{rfm} as discussed in Section~\ref{sec:rfms}) and their associated aspects, including architectures, challenges, learning paradigms and other core considerations as illustrated in Table~\ref{tab:survey-overview_rfm}.
There are several specialised surveys addressing particular facets of the \gls{vla} landscape, such as action tokenisation~\cite{zhongSurveyVisionLanguageActionModels2025}, dual-system architectures~\cite{cuiOpenHelixShortSurvey2025} or post-training strategies~\cite{xiangParallelsVLAModel2025}.
Some also discuss related model classes, like \glspl{llm}, \glspl{vlm} or \glspl{vfm}~\cite{maSurveyVisionLanguageActionModels2025, zhongSurveyVisionLanguageActionModels2025, zhouBridgingLanguageAction2025, huGeneralPurposeRobotsFoundation2024}. \\
Focusing on the \gls{rfm} perspective, Table~\ref{tab:survey-overview_rfm} compares the $11$ surveys ($20\%$ of the $55$ total) which specifically target the \gls{rfm} field.
Importantly, none of these works present an explicit industrial focus and almost all centre on \glspl{vla}, which forms only one distinct subset of \glspl{rfm}.
Merely two of those diverge from a \gls{vla}-centric approach: 
On the one hand, Zhou~et~al.~\cite{zhouBridgingLanguageAction2025} investigate language-conditioned manipulation, therefore bearing close resemblance to the \gls{vla} category.
On the other hand, Hu~et~al.~\cite{huGeneralPurposeRobotsFoundation2024} discuss the overarching \gls{rfm} paradigm, but primarily give an overview over current methodologies.
Consequently, although there is a myriad of recent surveys around \glspl{vla}, comprehensive investigation of the broader \gls{rfm} concept remains largely unaddressed, with virtually no attention given to further \gls{rfm} sub-categories -- such as Vision-Action models or other control-oriented and integrated \glspl{rfm}, as defined in Section~\ref{sec:rfms}.\\
Considering industrial relevant-reviews, Figure~\ref{fig:surveys_overview} and Table~\ref{tab:survey-overview_ind} reveal that only $2$ out of the $55$ analysed surveys ($3.6\%$) possess a clear industrial application focus.
Furthermore, neither emphasises the implications of \glspl{rfm} within industrial domains:
Fan~et~al.~\cite{fanVisionlanguageModelbasedHumanrobot2025} explicitly focuses on \glspl{vlm} in \gls{hrc} for smart manufacturing -- thus only a narrow subset of what we consider important for industrial robotics applications (see Section~\ref{sec:industrial-implications}) -- while also focusing solely on higher-level \glspl{vlm}, rather than on \glspl{rfm}. 
Ren~et~al.~\cite{renEmbodiedIntelligenceFuture2024} conceptually address general embodied intelligence, evaluating capabilities and limitations in industrial contexts, but they do not consider any specific models or what the industrial use case may imply for a given type of model. \\
Synthesising these findings, we expose a pronounced gap in the literature: given our corpus, there is currently no survey dedicated to the industrial application of \glspl{rfm} or implications of industrial use cases on \glspl{rfm}, not even within the restrictively narrower context of \glspl{vla}.
Industrially focused surveys are predominantly either too general or limited to specific facets (e.g. \glspl{vlm} in \gls{hrc}).
By contrast, our work diverges from the prevailing focus on \glspl{vla}' architectural or functional paradigms.
Instead, besides discussing taxonomy and giving an extensive overview over the \gls{rfm} landscape, we provide a detailed analysis of the attributes and capabilities demanded of \glspl{rfm} by industrial use cases, elucidating how such requirements should shape or be integrated into future \gls{rfm} design and how industrial maturity can be evaluated in given approaches -- a perspective unmet in current literature.
\section{Rise of Cobots in Industry}\label{sec:towards_cobots}
Recent advances in \glspl{rfm} are predominantly demonstrated alongside \glspl{cobot}, yet these developments are frequently confined to laboratory, table-top or kitchen settings.
To highlight the broader industrial significance of this trend, it is essential to first consider the evolution of industrial robotics and the transition toward more collaborative robotic systems.

\subsection{Landscape of Industrial Robotics: Tradition \& Transformation}\label{ssec:tc-landscape}
Industrial robots have long constituted a fundamental component of automated manufacturing, typically defined as automatically controlled, reprogrammable multipurpose manipulators, programmable in three or more axes and intended for use in industrial automation applications \cite{ISO10218120112011, guertlerWHENROBOTCOBOT2023}.
Core to their design is an incomplete-machine architecture, requiring the integration of additional tooling such as grippers or welding torches to fulfil specified tasks.
Their flexibility stems from complex series structures -- kinematic chains with multiple rotational degrees of freedom -- which provide mobility and adaptability across a broad range of production processes. \cite{guertlerWHENROBOTCOBOT2023, olszewskiModernIndustrialRobotics2020}
However, traditional industrial robots are primarily programmed for pre-defined, automated function and demand specialised expertise for complex reprogramming.
Due to their high speeds and significant payload capacities, they typically operate in designated spaces separated from human workers for safety reasons and are thus commonly deployed for heavy-duty, repetitive or hazardous industrial tasks.~\cite{borboniExpandingRoleArtificial2023, amiriStatisticalAnalysisCommercial2024, yangCollaborativeMobileIndustrial2019}
\\
To solve varying tasks within automation, several industrial robotic configurations have emerged, ranging from highly specialised parallel structures to general-purpose articulated arms.
They typically make use of revolute (R) or prismatic (P) joints and can be classified into six categories~\cite{amiriStatisticalAnalysisCommercial2024, roboticsInternationalFederationRobotics, muruScopingReviewEnergy2025}, which are shown in order of increasing similarity to \glspl{cobot} within Table~\ref{tab:robot-types}.
\setlength{\tabcolsep}{1.5pt}
\begin{table}[htp]
	\centering
	\renewcommand{\arraystretch}{1.65}
	\caption{Main industrial robot configuration categories.~\cite{amiriStatisticalAnalysisCommercial2024, brinkerComparativeStudySerialParallel2017, autsouAnalysisPossibleFaults2022,collinsDesignSphericalRobot2016,suriSCARAIndustrialAutomation2018},~\cite[pp.~72,~222,~452,~1393]{khatibSpringerHandbookRobotics2016}}
	\label{tab:robot-types}
	\footnotesize
	\begin{tabular}{
			>{\centering\arraybackslash}m{1.45cm}
			>{\centering\arraybackslash}m{1.55cm}
			>{\raggedright\arraybackslash}m{3.9cm}
		}
		\rowcolor{CoolGray!50}
		\cellcolor{Platinum!60}\textbf{\rule{0pt}{2.6ex}Type} &
		\textbf{\rule{0pt}{2.6ex}Config} &
		\textbf{\rule{0pt}{2.6ex}Comments} \\
		%------------------------------
		\cellcolor{CoolGray!50}Delta &
		(PPP) Parallel &
		\textbullet\ Several symmetric kinematic chains\par
		\textbullet\ High speed \& acceleration\par
		\textbullet\ Redundancy increases accuracy \\
		%\midrule
		\cellcolor{CoolGray!50}Cartesian &
		(PPP) ~Serial~ &
		\textbullet\ Perpendicular prismatic joints\par
		\textbullet\ Smooth linear movement\par
		\textbullet\ Rectangular workspace \\
		%\midrule
		\cellcolor{CoolGray!50}Cylindrical &
		(RPP) Serial &
		\textbullet\ Rotational base combined with two prismatic joints\par
		\textbullet\ Cylindrical workspace \\
		%\midrule
		\cellcolor{CoolGray!50}SCARA &
		(RRP) Serial &
		\textbullet\ Both revolute joints along the same axis\par
		\textbullet\ High accuracy \& repeatability\par
		\textbullet\ Planar positioning \\
		%\midrule
		\cellcolor{CoolGray!50}Spherical Arm &
		(RRP/RRR) Serial &
		\textbullet\ Revolute axes intersect at a common center point\par
		\textbullet\ Simplifies collision \& reachability analysis\par
		\textbullet\ Spherical workspace \\
		%\midrule
		\cellcolor{CoolGray!50}Articulated &
		(RRR) Serial &
		\textbullet\ Revolute joints' axes typically offset\par
		\textbullet\ Similar to human arm\par
		\textbullet\ Flexible 3D workspace \\
	\end{tabular}
\end{table}
\setlength{\tabcolsep}{6pt}
\\
Despite their varied forms and operational strengths, these traditional industrial robot types inherently lack the flexibility, adaptability, safe direct \gls{hrc} and \gls{hri} -- attributes now increasingly demanded by modern manufacturing, especially \glspl{sme}\cite{taesiCOBOTApplicationsRecent2023, johannsmeierHierarchicalHumanRobotInteractionPlanning2017, saleemReviewExternalSensors2025, tanApplicationOptimizationRobot2024}.
\textbf{\Glspl{cobot}} have emerged as a dedicated response to these limitations.
Initially introduced as intrinsically passive devices designed for operator guidance and constrained movement~\cite{colgateCobotsRobotsCollaboration1996, peshkinCobotArchitecture2001}, \glspl{cobot} have evolved into a distinct subcategory of industrial articulated robots engineered to operate safely and efficiently alongside humans.
Contemporary definitions emphasise their lower payloads, reduced speeds, rounded geometries and sensor suites for real-time contact or human detection, allowing safe intentional \gls{hrc}~\cite{khedrOverviewCobotsAdvanced2024, azizIntegrationCollaborativeRobots2024, olszewskiModernIndustrialRobotics2020, koppSuccessFactorsIntroducing2021}.
Importantly, \glspl{cobot} are tailored for flexible deployment, intuitive \gls{hri} and programming as well as rapid reconfiguration -- minimising downtime, setup costs and enable frequent application changes.
This makes them particularly attractive for \glspl{sme} requiring high product variability and small batch sizes.~\cite{khedrOverviewCobotsAdvanced2024, azizIntegrationCollaborativeRobots2024, olszewskiModernIndustrialRobotics2020, koppSuccessFactorsIntroducing2021, javaidSignificantApplicationsCobots2022}
The principal differentiator from traditional industrial robots lies in their functional ethos: while conventional robots are generally engineered to replace human involvement -- automating entire processes within isolated environments -- \glspl{cobot} are expressly intended to enhance human capability by enabling joint execution of complex, variable, or ergonomically challenging tasks.
Rather than aiming for maximal speed or payload, cobots prioritise safe cooperation, ease of integration and efficiency gains through shared workflows.~\cite{borboniExpandingRoleArtificial2023}
\subsection{Productive Cobot Applications \& Domains}\label{ssec:tc-applications}
Due to their increased flexibility and safety compared to traditional industrial robots, modern \glspl{cobot} have exceeded their origins in laboratory and demonstrator settings to become integral elements across a wide spectrum of real-world industrial and commercial domains, from manufacturing floors to healthcare and logistics: \\
\textbf{Manufacturing and Assembly:}
\Glspl{cobot} are frequently deployed in the manufacturing sector, particularly in tasks that require consistent precision alongside adaptability, such as assembly, pick-and-place operations, machine tending, quality inspection, palletising, packaging, general material handling and welding or gluing~\cite{kakadeApplicationsCollaborativeRobots2023, georgeCobotChroniclesEvaluating2023, taesiCOBOTApplicationsRecent2023}. 
Concrete examples include the use of Universal Robots' UR10e \glspl{cobot} at PSA and Fiat Chrysler automotive assembly plants for activities such as door, hood and soft-top assembly, alongside riveting and mounting~\cite{kakadeApplicationsCollaborativeRobots2023}.
Further applications include repetitive pick-and-place tasks in plant nurseries, automated packaging in board manufacturing or assistance in soldering, drilling and arc welding processes in metal working environments.~\cite{kakadeApplicationsCollaborativeRobots2023, taesiCOBOTApplicationsRecent2023, elzaatariCobotProgrammingCollaborative2019}. \\
\textbf{Quality Control and Inspection:}
By leveraging advanced sensor and computer vision systems, \glspl{cobot} perform real-time inspection of part dimensions, assembly integrity and defect detection.
In electronics manufacturing, for example, UR3s are efficiently utilised for quality inspection of automotive touchscreen panels~\cite{kakadeApplicationsCollaborativeRobots2023}.
These inspection roles leverage the \glspl{cobot}' repeatability, adaptable end-effectors and their capacity to be easily reconfigured for frequent product changes.~\cite{georgeCobotChroniclesEvaluating2023, taesiCOBOTApplicationsRecent2023, kakadeApplicationsCollaborativeRobots2023} \\
\textbf{Material Handling, Logistics, and Mobile Manipulation:}
Industrial environments frequently integrate \gls{cobot} arms with \gls{amr} platforms, resulting in mobile manipulators that can transport goods between workstations, feed machines and thus enable flexible and precise in-factory logistics.
Concrete use cases include the movement of assembly components using MiR200 \glspl{amr} in appliance manufacturing~\cite{kakadeApplicationsCollaborativeRobots2023}, as well as the management of heavy, repetitive, or ergonomically challenging transport tasks, thus enabling more agile and just-in-time production processes.~\cite{kakadeApplicationsCollaborativeRobots2023,javaidSignificantApplicationsCobots2022, yangCollaborativeMobileIndustrial2019}. \\
\textbf{Healthcare:} \Glspl{cobot} support patient handling, rehabilitation, medication dispensing and precise surgical assistance. 
Applications include robotic surgery, manipulation of medical tools and remote ultrasound or endoscope positioning~\cite{senCOBOTCOLLABORATIVEROBOT2024,saleemReviewExternalSensors2025, taesiCOBOTApplicationsRecent2023}. \\
\textbf{Agriculture:} Tasks include precise planting, harvesting and sorting in harsh or unpredictable environments.
During periods of high-volume workloads, continuous 24/7 operation can reduce reliance on manual seasonal labour for highly repetitive tasks.~\cite{kakadeApplicationsCollaborativeRobots2023, senCOBOTCOLLABORATIVEROBOT2024} \\
\textbf{Retail and Service:} \Glspl{cobot} assist with shelf stocking, object retrieval and customer engagement, thereby improving service quality and operational efficiency in commerce and hospitality sectors~\cite{saleemReviewExternalSensors2025, senCOBOTCOLLABORATIVEROBOT2024}. \\
\textbf{Military, Security and Inspection:} To increase throughput and operator safety, \glspl{cobot} equipped with specialised sensors, are employed in luggage scanning, dangerous substance detection, bomb disposal, facility inspection, cleaning and delicate manipulation tasks in environments such as spacecraft maintenance, hazardous material operations, airport and public security contexts.~\cite{senCOBOTCOLLABORATIVEROBOT2024,taesiCOBOTApplicationsRecent2023, saleemDesignFabricationSix2024}
In particular, \glspl{cobot} have been employed to inspect and clean the Space Rider orbiter’s surfaces during extended missions~\cite{saleemReviewExternalSensors2025}, for bomb defusing during military operations using specialised tools~\cite{senCOBOTCOLLABORATIVEROBOT2024}, or for assistance of military personnel during border inspection~\cite{taesiCOBOTApplicationsRecent2023}.
\subsection{Drivers of the Industrial Shift Toward Cobots}\label{ssec:tc-drivers}
As implied by the adoption of \glspl{cobot} across diverse sectors outlined in Section~\ref{ssec:tc-applications}, there has been a notable shift within industrial and commercial domains from exclusive reliance on manual labour or traditional industrial robots towards increasing adoption of \glspl{cobot}.
This shift is driven by a broad array of factors -- technological, organisational and economical -- that collectively highlight the growing necessity for adaptable, user-friendly and interactive automation solutions within diverse industrial domains.
Traditional industrial robots, with their high payload capacities, precision and speed, have long been essential for large-scale manufacturers focusing on high-volume, repetitive processes.
However, these systems, which are complex, expensive, and difficult to reconfigure, often prove unsuitable, especially for \glspl{sme} whose operational requirements centre around low-volume, high-mix production.~\cite{yangAutomationSMEProduction2023, javaidSignificantApplicationsCobots2022, sahanRoleCobotsIndustrial2023}
\Glspl{cobot} directly address these needs through their more flexible design, compact footprint and comparable ease of programming, enabling straightforward redeployment across varying tasks and locations within manufacturing environments \cite{javaidSignificantApplicationsCobots2022, kakadeApplicationsCollaborativeRobots2023, lefrancImpactCobotsAutomation2022}.
Notably, \glspl{cobot} claim to offer user-friendly interfaces and intuitive teach modes, thereby reducing the need for specialised engineering expertise and minimising integration time \cite{georgeCobotChroniclesEvaluating2023, lefrancImpactCobotsAutomation2022}.
Aside from increasing operational efficiency, \glspl{cobot} are intended to directly assist human workers by taking over repetitive, physically demanding, or hazardous tasks, thus enhancing workplace safety and allowing human operators to focus on assignments requiring higher cognitive skills and problem-solving abilities \cite{borboniExpandingRoleArtificial2023, koppSuccessFactorsIntroducing2021, javaidSignificantApplicationsCobots2022}.
In contrast to conventional industrial robotics, their design enables direct \gls{hrc}, allowing the combination of human creativity with robotic reliability and precision in shared workspaces \cite{koppSuccessFactorsIntroducing2021, lefrancImpactCobotsAutomation2022}.
Several safety features -- including real-time sensor integration, force/torque monitoring and intelligent behaviour adaptation -- are intended to enable \glspl{cobot} to operate closely with people, reducing the need for impractical and expensive barriers or isolated cages for a respective use cases~\cite{senCOBOTCOLLABORATIVEROBOT2024, olszewskiModernIndustrialRobotics2020, kakadeApplicationsCollaborativeRobots2023, georgeCobotChroniclesEvaluating2023}. 
In addition to optimising available production space, this also enables workflow modernisation and a reduction in ancillary costs. \\
The increasing demand for \glspl{cobot} is reflected in both industrial practice and research trends.
While the total volume of publications on traditional industrial robots remains higher, Khedr et al.~\cite{khedrOverviewCobotsAdvanced2024} showed in their survey, that the growth rate of \gls{cobot}-focused research has accelerated rapidly in recent years -- over 5-fold within a single decade -- underscoring their expanding role in advanced manufacturing and their rising relevance for future automation solutions~\cite{khedrOverviewCobotsAdvanced2024}.
This is also supported by the absolute number of global installations: 
Annual \gls{cobot} deployments have increased from 11,000 units in 2017 to 55,000 units in 2022, reflecting not only ongoing innovation but also expanding industrial uptake alongside continued strong deployment of conventional robots~\cite{amiriStatisticalAnalysisCommercial2024, grauRobotsIndustryPresent2021}.
Correspondingly, the International Federation of Robotics has noted a consistent increase in the proportion of \glspl{cobot} relative to traditional industrial robots~\cite{grauRobotsIndustryPresent2021}.
Beyond hardware, advanced technologies such as computer vision and adaptive control algorithms have extended \gls{cobot} capabilities, paving the way for more sophisticated and autonomous interaction within dynamic industrial settings~\cite{kakadeApplicationsCollaborativeRobots2023, dzedzickisAdvancedApplicationsIndustrial2021}.
This ongoing evolution aligns with the principles of Industry $4.0$, where digitalisation, interconnectivity and smart automation require flexible systems competent of both autonomous operation and collaborative behaviour.
Given sufficiently capable control methods, \glspl{cobot} could be increasingly integrated as key components of smart factories, to connecting with other industrial \gls{iot} devices and \gls{ai}-driven systems, enable data-driven optimisation, remote operation and rapid reconfiguration for changing market demands~\cite{georgeCobotChroniclesEvaluating2023, dzedzickisAdvancedApplicationsIndustrial2021, javaidSignificantApplicationsCobots2022, grauRobotsIndustryPresent2021}.
\subsection{Towards Intelligent \& Collaborative Automation}\label{ssec:tc-rfms}
Despite these advances, there remain specific challenges and requirements for \gls{cobot} deployment in industrial environments.
While \glspl{cobot} are effective at supplementing human labour and undertaking routine or ergonomic tasks, their ability to autonomously handle complex, cognitive, or dexterous processes is still limited~\cite{borboniExpandingRoleArtificial2023}.
Issues such as real-time hazard detection, continuous safe operation, as well as trustworthy \gls{hri} and \gls{hrc} call for further development of sophisticated control strategies and intelligent interfaces.
This is particularly pertinent regarding trust and transparency during both initial deployment and ongoing operation, which are regarded as critical success factors for the introduction of \glspl{cobot}~\cite{koppSuccessFactorsIntroducing2021, borboniExpandingRoleArtificial2023}.
In this context, \glspl{rfm} -- as introduced in Section~\ref{sec:rfms} -- have the potential to address many of these emerging challenges.
\glspl{rfm}, when integrated in industrial \gls{cobot} platforms, can provide powerful tools for intuitive instruction, autonomous decision-making and real-time adaptation by leveraging natural language inputs, multisensor fusion and advanced reasoning capabilities.
Both, \glspl{cobot}' productive use cases and  their current limitations form a perfect fit for \gls{rfm} integration, making the assessment of emerging \glspl{rfm}' industrial applicability crucial.
Future research must focus on ensuring that these models are not designed solely for controlled laboratory settings, but are robust and effective within the complex, variable and safety-critical landscapes of actual industrial operation.
In summary, the ongoing move from traditional industrial robots to \glspl{cobot} reflects a broader reorientation of manufacturing towards collaborative, intelligent and flexible automation.
This transition underscores two key aspects: (a) the industrial relevance and necessity of \glspl{rfm}, which have high potential to directly shape the capabilities, accessibility and trustworthiness of next-generation collaborative robotics; but conversely (b) \glspl{cobot} also create the deployment pressure that exposes the limitations and weak points of current \glspl{rfm} for productive use.
\section{Cobot \& Compute Hardware}\label{sec:cobot_capabilities_hardware}
To meaningfully assess the maturity and applicability of \glspl{rfm} in industry, it is essential to first understand the capabilities and typical platforms for deployment.
As discussed in Section~\ref{sec:towards_cobots}, contemporary \glspl{cobot} still inherently lack the ability to autonomously handle complex or cognitive tasks -- gaps that \glspl{rfm} are well positioned to address.
However, evaluating \glspl{rfm} without reference to the practical capabilities, such as reach, payload, accuracy and utilised perception or manipulation hardware of commonly used \glspl{cobot}, we would risk incomplete analysis.
Additionally, considerations such as power consumption and inference hardware, which are often overlooked in scientific studies, are critical for translating laboratory results to industrial applications.
Clear insight into typical industrial computing platforms and their real-world performance is thus necessary. \\
Therefore, this section provides an overview of the key hardware and system components required for the deployment of \glspl{cobot} in industrial settings.
We examine the quantitative capabilities and power consumption profiles of contemporary \glspl{cobot} in Section~\ref{ssec:cch-cp}, the characteristics and suitability of industrial edge computing platforms necessary for real-time control and inference in Section~\ref{ssec:cch-ipc}, and the essential sensing and manipulation hardware required for real-world robotic applications in Section~\ref{ssec:cch-shw}.
Through a combination of survey data and representative hardware comparison, we elucidate technical prerequisites for integrating \glspl{rfm} in cobotic environments, considering robot capability, computational infrastructure, sensor integration and operational efficiency.
\subsection{Cobot Capabilities \& Power Consumption}\label{ssec:cch-cp}
To highlight aforementioned prerequisites for \gls{rfm} integration and to assess \glspl{cobot} suitability across various industrial tasks, a quantitative understanding of contemporary \glspl{cobot}'s capabilities is crucial.
To this end, an extensive survey of \glspl{cobot} by Taesi et al.~\cite{taesiCOBOTApplicationsRecent2023} provides comprehensive data, which we have synthesised in Tables~\ref{tab:cobot-dpra} and~\ref{tab:cobot-pvc}, reporting minimum, maximum, mean, and median values across key specifications.
In the following, only the articulated anthropomorphic \glspl{cobot} of~\cite{taesiCOBOTApplicationsRecent2023} are considered, consistent with the typical definition of \glspl{cobot} as presented in Section~\ref{ssec:tc-landscape}.
\setlength{\tabcolsep}{2pt}
\begin{table}[h]
	\centering
	\renewcommand{\arraystretch}{1.2}
	\caption{Composed cobot configuration and capability statistics.~\cite{taesiCOBOTApplicationsRecent2023} \textit{Values are rounded to two decimal places.}}
	\label{tab:cobot-dpra}
	\begin{tabular}{lcccc}
		 \cellcolor{Platinum!60}$n=180$ & \cellcolor{CoolGray!50}\textbf{DoF} & \cellcolor{CoolGray!50}\textbf{Payload} & \cellcolor{CoolGray!50}\textbf{Reach} & \cellcolor{CoolGray!50}\textbf{Accuracy} \\
		\rowcolor{CoolGray!50}
		\cellcolor{Platinum!60}&&\textbf{(kg)}&\textbf{(mm)}&\textbf{(mm)}\\
		\cellcolor{CoolGray!50}\textbf{Min}  & 4 &0.3 & 280 & 0.01\\
		\cellcolor{CoolGray!50}\textbf{Max}  & 7 & 170& 2790 & 1\\
		\cellcolor{CoolGray!50}\textbf{Mean} & 6.09 & 8.56 & 970.12 & 0.09\\
		\cellcolor{CoolGray!50}\textbf{Median} & 6 & 5.5 & 910 & 0.05\\
	\end{tabular}
\end{table}
\\
Table~\ref{tab:cobot-dpra} aggregates statistics from $180$ articulated \gls{cobot} models, focusing on \glspl{dof}, payload, reach, and accuracy.
The majority of \glspl{cobot} feature a mean of $6.09$ \glspl{dof}, underscoring the dominance of six-axis articulated robots in collaborative applications.
Typical payload capacity spans a median of $5.5$kg (mean: $8.56$kg), clearly indicating that most commercially available \glspl{cobot} are optimised for handling moderate loads.
The average reach is approximately $910$mm (mean: $97$0mm) and positioning accuracy is typically very precise, ranging from a median value of $0.05$mm to a mean of $0.09$mm. 
Notably, there is a significant spread in both payload and reach, with extreme cases such as the Comau Aura, which offers a payload up to $170$kg and a reach of $2790$mm.
While the presence of high-payload outliers slightly biases the mean values upwards, these statistics capture the core capabilities of most \glspl{cobot} deployed in industrial contexts and confirm the suitability of \glspl{cobot} for applications prioritising moderate payloads, high repeatability and operational flexibility across a diverse range of industrial fields as specified in Section~\ref{ssec:tc-applications}.
\begin{table}[h]
	\centering
	\renewcommand{\arraystretch}{1.2}
	\caption{Composed peak cobot TCP velocity and power consumption statistics.~\cite{taesiCOBOTApplicationsRecent2023} \textit{Showing only values for typical cobots with a payload between $0.5$ and $20$kg as in the wattage comparison of Taesi et al.~\cite{taesiCOBOTApplicationsRecent2023}. Values are rounded to two decimal places.}}
	\label{tab:cobot-pvc}
	\begin{tabular}{lccc}
		 \cellcolor{Platinum!60}$n=63$ & \cellcolor{CoolGray!50}\textbf{Payload} & \cellcolor{CoolGray!50}\textbf{\shortstack{TCP\\Velocity}} & \cellcolor{CoolGray!50}\textbf{\shortstack{Power\\Consumption}\rule{0pt}{4.75ex}} \\
		\rowcolor{CoolGray!50}
		\cellcolor{Platinum!60}&\textbf{(kg)}&\textbf{(m/s)}&\textbf{(kW)}\\
		\cellcolor{CoolGray!50}\textbf{Min}    & 0.5 & 0.2 & 0.06\\
		\cellcolor{CoolGray!50}\textbf{Max}    & 20 & 6 & 0.9\\
		\cellcolor{CoolGray!50}\textbf{Mean}   & 6.69 & 1.68 & 0.31\\
		\cellcolor{CoolGray!50}\textbf{Median} & 5 & 1.1 & 0.25\\
	\end{tabular}
\end{table}
\setlength{\tabcolsep}{6pt}
\\
Table~\ref{tab:cobot-pvc} provides insight into the dynamic performance and power requirements of typical \glspl{cobot} rated for payloads between $0.5$ and $20$~kg -- as those more accurately represent the average \gls{cobot} identified in table~\ref{tab:cobot-dpra}.
The maximum \gls{tcp} velocity for these models ranges from $0.2$ to $6~$m/s, typically between $1.1$ and $1.68$~m/s.
Power consumption is remarkably restrained compared to conventional industrial robots: typically between $0.25$ and $0.31$~kW, with a maximum of $0.9$~kW for heavier-duty models within this class of \glspl{cobot} as shown in table~\ref{tab:cobot-pvc}.
This is in strong contrast to traditional industrial robots, whose average power usage has been estimated at around $3$~kW by Barnett et al.~\cite{barnettDirectIndirectImpacts2017}, with peak draws measured as high as $12$~kW (KUKA KR60HA)~\cite{uhlmannEnergyEfficientUsage2016} or over $15$~kW (KUKA KR210 R2700)~\cite{gadaletaExtensiveExperimentalInvestigation2021} and sustained values near $6$~kW under typical acceleration or velocity~\cite{uhlmannEnergyEfficientUsage2016, gadaletaExtensiveExperimentalInvestigation2021}.
The lower energy requirements of \glspl{cobot} yield significant operational cost advantages, facilitate use in environments with lower power infrastructure and enable deployment on mobile or battery-powered platforms.
Importantly, these comparatively low power requirements have direct effect on the hardware selection and computational strategy for \glspl{cobot} control, thus \glspl{rfm}:
Successful real-time deployment requires models that operate efficiently within the constrained energy budgets typical of cobotic systems, particularly for mobile or battery-powered applications. Achieving low-latency inference on hardware that adheres to strict industrial power limits underscores the importance of model efficiency.
\subsection{Sensing \& Manipulation Hardware}\label{ssec:cch-shw}
\textbf{Sensors} are essential to the functional operation of \glspl{cobot}, serving as the primary means for environmental perception and self-monitoring.
Their central role covers key industrial aspects such as \gls{hrc}, \gls{hri}, autonomous navigation, manipulation and real-time task execution~\cite{saleemReviewExternalSensors2025, liCommonSensorsIndustrial2019}.
A variety of sensor types are commonly deployed in both research and industrial cobotic settings~\cite{liCommonSensorsIndustrial2019, saleemReviewExternalSensors2025, yangAutomationSMEProduction2023, taesiCOBOTApplicationsRecent2023, yangCollaborativeMobileIndustrial2019, dangExtPerFCEfficient2D2023, azizIntegrationCollaborativeRobots2024, scholzSensorEnabledSafetySystems2025, georgeCobotChroniclesEvaluating2023, borboniExpandingRoleArtificial2023, elzaatariCobotProgrammingCollaborative2019}:
\begin{itemize}
	\item \textbf{Visual:} RGB (single/stereo), depth, \gls{ccd} and 2D/3D vision sensors
	\item \textbf{Range:} \gls{lidar}, \gls{radar}, \gls{sonar}, ultrasonic, \gls{tof}/\gls{pmd} and laser scanners
	\item \textbf{Dynamics:} \glspl{imu}, gyroscopes, accelerometers, force/torque sensors, tactile and pressure sensors
	\item \textbf{Position and Identification:} Encoders, \gls{rfid}, infrared, \gls{gps}, odometry and magnetic sensors
	\item \textbf{Specialised:} Acoustic, thermal and similar sensors
\end{itemize}
Advanced applications, such as \glspl{rfm}, can benefit from sensor fusion, combining modalities such as vision with proximity, inertial, or force/torque measurements to increase perception robustness and accuracy~\cite{saleemReviewExternalSensors2025}.
Configurations can involve redundant or complementary sensing to support safety, compliance, and reliable manipulation -- for example, using visual and tactile feedback for precise object handling, or adding force sensors to detect unwanted contact in collaborative settings.
\\
\textbf{End-effectors} constitute the physical interface for manipulation tasks and are tailored to specific application requirements.
As outlined in Section~\ref{ssec:tc-landscape}, the integration of these interfaces is essential for enabling an industrial robot/ \gls{cobot} to perform its designated tasks.
Concrete examples include grippers with configurable fingers and pads, suction cups, dual-stage end-effectors incorporating both gripping and suction capabilities or custom tooling such as drills and polishers for advanced operations~\cite{yangAutomationSMEProduction2023, elzaatariCobotProgrammingCollaborative2019}. 
Many end-effectors embed their own dedicated sensors~\cite{yangCollaborativeMobileIndustrial2019} -- such as force feedback for manipulation tasks or embedded vision modules for object detection and pose estimation.
Notably, the modularity of modern \gls{cobot} end-effectors allows quick adaptation to varying industrial tasks, which increases reusability and operational uptime.
Nonetheless, attaining this level of modularity assumes similar flexibility in action-space of the control algorithm.
\\
Overall, the selection and integration of sensors and end-effectors play a pivotal role in the industrial applicability of \glspl{cobot} and must be considered for deploying robust and efficient \glspl{rfm} in real-world scenarios.
Specifically, the sheer amount of sensing and manipulation hardware results in an even greater number of observation-/action-space permutations, further highlighting the need for multi-modality and output flexibility in modern control algorithms -- capabilities central to \glspl{rfm} (cf. Section~\ref{sec:rfms}).
\subsection{Industrial Edge PCs}\label{ssec:cch-ipc}
\setlength{\tabcolsep}{3pt}

\begin{table*}[htbp]
	\centering
	\renewcommand{\arraystretch}{1.4}
	\caption{Comparison of NVIDIA edge devices' AI performance in INT8 precision (dense)~\cite{BuyLatestJetson2025, JetsonT5000Modules2025, NVIDIARTXBlackwell2025}. \textit{Prices for all Jetson devices correspond to single-unit purchases from the same NVIDIA-specified retailer}~\cite{ArrowElectronicsConnect2025}\textit{, collected on the same date. Operational costs are estimated for 24/7 usage with 0.25\$/kWh. All values are rounded to two decimal places. Su~=~Super, DK~=~Developer~Kit, IND~=~Industrial}}
	\label{tab:edge}
	\scriptsize
	\begin{tabular}{l c c c c c c c c c}
		\rowcolor{CoolGray!50}
		\cellcolor{Platinum!60}\textbf{\shortstack{~\\Jetson\\Device\rule{0pt}{2ex}}} & \textbf{\shortstack{Price\\(\$) $\downarrow$\rule{0pt}{2ex}}} & \textbf{\shortstack{TOPS\\(INT8) $\uparrow$\rule{0pt}{2ex}}} & \textbf{\shortstack{Memory\\(GB) $\uparrow$}} & \textbf{\shortstack{Power\\(W) $\downarrow$\rule{0pt}{2ex}}} & \textbf{\shortstack{GOPS\\/\$ $\uparrow$\rule{0pt}{2ex}}} & \textbf{\shortstack{TOPS\\/W $\uparrow$\rule{0pt}{2ex}}} & \textbf{\shortstack{Memory\\(MB)/\$ $\uparrow$}} & \textbf{\shortstack{\$\\/year $\downarrow$\rule{0pt}{2ex}}} & \textbf{\shortstack{TOPS\\/y\$ $\uparrow$\rule{0pt}{2ex}}} \\
		%\hline
		%\toprule
		%\midrule
		%\bottomrule
		\cellcolor{CoolGray!50}Orin Nano 4GB & \cellcolor{MintGreen!60}229 & 34 & \cellcolor{LightRed!60}4 & 25 & 148.47 & 1.36 & 17.47 & 54.75 & 0.62 \\
		\cellcolor{CoolGray!50}Orin Nano 8GB & 249 & 67 & 8 & 25 & \cellcolor{MintGreen!60}269.08 & 2.68 & 32.13 & 54.75 & 1.22 \\
		\cellcolor{CoolGray!50}Orin Nano SuDK & 249 & 67 & 8 & 25 & \cellcolor{MintGreen!60}269.08 & 2.68 & 32.13 & 54.75 & 1.22 \\
		\cellcolor{CoolGray!50}Orin NX 8GB & 479 & 117 & 8 & 40 & 244.26 & 2.93 & 16.7 & 87.6 & 1.34 \\
		\cellcolor{CoolGray!50}Orin NX 16GB & 699 & 157 & 16 & 40 & 224.61 & 3.93 & 22.89 & 87.6 & 1.79 \\
		\cellcolor{CoolGray!50}Orin AGX 32GB & 999 & 200 & 32 & 40 & 200.2 & \cellcolor{MintGreen!60}5 & 32.03 & 87.6 & \cellcolor{MintGreen!60}2.28 \\
		\cellcolor{CoolGray!50}Orin AGX 64GB & 1799 & 275 & 64 & 60 & 152.86 & 4.58 & 35.58 & 131.4 & 2.09 \\
		\cellcolor{CoolGray!50}Orin AGX DK & 1999 & 275 & 64 & 60 & 137.57 & 4.58 & 32.02 & 131.4 & 2.09 \\
		\cellcolor{CoolGray!50}Orin AGX IND & 2349 & 248 & 64 & 75 & 105.58 & 3.31 & 27.25 & 164.25 & 1.51 \\
		\cellcolor{CoolGray!50}Xavier NX & 502 & \cellcolor{LightRed!60}21 & 8 & \cellcolor{MintGreen!60}15 & 41.83 & 1.4 & \cellcolor{LightRed!60}15.94 & \cellcolor{MintGreen!60}32.85 & 0.64 \\
		\cellcolor{CoolGray!50}Xavier NX 16GB & 579 & \cellcolor{LightRed!60}21 & 16 & 20 & 36.27 & 1.05 & 27.63 & 43.8 & 0.48 \\
		\cellcolor{CoolGray!50}Xavier AGX & 999 & 32 & 32 & 30 & 32.03 & 1.07 & 32.03 & 65.7 & 0.49 \\
		\cellcolor{CoolGray!50}Xavier AGX 64GB & 1399 & 32 & 64 & 30 & 22.87 & 10.7 & \cellcolor{MintGreen!60}45.75 & 65.7 & 0.49 \\
		\cellcolor{CoolGray!50}Xavier AGX IND & 1449 & 30 & 32 & 40 & \cellcolor{LightRed!60}20.7 & \cellcolor{LightRed!60}0.75 & 22.08 & 87.6 & \cellcolor{LightRed!60}0.34 \\
		\cellcolor{CoolGray!50}Thor T5000 & 3199 & \cellcolor{MintGreen!60}517 & \cellcolor{MintGreen!60}128 & \cellcolor{LightRed!60}130 & 161.61 & 3.98 & 40.01 & \cellcolor{LightRed!60}284.7 & 1.82 \\
		\cellcolor{CoolGray!50}Thor AGX DK & \cellcolor{LightRed!60}3499 & \cellcolor{MintGreen!60}517 & \cellcolor{MintGreen!60}128 & \cellcolor{LightRed!60}130 & 147.76 & 3.98 & 36.58 & \cellcolor{LightRed!60}284.7 & 1.82 \\
		\midrule
		\cellcolor{CoolGray!50}\textbf{Min} & \cellcolor{MintGreen!60}229 & \cellcolor{LightRed!60}21 & \cellcolor{LightRed!60}4 & \cellcolor{MintGreen!60}15 & \cellcolor{LightRed!60}20.7 & \cellcolor{LightRed!60}0.75 & \cellcolor{LightRed!60}15.94 & \cellcolor{MintGreen!60}32.85 & \cellcolor{LightRed!60}0.34 \\
		\cellcolor{CoolGray!50}\textbf{Max} & \cellcolor{LightRed!60}3499 & \cellcolor{MintGreen!60}517 & \cellcolor{MintGreen!60}128 & \cellcolor{LightRed!60}130 & \cellcolor{MintGreen!60}269.08 & \cellcolor{MintGreen!60}5 & \cellcolor{MintGreen!60}45.75 & \cellcolor{LightRed!60}284.7 & \cellcolor{MintGreen!60}2.28 \\
		\cellcolor{CoolGray!50}\textbf{Mean} & 1292.31 & 163.13 & 42.25 & 49.06 & 138.42 & 2.77 & 29.26 & 107.45 & 1.27 \\
		\cellcolor{CoolGray!50}\textbf{Median} & 999 & 92 & 32 & 40 & 148.11 & 2.8 & 32.02 & 87.6 & 1.28 \\
		\midrule
		\cellcolor{CoolGray!50}RTX 5090$^{a}$ & \textit{2600} & \cellcolor{MintGreen!60}838 & N.A. & \cellcolor{LightRed!60}575 & \cellcolor{MintGreen!60}322.31 & 1.46 & N.A. & \cellcolor{LightRed!60}1259.25 & 0.67 \\

	\end{tabular}
	
	\begin{flushleft}
	\scriptsize $^{a}$\textit{\footnotesize Pricing estimated due to fluctuation and local availability at time of writing. As Jetson Devices use low-power DDR memory (LPDDR) for the complete system and the RTX cards are supplied with dedicated graphics memory (GDDR), those values are not considered.}
	\end{flushleft}
\end{table*}

\setlength{\tabcolsep}{6pt}
%%%
Given the low-latency requirements of cobotic applications, computational platforms for real-time \gls{ai}-control and -perception are often realised as edge devices -- hardware installed as close to the robot and its data sources as feasible, often directly integrated into the \gls{cobot}’s controller or platform.
They are characterised by processing real-time data locally rather than relying on cloud or external datacentres.~\cite{samantaSurveyHardwareAccelerator2024}
This proximity is vital for latency-sensitive robotic applications, especially for mobile or remote environments, where network traffic must be kept to a minimum~\cite{woisetschlagerFederatedFineTuningLLMs2024}.
Alternative paradigms such as fog or cloud computing can provide scalable resources but come at the cost of additional infrastructure or third-party service requirements.
Fog computing, as defined by Yi et al.~\cite{yiFogComputingPlatform2015}, encompasses distributed and collaborative architectures involving heterogeneous devices close to the edge, but still depends on non-local hardware and stable networking.
Classical cloud computing, while enabling on-demand computational resources, is generally associated with higher latency, lower throughput and an increased risk of service interruption -- limitations that are particularly problematic for time-critical robotic control.~\cite{yiFogComputingPlatform2015, atlamFogComputingInternet2018, bossCLOUD_COMPUTING2007}
According to Ismail et al.~\cite{ismailCobotFleetManagement2020}, moving computation closer to the edge grows increasingly important for industrial deployments, due to a drop in compute costs and the operational advantage of minimising recurring networking expenses~\cite{ismailCobotFleetManagement2020}.
Accordingly, we provide an overview of currently available, representative edge devices to clarify relevant hardware requirements for industrial \gls{rfm} deployment and to establish a foundation for real-world inference evaluation and power consumption considerations.
For the quantitative hardware analysis presented in Table~\ref{tab:edge}, we selected NVIDIA as the representative manufacturer for several reasons. 
Their Jetson devices are among the most widely adopted platforms for edge AI in robotics research and industrial applications, offering a standardised and scalable portfolio (from the compact Nano to the new high-end Thor series). 
Most importantly, relevant performance metrics, pricing, and other specifications are consistently documented and publicly available for the entire product family. 
This transparency enables an accurate, fair, and reproducible comparison of hardware options and their suitability for running modern robotic control methods, predominantly based on \gls{ai} models.
While Table~\ref{tab:edge} concentrates on the Jetson portfolio, a range of alternative manufacturers -- such as Siemens, AMD (Xilinx), Intel, Advantech or Beckhoff among others~\cite{mystkowskaHardwarePlatformsEnabling2025} -- also provide relevant edge devices.
However, for many of these alternatives, comparable and up-to-date public data on prices, performance (in \gls{ai}-relevant metrics) and efficiency is often limited or unavailable.
We additionally report the specifications of a desktop-class \gls{gpu}, the RTX 5090, to provide context for typical platforms used in academic benchmarking of foundation and control models.
While the RTX 5090 is not an edge device, but a standalone \gls{gpu} intended for workstations and servers, it serves as a reference point for the upper bound of single-device computational performance in recent literature. Comparing the RTX 5090 to integrated Jetson devices highlights the trade-offs between desktop/workstation-grade and edge-class hardware.
Examining the comparative statistics in Table~\ref{tab:edge} reveals several significant trends relevant to both industrial and research applications.
The Orin Nano series stands out for its exceptionally low entry cost and high performance relative to purchase price, making it more suitable for cost-sensitive or large-scale deployments.
On the other hand, the entry-level Xavier NX -- while more expensive -- provides the lowest overall AI compute performance in the lineup.
However, its 8GB variant is noteworthy for having the lowest maximum power consumption across all compared models, potentially favoring applications where minimal energy draw is of essence.
The mid-tier, such as the Orin AGX 32GB, provide a balance between increased computational capability and moderate power requirements.
Notably, the Orin AGX 32GB emerges as the most efficient device in terms of compute power per operational cost, offering a strong compromise for applications prioritising both capability and efficiency.
At the high end of the spectrum, devices like the Jetson Thor series or workstation-class desktop \glspl{gpu} exhibit strong increases in computational throughput.
However, this heightened performance is directly linked to substantial rises in both energy consumption and annual operational costs -- a factor that can restrict their scalability and render them impractical for mobile, battery-powered, or energy-constrained robotics platforms.
Particularly, the RTX 5090 can consume more than twice the power of a median \gls{cobot} as shown in Table~\ref{tab:cobot-pvc}.
This contrast towards edge devices' power draw, underscores that desktop \glspl{gpu} are designed for short, intensive workloads or for environments where energy is abundant, whereas edge devices are optimized for consistent, energy-efficient use in real-world, often resource-constrained settings.
Thus, a trade-off emerges: deploying larger and more capable \glspl{rfm} is constrained by both compute and energy efficiency considerations, underscoring the importance of aligning model design with operational cost and industrial practicality. 
Additionally, industrial-class devices (IND) frequently employ features such as ECC-memory for increased reliability, which can raise both initial and operational expenses or lower computational power compared to standard research hardware.
\section{General \& RFM-specific Industrial Implications}\label{sec:industrial-implications}
Industrial deployment settings impose constraints and expectations that systematically differ from laboratory environments: they combine heterogeneous hardware, tight safety and compliance obligations, high reliability targets, and continuous operational pressures. 
This section therefore distils the recurring industrial viewpoints extracted from our industrial implication corpus (Section~\ref{ssec:methods-cinim}) -- aligned with our discussions in Sections~\ref{sec:towards_cobots} and~\ref{sec:cobot_capabilities_hardware} -- into an assessment-oriented structure that can be used to analyse how well current and future \glspl{rfm} comply with real-world industrial demands.
We proceed in three steps: 
First, Section~\ref{ssec:ii-gi} synthesises the corpus into eleven \emph{general industrial implications} that capture the dominant, domain-agnostic requirements reported across industrial robotics applications and that are supported by our \gls{cobot}- and hardware-related considerations.
Second, Section~\ref{ssec:ii-am} translates these implications into a structured set of \emph{attributes} that enable a more fine-grained discussion of where industrial constraints can manifest in practice within \glspl{rfm} and their surrounding ecosystem.
Third, Section~\ref{ssec:ii-cc} consolidates these considerations into a comprehensive criteria catalogue intended to support a consistent, transparent evaluation of industrial maturity across existing \glspl{rfm}.
\subsection{General Industrial Implications}\label{ssec:ii-gi}
To derive general, domain-agnostic implications from the industrial implication corpus (Section~\ref{ssec:methods-cinim}), we conducted an iterative qualitative synthesis across the $125$ manually reviewed works.
Concretely, we (1.) extracted statements that described recurring industrial requirements, constraints, and deployment challenges for robotic systems (both explicitly and as implied by described industrial case studies), (2.) grouped these excerpts into thematic clusters, and (3.) iteratively refined and consolidated the most prominent clusters until they formed a compact set of implications that was sufficiently expressive to capture the dominant industrial viewpoints while remaining actionable for later assessment of \glspl{rfm}.
Throughout this process, we noticed no explicitly contradictory themes, instead found the implications to be \emph{interdependent} rather than mutually exclusive: industrial authors frequently discuss the same practical challenge (e.g. human presence) through different lenses such as safety, interaction design, robustness, or perception, and our synthesis preserves these overlaps by separating \emph{what is demanded} from \emph{why it is demanded}. \\
The resulting set comprises eleven implications, each reflecting a recurring industrial expectation about deployed robotic applications and, by extension, about learning-based robot control and foundation-model-based approaches.
In the following, we motivate each detected implication by summarising the corresponding clusters from the literature corpus.
\runinheading{Implication 1: Adaptability \textup{\&}  Flexibility}
Across industrial domains, authors consistently describe variability as a defining property of real deployments: processes change, product variants proliferate, and boundary conditions shift, which in turn increases re-parameterisation effort and uncertainty~\cite{huangSurveyAIDrivenDigital2021,herreroEnhancedFlexibilityReusability2017,waseemEnhancingFlexibilitySmart2025}.
Multiple works contrast this with classical industrial automation, emphasising that traditional solutions are often designed for repetitive, tightly structured workflows and tend to be brittle under even minor process changes~\cite{tellaecheHumanRobotInteraction2015,renEmbodiedIntelligenceFuture2024,fanEmbodiedIntelligenceManufacturing2025}.
As a consequence, adaptability is framed not merely as a convenience but as a key enabler for future “fast and low-cost” production and for increasing personalisation~\cite{mohammadiaminMixedPerceptionApproachSafe2020,bonciHumanRobotPerceptionIndustrial2021,renEmbodiedIntelligenceFuture2024}.
The corpus further highlights that industrial environments are frequently dynamic, partially known, or unstructured, with unpredictable events and random disruptions being expected rather than exceptional~\cite{johannsmeierHierarchicalHumanRobotInteractionPlanning2017,liuRobotLearningSmart2022,waseemEnhancingFlexibilitySmart2025}.
This is intensified when humans are present in the workspace~\cite{johannsmeierHierarchicalHumanRobotInteractionPlanning2017,liuRobotSafeInteraction2018}.
Several authors therefore argue for robots and control methods that can adjust behaviour online, be reconfigurable across products and processes, and reduce costly reprogramming cycles~\cite{abuzaryaqoobmullaAIRoboticsDesigning2023,herreroEnhancedFlexibilityReusability2017,villaniSurveyHumanRobotInteraction2018,liuRobotSafeInteraction2018}.
In addition, sector-specific discussions (e.g. logistics/warehousing, outdoor industry) underscore that environmental variation (lighting, weather, clutter) can materially affect operation and must be handled robustly~\cite{liOptimizingAutomatedPicking2024,wunderleSensorFusionFunctional2022}.
Additionally, as outlined in Sections~\ref{sec:towards_cobots} and~\ref{ssec:cch-shw}, \glspl{cobot} are seeing growing adoption in industry due to their flexible design and ease of reconfiguration.
However, such adaptations must also be reflected within the applied control system.
Only methods that are intrinsically agnostic to changing tasks, embodiments or sensor setups and quality would enable \glspl{cobot} to realise their full intended potential.
Taken together, the literature motivates adaptability and flexibility as a primary industrial implication: systems are expected to cope with frequent task, process, and environment variation while avoiding extensive downtime and engineering overhead~\cite{calderon-cordovaDeepReinforcementLearning2024,abuzaryaqoobmullaAIRoboticsDesigning2023,fanEmbodiedIntelligenceManufacturing2025}.
\runinheading{Implication 2: Safety \textup{\&} Compliance}
Safety is consistently portrayed as a gating factor for industrial deployment, particularly in shared workspaces, physical interaction, and other high-consequence settings~\cite{johannsmeierHierarchicalHumanRobotInteractionPlanning2017,zhangOnlineRobotCollision2021,andresBlackboxExplainabilityObject2024}.
This is of especial relevance owing to the expanding implementation of \glspl{cobot}, as described in Section~\ref{sec:towards_cobots}:
Despite being intended to significantly reduce the need for physical separation between workers and robots, deploying \glspl{cobot} in shared workspaces requires an equally safety-aware control logic.
Such safety awareness spans both general considerations that injuries must be prevented in human-shared environments~\cite{johannsmeierHierarchicalHumanRobotInteractionPlanning2017,rodriguez-guerraHumanRobotInteractionReview2021} and more explicit arguments that intelligent or adaptive robots must be tested, certified, and equipped with appropriate safety features prior to deployment~\cite{abuzaryaqoobmullaAIRoboticsDesigning2023}.
Critically, multiple works connect safety to standards and certification frameworks as a prerequisite for industrial acceptability.
The literature cites the role of ISO standards for safe installation and operation of robotic systems (e.g. ISO10218) and for collaborative operation and contact-related constraints (e.g. ISO/TS15066), as well as further regional standards~\cite{othmanHumanRobotCollaborations2023,halmeReviewVisionbasedSafety2018,robla-gomezWorkingTogetherReview2017}.
Beyond formal compliance, several authors emphasise that safety must be maintained under dynamic and unpredictable conditions, including human behavioural variability and sensing uncertainty~\cite{renEmbodiedIntelligenceFuture2024,nahavandiMachineLearningMeets2023,sapkotaVisionLanguageActionModelsConcepts2025}.
Industrial discussions also note that safety mechanisms can reduce efficiency if they trigger overly conservative stops or interruptions, motivating approaches that integrate safety-aware reasoning and sensing to preserve productivity without compromising protection~\cite{duijkerenIndustrialPerspectiveMultiAgent2023,grauRobotsIndustryPresent2021,landiSafetyBarrierFunctions2019}.
Overall, the corpus motivates safety and compliance as a distinct implication: industrially viable systems must satisfy relevant safety norms and preserve safe operation under realistic uncertainty, including failures and \gls{ood} inputs~\cite{nahavandiMachineLearningMeets2023,liuRobotLearningSmart2022,sapkotaVisionLanguageActionModelsConcepts2025}.
\runinheading{Implication 3: HRI \textup{\&} HRC}
In line with the drivers for the rising adoption of \glspl{cobot} discussed in Section~\ref{ssec:tc-drivers}, a significant segment of the literature likewise envisions future industrial robotics as inherently human-centric, where robots and humans work in proximity or physical collaboration to combine human flexibility and cognitive skills with robotic repeatability and precision~\cite{linsCooperativeRoboticsMachine2021,liuSensorlessForceEstimation2021,pratiUseInteractionDesign2022}.
The literature uses established interaction taxonomies (e.g. differentiating coexistence, cooperation, and collaboration; and distinguishing remote, proximity, and physical interaction) to emphasise that industrial scenarios span multiple interaction regimes and thus demand correspondingly diverse interface and coordination solutions~\cite{rodriguez-guerraHumanRobotInteractionReview2021,bonciHumanRobotPerceptionIndustrial2021,nakhaeiniaSafeCloseProximityPhysical2015}.
Works addressing \gls{hrc} further highlight that direct physical collaboration is valued for its flexibility, but requires the robot to perceive contact and respond appropriately~\cite{mohammadiaminMixedPerceptionApproachSafe2020,hjorthHumanRobotCollaboration2022}.
Beyond physical coordination, authors repeatedly emphasise that communication quality and interface usability are decisive for acceptance and productivity.
This includes calls for more intuitive, comfortable, and natural interaction, including speech and other modalities, to reduce reliance on expert programming and to better support operators in real time~\cite{desimoneContextawareDataAugmentation2025,renEmbodiedIntelligenceFuture2024,weissFirstApplicationRobot2016}.
Several works discuss that current industrial control and programming workflows remain labour-intensive, motivating approaches that allow operators to instruct, supervise, and correct systems more directly~\cite{herreroEnhancedFlexibilityReusability2017,villaniSurveyHumanRobotInteraction2018,patricioFrameworkIntegratingRobotic2025}.
The literature also points out that collaborative scenarios require reactive methods that can handle uncertainty and unexpected behaviour to maintain fluid workflow between humans and robots~\cite{rodriguez-guerraHumanRobotInteractionReview2021,schreiterMultimodalInteractionIntention2025}.
Hence, \gls{hri} and \gls{hrc} emerge as a separate implication: industrial systems are expected to support human-centred collaboration with effective, natural communication and continuous operator involvement, particularly for complex, variable, and partially automatable tasks~\cite{johannsmeierHierarchicalHumanRobotInteractionPlanning2017,othmanHumanRobotCollaborations2023,fanVisionlanguageModelbasedHumanrobot2025}.
\runinheading{Implication 4: Robustness \textup{\&} Reliability}
Industrial environments are repeatedly characterised as uncertain, noisy, and prone to disturbances and component imperfections.
Authors discuss both internal uncertainties (e.g. unmodelled dynamics, friction effects, sampling delays) and external uncertainties (e.g. measurement errors and environmental noise) as persistent challenges for deployed systems~\cite{calderon-cordovaDeepReinforcementLearning2024,manonmaniLiteratureReviewAdvanced2025}.
This leads to calls for robust methodologies that can integrate perception, planning, and control under uncertainty, especially in complex systems such as warehouse automation or autonomous mobility~\cite{triantafyllouMethodologyApproachingIntegration2021a,johannsmeierHierarchicalHumanRobotInteractionPlanning2017}.
Reliability is also linked to monitoring and timely status updates in order to support reactive planning and practical operation~\cite{duijkerenIndustrialPerspectiveMultiAgent2023}.
Importantly, the corpus extends robustness beyond stochasticity to include resilience against faulty sensors and \gls{ood} inputs, noting that invalid sensor readings can induce unpredictable outputs in learning-based systems and may be particularly dangerous in safety-critical contexts~\cite{nahavandiMachineLearningMeets2023}.
Further, some works explicitly consider adversarial and cybersecurity threats as integral requirements for industrial robots~\cite{quartaExperimentalSecurityAnalysis2017,GuestEditorialEmbedded2019}.
Accordingly, robustness and reliability form a distinct implication: deployed systems are expected to maintain stable operation under uncertainty, disturbances, sensor degradation and security threats, while providing operationally meaningful status and fault awareness~\cite{andresBlackboxExplainabilityObject2024,sapkotaVisionLanguageActionModelsConcepts2025,quartaExperimentalSecurityAnalysis2017,nahavandiMachineLearningMeets2023}.
\runinheading{Implication 5: Precision \textup{\&} Accuracy}
While flexibility and generalisation are emphasised, the industrial corpus simultaneously stresses that robots derive value from high precision, repeatability, and quality-critical execution~\cite{linsCooperativeRoboticsMachine2021,abdullayevIntegrationArtificialIntelligence2025}.
Several works frame industrial competitiveness in terms of efficiency \emph{and} accuracy, reflecting that many tasks require both throughput and reliable outcome quality~\cite{prajapatiComputerVisionEnabled2024,mezgarTransformingExperimentalCobot2023}.
Moreover, accuracy is not restricted to actuation: sensing and perception accuracy is treated as foundational for safe and correct operation~\cite{quartaExperimentalSecurityAnalysis2017,shenProgressComprehensiveAnalysis2024}.
Shen et al.~\cite{shenProgressComprehensiveAnalysis2024} further note that current systems can exhibit insufficient grasping and positioning accuracy in demanding settings, motivating improved sensing and algorithms.
In the context of vision-/language-guided robot control, Fan et al.~\cite{fanVisionlanguageModelbasedHumanrobot2025} explicitly highlight the gap between household-style demonstrations and industrial requirements, including the need for higher precision and improved motion planning accuracy while retaining flexibility and generalisation.
Therefore, precision and accuracy form an implication in their own right: industrial deployment demands consistent, quality-aligned execution and perception accuracy, often under noisy or varying sensing and environmental conditions, while preserving the required throughput to enable efficient workflows~\cite{calderon-cordovaDeepReinforcementLearning2024,fanVisionlanguageModelbasedHumanrobot2025,prajapatiComputerVisionEnabled2024,mezgarTransformingExperimentalCobot2023}.
\runinheading{Implication 6: Real-Time Performance}
Real-time capability is presented as a pervasive constraint affecting sensing, decision-making, interaction and control.
The corpus references execution-time considerations (e.g. action-level and task-level timing)~\cite{johannsmeierHierarchicalHumanRobotInteractionPlanning2017} and highlights low-level requirements such as real-time communication and motion control in integrated robotic systems~\cite{tanApplicationOptimizationRobot2024}.
For industrial autonomy more broadly, authors emphasise real-time sensing and processing needs, including the acquisition of real-time environmental information and the integration of multi-modal signals to enable both sudden responses and higher-level reasoning~\cite{saleemReviewExternalSensors2025,yangEnhancingToFSensor2024,renEmbodiedIntelligenceFuture2024}.
Several works connect real-time feasibility to computational and resource constraints, especially for edge deployment, and note that data-driven methods are often limited by computational complexity and large training/inference requirements~\cite{desimoneContextawareDataAugmentation2025,chenRoleMachineLearning2024,dIntelligentControlAlgorithms2025}.
Indeed, as illustrated in Table~\ref{tab:edge}, even the most capable contemporary edge-devices lag significantly behind modern workstation \glspl{gpu} in computational power, as their design prioritises reasonable energy consumption.
Nevertheless, real-time performance must still be achieved on these resource-constrained platforms.
Besides computational concerns, real-time planning in dynamic scenes is explicitly identified as an unresolved challenge in vision-language(-action) discussions, while recent work also points to the possibility of smaller models enabling real-time inference in constrained settings~\cite{fanVisionlanguageModelbasedHumanrobot2025,sapkotaVisionLanguageActionModelsConcepts2025}.
As a result, real-time performance emerges as an implication spanning the full system: industrial robots must perceive, decide, communicate, and act within tight latency bounds despite environmental complexity and limited compute budgets~\cite{youEvolutionIndustrialRobots2024,manonmaniLiteratureReviewAdvanced2025}.
\runinheading{Implication 7: Cost-Effectiveness \textup{\&} Integration Capabilities}
A recurring industrial framing is that automation must be economically justified, including for low-volume production and for \glspl{sme}.
Authors argue that solutions must reduce manual effort, errors, and operational costs, and that high system cost remains a barrier to wider adoption of advanced autonomy~\cite{duijkerenIndustrialPerspectiveMultiAgent2023,prajapatiComputerVisionEnabled2024,weissFirstApplicationRobot2016}.
Multiple works explicitly discuss that frequent product changes increase reprogramming effort and costs, motivating approaches that reduce commissioning overhead and ease reconfiguration~\cite{herreroEnhancedFlexibilityReusability2017,abuzaryaqoobmullaAIRoboticsDesigning2023}.
Beyond acquisition cost, Triantafyllou et al.~\cite{triantafyllouMethodologyApproachingIntegration2021a} stress integration as a practical obstacle: deploying state-of-the-art research components in industrial systems requires coherent integration methodologies, yet community-wide standards and guidelines are described as lacking.
This concern is also reflected in specific domains such as AGVs/AMRs, where robust integration into factory architectures is highlighted as necessary for smart manufacturing success~\cite{oyekanluReviewRecentAdvances2020}.
Further, authors link industrial adoption to scalability and cost-effectiveness of intelligent control, as well as to the ability to operate with cost-sensitive hardware and sensing configurations~\cite{pelletiIntelligentControlSystems2024,linderCrossModalAnalysisHuman2021}, aligning with our observation that an efficient deployment of control methods on low-power edge devices is a necessity, as modern high-tier \glspl{gpu} can consume more than twice the power of a median \gls{cobot} itself (cf. Section~\ref{ssec:cch-ipc}).
Hence, cost-effectiveness and integration capabilities emerge as an implication: industrial solutions must be economically viable and practically integrable into existing production environments, including under heterogeneous, resource-limited hardware and organisational constraints~\cite{grauRobotsIndustryPresent2021,fanVisionlanguageModelbasedHumanrobot2025,pelletiIntelligentControlSystems2024}.
\runinheading{Implication 8: Explainability \textup{\&} Trust}
Authors repeatedly link the adoption of \gls{ai}-driven robotics to transparency, interpretability, and operator trust -- particularly in settings where failures have severe consequences.
Alt et al.~\cite{altHumanAIInteractionIndustrial2024} argue that explainability mechanisms, together with suitable user interfaces for presenting explanations, are key enablers for practical uptake of \gls{ai} in industry.
For safety- and risk-sensitive deployments, the limited adoption of perception models is explicitly attributed to the difficulty of understanding and trusting model outputs under potential failure consequences, motivating explainable approaches beyond simple classification settings~\cite{andresBlackboxExplainabilityObject2024}.
Trust is also discussed as a determinant of acceptance in collaborative work, with concerns that fear of failure can reduce productivity and that appropriate trust levels are important success factors~\cite{rodriguez-guerraHumanRobotInteractionReview2021,koppSuccessFactorsIntroducing2021,mezgarTransformingExperimentalCobot2023}.
Related literature additionally emphasises psychological and ergonomic aspects -- operators should feel comfortable and safe, and interfaces should provide visibility into system status and feedback about actions and responses~\cite{robla-gomezWorkingTogetherReview2017,pratiUseInteractionDesign2022}.
Empirical observations further suggest that limitations in sensing and flexibility can negatively affect perceived safety and trust in industrial collaboration~\cite{weissFirstApplicationRobot2016}.
Therefore, explainability and trust form an implication: industrial systems are expected not only to perform well, but also to provide comprehensible accounts of their behaviour and status to support acceptance, oversight, and safe collaboration~\cite{patricioFrameworkIntegratingRobotic2025,altHumanAIInteractionIndustrial2024}.
\runinheading{Implication 9: Sensor Fusion \textup{\&} Perception}
Authors frequently describe perception as a central capability limiting automation scope, particularly in shared industrial workspaces and variable environments.
Seleem et al.~\cite{saleemReviewExternalSensors2025} explicitly state that no single sensing modality is sufficient to handle varying complex situations in real time, motivating multi-modal sensing and fusion.
This includes combining internal and external sensors for detecting humans and obstacles~\cite{saleemReviewExternalSensors2025,oyekanluReviewRecentAdvances2020} and using complementary sensing such as vision and tactile feedback to extend perception and improve safe collaboration~\cite{mohammadiaminMixedPerceptionApproachSafe2020,nahavandiMachineLearningMeets2023}.
The corpus further connects sensor fusion to robustness: occlusions, dust, lighting variation, and sensor degradation can impair single-sensor solutions, and multiple works describe fusion or multi-sensor setups as mitigation strategies~\cite{halmeReviewVisionbasedSafety2018,menolottoMotionCaptureTechnology2020}.
At the same time, perception is framed as a prerequisite for real-time \gls{hrc}, requiring holistic scene understanding that includes object recognition, environment parsing, and human recognition~\cite{nahavandiMachineLearningMeets2023,bonciHumanRobotPerceptionIndustrial2021}.
Authors also highlight that advanced industrial tasks often require customised toolings and additional sensors, implying that control systems must accommodate heterogeneous and evolving observation spaces~\cite{dzedzickisAdvancedApplicationsIndustrial2021,makulaviciusIndustrialRobotsMechanical2023}.
Hence, when implemented on \glspl{cobot}, industry requires control methods that can accommodate the diverse and dynamic modalities arising from the sensor configurations commonly used in conjunction with \glspl{cobot} (cf. Section~\ref{ssec:cch-shw}).
Consequently, sensor fusion and perception emerge as an implication: industrial deployment demands multi-modal, robust perception pipelines that can support safe, reliable operation and collaboration under real-world sensing limitations~\cite{renEmbodiedIntelligenceFuture2024,wunderleSensorFusionFunctional2022}.
\runinheading{Implication 10: Standardised Benchmarking \textup{\&} Evaluation}
Multiple works emphasise the need for replicable, standardised performance assessment to support adoption and comparability across systems.
For example, standard test methods are described as necessary so that reported measurements can be reproduced with limited cost and effort~\cite{oyekanluReviewRecentAdvances2020}.
At the same time, Yang et al.~\cite{yangCollaborativeMobileIndustrial2019} identify a lack of unified standards and metrics for complex, component-rich industrial robotic systems, which complicates evaluation and hampers dissemination.
Further, Sarathkumar et al.~\cite{dIntelligentControlAlgorithms2025} argue more generally for benchmark and protocol development for intelligent control algorithms, including best practices that improve interoperability and reliability.
In manufacturing-focused discussions, there is an explicit expectation for training and evaluation systems that can replay manufacturing tasks precisely and provide suitable interfaces, reflecting dissatisfaction with benchmarks that do not reflect industrial task realities~\cite{liuRobotLearningSmart2022}.
Complementary perspectives highlight the role of virtualisation and digital twins as an enabler for testing and monitoring industrial systems~\cite{grauRobotsIndustryPresent2021,othmanHumanRobotCollaborations2023}.
In addition, in our own prior work we argue for using multiple, distinct benchmarks to better characterise generalisation and distribution shifts rather than relying on a single testbed~\cite{kubePerformanceExplainingGeneralisation2025}.
Overall, the literature motivates standardised benchmarking and evaluation as an implication: credible industrial adoption requires reproducible assessments that meaningfully reflect industrial tasks, conditions, and system complexity~\cite{dIntelligentControlAlgorithms2025,liuRobotLearningSmart2022}.
\runinheading{Implication 11: Data Requirements \textup{\&} Usage}
Finally, a consistent theme is that industrial robotics is data-constrained and that learning-based methods must address both data scarcity and deployment gaps.
Several works explicitly note limited labelled data in industrial domains and the difficulty of building representative datasets~\cite{desimoneContextawareDataAugmentation2025,linderCrossModalAnalysisHuman2021}.
Related observations include that small datasets limit the strength of conclusions for specialised industrial use cases, underscoring how narrow and expensive data collection can be in practice~\cite{andresBlackboxExplainabilityObject2024}.
In parallel, authors caution that many advanced approaches demand large volumes of training data and substantial computational resources, which can be misaligned with agile manufacturing contexts~\cite{fanEmbodiedIntelligenceManufacturing2025,dIntelligentControlAlgorithms2025}.
Calderón-cordova et al.~\cite{calderon-cordovaDeepReinforcementLearning2024} also discuss the importance of bridging simulated and real-world conditions, noting limitations of simulation fidelity and the resulting transfer gap for vision-based inputs.
Digital-twin perspectives reinforce that data infrastructure is foundational for industrial modelling and services~\cite{huangSurveyAIDrivenDigital2021}.
For future embodied intelligence and mass personalisation, Ren et al.~\cite{renEmbodiedIntelligenceFuture2024} argue for continuous adaptation over time, enabled by automatic data collection and continual learning.
In the specific context of vision-language(-action) approaches, the literature highlights the need for large-scale multimodal data and sim-to-real adaptation strategies to improve transfer and robustness, alongside calls to reduce data and compute requirements for industrial deployment~\cite{sapkotaVisionLanguageActionModelsConcepts2025,fanVisionlanguageModelbasedHumanrobot2025}.
Finally, data-efficient methods, such as few-shot learning, are explicitly motivated as a practical necessity in industrial settings~\cite{tsoumplekasEvaluatingEnergyEfficiency2024}.
Thus, data requirements and usage form a distinct implication: industrially relevant learning-based robotics must contend with limited, costly, and heterogeneous data while supporting transfer and adaptation under realistic deployment gaps~\cite{linderCrossModalAnalysisHuman2021,calderon-cordovaDeepReinforcementLearning2024,tsoumplekasEvaluatingEnergyEfficiency2024,sapkotaVisionLanguageActionModelsConcepts2025}.
\runinheading{Synthesis \textup{\&} relevance for industrial assessment of RFMs}
In summary, the implication corpus does not merely enumerate desirable properties, rather it repeatedly frames them as adoption bottlenecks and as reasons why laboratory-grade learning-based approaches often fail to translate into production environments~\cite{liuRobotLearningSmart2022,weissFirstApplicationRobot2016,fanEmbodiedIntelligenceManufacturing2025}.
The eleven proposed implications above therefore serve as the organising structure for the remainder of this section: they are used to discuss which attributes are required for \glspl{rfm} to be credible candidates for industrial deployment (Section~\ref{ssec:ii-am}), and motivate a later, explicit assessment framework for industrial readiness that is grounded in the recurring requirements identified in the reviewed literature (Section~\ref{ssec:ii-cc}).
\subsection{Attributes of Industry-Grade RFMs}\label{ssec:ii-am}
To operationalise the eleven industrial implications from Section~\ref{ssec:ii-gi} for the \gls{rfm} paradigm, we performed a second synthesis step that translates the largely qualitative, application-facing observations of the implication corpus into a structured set of \emph{model- and deployment-relevant viewpoints}.
The motivation for this step is that the implications themselves describe \emph{what} industry expects (e.g. safe collaboration, robustness under uncertainty, or real-time feasibility), but do not yet specify \emph{where} in an \gls{rfm}-based system such expectations could manifest. \\
We therefore analysed each implication individually through a set of derived attributes that capture the principal degrees of freedom of industrial \gls{rfm} deployment.
This attribute set was designed to be sufficiently broad to cover both implications regarding embodied-operation (e.g. hardware, sensing, actions) and organisation (e.g. documentation, compliance), while remaining granular enough to express typical failure modes and deployment bottlenecks discussed across the industrial robotics literature (Section~\ref{ssec:ii-gi}). \\
In total, we consolidated $47$ attributes as shown in Table~\ref{tab:attributes-overview}.
\colorlet{AttrRow}{CoolGray!50}
\begin{table*}[t]
	\centering
	\renewcommand{\arraystretch}{1.5}
	\caption{Consolidated list of the 47 attributes considered for each implication.}
	\label{tab:attributes-overview}
	\footnotesize
	\setlength{\tabcolsep}{3.5pt}
	\begin{tabular}{
			>{\raggedright\arraybackslash}p{0.24\textwidth}
			>{\raggedright\arraybackslash}p{0.24\textwidth}
			>{\raggedright\arraybackslash}p{0.24\textwidth}
			>{\raggedright\arraybackslash}p{0.24\textwidth}
		}
		\rowcolor{AttrRow} \textbf{(1)} Hardware &
		\textbf{(13)} Security &
		\textbf{(25)} Actions/ Outputs &
		\textbf{(37)} Precision \& Accuracy \\
		\rowcolor{AttrRow} \textbf{(2)} Tasks &
		\textbf{(14)} Detection, Recognition \& Understanding &
		\textbf{(26)} Skills &
		\textbf{(38)} Trust \\
		\rowcolor{AttrRow} \textbf{(3)} Environment &
		\textbf{(15)} Multimodal Inputs/ Outputs &
		\textbf{(27)} Observation/ Input &
		\textbf{(39)} Goal Specification \& Interpretation \\
		\rowcolor{AttrRow} \textbf{(4)} Objects/ Materials &
		\textbf{(16)} Integration &
		\textbf{(28)} External Influences/ Disturbances &
		\textbf{(40)} Solution Strategy \\
		\rowcolor{AttrRow} \textbf{(5)} Instructability/ Commanding &
		\textbf{(17)} Safety &
		\textbf{(29)} Reaction &
		\textbf{(41)} Offline Adaptation (Finetuning/ Retraining) \\
		\rowcolor{AttrRow} \textbf{(6)} Distractors/ Distractions &
		\textbf{(18)} Compliance &
		\textbf{(30)} Validation/ Benchmark/ Evaluation &
		\textbf{(42)} Energy \\
		\rowcolor{AttrRow} \textbf{(7)} Adaptability/ Generalisation/ Flexibility &
		\textbf{(19)} Supervision &
		\textbf{(31)} Realtime Performance \& Inference Frequency &
		\textbf{(43)} Data \\
		\rowcolor{AttrRow} \textbf{(8)} Online Adaptation/ Learning &
		\textbf{(20)} Fault Tolerance \& Recovery/ Safeguards &
		\textbf{(32)} Noise &
		\textbf{(44)} Documentation/ Open-Source \\
		\rowcolor{AttrRow} \textbf{(9)} Internal Interfaces/ Representations &
		\textbf{(21)} Reasoning/ Intention &
		\textbf{(33)} Autonomy &
		\textbf{(45)} (Self-)Awareness/ Assessment \& Monitoring \\
		\rowcolor{AttrRow} \textbf{(10)} ID/ OOD Robustness &
		\textbf{(22)} Model Feedback to User &
		\textbf{(34)} Consistency \& Stability &
		\textbf{(46)} Memory \\
		\rowcolor{AttrRow} \textbf{(11)} Adversarial Robustness &
		\textbf{(23)} External Guidance/ Correction to Model &
		\textbf{(35)} Explainability/ Insights &
		\textbf{(47)} Anomaly/ Fault/ Error Detection \& Recognition \\
		\rowcolor{AttrRow} \textbf{(12)} Authorisation &
		\textbf{(24)} Interaction/ Collaboration &
		\textbf{(36)} Repeatability &
		\\
	\end{tabular}
	\setlength{\tabcolsep}{6pt}
\end{table*}
Using these attributes, we conducted an iterative mapping procedure in which we revisited each of the eleven implications and asked, for each attribute, what the implication would demand from an \gls{rfm}-based robotic system when that system is viewed through the lens of the respective attribute.
This yields $11\times47=517$ implication-attribute combinations, forming a conceptual matrix in which each cell captures the relevant industrial concerns and constraints for that attribute under the given implication, and their specific interpretation in terms of \gls{rfm} development and deployment. \\
This matrix-based perspective is particularly important because the implications are interdependent: for industrial deployments, safety is not separable from perception, and real-time feasibility is not separable from interaction or robustness.
Similarly, issues such as authorisation, adversarial robustness, and fault handling naturally span multiple implications (e.g. safety, robustness, trust, and integration).
By enforcing a systematic traversal across all implication-attribute pairs, we aimed to reduce the risk of under-representing industrial requirements that might otherwise be subsumed under a single headline implication. \\
The primary outcome of this step is a consolidated, implication-grounded view of what it would mean for \glspl{rfm} to be \emph{industry-grade}.
Rather than proposing a single, monolithic list of requirements, the attribute set enables a structured discussion of how industrial expectations manifest at different layers -- ranging from sensing and action interfaces, over adaptation behaviour, to governance-related aspects such as compliance and documentation. \\
At the same time, the full matrix is too extensive to be reproduced verbatim in this work without disproportionate volume.
We therefore use the attribute set as a compact, transparent intermediate representation: it makes explicit which viewpoints were considered during the synthesis of the proposed criteria catalogue (Section \ref{ssec:ii-cc}), and it provides traceable coverage of the industrial concerns extracted from the implication corpus (Section~\ref{ssec:ii-gi}).
In the succeeding subsection, these structured considerations are subsequently distilled into an explicit assessment framework for industrial readiness, enabling a consistent evaluation of existing \glspl{rfm} against the industrial expectations highlighted throughout the literature.
\subsection{Evaluating Industrial Maturity: A Criteria Catalogue}\label{ssec:ii-cc}
The eleven implications in Section~\ref{ssec:ii-gi} describe \emph{what} industrial settings recurrently demand from deployed robotic systems.
However, for \glspl{rfm}, these expectations must be made concrete at the level of model capabilities, interfaces, and the surrounding deployment stack.
Building on the $517$ implication-attribute combinations from Section~\ref{ssec:ii-am}, we therefore translated the implication-grounded requirements into an explicit criteria catalogue that can be used to reason about, compare, and ultimately rate the industrial readiness of existing and future \glspl{rfm}. \\
For each implication-attribute combination, we collected and refined the key questions that an industrial deployment would have to answer (e.g. what happens under distribution shifts, how failures are detected and handled, how human input is interpreted and verified, and which artefacts are required for audit and integration).
We iteratively consolidated these questions into a \textbf{catalogue of 149 criteria (C)  -- shown in the Appendix, Table~\ref{tab:app-cc} --} spanning all eleven general industrial implications (I).
While comprehensive, we do not consider this catalogue final: it is intended as a living foundation that can be extended as additional industrial constraints or regulatory expectations emerge.
\runinheading{Criteria scope: RFMs \textup{\&} their deployment ecosystem}
Although most criteria target \glspl{rfm} directly, industrial readiness is not determined by the model alone.
Industrial deployments require an ecosystem -- robot software stack, interfaces, safety infrastructure, monitoring, logging, and documentation -- that must satisfy similar constraints.
Consequently, the catalogue includes criteria that (a) are primarily \gls{rfm}-centric (e.g. adaptive behaviour under changing observations/actions), (b) are primarily ecosystem-centric (e.g. the presence and maintenance of accessible documentation on training data composition and transfer behaviour), or (c) require joint support by model and ecosystem (e.g. a model producing structured signals that can be persistently logged and leveraged by the surrounding system).
\runinheading{Intentional redundancy across implications}
Because the industrial implications are interdependent (cf. Section~\ref{ssec:ii-am}), certain criteria re-appear \emph{deliberately} across different implications, albeit with different emphasis.
We therefore avoided merging superficially similar criteria across implications when doing so would remove an aspect that is essential for assessing maturity with respect to a particular implication.
For example, hot-swapping appears under adaptability (I1-C2), integration/cost (I7-C4), and perception (I9-C12), but each instance stresses a different industrial concern: autonomous adjustment of observation/action spaces, recognition and handling of various interface changes during commissioning, or modularity of the observation space alone and its documentation.
Retaining these criteria within their respective implications preserves the ability to diagnose \emph{why} an \gls{rfm} is immature in a specific industrial dimension rather than merely concluding that it "lacks hot-swapping support".
\runinheading{Key paradigms reflected in the catalogue}
Many of the catalogue's central motifs are rarely discussed \emph{jointly} in prior work and often not yet implemented in current \glspl{rfm}, however, most are already being investigated as distinct research directions scattered across the \gls{rfm}, \gls{vla}/\gls{llm} or general robotics control landscapes.
We therefore use the following paradigms as \emph{anchor points} to illustrate that a substantial subset of the catalogue’s requirements is directly supported by emerging literature, and that our contribution lies in consolidating these partially disconnected strands and enriching them from an industrial viewpoint to craft a single, implication-grounded assessment framework.\\
Central to criteria on multi-hardware support, hot-swapping, and dimensionality adaptation is the treatment of changing observation and action interfaces.
Recent work has begun to explicitly adapt \glspl{vla} to new embodiments by learning unified action representations and steering pre-trained policies towards platform-specific action distributions during downstream adaptation~\cite{zhangAlignThenstEerAdaptingVisionLanguage2025} or by enabling architectures to handle differing sized action and observation spaces~\cite{bohlingerOnePolicyRun2025}.
These approaches align closely with our emphasis that industrial deployment requires transfer across morphologies, whereas a natural extension emerges in runtime resilience to interface changes and sensor/actuator substitutions.\\
Changing action- and observation spaces naturally motivate concerns about handling \gls{ood} scenarios.
Challenging industrial environments are indeed characterised by disturbances, clutter, and changes that are not exhaustively covered during training, requiring robustness against distribution shifts, anomalies, and deliberate manipulation.
Accordingly, our catalogue places repeated emphasis on online detection, confidence signalling, and fallback behaviour.
Complementary to this, emerging evidence suggests that current \gls{vla} systems may inherit security weaknesses known from \glspl{llm} and \glspl{vlm}, amplifying their consequences. 
Both textual attacks that can persist over long horizons and vision-side attacks (e.g. through disrupting patch exposure) have been shown to substantially degrade performance or even obtain absolute control authority of \glspl{vla}, highlighting that adversarial robustness cannot be treated as an optional add-on for physically embodied systems.~\cite{jonesAdversarialAttacksRobotic2025, xuModelagnosticAdversarialAttack2025}
In parallel, anomaly prediction has been explored as a proactive mechanism, for instance by forecasting action outcomes and anticipating whether they lead to anomalous states~\cite{liuMultiTaskInteractiveRobot2024}.
This combination motivates why our catalogue repeatedly requires threat/attack detection, anomaly handling, and explicit recovery pathways: in industrial settings, models must differentiate between benign novelty and adversarial or safety-critical deviations.\\
These mechanisms also depend on the system’s ability to quantify and communicate uncertainty in a usable manner.
We therefore include multiple confidence-related criteria (spanning operational readiness, safety margin, real-time compliance, perceptual certainty, and \gls{vpa} fulfilment), reflecting the view that trust requires calibrated estimates rather than raw success rates.
This is supported by work that explore confidence calibration in \glspl{vla} for uncertainty quantification~\cite{zolloConfidenceCalibrationVisionLanguageAction2025}, as well as methods that align uncertainty to trigger help-seeking behaviour with statistical guarantees for \gls{llm}-based planning settings~\cite{renRobotsThatAsk2023}.
Such capabilities enable explicit halting, logging, clarification requests, or escalation to supervisors when the system’s own predicted reliability degrades.\\
Confidence also naturally links to interactive clarification and feedback, repeatedly mentioned in our catalogue, which are essential for industrial \gls{hri}/\gls{hrc} where instructions can be ambiguous, incomplete, or safety-constrained.
Recent approaches explicitly reason about linguistic ambiguity grounded in the scene and generate targeted follow-up questions, thereby reducing downstream execution failures caused by \gls{nl} command-misinterpretation~\cite{chisariRoboticTaskAmbiguity2025}.
Conversely, other works demonstrate that \gls{nl} can serve as an intuitive channel for fine-grained trajectory modification and incremental corrections, suggesting practical routes towards mid-execution trajectory adjustability and operator-in-the-loop refinement as demanded by several criteria~\cite{buckerLATTELAnguageTrajectory2023,shiYellYourRobot2024}.
More broadly, collaboration frameworks that incorporate expert interventions during deployment indicate that interactive supervision can simultaneously improve immediate reliability and collect data for future refinement~\cite{xiangVLAModelExpertCollaboration2025,liuMultiTaskInteractiveRobot2024}.\\
A further recurring motif within our catalogue is the need for explicit supervisory or verification layers that monitor execution quality and safety independently of the core policy.
This direction is reflected by Zhou et al.~\cite{zhouCodeasMonitorConstraintawareVisual2025} that leverage \glspl{vlm} for reactive and proactive failure detection, and methods designed for general failure detection across multiple tasks rather than narrow, task-specific detectors~\cite{guSAFEMultitaskFailure2025}, which is essential for task-agnostic policies.
Related efforts target the gap between semantic failure understanding and actionable correction by explicitly bridging high-level reflection and low-level action adjustment~\cite{xiaPhoenixMotionbasedSelfReflection2025}.
Such results substantiate the repeated insistence on decoupled safety gating, redundant safety mechanisms, fault categorisation, and explainable recovery behaviour: industrial assurance typically requires independent checks, predictable intervention pathways, and evidence of consistent handling across task variation.\\
Finally, several works reinforce why the catalogue treats memory, data governance, and documentation as recurring concerns.
Methods that add lightweight memory modules to \glspl{vla} highlight that exploiting historical context can materially improve long-horizon performance~\cite{kooHAMLETSwitchYour2025}, while other frameworks explicitly summarise prior experience to support failure reasoning and replanning~\cite{liuREFLECTSummarizingRobot2023}.
At the same time, privacy-preserving training mechanisms for \glspl{vla}~\cite{miaoFedVLAFederatedVisionLanguageAction2025} imply that industrial adoption may be constrained by data protection requirements, not by model capability alone.
In addition, dataset construction efforts that deliberately incorporate failures and recoveries imply that training data must cover various edge- or failure-cases to enable robust deployment behaviour~\cite{linFailSafeReasoningRecovery2025}.
These developments collectively support the inclusion of criteria on persistent logging, memory mechanisms, data composition variation and transparency, privacy compliance, and learning from faults and incidents.\\
Taken together, these strands illustrate that many catalogue requirements correspond to active research directions -- the industrial immaturity is less the absence of isolated technical proposals than the lack of their systematic combination into a deployable, auditable, and safety-governed \gls{rfm} ecosystem.
The criteria catalogue therefore consolidates these partially disconnected lines into a unified framework that is directly grounded in the implication corpus and tailored to the operational constraints of industrial deployment, serving as a structured bridge between industrial expectations and \gls{rfm} assessment: it translates the implication-level requirements into concrete items that can be used to analyse where current \glspl{rfm} are industrial mature, where they are incomplete, and which research directions are likely necessary for credible industrial uptake.
In the survey section that follows, we use this framework as a reference lens to discuss existing \glspl{rfm} and to highlight gaps between current capabilities and the implication-derived expectations.
\section{RFM Overview \& Evaluation}\label{sec:survey}
This section provides a structured overview of the current landscape of manipulation-capable \glspl{rfm} and assesses their industrial readiness using the implication-grounded criteria catalogue introduced in Section~\ref{ssec:ii-cc}.
For transparency and overview, we report the full corpus of all $324$ surveyed models in the Appendix (Table~\ref{tab:rfms-all}). \\
We evaluate each model against the 149 catalogue criteria, enabling both an overall maturity score per paper and implication-wise maturity profiles.
We describe the resulting evaluation methodology and procedure in Section~\ref{ssec:survey-proc}, including the design decisions required to scale the assessment to the full corpus, corresponding to 324×149=48,276 criterion-level decisions. \\
We additionally report an agreement and consistency check of our evaluation pipeline against expert judgements on a small ground-truth subset in Section~\ref{ssec:survey-proc}.
Finally, we summarise the strongest-performing \glspl{rfm} (Section~\ref{ssec:survey-highest}) and provide a broader industrial applicability analysis over the full corpus and the Top-50 subset (Section~\ref{ssec:survey-overview}).
The combination of those results is used to characterise the current industrial maturity of \glspl{rfm}, highlight which industrial implications are most and least addressed, and finally provide an evidence-based baseline for the synthesis and outlook in Section~\ref{sec:summary}.
\subsection{Evaluation Procedure}\label{ssec:survey-proc}
To assess industrial readiness across the manipulation-capable \gls{rfm} landscape, we evaluated all 324 manipulation-focused or generalist \glspl{rfm} contained in the models sub-corpus of our main \gls{rfm}-related corpus (out of 341 total model publications; cf. Section~\ref{ssec:methods-crfm}).
Each paper was assessed against the 149 criteria of our implication-grounded catalogue (cf. Section~\ref{ssec:ii-cc}, Appendix~\ref{tab:rfms-all}), yielding $324\times149=48,276$ criterion-level decisions.
A fully manual expert annotation of this scale is infeasible within a fixed time frame and would furthermore restrict the analysis to a small, subjectively filtered subset, limiting the validity of corpus-level conclusions.
We therefore adopted an automated, \gls{llm}-based evaluation pipeline to (a) enable a complete-corpus assessment, (b) reduce reviewer-dependent variability via a fixed and mostly repeatable procedure, and (c) obtain both a global maturity score per paper (fraction of fulfilled criteria) and implication-wise maturity profiles (fraction of fulfilled criteria per implication). \\
Our evaluation was implemented as an objective, structured pipeline that leverages an \gls{llm} (GPT-5.1 in our case) to extract evidence from the full text of a paper and to decide whether a single criterion is fulfilled by the proposed model and/or its deployment ecosystem (as defined by the criterion; cf. Section~\ref{ssec:ii-cc}).
A key design goal was to minimise cross-contamination between papers and between criteria.
Accordingly, each \emph{single} evaluation run was performed under two strict isolation constraints: (a) the \gls{llm} received exactly one criterion at a time (and was not informed about any other criteria), and (b) the retrieval context contained exactly one paper (including figures), provided through a dedicated vector store that was created for that evaluation run only.
After the decision for that criterion-paper pair, the vector store was deleted, the session was closed, and the next criterion-paper pair was evaluated in a fresh run.
This enforced independence across the $48,276$ evaluations and prevented leakage of information or implicit calibration across different models or criteria.
Importantly, \textbf{all papers were processed using private, Siemens-internal infrastructure}.
In particular, no papers were uploaded to public servers, and the evaluation was executed on private systems where the processed documents are not used for any training/ learning.
This ensured that the large-scale assessment did not create any disclosure or copyright-related risks.
\\
For every paper-criterion pair, the \gls{llm} was required to produce a structured output with the following fields: a binary judgement of criterion fulfilment, a flag indicating whether the property was directly stated, a confidence score in the range $[0, 1]$ indicating explicitness and evidence strength, a boolean evidence-availability signal, textual extracted quotes (or an explicit statement that no relevant information was found), and a brief rationale connecting evidence to the decision.
Importantly, we instructed the \gls{llm} to decide conservatively: if a criterion was not supported clearly and in-scope by the paper, it was to be marked as \emph{not} fulfilled.
This makes the resulting maturity estimates intentionally cautious, particularly for criteria that are only partially addressed, or for capabilities that are not sufficiently documented in the publication.
The complete corpus run required over $5$ days under rate limits, further underscoring why a purely manual evaluation is practically unattainable at this scale.
%%%
\runinheading{Expert Agreement Comparison}
Because the pipeline is used to draw corpus-level conclusions, we evaluated its reliability against expert judgements on a small ground-truth subset.
Specifically, we created three blind expert-rated catalogue evaluations: two models were annotated by an expert prior to running the automated pipeline, and one additional model was annotated after the automated run had completed (the best-scoring model according to the \gls{llm} results).
These three papers serve as a check on whether the procedure generates results with sufficient accuracy to draw corpus-level conclusions.
\begin{table*}[t]
	\centering
	\caption{Comparison between expert- and LLM-based evaluation results per implication for three representative RFMs.}
	\label{tab:gtvseval}
	\renewcommand{\arraystretch}{1.3}
	\setlength{\tabcolsep}{8pt}
	
	\begin{tabular}{>{\centering\arraybackslash}m{2.2cm}
			>{\centering\arraybackslash}m{1.85cm}
			>{\centering\arraybackslash}m{1.6cm}
			>{\centering\arraybackslash}m{1.6cm}
			>{\centering\arraybackslash}m{1.6cm}
			>{\centering\arraybackslash}m{1.6cm}
			>{\centering\arraybackslash}m{1.6cm}}
		
		\rowcolor{CoolGray!50}
		\multicolumn{1}{>{\centering\arraybackslash\cellcolor{Platinum!60}}m{2.2cm}}{} &
		\multicolumn{2}{c}{Gemini Robotics 1.5~\cite{teamGeminiRobotics152025}} &
		\multicolumn{2}{c}{OpenVLA~\cite{kimOpenVLAOpenSourceVisionLanguageAction2024}} &
		\multicolumn{2}{c}{$\pi_0$~\cite{black$p_0$VisionLanguageActionFlow2024}} \\
		
		\rowcolor{Platinum!60}
		\cellcolor{CoolGray!50}Implication &
		Expert & LLM &
		Expert & LLM &
		Expert & LLM \\

		\cellcolor{CoolGray!50} I1  & 9&8 & 3&4 & 1&1 \\
		\cellcolor{CoolGray!50} I2  & 1&1 & 0&0 & 0&0 \\
		\cellcolor{CoolGray!50} I3  & 1&1 & 0&0 & 0&0 \\
		\cellcolor{CoolGray!50} I4  & 1&1 & 1&1 & 0&0 \\
		\cellcolor{CoolGray!50} I5  & 0&0 & 0&0 & 0&0 \\
		\cellcolor{CoolGray!50} I6  & 0&0 & 0&0 & 0&0 \\
		\cellcolor{CoolGray!50} I7  & 0&0 & 0&0 & 0&0 \\
		\cellcolor{CoolGray!50} I8  & 1&1 & 0&0 & 0&0 \\ 
		\cellcolor{CoolGray!50} I9  & 1&2 & 0&0 & 1&0 \\ 
		\cellcolor{CoolGray!50} I10 & 2&3 & 5&5 & 1&1 \\ 
		\cellcolor{CoolGray!50} I11 & 2&1 & 0&1 & 1&1 \\
		% SUM
		\hline
		\cellcolor{CoolGray!50} $\sum$ & 18&18 & 9&11 & 4&3 \\
	\end{tabular}
\end{table*}
Table~\ref{tab:gtvseval} shows absolute values, specifying how many criteria per paper-implication pair have been marked as fulfilled by expert and \gls{llm}.
The table reveals that the \gls{llm}-based procedure closely reproduces the implication-level patterns of the expert assessments for these representative cases.
In particular, the qualitative conclusion is preserved across all three models: Gemini~Robotics~1.5~\cite{teamGeminiRobotics152025} attains substantially higher industrial maturity than OpenVLA~\cite{kimOpenVLAOpenSourceVisionLanguageAction2024}, while $\pi_0$~\cite{black$p_0$VisionLanguageActionFlow2024} remains limited across most implications.
\begin{table}[t]
	\centering
	\small
	\renewcommand{\arraystretch}{1.3}
	\setlength{\tabcolsep}{4pt}
	\caption{LLM-expert agreement on criterion fulfillment.
	\textit{Reporting the number of N evaluations considered, the overall Accuracy (Acy), $F_1$-score, Cohen's kappa agreement ($\kappa$) and false positive/ negative rates (FPR/FNR).}}
	\label{tab:llm_expert_agreement}
	\begin{tabular}{cccccc}
		\rowcolor{CoolGray!50}
		\cellcolor{Platinum!60}Setting & $N$ & Acy & F1 & $\kappa$ & FPR/FNR \\
		\cellcolor{CoolGray!50}All & 447 & 0.966 & 0.762 & 0.744 & 1.9\%/22.6\% \\
		\cellcolor{CoolGray!50}InfoFound & 304 & 0.951 & 0.762 & 0.734 & 2.9\%/22.6\% \\
	\end{tabular}
\end{table}
\\
Across the three expert-annotated papers, Table~\ref{tab:llm_expert_agreement} reports strong criterion-level agreement \mbox{($N=447$)}: the \gls{llm} reaches an accuracy of $0.966$ and an $F_1$~score of $0.762$ for the positive (fulfilled) class, with substantial agreement beyond chance (Cohen’s $\kappa=0.744$).
Crucially for the maturity estimator, the false-positive rate is low ($FPR=1.9$\%), indicating that the \gls{llm} rarely over-attributes industrial capabilities that the expert does not confirm.
The false-negative rate is higher ($FNR=22.6$\%), consistent with our conservative decision policy. 
Consequently, the derived maturity values should be interpreted as a cautious, lower-bound estimate rather than an optimistic readiness score.
Restricting the analysis to those cases where the \gls{llm} reports information found as \emph{true}, (Table~\ref{tab:llm_expert_agreement}, \textit{InfoFound}) does not increase the $F_1$~score and slightly reduces $\kappa$.
This supports our intention that the information-found flag primarily captures evidence availability in the text, rather than serving as a reliable filter for correctness.
\begin{table}[t]
	\centering
	\small
	\renewcommand{\arraystretch}{1.3}
	\setlength{\tabcolsep}{3.2pt}
	\caption{Per-paper agreement. 
		\textit{Pos\% reports the expert positive rate and $J^+$ the Jaccard similarity index, computed only for the positive class to ignore the dominating negatives.}}
	\label{tab:llm_expert_agreement_perpaper}
	\begin{tabular}{cccccc}
		\rowcolor{CoolGray!50}
		\cellcolor{Platinum!60}Paper & Pos\% & Acy & F1 & $\kappa$ & $J^+$\\
		\cellcolor{CoolGray!50}Gemini 1.5~\cite{teamGeminiRobotics152025} & 0.121 & 0.960 & 0.833 & 0.810 & 0.714 \\
		\cellcolor{CoolGray!50}OpenVLA~\cite{kimOpenVLAOpenSourceVisionLanguageAction2024} & 0.060 & 0.973 & 0.800 & 0.786 & 0.667 \\
		\cellcolor{CoolGray!50}$\pi_0$~\cite{black$p_0$VisionLanguageActionFlow2024} & 0.027 & 0.966 & 0.286 & 0.269 & 0.167 \\
	\end{tabular}
\end{table}
\\
At the per-paper level (Table~\ref{tab:llm_expert_agreement_perpaper}), agreement between expert and \gls{llm} is high for both Gemini~Robotics~1.5 and OpenVLA, especially supported by strong $\kappa$ and $J^+$ (Jaccard similarity for fulfilled cases).
For $\pi_0$, the expert positive rate at only $2.7\%$ is notably low, with the corresponding $F_1$, $\kappa$ and $J^+$ substantially lower than for the other two papers.
This underlines that even a small number of disagreements can dominate positive-class metrics when positives are rare, however, also explains the higher false negative rates reported in Table~\ref{tab:llm_expert_agreement}.
Therefore, per-paper metrics should be read with care in \textit{very} low-positivity cases.\\
Nevertheless, lower agreement cases as reported for $\pi_0$ do not necessarily reflect substantial judgment errors, rather often report interpretation differences: e.g. Black et al.~\cite{black$p_0$VisionLanguageActionFlow2024} include a brief evaluation of language-following capabilities and report an increased task success rate when providing intermediate language commands to the policy. 
The expert acknowledged this as sufficient for I1-C11 ("Mid-Execution Feedback Responsiveness"), whereas the \gls{llm} argued that only an increase in success rate does not necessarily imply the requested ability of course corrections.\\
In fact, we observed a few patterns within our criteria that may inflate reported disagreement, while not necessarily introducing hard errors. 
As stated for the above example, some criteria may leave some room for interpretation, even among experts: regarding the "Low-Data Adaptation \& Few-Shot Requirements" criterion (I11-C1), OpenVLA reports requiring on average $10$-$150$ episodes for downstream task adaptation~\cite{kimOpenVLAOpenSourceVisionLanguageAction2024} -- the \gls{llm} interpreted this as few-shot and marked the criterion fulfilled, whereas our expert judgement considered up to $150$ episodes too demanding for an industrial-ready "low-data adaptation" interpretation.
This implies that some criteria may need stricter boundaries to ensure higher evaluator agreement.
In contrast, other criteria are inherently stringent (e.g. formulations that imply guarantees such as "always"), which leads the \gls{llm} to often mark such criteria as not fulfilled, even when the capability is basically implemented, but the paper does not state a near 100\% guarantee of adherence. \\
Taken together, the agreement comparison indicates that the proposed pipeline provides sufficiently reliable, conservative estimates for our goal of scaling industrial maturity assessment to hundreds of \glspl{rfm}.
The low over-claiming rate supports its use for comparing models and identifying implication-level blind spots in the current state of the art, while the elevated miss rate and the interpretative latitude of some criteria motivate caution when analysing individual, fine-grained criterion outcomes.
Accordingly, in the following, we treat criterion-level results as indicative and focus primarily on implication-level trends and cross-corpus patterns.
\subsection{Highest Rated RFMs}\label{ssec:survey-highest}
\begin{figure}[hbt]
	\centering
	\includegraphics[width=0.5\textwidth, trim=55pt 15pt 15pt 10pt, clip]{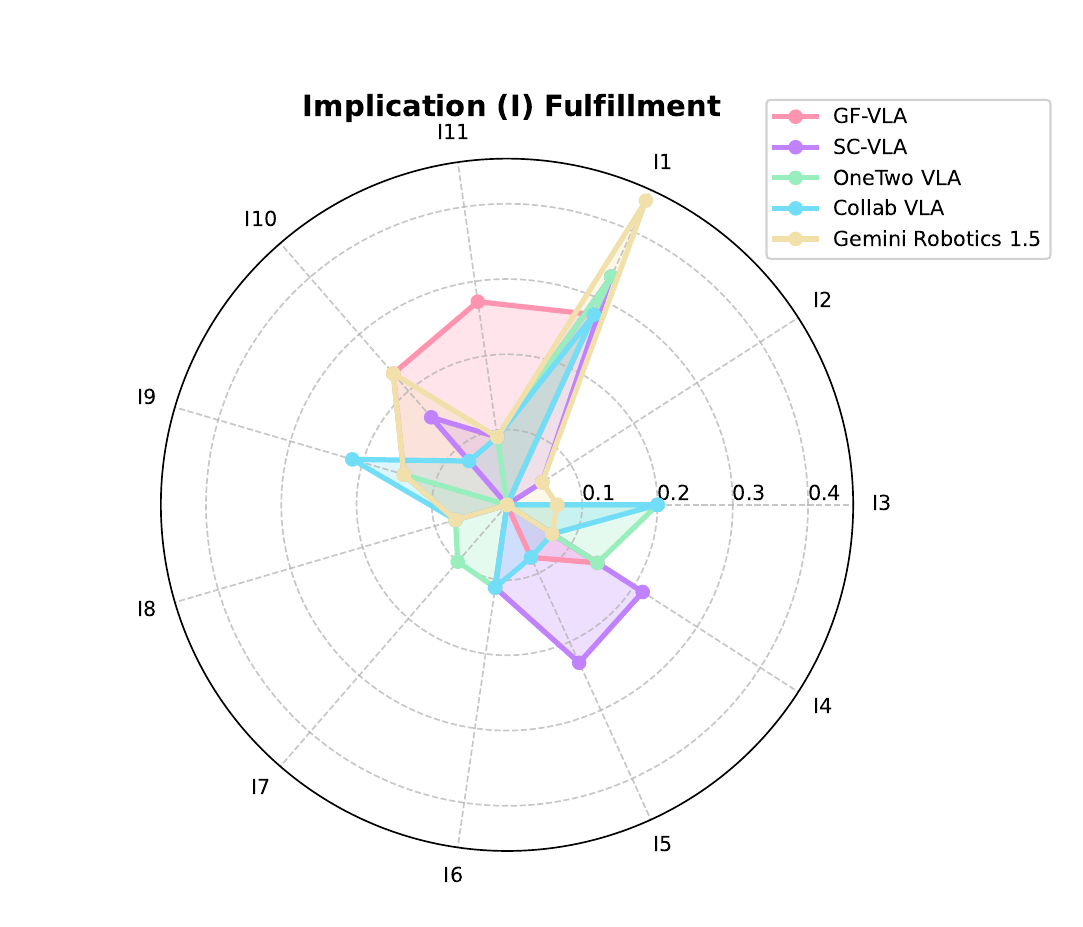}  % plotted with 0.25 opacity
	\caption{Proportional implication fulfillment. 
	\textit{Depicting the five overall highest-rated RFMs (see Table~\ref{tab:rfms-highest} for detailed values).}
	}
	\label{fig:spc_best5}
\end{figure}

Across the 324 evaluated \glspl{rfm} (cf. Table~\ref{tab:rfms-all}), the overall highest-rated models, as depicted in Table~\ref{tab:rfms-highest}, reach total maturity scores of only $0.11-0.12$, i.e. they satisfy roughly one out of ten catalogue criteria on average.
Although possibly lower-bound, this pronounced ceiling effect is informative in itself: even the strongest current approaches cover only a narrow subset of the industrial requirements captured by our implication-grounded framework, and thus no single model can yet be regarded as comprehensively industry-grade.
\begin{table*}[t]
	\centering
	\renewcommand{\arraystretch}{1.35}
	\small
	\caption{Overview of the five highest rated RFMs (\textbf{total}) of the $324$ evaluated.
		\textit{Reporting their proportional industrial applicability ratings for each individual implication, and a total score over all implications.
		Per-criterion ratings are created utilising the publication's context, by leveraging our LLM-based pipeline as described in Section~\ref{ssec:survey-proc}. Highest values within the five RFMs for each implication are shown in green, lowest in red. Values are rounded to two decimal places.}}
	\label{tab:rfms-highest}
	\begin{tabular}{@{}
			>{\centering\arraybackslash}m{0.9cm}
			>{\centering\arraybackslash}m{1.1cm}
			>{\centering\arraybackslash}m{1.1cm}
			*{11}{>{\centering\arraybackslash}m{0.62cm}}
			@{}}
		\rowcolor{CoolGray!50}
		\cellcolor{Platinum!60}\textbf{\rule{0pt}{2.5ex}Ref} &
		\textbf{\rule{0pt}{2.5ex}Model} &
		\textbf{\rule{0pt}{2.5ex}Total~$\downarrow$} &
		\textbf{\rule{0pt}{2.5ex}I1} &
		\textbf{\rule{0pt}{2.5ex}I2} &
		\textbf{\rule{0pt}{2.5ex}I3} &
		\textbf{\rule{0pt}{2.5ex}I4} &
		\textbf{\rule{0pt}{2.5ex}I5} &
		\textbf{\rule{0pt}{2.5ex}I6} &
		\textbf{\rule{0pt}{2.5ex}I7} &
		\textbf{\rule{0pt}{2.5ex}I8} &
		\textbf{\rule{0pt}{2.5ex}I9} &
		\textbf{\rule{0pt}{2.5ex}I10} &
		\textbf{\rule{0pt}{2.5ex}I11} \\
	\cellcolor{CoolGray!50}\cite{teamGeminiRobotics152025} & Gemini Robotics 1.5 & 0.12 & \cellcolor{MintGreen!60}0.44 & \cellcolor{MintGreen!60}0.06 & 0.07 & \cellcolor{LightRed!60}0.07 & \cellcolor{LightRed!60}0.00 & \cellcolor{LightRed!60}0.00 & \cellcolor{LightRed!60}0.00 & \cellcolor{MintGreen!60}0.07 & 0.14 & \cellcolor{MintGreen!60}0.23 & \cellcolor{LightRed!60}0.09\\
	\cellcolor{CoolGray!50}\cite{sunCollabVLASelfReflectiveVisionLanguageAction2025} & Collab VLA & 0.11 & \cellcolor{LightRed!60}0.28 & \cellcolor{LightRed!60}0.00 & \cellcolor{MintGreen!60}0.20 & \cellcolor{LightRed!60}0.07 & 0.08 & \cellcolor{MintGreen!60}0.11 & \cellcolor{LightRed!60}0.00 & \cellcolor{MintGreen!60}0.07 & \cellcolor{MintGreen!60}0.21 & 0.08 & \cellcolor{LightRed!60}0.09\\
	\cellcolor{CoolGray!50}\cite{liInformationTheoreticGraphFusion2025} & GF-VLA & 0.11 & \cellcolor{LightRed!60}0.28 & \cellcolor{LightRed!60}0.00 & \cellcolor{LightRed!60}0.00 & 0.14 & 0.08 & \cellcolor{LightRed!60}0.00 & \cellcolor{LightRed!60}0.00 & \cellcolor{MintGreen!60}0.07 & 0.14 & \cellcolor{MintGreen!60}0.23 & \cellcolor{MintGreen!60}0.27\\
	\cellcolor{CoolGray!50}\cite{linOneTwoVLAUnifiedVisionLanguageAction2025} & OneTwo VLA & 0.11 & 0.33 & \cellcolor{LightRed!60}0.00 & \cellcolor{MintGreen!60}0.20 & 0.14 & \cellcolor{LightRed!60}0.00 & \cellcolor{MintGreen!60}0.11 & \cellcolor{MintGreen!60}0.10 & \cellcolor{MintGreen!60}0.07 & 0.14 & \cellcolor{LightRed!60}0.00 & \cellcolor{LightRed!60}0.09\\
	\cellcolor{CoolGray!50}\cite{liSelfCorrectingVisionLanguageActionModel2025} & SC-VLA & 0.11 & 0.33 & \cellcolor{MintGreen!60}0.06 & \cellcolor{LightRed!60}0.00 & \cellcolor{MintGreen!60}0.21 & \cellcolor{MintGreen!60}0.23 & \cellcolor{MintGreen!60}0.11 & \cellcolor{LightRed!60}0.00 & \cellcolor{LightRed!60}0.00 & \cellcolor{LightRed!60}0.00 & 0.15 & \cellcolor{LightRed!60}0.09\\
	\end{tabular}
\end{table*}
\begin{table*}[t]
	\centering
	\renewcommand{\arraystretch}{1.35}
	\small
	\caption{Overview of the highest rated RFMs \textbf{per implication} of the $324$ total evaluated.
		\textit{Per-criterion ratings are created utilising the publication's context, by leveraging our LLM-based pipeline as described in Section~\ref{ssec:survey-proc}. Highest values within the respective implication are highlighted in green. Values are rounded to two decimal places. The entry "Multiple" represents the fact that several works have the similar highest score of 0.11 for I6, which are thus consolidated in one row.}}
	\label{tab:rfms-hpi}
	
	\begin{tabular}{@{}
			>{\centering\arraybackslash}m{0.9cm}
			>{\centering\arraybackslash}m{1.1cm}
			>{\centering\arraybackslash}m{1.1cm}
			*{11}{>{\centering\arraybackslash}m{0.62cm}}
			@{}}
		\rowcolor{CoolGray!50}
		\cellcolor{Platinum!60}\textbf{\rule{0pt}{2.5ex}Ref} &
		\textbf{\rule{0pt}{2.5ex}Model~$\uparrow$} &
		\textbf{\rule{0pt}{2.5ex}Total} &
		\textbf{\rule{0pt}{2.5ex}I1} &
		\textbf{\rule{0pt}{2.5ex}I2} &
		\textbf{\rule{0pt}{2.5ex}I3} &
		\textbf{\rule{0pt}{2.5ex}I4} &
		\textbf{\rule{0pt}{2.5ex}I5} &
		\textbf{\rule{0pt}{2.5ex}I6} &
		\textbf{\rule{0pt}{2.5ex}I7} &
		\textbf{\rule{0pt}{2.5ex}I8} &
		\textbf{\rule{0pt}{2.5ex}I9} &
		\textbf{\rule{0pt}{2.5ex}I10} &
		\textbf{\rule{0pt}{2.5ex}I11} \\
		
		\cellcolor{CoolGray!50}\cite{sunCollabVLASelfReflectiveVisionLanguageAction2025} & Collab VLA & 0.11 & 0.28 & 0.00 & 0.20 & 0.07 & 0.08 & 0.11 & 0.00 & 0.07 & \cellcolor{MintGreen!60}0.21 & 0.08 & 0.09\\
		\cellcolor{CoolGray!50}\cite{jonesSightFinetuningGeneralist2025} & FuSe & 0.05 & 0.06 & 0.00 & 0.00 & 0.00 & 0.00 & 0.00 & 0.00 & 0.00 & \cellcolor{MintGreen!60}0.21 & 0.00 & 0.27\\
		\cellcolor{CoolGray!50}\cite{teamGeminiRobotics152025} & Gemini Robotics 1.5 & 0.12 & \cellcolor{MintGreen!60}0.44 & 0.06 & 0.07 & 0.07 & 0.00 & 0.00 & 0.00 & 0.07 & 0.14 & 0.23 & 0.09\\
		\cellcolor{CoolGray!50}\cite{huangGraphCoTVLA3DSpatialAware2025} & GraphCoT VLA & 0.05 & 0.06 & 0.00 & 0.00 & 0.00 & 0.00 & 0.00 & 0.00 & \cellcolor{MintGreen!60}0.21 & 0.14 & 0.00 & 0.09\\
		\cellcolor{CoolGray!50}\cite{liHAMSTERHierarchicalAction2025} & HAMSTER & 0.07 & 0.11 & 0.00 & 0.00 & 0.00 & 0.00 & 0.00 & 0.10 & 0.00 & 0.07 & 0.23 & \cellcolor{MintGreen!60}0.36\\
		\cellcolor{CoolGray!50}\cite{wuHiBerNACHierarchicalBrainemulated2025} & HiBer NAC & 0.10 & 0.22 & 0.11 & 0.00 & 0.00 & 0.08 & 0.11 & 0.10 & 0.00 & \cellcolor{MintGreen!60}0.21 & 0.15 & 0.09\\
		\cellcolor{CoolGray!50}\cite{lynchLanguageConditionedImitation2021} & LCIL & 0.07 & 0.11 & 0.00 & \cellcolor{MintGreen!60}0.27 & 0.00 & 0.00 & 0.11 & 0.00 & 0.00 & 0.00 & 0.08 & 0.18\\
		\cellcolor{CoolGray!50}\cite{chenNanoVLARoutingDecoupled2025} & NanoVLA & 0.07 & 0.11 & 0.00 & 0.00 & 0.07 & 0.08 & 0.00 & \cellcolor{MintGreen!60}0.20 & 0.00 & 0.00 & 0.31 & 0.00\\
		\cellcolor{CoolGray!50}\cite{kimOpenVLAOpenSourceVisionLanguageAction2024} & OpenVLA & 0.07 & 0.22 & 0.00 & 0.00 & 0.07 & 0.00 & 0.00 & 0.00 & 0.00 & 0.00 & \cellcolor{MintGreen!60}0.38 & 0.09\\
		\cellcolor{CoolGray!50}\cite{driessPaLMEEmbodiedMultimodal2023} & PaLM-E & 0.09 & 0.33 & 0.00 & 0.00 & \cellcolor{MintGreen!60}0.29 & 0.00 & 0.00 & 0.00 & 0.00 & 0.07 & 0.00 & 0.18\\
		\cellcolor{CoolGray!50}\cite{liSelfCorrectingVisionLanguageActionModel2025} & SC-VLA & 0.11 & 0.33 & 0.06 & 0.00 & 0.21 & \cellcolor{MintGreen!60}0.23 & 0.11 & 0.00 & 0.00 & 0.00 & 0.15 & 0.09\\
		\cellcolor{CoolGray!50}\cite{liuMultiTaskInteractiveRobot2024} & Sirius-Fleet & 0.10 & 0.39 & \cellcolor{MintGreen!60}0.22 & 0.00 & 0.14 & 0.00 & 0.00 & 0.00 & 0.00 & 0.00 & 0.08 & 0.09\\
		\cellcolor{CoolGray!50}\cite{shukorSmolVLAVisionLanguageActionModel2025} & Smol VLA & 0.04 & 0.06 & 0.00 & 0.00 & 0.07 & 0.00 & 0.00 & \cellcolor{MintGreen!60}0.20 & 0.00 & 0.00 & 0.15 & 0.00\\
		\cellcolor{CoolGray!50}\cite{buUniVLALearningAct2025} & UniVLA & 0.07 & 0.11 & 0.00 & 0.00 & 0.00 & 0.08 & 0.00 & 0.00 & 0.07 & 0.00 & 0.23 & \cellcolor{MintGreen!60}0.36\\
		\cellcolor{CoolGray!50}-- & \textit{Multiple} & -- & -- & -- & -- & -- & -- & \cellcolor{MintGreen!60}0.11 & -- & -- & -- & -- & --\\
	\end{tabular}
\end{table*}
\\
A commonality among the five best models is that they do not merely present a monolithic visuomotor policy, but explicitly introduce and focus on single mechanisms that address failure modes and operational bottlenecks frequent to industrial settings: structured reasoning, human-in-the-loop guidance, self-correction, and improved perception/data utilisation.
Accordingly, their maturity profiles exhibit pronounced peaks on a small number of implications while remaining low elsewhere (cf. Table~\ref{tab:rfms-highest} and  Figure~\ref{fig:spc_best5}).
This implies that recent \gls{rfm} work seems to advance one or two enabling dimensions at a time, but do not yet integrate the broader industrial requirements captured by our catalogue.
\\
Specifically, Gemini~Robotics~1.5~\cite{teamGeminiRobotics152025} achieves the highest total score ($0.12$) and the strongest value on adaptability/flexibility (I$1$: $0.44$).
A plausible reason for this scoring is that it combines two complementary components (Gemin~Robotics~1.5 as a \gls{vla}, and Gemini~Robotics~1.5-ER as a \gls{vlm}), which can be combined to an agentic framework, rather than relying on a single controller in isolation.
From an industrial perspective, this implies that near-future deployments may demand layered systems where high-level reasoning, verification, and interaction are distinct from low-level action generation.
As Gemini’s leading score is not driven by uniformly higher values across all implications, it suggests that hybrid or multi-component designs may only be one part of the puzzle towards industrial maturity, primarily aiding flexibility.\\
In general, we noticed a strong consistency in positive evaluator agreement across the $5$ highest ranked \glspl{rfm}:
CollabVLA~\cite{sunCollabVLASelfReflectiveVisionLanguageAction2025} attains the highest scores within this group for \gls{hri}/\gls{hrc} (I$3$: $0.20$) and sensor fusion/perception (I$9$: $0.21$).
Those implications highly correlate, since perception of human/ operator detection is a prerequisite for \gls{hrc} (cf. Section~\ref{ssec:ii-gi}).
Importantly, this concentration directly matches the paper’s explicit objective of moving from a closed-loop visuomotor policy towards a collaborative agent by incorporating human guidance into action generation~\cite{sunCollabVLASelfReflectiveVisionLanguageAction2025}.
An equally strong agreement is obvious in data requirements and usage (I$11$: $0.27$) for \mbox{GF-VLA}~\cite{liInformationTheoreticGraphFusion2025}.
The authors focus on learning from human demonstrations aiming to improve generalisation beyond low-level trajectory imitation, framing the efficient utilisation of demonstration data as their main goal~\cite{liInformationTheoreticGraphFusion2025}.
Similar to Gemini, OneTwoVLA~\cite{linOneTwoVLAUnifiedVisionLanguageAction2025} shows high flexibility  (I$1$: $0.33$) and also employs more than one component, however, explicitly distinguishes between a fast acting mode (System One) and a slower reasoning mode (System Two) and switches between them during execution~\cite{linOneTwoVLAUnifiedVisionLanguageAction2025}.
\mbox{SC-VLA}~\cite{liSelfCorrectingVisionLanguageActionModel2025}'s high values on robustness/reliability (I$4$: $0.21$) and precision/accuracy (I$5$: $0.23$) are equally well explained by the model's main intention: 
Li et al.~\cite{liSelfCorrectingVisionLanguageActionModel2025} integrate a fast action prediction with a slower reflection mechanism to correct failed actions for more robust manipulation.
Such failure recovery could naturally translate to a more robust system. \\
Taken together, Table~\ref{tab:rfms-highest} and Table~\ref{tab:rfms-hpi} illustrate a consistent trend: the strongest models predominantly optimise for one or two identified bottlenecks, and this intent is reflected by implication-specific peaks in the evaluation.
However, even the best per-implication results remain limited in absolute terms: the highest proportional fulfilment observed across all implications is $0.44$ by Gemini~Robotics~1.5 on I$1$, and this is the only value over all evaluated \glspl{rfm} to exceed $0.4$ at all (cf. Table~\ref{tab:rfms-all}).
This indicates that, while individual industrial concerns are beginning to be addressed, current \glspl{rfm} do not yet attempt -- nor achieve -- integrated coverage across the full set of industrial implications.
In the following subsection, we therefore move beyond the top-ranked models to provide a broader view of how industrial maturity is distributed across the full \gls{rfm} landscape, and which implications remain systematically underrepresented.
\subsection{Industrial Applicability Overview}\label{ssec:survey-overview}
\begin{figure}[hbt]
	\centering
	\includegraphics[width=0.5\textwidth, trim=75pt 10pt 40pt 10pt, clip]{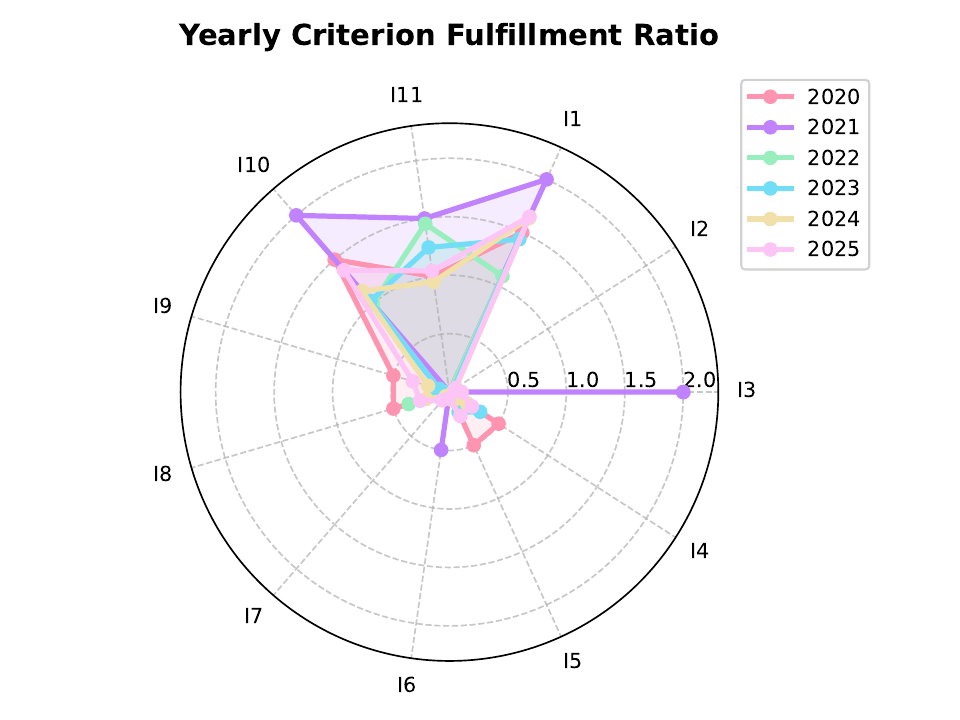}  % plotted with 0.15 opacity
	\caption{Normalised per-implication criteria coverage.
		\textit{Each value represents the sum over all criteria fulfilled within the given implication} \& \textit{year, divided by the total number of papers published in the respective year.}}
	\label{fig:yearly_impl_fulf}
\end{figure}

To provide a corpus-level perspective beyond the highest-ranked models (Section~\ref{ssec:survey-highest}), we analysed the distribution of implication-wise fulfilment across all $324$ evaluated \glspl{rfm} (cf. Table~\ref{tab:rfms-all}) and the stronger Top-50 model subset, leveraging the total and per-implication industrial maturity scores.
We report two complementary views: (a) a brief temporal stability overview of which implications are addressed (Figure~\ref{fig:yearly_impl_fulf}), and (b) breadth-versus-depth of coverage per implication for the full corpus and the Top-50 (Figure~\ref{fig:1p-acc}, with exact values in Table~\ref{tab:eval-metrics}).
Finally, we contextualise these trends by inspecting which individual criteria are most frequently fulfilled (Table~\ref{tab:crit-imp-fulfilled}) and by quantifying whether criteria are widely discussed yet rarely met (Table~\ref{tab:gap-top3}).
\runinheading{Temporal Stability Overview}
Figure~\ref{fig:yearly_impl_fulf} shows the normalised, per-implication criteria coverage aggregated by publication year.
Across the years depicted, the overall profile is highly consistent: the same small subset of implications dominates the addressed space, while the remaining implications remain persistently low.
In particular, the corpus repeatedly concentrates on adaptability/flexibility (I1), benchmarking/evaluation (I10), and data requirements/usage (I11), whereas safety/compliance (I2), \gls{hri}/\gls{hrc} (I3), real-time performance (I6), and cost-effectiveness/integration (I7) remain comparatively underrepresented.
The absence of a clear, monotonic rise in any single implication suggests that current \gls{rfm} research is not (yet) converging towards progressively more holistic industrial coverage at the implication level -- rather, the field appears to deepen within a relatively stable set of research emphases.
\runinheading{Breadth vs. Depth Analysis}
A central question to our industrial applicability analysis is, whether the literature does not mention an implication at all, merely mentions it, or provides substantive, multi-criterion coverage.
We therefore contrast two metrics per implication: (a) $1_+$-Coverage, the share of papers meeting at least one criterion, representing \emph{breadth} of implication considerations, and (b) \gls{acc}, the average fraction of criteria met per paper, implying consideration \emph{depth}.\\
The combined view of Figure~\ref{fig:1p-acc} and Table~\ref{tab:eval-metrics}, reveals a consistent breadth-depth mismatch across the current \gls{rfm} landscape:
\begin{figure*}[t]
	\centering
	\begin{subfigure}[t]{0.49\textwidth}
		\centering
		\includegraphics[width=\linewidth, trim=90pt 15pt 55pt 15pt, clip]{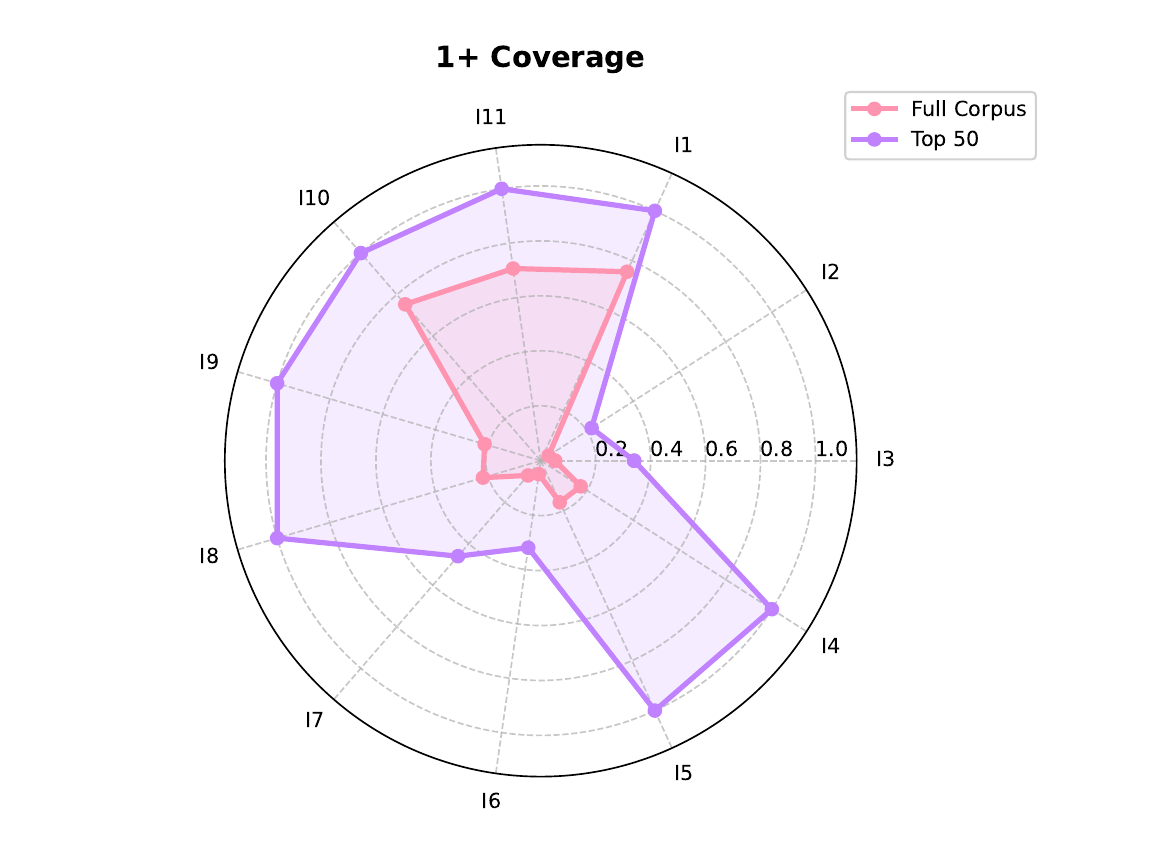}
		\caption{$1_+$-Coverage over evaluated RFMs: 
			\textit{The metric specifies the percentage of works that meet at least one criterion for the respective implication (I) within the given (sub)corpus.}}
		\label{fig:1p-coverage}
	\end{subfigure}\hfill
	\begin{subfigure}[t]{0.49\textwidth}
		\centering
		\includegraphics[width=\linewidth, trim=70pt 10pt 40pt 10pt, clip]{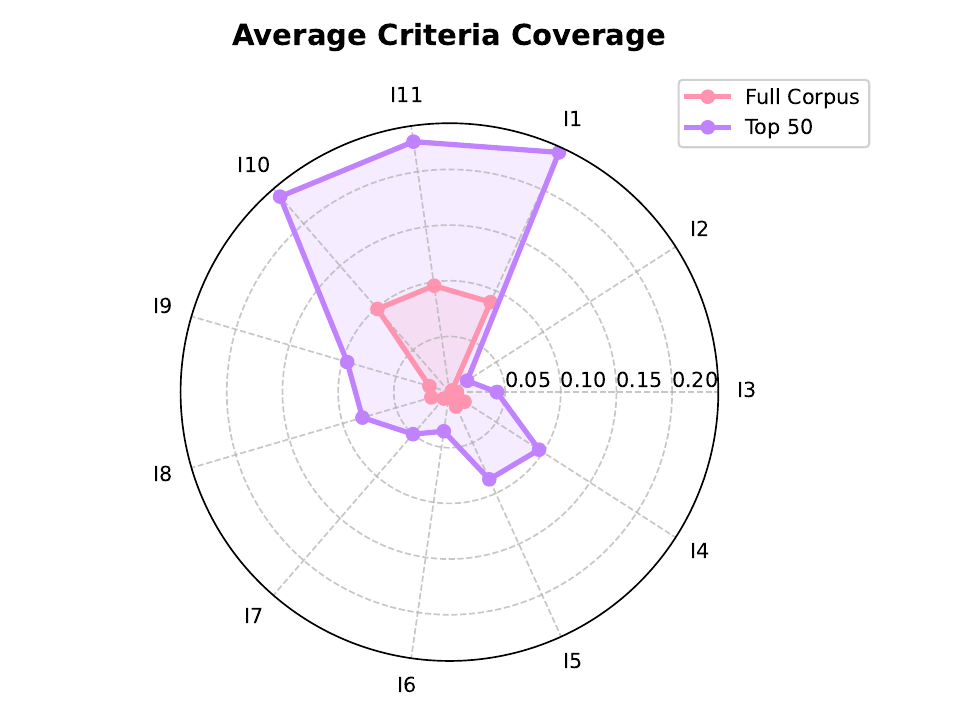}
		\caption{\gls{acc} for evaluated RFMs: 
			\textit{\gls{acc} specifies the average percentage of criteria each paper meets for the respective implication (I) within the given (sub)corpus.}}
		\label{fig:rfm-acc}
	\end{subfigure}
	\caption{Implication-specific evaluation coverage across the full set of 324 RFMs (Full Corpus) and the Top-50 subset.}
	\label{fig:1p-acc}
\end{figure*}
\begin{table*}[t]
	\centering
	\renewcommand{\arraystretch}{1.35}
	\caption{Detailed comparison of $1_+$-Coverage and \gls{acc} between the full evaluated corpus ($324$) and the Top 50:
		\textit{For visualisation, see Figure~\ref{fig:1p-acc}. Values are rounded to three decimal places.}}
	\label{tab:eval-metrics}
	\small
	\begin{tabular}{c c *{11}{c}}
		\rowcolor{CoolGray!50}
		\cellcolor{CoolGray!50}Metric & Corpus & I1 & I2 & I3 & I4 & I5 & I6 & I7 & I8 & I9 & I10 & I11 \\
		\multirow{2}{*}{1+}  & \cellcolor{Platinum!60}Full & 0.756 & 0.034 & 0.052 & 0.172 & 0.167 & 0.049 & 0.071 & 0.219 & 0.213 & 0.753 & 0.707 \\
		& \cellcolor{CoolGray!50}Top 50 & 1.0 & 0.22 & 0.34 & 1.0 & 1.0 & 0.32 & 0.46 & 1.0 & 1.0 & 1.0 & 1.0 \\
		\multirow{2}{*}{ACC} & \cellcolor{Platinum!60}Full & 0.089 & 0.003 & 0.007 & 0.016 & 0.014 & 0.005 & 0.008 & 0.017 & 0.019 & 0.099 & 0.097 \\
		& \cellcolor{CoolGray!50}Top 50 & 0.238 & 0.019 & 0.043 & 0.096 & 0.086 & 0.036 &  0.05 & 0.081 & 0.097 & 0.232 & 0.227 \\
	\end{tabular}
\end{table*}
While many papers touch a subset of implications, far fewer demonstrate multi-criterion, implementation-relevant coverage within those implications (note scale differences of Figure~\ref{fig:1p-coverage} and Figure~\ref{fig:rfm-acc}).
Concretely, the \textbf{full corpus} exhibits strong breadth for only three implications: Adaptability/flexibility (I$1$: ($1_+$=$0.756$)), benchmarking/evaluation (I$10$: ($1_+$=$0.753$)), and data requirements/usage (I$11$: ($1_+$=$0.707$)).
Yet even for these leading implications, depth remains low: their \gls{acc} values stay below ($0.10$) (I$10$: ($0.099$), I$11$: ($0.097$), I$1$: ($0.089$)).
This pattern indicates that on average, papers satisfy only a small fraction of the criteria within the implications they most frequently address. 
Additionally, several implications central to industrial deployment remain already weak at the $1_+$-Criterion level -- most notably safety/compliance (I$2$: ($1_+$=$0.034$)), \gls{hri}/\gls{hrc} (I$3$: ($1_+$=$0.052$)), real-time performance (I$6$: ($1_+$=$0.049$)), and cost-effectiveness/integration (I$7$: ($1_+$=$0.071$)).
The persistently low coverage for these implications aligns with the temporal stability observation in Figure~\ref{fig:yearly_impl_fulf}: the research field’s emphasis appears comparatively stable over time and does not show a clear shift towards holistic industrial requirement satisfaction at the implication level. \\
Restricting attention to the \textbf{Top-50} models strengthens the same overall picture rather than changing it.
Breadth becomes near-universal for seven implications (I$1$, I$4$, I$5$, I$8$, I$9$, I$10$, I$11$ all at ($1_+$=$1.0$)), suggesting that higher-performing \glspl{rfm} increasingly include at least some features, experiments, or claims related to these dimensions.
However, the remaining implications still lag even in the Top-50 subset (I$2$: ($0.22$), I$3$: ($0.34$), I$6$: ($0.32$), I$7$: (0$.46$)).
Depth improves substantially but remains far from comprehensive: the highest Top-$50$ \gls{acc} values are again concentrated in I$1$ ($0.238$), I$10$ ($0.232$), and I$11$ ($0.227$), whereas safety/compliance (I$2$: $0.019$) and real-time performance (I$6$: $0.036$) remain particularly low.
Thus, even among the strongest papers, industrial maturity is characterised more by selective advances than by broad, high-coverage satisfaction across implications.
\begin{table*}[t]
	\centering
	\renewcommand{\arraystretch}{1.35}
	\small
	\caption{Top 10 most frequently fulfilled \textbf{criteria} across all papers. \textit{Reporting each criterion’s implication (I1-I11), absolute count and rate of fulfillment, explicitness among fulfilled cases (Expl(fulf)), evidence availability (Evid), and mean confidence when fulfilled (ConfFul). Values are rounded to four decimal places.}}
	\label{tab:crit-imp-fulfilled}
	
	\begin{tabular}{@{}
			>{\raggedright\arraybackslash}m{6cm}
			>{\centering\arraybackslash}m{0.8cm}
			>{\centering\arraybackslash}m{1.2cm} 
			>{\centering\arraybackslash}m{1cm}
			>{\centering\arraybackslash}m{1.4cm}
			>{\centering\arraybackslash}m{0.9cm}
			>{\centering\arraybackslash}m{1.2cm}
			@{}}
		
		\rowcolor{CoolGray!50}
		\cellcolor{Platinum!60}\textbf{\rule{0pt}{2.5ex}Criterion} &
		\textbf{\rule{0pt}{2.5ex}Impl} &
		\textbf{\rule{0pt}{2.5ex}Fulfilled} &
		\textbf{\rule{0pt}{2.5ex}Rate}~$\downarrow$&
		\textbf{\rule{0pt}{2.5ex}Expl(fulf)} &
		\textbf{\rule{0pt}{2.5ex}Evid} &
		\textbf{\rule{0pt}{2.5ex}ConfFul} \\
		\cellcolor{CoolGray!50}Multi-Task Generalisation & I1 & 160 & 0.4938 & 0.2812 & 0.9938 & 0.8605 \\
		\cellcolor{CoolGray!50}Flexible Data Type Acceptance & I11 & 158 & 0.4877 & 0.3671 & 0.9907 & 0.8693 \\
		\cellcolor{CoolGray!50}Repeatability \& Reproducibility & I10 & 145 & 0.4475 & 0.2483 & 0.9815 & 0.8474 \\
		\cellcolor{CoolGray!50}Low-Data Adaptation \& Few-Shot Requirements & I11 & 136 & 0.4198 & 0.8088 & 0.9907 & 0.9102 \\
		\cellcolor{CoolGray!50}Module \& Skill-Specific Benchmarking & I10 & 129 & 0.3981 & 0.0698 & 0.9938 & 0.8517 \\
		\cellcolor{CoolGray!50}Generalisation to Environmental Variation & I1 & 101 & 0.3117 & 0.5446 & 0.9815 & 0.8739 \\
		\cellcolor{CoolGray!50}Generalisation \& Adaptation Assessment & I10 & 90 & 0.2778 & 0.1111 & 1.0000 & 0.8676 \\
		\cellcolor{CoolGray!50}Autonomous Handling of Distractors/Obstacles & I1 & 87 & 0.2685 & 0.5057 & 0.9568 & 0.8655 \\
		\cellcolor{CoolGray!50}Performance Benchmarking \& OOD Feedback & I8 & 63 & 0.1944 & 0.2063 & 0.9877 & 0.8367 \\
		\cellcolor{CoolGray!50}Online/Mid-Execution Adaptation & I1 & 49 & 0.1512 & 0.5306 & 0.9907 & 0.8771 \\
	\end{tabular}
\end{table*}
\begin{table*}[t]
	\centering
	\renewcommand{\arraystretch}{1.35}
	\small
	\caption{Evidence-Fulfilled-Gap analysis: 
		\textit{Reporting the gap score $g_c$ alongside its evidence availability (EvidRate) and full-corpus fulfillment rate (FulfillRate) for each implication's criterion with the highest evidence rate. Values are rounded to four decimal places.}}
	\label{tab:gap-top3}
	
	\begin{tabular}{@{}
			>{\centering\arraybackslash}m{1.1cm}
			>{\raggedright\arraybackslash}m{8.0cm}
			>{\centering\arraybackslash}m{1.2cm}
			>{\centering\arraybackslash}m{1.6cm}
			>{\centering\arraybackslash}m{1.8cm}
			@{}}
		\rowcolor{CoolGray!50}
		\cellcolor{Platinum!60}\textbf{\rule{0pt}{2.5ex}Impl~$\uparrow$} &
		\textbf{\rule{0pt}{2.5ex}Criterion} &
		\textbf{\rule{0pt}{2.5ex}$g_c$} &
		\textbf{\rule{0pt}{2.5ex}EvidRate~$\downarrow$} &
		\textbf{\rule{0pt}{2.5ex}FulfillRate} \\
		\cellcolor{CoolGray!50}I1 & Multi-Task Generalisation & 0.5000 & 0.9938 & 0.4938 \\
		\cellcolor{CoolGray!50}I2 & Reactive, Real-Time Safety & 0.8086 & 0.8117 & 0.0031 \\
		\cellcolor{CoolGray!50}I3 & Collaborative Adaptation & 0.8735 & 0.8981 & 0.0247 \\
		\cellcolor{CoolGray!50}I4 & Environmental Robustness \& Adaptation & 0.9475 & 0.9907 & 0.0432 \\
		\cellcolor{CoolGray!50}I5 & Consistency Over Time \& Contexts & 0.8796 & 0.9969 & 0.1173 \\
		\cellcolor{CoolGray!50}I6 & Goal-Responsive Solution Strategies & 0.9352 & 0.9722 & 0.0370 \\
		\cellcolor{CoolGray!50}I7 & Flexible Input/Output Adaptation & 0.9815 & 0.9815 & 0.0000 \\
		\cellcolor{CoolGray!50}I8 & Performance Benchmarking \& OOD Feedback & 0.7932 & 0.9877 & 0.1944 \\
		\cellcolor{CoolGray!50}I9 & Multimodal Sensor Acceptance \& Fusion & 0.9660 & 1.0000 & 0.0340 \\
		\cellcolor{CoolGray!50}I10 & Generalisation \& Adaptation Assessment & 0.7222 & 1.0000 & 0.2778 \\
		\cellcolor{CoolGray!50}I11 & Low-Data Adaptation \& Few-Shot Requirements & 0.5710 & 0.9907 & 0.4198 \\
	\end{tabular}
\end{table*}
\\
Table~\ref{tab:crit-imp-fulfilled} sharpens this conclusion by showing which capabilities are most frequently met.
The ten most commonly fulfilled criteria cluster heavily in the same three implications that dominate in depth (I$1$, I$10$, I$11$).
The top entries -- Multi-Task Generalisation (I$1$: $0.4938$), Flexible Data Type Acceptance (I$11$: $0.4877$), Repeatability \& Reproducibility (I$10$: $0.4475$), and Low-Data Adaptation \& Few-Shot Requirements (I$11$: $0.4198$) -- indicate that the present \gls{rfm} literature prioritises generality across tasks, flexible use of available (and heterogeneous) data, and benchmark-driven reporting.
At the same time, the explicitness values for several of these frequently fulfilled criteria are relatively low: e.g. Multi-Task Generalisation~($0.2812$) and Repeatability \& Reproducibility~($0.2483$), suggesting that fulfilment is often inferred from experimental set-ups or evaluation sections rather than being framed as a directly stated requirement.
The comparatively high explicitness for Low-Data Adaptation~($0.8088$) is consistent with this topic being frequently highlighted explicitly as a contribution, though -- as discussed in Section~\ref{ssec:survey-proc} -- the interpretation latitude of this criterion may also inflate fulfilment rates.
In general, however, the most fulfilled criteria support what we would expect of \glspl{rfm}: systems adaptive to multiple tasks that can digest various forms of data for behaviour modulation.
\\
Finally, Table~\ref{tab:gap-top3} indicates that many criteria are discussed in the current literature, however, most seem to be not, or only partially fulfilled. 
The gap~score~$g_c$ ($\mathrm{EvidRate}_c-\mathrm{FulfillRate}_c$) is large for the top-evidence criterion in almost every implication, frequently exceeding ($0.8$).
As EvidRate reflects the rate of evidence found in the given texts, these large gaps imply that many criteria-relevant aspects are widely mentioned or at least documentable, yet rarely satisfied under the (binary, conservative) fulfilment policy of our evaluation.
In combination with the low \gls{acc} values in Table~\ref{tab:eval-metrics}, this points to a broader tendency: the corpus increasingly acknowledges industrial constraints and desiderata, but typical papers still implement or document only limited, criterion-incomplete solutions -- particularly for implications tied to deployment integration~(I$7$), safety/ compliance~(I$2$), real-time performance~(I$6$), and \gls{hrc}/\gls{hri}~(I$3$).
\section{Summary \& Discussion}\label{sec:summary}
Industrial robotics is undergoing a discernible transition from isolated, highly optimised automation towards flexible, collaborative deployment paradigms in which humans and robots share workspaces and responsibilities.
A driver of this shift is the uptake of \glspl{cobot}, particularly in high-mix, low-volume settings and in \glspl{sme}, where traditional automation is often economically or organisationally impractical due to commissioning effort, frequent production changes, and the need for specialised expertise (cf. Section~\ref{sec:towards_cobots}).
While modern \glspl{cobot} provide the hardware foundations for such a transition -- safe physical design, appropriate payload and sizing, comparatively low operating power -- they do not, by themselves, resolve the central bottleneck: robust, deployment-ready intelligence that can flexibly adapt behaviour to changing tasks, environments, and multimodal input without imposing prohibitive engineering overhead. \\
Our synthesis therefore frames \glspl{rfm} as a promising enabling technology for this industrial transformation, because they target precisely the dimensions in which classical control and even specialised learning-based approaches remain limited: multi-task competence, cross-domain generalisation, and instruction- or demonstration-driven/ low-effort reconfiguration (cf. Section~\ref{sec:rfms}).
However, the same factors that make industrial \gls{cobot} deployments attractive -- shared workspaces, continuous operation, heterogeneous actuation/ sensing stacks and cost pressure -- also amplify the demands on any learning-based control paradigm.
In particular, typical \glspl{cobot} exhibit high mechanical precision and repeatability, moderate payload capability, and low median power draw in the widely deployed $0.5$-$20$kg class (cf. Section~\ref{ssec:cch-cp}).
These characteristics imply that industrial value is frequently limited not by actuation quality but by perception and decision-making under real-world variability, and they simultaneously constrain \gls{rfm} deployment to low-latency, energy-aware inference on industrial edge platforms rather than workstation-grade hardware (cf. Section~\ref{ssec:cch-ipc}).
The practical prevalence of multi-sensor configurations, sensor fusion, and modular end-effectors further highlights that industrial intelligence must be interface-flexible: it must tolerate changing observation and action spaces that arise from tool changes, sensor substitutions, and integration into existing automation infrastructure (cf. Section~\ref{ssec:cch-shw}).
To capture these deployment realities systematically, we distilled eleven general interdependent implications (Section~\ref{ssec:ii-gi}) from industrial literature and translated them into a structured assessment framework (Section~\ref{ssec:ii-cc}). \\
The key contribution of this step is not merely a list of desirable properties, but a traceable operationalisation: 
As foundation for our assessment framework, we started with $11$ implications of what industry generally demands from robotic applications, extracted $47$ deployment-relevant attributes to create an implication-attribute matrix of $11\times47=517$ considerations, to finally retrieve an implication-grounded, assessable criteria catalogue of $149$ concrete items spanning both model capabilities and ecosystem requirements (cf. Section~\ref{sec:industrial-implications}).
Our observations clearly show that raw success rates alone are not sufficient for industrial application.
Rather, capability is inseparable from deployability: 
safety depends on perception and verification, real-time feasibility depends on compute constraints and system integration, and trust depends on observable and auditable behaviour rather than raw benchmark success.
When applying this framework to the manipulation-capable \gls{rfm} landscape, our evaluation results paint a consistent picture: the field is progressing rapidly, e.g. in research breadth, yet remains at an early stage of industrial maturity (cf. Section~\ref{sec:survey}).
Across the $324$ discovered and evaluated models and $48,276$ criterion-level decisions, even the top-ranked \glspl{rfm} reach only $0.11$-$0.12$ overall maturity -- roughly one out of ten criteria met on average (cf. Section~\ref{ssec:survey-highest}) and most models only score (comparably) high on a maximum of two to three varying implications.
This implies that the primary limiting factor for industrial deployment is not the absence of isolated technical ideas, but the incomplete and non-systematic integration of those ideas into comprehensive, deployable \gls{rfm} stacks.
The frontier models further indicate how progress is currently realised.
The five strongest systems exhibit implication-wise "peaks" rather than balanced coverage, typically by introducing targeted mechanisms such as structured reasoning, human-in-the-loop interaction, self-correction, or improved data utilisation (cf. Section~\ref{ssec:survey-highest}).
Gemini~Robotics~$1.5$, for example, leads overall and reaches the highest observed implication fulfilment on adaptability/flexibility (I$1$: $0.44$), yet remains near zero on multiple deployment-critical implications, underscoring that hybrid or layered architectures may advance flexibility without automatically closing safety, integration, or real-time gaps.
Similarly, models that score comparatively well in \gls{hri}/\gls{hrc} and perception emphasise collaborative mechanisms and richer observation handling, but do not simultaneously satisfy the broader criteria spectrum required for sustained industrial operation.
Taken together, the current \gls{rfm} state of the art appears to advance by isolating and optimising individual bottlenecks, yet comprehensive "industry-grade" integration, by contrast, remains largely unaddressed. 
This interpretation is reinforced when expanding evaluation to the full corpus. 
Over time, the implication profile remains strikingly stable: research consistently concentrates on adaptability/flexibility (I$1$), benchmarking/evaluation (I$10$), and data requirements/usage (I$11$), while safety/compliance (I$2$), \gls{hri}/\gls{hrc} (I$3$), real-time performance (I$6$), and cost-effectiveness/integration (I$7$) remain persistently underrepresented (cf. Section~\ref{ssec:survey-overview}).
The breadth-depth analysis  of our evaluation sharpens the same conclusion:
In the full corpus, several implications are frequently "touched", yet few papers satisfy more than a small fraction of criteria within those implications.
Even the Top-$50$ subset does not convert their breadth into comprehensive satisfaction: depth remains modest even for leading implications (I$1$/I$10$/I$11$), and many deployment-critical implications (I$2$/I$3$/I$6$/I$7$) remain relatively weak both in breadth and depth (cf. Section~\ref{ssec:survey-overview}).
Hence, current progress is better characterised as wider acknowledgement of industrial concerns than as demonstrable, multi-criterion completion of those concerns. \\
These findings suggest several concrete priorities for future work towards industry-grade \glspl{rfm}:
First, the field must move from isolated capability demonstrations towards deployable, layered systems in which planning, memory, action generation, verification, and safety gating are treated as central, independently testable components -- an architectural direction implicitly supported by high-performing models, yet not broadly matured across the corpus (cf. Section~\ref{ssec:survey-highest}).
Second, industrial adoption may require bridging the persistent underrepresentation of safety/compliance, real-time feasibility, and integration concerns by developing evaluation protocols and artefacts that connect \gls{rfm} capabilities to operational constraints: low-latency inference on edge hardware, structured logging and monitoring, and explicit handling of changing sensor and tooling interfaces (cf. Sections~\ref{ssec:cch-ipc} and~\ref{ssec:ii-cc}).
Third, the strong and stable emphasis on benchmarking (I$10$) should be leveraged to improve industrial relevance rather than to merely report success rates in a diversity of benchmarks: evaluations must reflect deployment realities such as distribution shifts, failure recovery, human supervision, and long-horizon operation, and must enable diagnosing why a model fails under industrial conditions rather than only whether it succeeds on curated tasks.
Finally, the large evidence-fulfilment gaps (cf. Section~\ref{ssec:survey-overview}) indicate an opportunity and a requirement: many industrial desiderata are already recognised by the research community, yet the step from recognition to robust implementation, documentation, and validation remains a dominant obstacle. \\
In conclusion, \glspl{rfm} constitute a credible technological pathway to unlock the full potential of \gls{cobot}-centred automation: they directly target the flexibility, reconfigurability, and intuitive instructioning demanded by modern industrial production, especially in cost- and changeover-sensitive settings (cf. Section~\ref{sec:towards_cobots}).
Nevertheless, our corpus-scale evaluation demonstrates that current \gls{rfm} research is still predominantly laboratory-oriented when measured against deployment-grounded industrial requirements.
The presented implication synthesis and criteria catalogue provide a transparent \emph{industrial readiness} reference that can be used both to evaluate future \glspl{rfm} and to guide their design towards auditable, safe, real-time-capable and integration-ready systems.
Closing the identified gaps -- particularly in safety/compliance, \gls{hri}/\gls{hrc}, real-time performance, and cost-effectiveness/integration capabilities -- therefore represents the central research agenda for translating the rapidly advancing field of \glspl{rfm} into reliable industrial autonomy.
\bibliography{full-bibliography}
\begin{appendices}
\newlength{\CCWidth}
\setlength{\CCWidth}{12.5cm}
\section{Related Works}\label{app:rl}
\onecolumn
\renewcommand{\arraystretch}{1.4}
\small

\begin{longtable}{@{}
		>{\centering\arraybackslash}m{0.9cm}
		>{\raggedright\arraybackslash}m{12.5cm}
		>{\centering\arraybackslash}m{1.6cm}
		@{}}
	\caption{Full list of identified related works/ surveys.
		\textit{Total 55 publications from survey corpus as depicted in Figure~\ref{fig:surveys_overview}.}} 
	\label{tab:app-rl}\\
	
	\rowcolor{CoolGray!50}
	\cellcolor{Platinum!60}\textbf{\rule{0pt}{2.5ex}Ref} &
	\textbf{\rule{0pt}{2.5ex}Title} &
	\textbf{\rule{0pt}{2.5ex}Year~$\downarrow$} \\
	\endfirsthead
	
	\rowcolor{CoolGray!50}
	\cellcolor{Platinum!60}\textbf{\rule{0pt}{2.5ex}Ref} &
	\textbf{\rule{0pt}{2.5ex}Title} &
	\textbf{\rule{0pt}{2.5ex}Year} \\
	\endhead
	
	\multicolumn{3}{r}{\footnotesize\textit{continued on next page}}\\
	\endfoot
	
	\endlastfoot
	\cellcolor{CoolGray!50}\cite{sunReviewEmbodiedGrasping2025} & A Review of Embodied Grasping & 2025 \\
	\cellcolor{CoolGray!50}\cite{yuanSurveyBehaviorFoundation2025} & A Survey of Behavior Foundation Model: Next-Generation Whole-Body Control System of Humanoid Robots & 2025 \\
	\cellcolor{CoolGray!50}\cite{maSurveyVisionLanguageActionModels2025} & A Survey on Vision-Language-Action Models for Embodied AI & 2025 \\
	\cellcolor{CoolGray!50}\cite{zhongSurveyVisionLanguageActionModels2025} & A Survey on Vision-Language-Action Models: An Action Tokenization Perspective & 2025 \\
	\cellcolor{CoolGray!50}\cite{gaoTaxonomyEvaluatingGeneralist2025} & A Taxonomy for Evaluating Generalist Robot Policies & 2025 \\
	\cellcolor{CoolGray!50}\cite{liuAligningCyberSpace2025} & Aligning Cyber Space with Physical World: A Comprehensive Survey on Embodied AI & 2025 \\
	\cellcolor{CoolGray!50}\cite{xuAnatomyVisionLanguageActionModels2025} & An Anatomy of Vision-Language-Action Models: From Modules to Milestones and Challenges & 2025 \\
	\cellcolor{CoolGray!50}\cite{liApplicationsLargeLanguage2025} & Applications of Large Language Models and Multimodal Large Models in Autonomous Driving: A Comprehensive Review & 2025 \\
	\cellcolor{CoolGray!50}\cite{zhouBridgingLanguageAction2025} & Bridging Language and Action: A Survey of Language-Conditioned Robot Manipulation & 2025 \\
	\cellcolor{CoolGray!50}\cite{lisondraEmbodiedAIFoundation2025} & Embodied AI with Foundation Models for Mobile Service Robots: A Systematic Review & 2025 \\
	\cellcolor{CoolGray!50}\cite{xuEmbodiedAIBridging2025} & Embodied AI: Bridging Simulation and Reality in Robotics & 2025 \\
	\cellcolor{CoolGray!50}\cite{fengEmbodiedAILLMs2025} & Embodied AI: From LLMs to World Models & 2025 \\
	\cellcolor{CoolGray!50}\cite{chenExploringEmbodiedMultimodal2025} & Exploring Embodied Multimodal Large Models: Development & 2025 \\
	\cellcolor{CoolGray!50}\cite{khanFoundationModelDriven2025} & Foundation Model Driven Robotics: A Comprehensive Review & 2025 \\
	\cellcolor{CoolGray!50}\cite{suiGroundingManipulationCase2025} & From Grounding to Manipulation: Case Studies of Foundation Model Integration in Embodied Robotic Systems & 2025 \\
	\cellcolor{CoolGray!50}\cite{zhangGenerativeArtificialIntelligence2025} & Generative Artificial Intelligence in Robotic Manipulation: A Survey & 2025 \\
	\cellcolor{CoolGray!50}\cite{liuIntegratingLargeLanguage2025} & Integrating Large Language Models into Robotic Autonomy: A Review of Motion & 2025 \\
	\cellcolor{CoolGray!50}\cite{mehtaLargeLanguageModels2025} & Large Language Models and 3D Vision for Intelligent Robotic Perception and Autonomy & 2025 \\
	\cellcolor{CoolGray!50}\cite{liangLargeModelEmpowered2025} & Large Model Empowered Embodied AI: A Survey on Decision-Making and Embodied Learning & 2025 \\
	\cellcolor{CoolGray!50}\cite{shaoLargeVLMbasedVisionLanguageAction2025} & Large VLM-based Vision-Language-Action Models for Robotic Manipulation: A Survey & 2025 \\
	\cellcolor{CoolGray!50}\cite{borazjaniMultiModalMultiTaskM3T2025} & Multi-Modal Multi-Task (M3T) Federated Foundation Models for Embodied AI: Potentials and Challenges for Edge Integration & 2025 \\
	\cellcolor{CoolGray!50}\cite{hanMultimodalFusionVisionLanguage2025} & Multimodal Fusion and Vision-Language Models: A Survey for Robot Vision & 2025 \\
	\cellcolor{CoolGray!50}\cite{liuNeuralBrainNeuroscienceinspired2025} & Neural Brain: A Neuroscience-inspired Framework for Embodied Agents & 2025 \\
	\cellcolor{CoolGray!50}\cite{sartorNeuralScalingLaws2025} & Neural Scaling Laws in Robotics & 2025 \\
	\cellcolor{CoolGray!50}\cite{cuiOpenHelixShortSurvey2025} & OpenHelix: A Short Survey & 2025 \\
	\cellcolor{CoolGray!50}\cite{xiangParallelsVLAModel2025} & Parallels Between VLA Model Post-Training and Human Motor Learning: Progress & 2025 \\
	\cellcolor{CoolGray!50}\cite{zhangPureVisionLanguage2025} & Pure Vision Language Action (VLA) Models: A Comprehensive Survey & 2025 \\
	\cellcolor{CoolGray!50}\cite{wangEmbodiedAGIReview2025} & Toward Embodied AGI: A Review of Embodied AI and the Road Ahead & 2025 \\
	\cellcolor{CoolGray!50}\cite{salimpourEmbodiedAgenticAI2025} & Towards Embodied Agentic AI: Review and Classification of LLM- and VLM-Driven Robot Autonomy and Interaction & 2025 \\
	\cellcolor{CoolGray!50}\cite{liuGeneralistRobotPolicies2025} & Towards Generalist Robot Policies: What Matters in Building Vision-Language-Action Models & 2025 \\
	\cellcolor{CoolGray!50}\cite{fanVisionlanguageModelbasedHumanrobot2025} & Vision-Language Model-Based Human-Robot Collaboration for Smart Manufacturing: A State-of-the-Art Survey & 2025 \\
	\cellcolor{CoolGray!50}\cite{sapkotaVisionLanguageActionModelsConcepts2025} & Vision-Language-Action Models: Concepts, Progress, Applications and Challenges & 2025 \\
	\cellcolor{CoolGray!50}\cite{liWhatFoundationModels2025} & What Foundation Models Can Bring for Robot Learning in Manipulation : A Survey & 2025 \\
	\cellcolor{CoolGray!50}\cite{gaoSurveyFoundationModels2024} & A Survey for Foundation Models in Autonomous Driving & 2024 \\
	\cellcolor{CoolGray!50}\cite{jeongSurveyRobotIntelligence2024} & A Survey of Robot Intelligence with Large Language Models & 2024 \\
	\cellcolor{CoolGray!50}\cite{cohenSurveyRoboticLanguage2024} & A Survey of Robotic Language Grounding: Tradeoffs between Symbols and Embeddings & 2024 \\
	\cellcolor{CoolGray!50}\cite{kimSurveyIntegrationLarge2024} & A Survey on Integration of Large Language Models with Intelligent Robots & 2024 \\
	\cellcolor{CoolGray!50}\cite{xuSurveyRoboticsFoundation2024} & A Survey on Robotics with Foundation Models: Toward Embodied AI & 2024 \\
	\cellcolor{CoolGray!50}\cite{banjanovic-mehmedovicAdvancementsRoboticIntelligence2024} & Advancements in Robotic Intelligence: The Role of Computer Vision & 2024 \\
	\cellcolor{CoolGray!50}\cite{sanghaiAdvancesTransformersRobotic2024} & Advances in Transformers for Robotic Applications: A Review & 2024 \\
	\cellcolor{CoolGray!50}\cite{linEmbodiedAILarge2024} & Embodied AI with Large Language Models: A Survey and New HRI Framework & 2024 \\
	\cellcolor{CoolGray!50}\cite{renEmbodiedIntelligenceFuture2024} & Embodied Intelligence Toward Future Smart Manufacturing in the Era of AI Foundation Model & 2024 \\
	\cellcolor{CoolGray!50}\cite{naderiFoundationModelsAutonomous2024} & Foundation Models for Autonomous Robots in Unstructured Environments & 2024 \\
	\cellcolor{CoolGray!50}\cite{moroncelliIntegratingReinforcementLearning2024} & Integrating Reinforcement Learning with Foundation Models for Autonomous Robotics: Methods and Perspectives & 2024 \\
	\cellcolor{CoolGray!50}\cite{wangLargeLanguageModels2024} & Large Language Models for Robotics: Opportunities & 2024 \\
	\cellcolor{CoolGray!50}\cite{adebayoLeveragingFoundationModels2024} & Leveraging Foundation Models in Robotics: Transforming Task Planning and Contextual Execution & 2024 \\
	\cellcolor{CoolGray!50}\cite{atuhurraLeveragingLargeLanguage2024} & Leveraging Large Language Models in Human-Robot Interaction: A Critical Analysis of Potential and Pitfalls & 2024 \\
	\cellcolor{CoolGray!50}\cite{kawaharazukaRealworldRobotApplications2024} & Real-World Robot Applications of Foundation Models: A Review & 2024 \\
	\cellcolor{CoolGray!50}\cite{pashangpourFutureIntelligentHealthcare2024} & The Future of Intelligent Healthcare: A Systematic Analysis and Discussion on the Integration and Impact of Robots Using Large Language Models for Healthcare & 2024 \\
	\cellcolor{CoolGray!50}\cite{huGeneralPurposeRobotsFoundation2024} & Toward General-Purpose Robots via Foundation Models: A Survey and Meta-Analysis & 2024 \\
	\cellcolor{CoolGray!50}\cite{mccarthyGeneralistRobotLearning2024} & Towards Generalist Robot Learning from Internet Video: A Survey & 2024 \\
	\cellcolor{CoolGray!50}\cite{googledeepmindDemonstratingLargeLanguage2023} & Demonstrating Large Language Models on Robots & 2023 \\
	\cellcolor{CoolGray!50}\cite{firooziFoundationModelsRobotics2023} & Foundation Models in Robotics: Applications & 2023 \\
	\cellcolor{CoolGray!50}\cite{zengLargeLanguageModels2023} & Large Language Models for Robotics: A Survey & 2023 \\
	\cellcolor{CoolGray!50}\cite{xiaoRobotLearningEra2023} & Robot Learning in the Era of Foundation Models: A Survey & 2023 \\
\end{longtable}
\twocolumn
\clearpage
\section{Literature Research}\label{app:lr}
\begin{table*}[t]
	\caption{All eleven ASQS Queries used for the RFM-related (main) corpus retrieval process as described in Section~\ref{ssec:methods-crfm}.
	\textit{Characters are escaped as used within the Python Script, each query is shown before being translated for the respective database.}}
	\label{tab:rfm-queries}
		\begin{minipage}{\linewidth}
			\begin{lstlisting}[style=searchquery,linewidth=\linewidth]
["(robot* OR robotic) AND (general* OR generalist OR generalisation OR generalization OR foundation OR \"multi-task\" OR multitask OR \"multi task\" OR \"pre-trained\" OR pretrained OR \"pre trained\") AND (control OR manipulation OR action OR learning) AND (transformer OR model* OR policy OR agent* OR system) AND (\"vision-based\" OR \"vision based\" OR \"vision-language\" OR \"vision language\" OR \"language-guided\" OR \"language guided\" OR \"natural language\" OR \"natural-language\" OR \"multi-modal\" OR \"multi modal\" OR multimodal)",

"(robot* OR robotic) AND \"foundation model*\"~3",

"(robot* OR robotic) AND (\"vision-language-action\" OR \"vision language action\"~3) AND model*",

"(robotic OR robot*) AND (\"foundation model\" OR \"general-purpose model\"~3 OR \"large-scale AI\"~3 OR \"pretrained AI\" OR \"generalist AI\") AND (\"multi-modal\" OR multimodal OR \"diverse datasets\" OR \"robot demonstration\" OR \"shared representation\" OR \"transfer learning\") AND (generalization OR versatility OR transferability OR adaptation OR \"broad capabilities\"~3) AND (perception OR \"decision-making\" OR actuation) NOT (\"specialized task\" OR \"specific kinematics\" OR \"highly-specialized\")",

"(robot* OR robotic OR robotics) AND (\"foundation model\" OR \"general-purpose model\" OR \"large-scale model\" OR \"generalist AI\") AND (multimodal OR \"multi-modal\" OR \"diverse datasets\" OR \"robot demonstrations\" OR \"sensor data\" OR \"visual data\") AND (control OR perception OR actuation OR \"decision-making\" OR intelligence) AND (generalization OR transfer OR \"adaptable intelligence\" OR \"transferable capabilities\") NOT (specialized OR \"specific kinematics\" OR \"specific tasks\" OR \"highly-specialized\")",

"(robot* OR robotic) AND (\"foundation model*\" OR \"pretrained model*\" OR versatile OR transferable OR generalist OR \"shared representation*\") AND (multimodal OR \"robot sensors\" OR \"sensor data\" OR (image OR text OR dataset*)) AND (perception OR intelligence OR actuation OR \"robot behavior\") AND (generalization OR adaptation OR versatility) NOT (specific OR specialized OR \"highly-specialized task\" OR kinematic*)",

"(\"robot* foundation model\"~3 OR \"robot* generalist model\"~3 OR \"general-purpose robot* model\"~3) AND ((multimodal OR \"multi-modal\" OR \"multi modal\") AND (pretrained OR \"pre-trained\" OR \"pre trained\")) AND (generalization OR generalisation OR transfer OR adaptability OR versatility OR \"broad capabilities\") AND (perception OR \"decision-making\" OR actuation OR control) NOT (\"task-specialized\" OR \"specific robot kinematics\" OR \"highly-specialized model*\")",

"(\"vision-language-action\" OR \"vision language action\"~3 OR VLA OR (vision AND language AND action)) AND ((model OR architecture OR system) AND (multimodal OR \"multi-modal\" OR \"multi modal\")) AND (robot* OR robotics) AND (manipulat* OR navigat* OR control OR action OR behavior) AND (versatile OR \"general-purpose\" OR general*) NOT (\"single-modality\" OR specialized OR narrow)",

"(\"vision-language-action\"~3 OR \"vision NEAR/3 language NEAR/3 action\" OR VLA OR (multimodal AND (vision OR perception) AND language AND action)) AND (robot* OR robotics OR robotic) AND (model OR system OR architecture) AND (control OR behavior OR movement OR manipulation OR navigation) AND (\"general-purpose\" OR versatile OR integration) NOT (\"single-modality\" OR \"perception-only\" OR \"language-only\" OR narrow OR specialized)",

"(multimodal AND (vision AND language AND action)) AND (robot* OR robotics) AND (\"control signals\" OR behaviors OR actions OR navigation OR manipulation) AND (versatile OR \"general-purpose\" OR integrated) NOT (narrow OR \"single-modality\" OR specialized)",

"(\"vision-language-action\" OR \"vision language action\"~3 OR VLA) AND (robot* OR robotics OR \"robotic system\") AND (AI OR \"artificial intelligence\" OR \"machine learning\") AND (control OR behavior OR action OR manipulation OR navigation) AND (multimodal OR \"multi-modal\" OR integrated OR \"multi modal\") AND (model OR framework) NOT (\"perception only\" OR \"single modality\" OR \"language modeling alone\" OR \"narrow task\")"]
			\end{lstlisting}
		\end{minipage}%
\end{table*}
\begin{table*}[t]
	\caption{All seven ASQS Queries used for the industrial implication corpus retrieval process as described in Section~\ref{ssec:methods-cinim}.
		\textit{Characters are escaped as used within the Python Script, each query is shown before being translated for the respective database.}}
	\label{tab:impl-queries}
	\begin{minipage}{\linewidth}
		\begin{lstlisting}[style=searchquery,linewidth=\linewidth]
["(robot OR robotics) AND (industrial OR \"industrial application\"~3 OR \"industrial use-case\"~3 OR \"applied robotics\"~3 OR \"industrial environment\"~2) AND (review OR analysis OR viewpoint OR guideline OR perspective) AND (hardware OR sensor OR \"compute resource\"~2 OR integration OR compliance OR standards OR workflow)",

"(robotics system* OR \"industrial robotics\" OR \"applied robot*\"~2) AND (design OR performance OR deployment OR scalability OR robustness OR reliability OR safety OR operator interaction) AND (industrial needs OR \"industrial requirement*\"~2 OR \"industrial-focused\" OR \"real-world application\"~3)",

"(robotics OR robot* OR \"robotic systems\") AND (industrial OR \"industry-specific\" OR \"industrial automation\") AND (design OR deployment OR \"performance requirement*\"~2) AND (\"machine learning\" OR ML OR \"artificial intelligence\" OR AI) AND (analyze OR review OR implication* OR considerations)",

"(\"industrial robots\" OR robotics OR \"robot hardware\" OR \"robotic systems\") AND (sensor* OR \"compute resources\" OR integration OR standards OR compliance OR regulatory) AND (manufacturing OR logistics OR assembly OR inspection OR \"quality control\") AND (property OR characteristic OR \"industrial requirements\")",

"(\"industrial deployment\" OR \"industrial automation\" OR \"real-world deployment\") AND (\"robotics\" OR \"robot\") AND (\"challenges\" OR scalability OR robustness OR reliability OR safety OR \"operator interaction\") AND (\"machine learning\" OR AI)",

"(robotics OR AI) AND (review OR \"case study\" OR analysis) AND (industrial OR \"manufacturing environments\" OR logistics OR \"assembly line\" OR warehouse OR inspection OR \"quality assurance\") AND (constraints OR \"sector-specific needs\" OR adaptability)",

"(industrial robotics OR \"robotic system\") AND (hardware OR \"compute resources\" OR sensor OR \"sensor types\" OR integration OR \"regulatory compliance\" OR standards OR \"performance requirements\") AND (manufacturing OR logistics OR factory OR automotive OR electronics OR warehousing) NOT (academic OR prototype)"]
		\end{lstlisting}
	\end{minipage}%
\end{table*}
\clearpage
\section{Industrial Implications}\label{app:ii}
\onecolumn

\renewcommand{\arraystretch}{1.35}
\small

\begin{longtable}{@{}
		>{\centering\arraybackslash}m{2.5cm}
		>{\raggedright\arraybackslash}m{\CCWidth}
		@{}}
	\caption{Full criteria catalogue for industrial applicability rating as discussed in Section~\ref{ssec:ii-cc}.
		\textit{Derived from the general industrial implications (see Section~\ref{ssec:ii-gi})}}
	\label{tab:app-cc}\\
	
	\rowcolor{CoolGray!50}
	\cellcolor{Platinum!60}\textbf{\rule{0pt}{2.5ex}Implication} &
	\multicolumn{1}{c}{\textbf{\rule{0pt}{2.5ex}Category/ Criterion}}\\
	\endfirsthead
	
	\rowcolor{CoolGray!50}
	\cellcolor{Platinum!60}\textbf{\rule{0pt}{2.5ex}Implication} &
	\multicolumn{1}{c}{\textbf{\rule{0pt}{2.5ex}Category/ Criterion}}\\
	\endhead
		\cellcolor{CoolGray!50}\textbf{I1:} &\multicolumn{1}{>{\centering\arraybackslash\columncolor{Platinum!60}}m{\CCWidth}}
		{A. Multi-Hardware and Configuration Adaptability}\\
		\cellcolor{CoolGray!50}\textbf{Adaptability \& Flexibility}&1. \textit{Multi-Hardware Support:} Operates with different kinematics, sensors, and end-effectors without retraining.\\
		\cellcolor{CoolGray!50}&2. \textit{Hot-Swappable} \& \textit{Dimensionality Adaptation:} Sensors/actuators can be replaced, added, or removed during operation. Automatically detects and configures input/output channels at runtime according to connected hardware and environmental context: Flexibly adapts the dimensionality, type and interpretation without manual intervention or system restart.\\
		\cellcolor{CoolGray!50}&\multicolumn{1}{>{\centering\arraybackslash\columncolor{Platinum!60}}m{\CCWidth}}
		{B. Task Adaptability and Strategy Flexibility}\\
		\cellcolor{CoolGray!50}&3. \textit{Multi-Task Generalisation:} Handles multiple task types/categories, including unseen variations thereof, without retraining. Adapts strategies dynamically for task variations and can switch between tasks online.\\
		\cellcolor{CoolGray!50}&4. \textit{Restricted Autonomous Task Switching:} Only switches between distinct tasks upon explicit user instruction: autonomy remains bounded, so adherence to the operator’s goal stays primary (may autonomously adapt solution approaches/ strategies).\\
		\cellcolor{CoolGray!50}&5. \textit{Strategy-Goal Alignment:} Any adaptation of solution strategies ensures adherence to the commanded task goal: deviating too far from user instructions triggers feedback, halt, or request for approval.\\
		\cellcolor{CoolGray!50}&\multicolumn{1}{>{\centering\arraybackslash\columncolor{Platinum!60}}m{\CCWidth}}{C. Environmental Adaptation and Robustness}\\
		\cellcolor{CoolGray!50}&6. \textit{Generalisation to Environmental Variation:} Maintains consistent performance across diverse environmental conditions (lighting, weather, background, friction, etc.), object/material variations, and scene clutter, with minimal degradation.\\
		\cellcolor{CoolGray!50}&7. \textit{Autonomous Handling of Distractors/Obstacles:} Identifies and robustly manages/ handles unknown, distracting, or non-required objects in the scene or obstacles along the path/ trajectory (including incorporation into task strategy if needed).\\
		\cellcolor{CoolGray!50}&8. \textit{Online Environment Adaptation} \& \textit{Response:} Recognises and continuously adapts to environmental changes, such as errors, faults, sensor degradation or external influences during operation: e.g. logging, halting, or feedback-requesting as confidence drops.\\
		\cellcolor{CoolGray!50}&9. \textit{Autonomous Fallbacks:} Capable of autonomously choosing context-appropriate error/fault responses (e.g. request feedback, switch input modalities, logging, etc.) and behaviour (e.g. continue operation, transition to safe state, adapt strategy, etc.) without requiring predefined fallback behaviour.\\
		\cellcolor{CoolGray!50}&\multicolumn{1}{>{\centering\arraybackslash\columncolor{Platinum!60}}m{\CCWidth}}{D. Instruction and Feedback Flexibility}\\
		\cellcolor{CoolGray!50}&10. \textit{Multimodal Command Interface:} Accepts user-input (task instructions, correction/feedback, etc.) via diverse modalities (natural language in multiple languages, images, teaching by demonstration, video, etc.).\\
		\cellcolor{CoolGray!50}&11. \textit{Mid-Execution Feedback Responsiveness:} Remains receptive to user feedback and correction throughout execution, enabling course correction and adjustment without re-initiating trajectories.\\
		\cellcolor{CoolGray!50}&\multicolumn{1}{>{\centering\arraybackslash\columncolor{Platinum!60}}m{\CCWidth}}{E. Self-Flexibility and Adaptive Behaviour}\\
		\cellcolor{CoolGray!50}&12.	\textit{Operational Adaptation Confidence:} Continuously outputs a confidence score reflecting its present predicted ability to execute the instructed task using the currently (adapted) solution strategy, in the current environment, with the currently active hardware configuration and sensor context. The score may be used to trigger automatic logging, operator feedback requests, or safe halting if confidence drops below a set threshold, ensuring transparent awareness and operational safety during dynamic adaptation.\\
		\cellcolor{CoolGray!50}&13.	\textit{Autonomous Incorporation of Skills:} Dynamically selects and integrates known robot skills (e.g. action primitives) as needed for current task/environment into the current solution strategy (like tool-calling).\\
		\cellcolor{CoolGray!50}&14.	\textit{Online/Mid-Execution Adaptation:} Integrates operator/environmental feedback, (input/ task)-quality degradation and detected faults dynamically during operation, adapting strategy, input importance or output.\\
		\cellcolor{CoolGray!50}&15.	\textit{Explained Adaptation Decisions:} Explains, upon request, its performed/ intended adaptation of strategies, handling of distractors, input importance/quality (for each modality), and response to uncertainties or noise.\\
		\cellcolor{CoolGray!50}&16.	\textit{Memory of Previous Faults/Distractors:} Remembers prior encountered variations, distractors, faults, anomalies, environmental changes and corresponding solutions, reducing the need for repeated user intervention for recurring scenarios.\\
		\cellcolor{CoolGray!50}&17.	\textit{Energy/System-Aware Adaptation:} Adapts solution strategies and predicted actions based on real-time resource constraints and consumption (e.g. battery charge, system health, energy/ compute utilisation).\\
		\cellcolor{CoolGray!50}&18.	\textit{Self-Evaluation} \& {Command Adherence:} Continuously self-evaluates its own adapted behaviour and adherence to user command(s): Flags or readjusts if deviation from expected outcome or operator intent is detected.\\
		\cellcolor{CoolGray!50}&\cellcolor{CoolGray!50}\\
		\cellcolor{CoolGray!50}\textbf{I2:}&\multicolumn{1}{>{\centering\arraybackslash\columncolor{Platinum!60}}m{\CCWidth}}{A. Verification, Standards, and Certification}\\
		\cellcolor{CoolGray!50}\textbf{Safety \& Compliance}&1. \textit{Standards-Compliant:} Behaviour and safety features are checked to align with relevant industrial standards, evaluated by reproducible tests.\\
		\cellcolor{CoolGray!50}&2. \textit{Integrable Safety Systems:} System interfaces and safety signals are compatible with industrial plant safety infrastructure, supporting seamless integration and monitoring (e.g. model-generated safety stop can be directly used to stop respective production line).\\
		\cellcolor{CoolGray!50}&3. \textit{Safety Documentation} \& \textit{Traceability:} All safety features, integration, and reasoning are well documented, including considered safety rules, internal processes, output explanations, and verified compliance with standards.\\
		\cellcolor{CoolGray!50}&\multicolumn{1}{>{\centering\arraybackslash\columncolor{Platinum!60}}m{\CCWidth}}{B. Safety Reasoning, Adaptation, and Confidence}\\
		\cellcolor{CoolGray!50}&4. \textit{Safety Confidence Output:} Outputs a confidence score estimating the safety margin of each predicted operation and for the complete task before starting operation (in the current environment). The confidence score quantifies the system's estimated likelihood that all specified safety requirements are currently satisfied, using sensor/hardware/context data.\\
		\cellcolor{CoolGray!50}&5. \textit{Persistence of Safety Capabilities in Adaptation:} Safety rules/modules remain enforced, and system safety is verified after any model adaptation, retraining, or finetuning $\to$ standalone safety logic cannot be bypassed by learning changes.\\
		\cellcolor{CoolGray!50}&6. \textit{Reactive, Real-Time Safety:} System performs continuous, real-time evaluation of safety (considering sensor noise, hardware degradation, dynamic obstacles, harmful objects, \gls{ood}, and environment changes) -- e.g. leveraging confidence -- with immediate adaptation (halting/ triggering stop, slowing, or replanning) if risks arise.\\
		\cellcolor{CoolGray!50}&7. \textit{Proactive, Plan-Level Safety:} Safety reasoning is incorporated into high-level planning/solution strategies (not only reactive), evaluating alternative actions and solutions by their associated predicted safety confidence.\\
		\cellcolor{CoolGray!50}&\multicolumn{1}{>{\centering\arraybackslash\columncolor{Platinum!60}}m{\CCWidth}}{C. Safety Rules, Enforcement, and Supervisory Control}\\
		\cellcolor{CoolGray!50}&8. \textit{Rule-Adherence} \& \textit{Autonomous Rejection:} The system always respects predefined safety/behavioural rules and can autonomously reject operator goals/commands that pose unacceptable risk (e.g. “grab knife and throw”), e.g. supervised by independently certified controllers/modules.\\
		\cellcolor{CoolGray!50}&9. \textit{Command Compliance with Online-Safety:} System adheres strictly to operator commands as long as safety requirements are not compromised: Command execution is continuously monitored and suspended/modified if unacceptable risk is detected, providing transparent feedback (to the user/ logs/ supervisor/ etc.).\\
		\cellcolor{CoolGray!50}&10. \textit{Redundant, Independent Safety Mechanisms:} Employs multiple, independent (possibly redundant and diverse-modality) mechanisms for safety assessment and enforcement (e.g. dual model prediction, dedicated supervisor, independent safety layers).\\
		\cellcolor{CoolGray!50}&11.	\textit{Decoupled, Real-Time Safety Control:} Safety checks and action execution operate independently: a "safety semaphore" or gating mechanism ensures actions are only predicted/executed when confidence exceeds threshold.\\
		\cellcolor{CoolGray!50}&\multicolumn{1}{>{\centering\arraybackslash\columncolor{Platinum!60}}m{\CCWidth}}{D. Operator/User and External Interfaces}\\
		\cellcolor{CoolGray!50}&12.	\textit{Authorised Operator Detection} \& \textit{Override:} Only authorised/ recognised operators may control safety overrides or cause safety-critical actions; interfaces securely verify operator identity and permissions.\\
		\cellcolor{CoolGray!50}&13.	\textit{External Safety Integration:} Safety outputs/signals are shareable with other robots/systems in plant (including alarms, safe state signals, zone definitions, and status reports).\\
		\cellcolor{CoolGray!50}&\multicolumn{1}{>{\centering\arraybackslash\columncolor{Platinum!60}}m{\CCWidth}}{E. Fault Detection and Recovery}\\
		\cellcolor{CoolGray!50}&14.	\textit{Fault Detection} \& \textit{Sensing:} Actively detects major faults or safety violations. Distinguishes between sensor fault and safety/ environmental hazard and the degree of severity.\\
		\cellcolor{CoolGray!50}&15.	\textit{Safe, Autonomous Recovery:} Recovers autonomously to a safe state after detecting major faults, safety loss, low safety-related confidence in own actions or accidents.\\
		\cellcolor{CoolGray!50}&16.	\textit{Learning from Safety Events:} Continuously adapts and improves safety confidence/logic or recovery behaviour based on feedback from operator, supervisor modules, comparison of actual vs. predicted safety outcomes and past incidents.\\
		\cellcolor{CoolGray!50}&\multicolumn{1}{>{\centering\arraybackslash\columncolor{Platinum!60}}m{\CCWidth}}{F. Transparency and Logging}\\
		\cellcolor{CoolGray!50}&17.	\textit{Transparent Safety Reasoning:} The system can explain its safety confidence, risk assessment, and chosen/blocked actions to operators, both live and retrospectively (with logs/reports or upon querying).\\
		\cellcolor{CoolGray!50}&18.	\textit{Persistent Safety Logs:} All safety-relevant events, decisions, adaptations, actions, and outcomes are logged, both user-readable and for self-improvement.\\
		\cellcolor{CoolGray!50}&\cellcolor{CoolGray!50}\\
		\cellcolor{CoolGray!50}\textbf{I3:}&\multicolumn{1}{>{\centering\arraybackslash\columncolor{Platinum!60}}m{\CCWidth}}{A. Communication Modalities \& Operator Interfaces}\\
		\cellcolor{CoolGray!50}\shortstack{\textbf{HRI} \\ \textbf{\&} \\ \textbf{HRC}\\~\\~\\~\\~}&1. \textit{Multimodal Instruction} \& \textit{Output:} Accepts and generates instructions, queries, and feedback through multiple operator interfaces (speech, text, gesture, visual). Chooses the most suitable modalities based on available hardware, environmental context (e.g. noise level, lighting), operator preferences, and task demands/ instruction.\\
		\cellcolor{CoolGray!50}&2. \textit{Instruction Confidence:} For every operator command or query, provides a confidence score reflecting certainty in both (a) correct recognition of the input (e.g. speech recognition, gesture identification), and (b) correct interpretation of the operator’s intent. When confidence is low, proactively seeks clarification or reframing from the operator, and communicates the reason for uncertainty (ambiguity, low signal, novel command, etc.).\\
		\cellcolor{CoolGray!50}&3. \textit{Intuitive Feedback Acceptance:} Integrates operator clarifications, corrections, or input refinements (e.g. specifying objects, adjusting approach, clarifying intent) online, into the ongoing action interpretation or prediction – adapting its immediate responses and future understanding, mostly without pausing or disrupting the main task flow.\\
		\cellcolor{CoolGray!50}&\multicolumn{1}{>{\centering\arraybackslash\columncolor{Platinum!60}}m{\CCWidth}}{B. Responsiveness \& Interaction Control}\\
		\cellcolor{CoolGray!50}&4. \textit{Real-time Interruptibility:} Allows operators/ users to instantly and directly pause, modify, query, or halt ongoing execution or task, providing full control and safe intervention across all interfaces and modalities.\\
		\cellcolor{CoolGray!50}&5. \textit{Continuous Responsiveness:} Maintains complete HRI/HRC accessibility and interaction throughout all operational phases (idle, executing, transitioning, recovering), ensuring operators can always intervene, instruct, or query – never leaving the system "locked out" or unresponsive.\\
		\cellcolor{CoolGray!50}&\multicolumn{1}{>{\centering\arraybackslash\columncolor{Platinum!60}}m{\CCWidth}}{C. Operator Management, Authorisation \& Privacy}\\
		\cellcolor{CoolGray!50}&6. \textit{Active Operator Exclusivity:} Guarantees that only one authorised/ known operator/ user is "active" and in control at any time. Enforces explicit, secure handover or switching protocols, and reliably ignores unverified or distracting bystanders in both input and output.\\
		\cellcolor{CoolGray!50}&7. \textit{Operator Identification} \& \textit{Authorisation:} Identifies and responds exclusively to securely recognised and authorised personnel/ user. Actively prevents unauthorised commands, even under disruptive or noisy conditions, and logs all identity-related access attempts.\\
		\cellcolor{CoolGray!50}&8. \textit{Operator Data Privacy:} Encrypts and protects operator-specific data and interaction history; permits authorised deletion or review in accordance with relevant privacy standards. Ensures only authorised personnel can access, modify, or request deletion of such data.\\
		\cellcolor{CoolGray!50}&9. \textit{Learned Operator Preferences:} Adapts interface layout, communication style, behaviour, or task sequencing to individual operators/ users – learning and persisting preferences across sessions for efficiency and comfort unless instructed otherwise.\\
		\cellcolor{CoolGray!50}&\multicolumn{1}{>{\centering\arraybackslash\columncolor{Platinum!60}}m{\CCWidth}}{D. Human-Robot Collaboration \& Behaviour Adaptation}\\
		\cellcolor{CoolGray!50}&10. \textit{Human Presence Awareness in Workspace:} Continuously senses human presence and proximity throughout and beyond the workspace using multiple sensor modalities (vision, audio, proximity, etc.). Dynamically adapts operation speed, safety margins, and communication in response to detected changes (especially in cases of unexpected entries, operator absence, etc.).\\
		\cellcolor{CoolGray!50}&11. \textit{Collaborative Adaptation:} Reactively modulates solution strategies, action plans, and interaction traits in direct response to real-time operator feedback, demonstrations, or correction. Behaves as an adaptive teammate, tailoring technical and social behaviours to maximise collaboration efficiency.\\
		\cellcolor{CoolGray!50}&12.	\textit{Human-Centric Adaptation:} Dynamically and autonomously adjusts behaviour, speed, and language/output based on human presence, operator skill/preference, perceived operator condition, task safety factors, and perceived risk.\\
		\cellcolor{CoolGray!50}&\multicolumn{1}{>{\centering\arraybackslash\columncolor{Platinum!60}}m{\CCWidth}}{E. Reasoning, Standards \& Exception Handling}\\
		\cellcolor{CoolGray!50}&13.	\textit{Transparent Reasoning Explanation:} Provides operator-friendly, on-demand explanations during operation of its own reasoning, uncertainty, behaviour, and decision-making – using appropriate modalities (natural language, images, charts) and tailoring explanations to the operator’s degree of expertise.\\
		\cellcolor{CoolGray!50}&14.	\textit{Industrial HRI/HRC Standards Compliance:} Adheres to relevant industrial safety, communication, and collaboration standards and best-practices for HRI and HRC; supports integration with company-specific norms, exception rules, and documentation requirements.\\
		\cellcolor{CoolGray!50}&15.	\textit{Exception Handling} \& \textit{Safe State Transition:} Reliably recognises faults, abnormal human behaviours, and communication breakdowns during HRI/HRC. Notifies the operator, explains the situation, and if necessary, transitions to a pre-defined safe state.\\
		\cellcolor{CoolGray!50}&\cellcolor{CoolGray!50}\\
		\cellcolor{CoolGray!50}\textbf{I4:}&\multicolumn{1}{>{\centering\arraybackslash\columncolor{Platinum!60}}m{\CCWidth}}{A. Fault Detection, Diagnosis \& Monitoring}\\
		\cellcolor{CoolGray!50}\shortstack{\textbf{Robustness} \\ \textbf{\&} \\ \textbf{Reliability}\\~\\~\\~\\~\\~\\~\\~}&1. \textit{Multi-Level Fault} \& \textit{Anomaly Detection:} Continuously detects and logs faults, anomalies, and degradations not only in system hardware and sensors, but also in environment, manipulated objects, operator actions, and internal representations – utilising diverse sensor inputs, automatic evaluation, supervisor components or confidence scores to spot issues before failures occur.\\
		\cellcolor{CoolGray!50}&2. \textit{Reproducible} \& \textit{Benchmarked Fault Diagnosis:} Integrated fault/anomaly detection delivers consistent, reproducible evaluations under matching conditions, with detection precision benchmarked against standardised scenarios.\\
		\cellcolor{CoolGray!50}&3. \textit{Threat} \& \textit{Attack Detection:} Autonomously monitors all inputs and interface signals for security threats, adversarial manipulations, and unauthorised commands, triggering alerts or adapting strategies or behaviour to mitigate risk and reduce attack impact.\\
		\cellcolor{CoolGray!50}&\multicolumn{1}{>{\centering\arraybackslash\columncolor{Platinum!60}}m{\CCWidth}}{B. Error Handling, Recovery \& Adaptation}\\
		\cellcolor{CoolGray!50}&4. \textit{Automated Recovery} \& \textit{Safe-State Transition:} Autonomously initiates recovery routines or shifts into a safe state when critical faults, severe anomalies, unrecoverable disturbances, or low confidences are detected (prioritisation of safety and system integrity over performance and task completion).\\
		\cellcolor{CoolGray!50}&5. \textit{Graceful Degradation} \& \textit{Strategy Adaptation:} Adjusts execution plans, sensor usage, and action strategies in real-time when encountering partial sensor failures, orientation/config changes, or slightly degraded environmental conditions, striving to maintain performance without outright task failure.\\
		\cellcolor{CoolGray!50}&6. \textit{Autonomous Fault Categorisation} \& \textit{Severity Response:} Recognises, categorises, and prioritises faults by severity, autonomously selecting and executing the most appropriate recovery behaviour, fallback routine, or adaptation strategy.\\
		\cellcolor{CoolGray!50}&7. \textit{Robust Redundancy} \& \textit{Modality Management:} Ensures reliable operation and fault tolerance by accepting, integrating, and dynamically switching between redundant input and output modalities. When any channel degrades or fails, the system autonomously maintains task progress through alternative modalities, adapting detection, execution, and communication strategies to suit the current hardware, environment, and operator constraints.\\
		\cellcolor{CoolGray!50}&\multicolumn{1}{>{\centering\arraybackslash\columncolor{Platinum!60}}m{\CCWidth}}{C. Robustness to Disturbance, Environment \& Variation}\\
		\cellcolor{CoolGray!50}&8. \textit{Environmental Robustness} \& \textit{Adaptation:} Operates effectively across varying environmental conditions (e.g. lighting, noise, object/material variation), detecting and alerting for changes or disturbances that could risk execution reliability.\\
		\cellcolor{CoolGray!50}&9. \textit{Disturbance Resilience} \& \textit{Continuous Operation:} Maintains task progress and adherence to commands even when exposed to distractors or minor disturbances, recovering or adapting autonomously when possible. (e.g. arm pushed aside, etc.)\\
		\cellcolor{CoolGray!50}&\multicolumn{1}{>{\centering\arraybackslash\columncolor{Platinum!60}}m{\CCWidth}}{D. Learning, Feedback \& Continuous Improvement}\\
		\cellcolor{CoolGray!50}&10.	\textit{Adaptive Error Learning} \& \textit{Memorisation:} Continuously learns from past faults, errors, environmental changes, and operator feedback –memorising causes and resolutions to improve future fault/ error detection, prediction, and recovery.\\
		\cellcolor{CoolGray!50}&11.	\textit{Online Confidence} \& \textit{Reliability Assessment:} Continuously generates a clear reliability confidence score for further operation, quantifying the risk and likelihood of fault occurrence and successful execution.\\
		\cellcolor{CoolGray!50}&\multicolumn{1}{>{\centering\arraybackslash\columncolor{Platinum!60}}m{\CCWidth}}{E. Supervision, Oversight \& Trust}\\
		\cellcolor{CoolGray!50}&12. \textit{Supervisor Oversight} \& \textit{Fault Review: }Employs dedicated supervisor components to monitor for faults in execution, hardware/sensor input, and operator commands, critically evaluating instructions and requesting operator engagement for suspected errors.\\
		\cellcolor{CoolGray!50}&13. \textit{Logging, Explanation} \& \textit{Operator Guidance:} Logs and communicates faults, anomalies, and noise through varied modalities, providing operators with actionable explanations (what/why/when/how to resolve) and performance insights for transparency and trust.\\
		\cellcolor{CoolGray!50}&14. \textit{Consistent} \& \textit{Documented Fault Handling:} Ensures both fault/anomaly-detection and response-behaviours remain reproducible and consistent across time and conditions, with all capabilities and accepted modalities documented and evaluated.\\
		\cellcolor{CoolGray!50}&\cellcolor{CoolGray!50}\\
		\cellcolor{CoolGray!50}\textbf{I5:}&\multicolumn{1}{>{\centering\arraybackslash\columncolor{Platinum!60}}m{\CCWidth}}{A. Specification Compliance}\\
		\cellcolor{CoolGray!50}\textbf{Precision \& Accuracy}&1. \textit{v/P}\&\textit{A Instructability:} For every task specification, the system can be instructed to meet an explicit \gls{vpa} target.\\
		\cellcolor{CoolGray!50}&2. \textit{Outcome Logging} \& \textit{Failure Reasoning:} Gathers/ creates and logs all available \gls{vpa}-relevant task execution metrics (such as velocity and, if sensor capabilities permit, achieved precision \& accuracy) after each run. If direct measurement is not possible, the system leverages predicted values, confidence scores, or proxy indicators. For any deviation or failure to meet specified standards, the system documents and explains likely reasons based on the given information.\\
		\cellcolor{CoolGray!50}&3. \textit{User-Defined Trade-offs:} Supports explicit operator input for \gls{vpa}-trade-offs (e.g. maximising speed at the cost of precision, or vice versa if either cannot be ensured), before or during execution and ensures system behaviour adapts accordingly.\\
		\cellcolor{CoolGray!50}&\multicolumn{1}{>{\centering\arraybackslash\columncolor{Platinum!60}}m{\CCWidth}}{B. Evaluation, Reporting \& Confidence}\\
		\cellcolor{CoolGray!50}&4. \textit{v/P}\&\textit{A-Confidence Score} \& \textit{Explanation:} For every execution, outputs a confidence score estimating the predicted likelihood that required \gls{vpa} will be achieved, factoring in task, object/material, environment, and other sensor data. When low confidence is predicted, system can explain likely causes (e.g. out-of-distribution scenario, sensor noise, disturbances).\\
		\cellcolor{CoolGray!50}&5. \textit{Achievable v/P}\&\textit{A-Proposals:} Alongside the confidence score and explanation, the system can propose and explain alternative \gls{vpa}-combinations likely to succeed in the given context (task, object, environment), helping users adjust requirements if needed.\\
		\cellcolor{CoolGray!50}&6. \textit{Transparent Documentation of v/P}\&\textit{A Capabilities:} Achievable \gls{vpa}-combinations for every tested task, object, and context is provided alongside the model (including limitations, known influence factors, etc.).\\
		\cellcolor{CoolGray!50}&\multicolumn{1}{>{\centering\arraybackslash\columncolor{Platinum!60}}m{\CCWidth}}{C. Consistency \& Disturbance Handling}\\
		\cellcolor{CoolGray!50}&7. \textit{Consistency Over Time} \& \textit{Contexts:} Demonstrates consistent \gls{vpa} across known tasks, objects, and environments, and under repeated test conditions.\\
		\cellcolor{CoolGray!50}&8. \textit{Disturbance} \& \textit{Uncertainty Management:} Identifies and factors in disturbances, distractors, sensor anomalies, or environmental changes when predicting, achieving, and reporting \gls{vpa} metrics. Explains in real-time what current conditions may affect \gls{vpa} and adapts behaviour accordingly.\\
		\cellcolor{CoolGray!50}&9. \textit{Autonomous v/P}\&\textit{A Adaptation:} Automatically adapts its own \gls{vpa}-settings/ targets and solution strategies in response to changing conditions or goals (e.g. sensor quality, energy saving) within the user-defined bounds (e.g. reduces speed within user-defined limits to save energy).\\
		\cellcolor{CoolGray!50}&\multicolumn{1}{>{\centering\arraybackslash\columncolor{Platinum!60}}m{\CCWidth}}{D. Learning, Feedback \& Continuous Improvement}\\
		\cellcolor{CoolGray!50}&10. \textit{v/P}\&\textit{A Learning from Experience and Feedback:} Memorises and utilises past \gls{vpa}-failures (e.g. slower than/ not as precise as demanded) and successful adaptations of \gls{vpa} to improve prediction confidence, action planning, and further relevant decisions for similar future tasks.\\
		\cellcolor{CoolGray!50}&11. \textit{HRI/HRC Integration:} Accepts \gls{vpa}-related user/operator feedback during or after execution to update confidence or task strategy and leverages help from operators or other robots to meet \gls{vpa} requirements.\\
		\cellcolor{CoolGray!50}&\multicolumn{1}{>{\centering\arraybackslash\columncolor{Platinum!60}}m{\CCWidth}}{E. External/ Separate Component Integration}\\
		\cellcolor{CoolGray!50}&12. \textit{Skill Integration with Known v/P}\&\textit{A:} Incorporates predefined and learned skills/tools for improved \gls{vpa} outcomes, selecting solutions with documented effectiveness whenever possible.\\
		\cellcolor{CoolGray!50}&13.	\textit{Supervision} \& \textit{Verification Module:} Includes separated modules or capabilities to explicitly verify actual outcome \gls{vpa} during and post-execution, not just prediction. Alerts/warns model/operator if outcome may not meet requirements, explaining why and offering alternatives.\\
		\cellcolor{CoolGray!50}&\cellcolor{CoolGray!50}\\
		\cellcolor{CoolGray!50}\textbf{I6:}&\multicolumn{1}{>{\centering\arraybackslash\columncolor{Platinum!60}}m{\CCWidth}}{A. Real-Time Robustness \& Generality}\\
		\cellcolor{CoolGray!50}\shortstack{\textbf{Real-time}\\ \textbf{Performance}\\~\\~\\~\\~\\~\\~\\~\\~\\~\\~\\~}&1. \textit{Task, Input} \& \textit{Environment Independence:} Meets real-time requirements (e.g. inference, control, responsiveness) regardless of task complexity, environment, objects, input modality, number or dimensionality of in/outputs, or presence of distractors, \gls{ood} conditions, or hardware degradation – demonstrated through benchmarked worst-case test scenarios on a given hardware-setup.\\
		\cellcolor{CoolGray!50}&2. \textit{System-wide Real-time Compliance:} Real-time operation is maintained holistically across all relevant system components – perception, reasoning, actuation, sensing, communication, feedback, and logging – rather than only for a subset.\\
		\cellcolor{CoolGray!50}&\multicolumn{1}{>{\centering\arraybackslash\columncolor{Platinum!60}}m{\CCWidth}}{B. Adaptive Resource \& Real-Time Management}\\
		\cellcolor{CoolGray!50}&3. \textit{Dynamic Resource Management} \& \textit{Determinism:} Autonomously adapts compute usage (e.g. through output/inference frequency modulation) to maintain user- or norm-specified real-time targets. When under load or during resource shortages, it disables or degrades non-critical functions to uphold core real-time properties and safety measures. Benchmarked under varying hardware constraints and external disturbances.\\
		\cellcolor{CoolGray!50}&4. \textit{Autonomous Adaptation} \& \textit{Safety Response:} Upon detecting real-time violations or imminent deadline risks, triggers fallback routines, prioritises safety-critical functionality, and can alert operators or other systems as needed. Adaptation and prioritisation logic is explainable and logged.\\
		\cellcolor{CoolGray!50}&5. \textit{Goal-Responsive Solution Strategies:} Comprehends and adapts to new or updated user goals and instructions "in real time" based on current workload – ensuring planning and execution adaptation meets pre-defined latency requirements.\\
		\cellcolor{CoolGray!50}&\multicolumn{1}{>{\centering\arraybackslash\columncolor{Platinum!60}}m{\CCWidth}}{C. Monitoring, Reporting, \& Continuous Improvement}\\
		\cellcolor{CoolGray!50}&6. \textit{Live Metrics, Logging} \& \textit{Explainability:} Continuously monitors, reports, and logs real-time metrics (e.g. frequencies, latencies, dropped cycles, deadline adherence), makes these accessible, and provides clear explanations for performance drops, adaptation events, or system changes (e.g. what, why, when a module was throttled/disabled).\\
		\cellcolor{CoolGray!50}&7. \textit{Real-Time Confidence/Alerting:} Continuously outputs a confidence score predicting current and future compliance with real-time constraints (factoring in compute status, history, planned actions, hardware utilisation). Issues alerts or recommendations if violation risk increases.\\
		\cellcolor{CoolGray!50}&8. \textit{Historical Data Utilisation:} Integrates historical real-time performance/logs and adaptation outcomes to optimise future resource use, prediction, confidence and adaptation strategies, coupled with continued evaluation.\\
		\cellcolor{CoolGray!50}&9. \textit{Transparent Documentation} \& \textit{Compliance:} Maintains up-to-date, accessible documentation on real-time system capabilities, adaptation mechanisms, hardware requirements (e.g. which hardware-specifications are required for which set of enabled functionalities) with evaluated compliance. All possible real-time related adaptations, violation cases and countermeasures are documented.\\
		\cellcolor{CoolGray!50}&\cellcolor{CoolGray!50}\\
		\cellcolor{CoolGray!50}\textbf{I7:}&\multicolumn{1}{>{\centering\arraybackslash\columncolor{Platinum!60}}m{\CCWidth}}{A. Hardware \& Operational Cost-Efficiency}\\
		\cellcolor{CoolGray!50}\textbf{Cost-effectiveness \& Integration Capabilities}&1. \textit{Low-Cost Hardware Readiness:} Delivers reliable robot control and core model capabilities on commercially available, affordable compute hardware (e.g. edge devices, embedded platforms), supporting lower-spec sensors and actuators.\\
		\cellcolor{CoolGray!50}&2. \textit{Dynamic Resource Allocation:} Dynamically adjusts its resource usage – such as modulating inference frequency, simplifying/ deactivating optional model operations, adapting velocity, etc. – minimising energy use or execution duration given current environmental factors (e.g. current required throughput) to reduce operating/ production-line costs.\\
		\cellcolor{CoolGray!50}&\multicolumn{1}{>{\centering\arraybackslash\columncolor{Platinum!60}}m{\CCWidth}}{B. Universal \& Modular Integration}\\
		\cellcolor{CoolGray!50}&3. \textit{Universal Integration without Retraining:} Seamlessly integrates with diverse robot embodiments, kinematics, sensor and actuator types, and plant automation systems, without requiring model retraining, software changes, or adaptation for each hardware configuration change. Expected performance is preserved wherever hardware capacities allow.\\
		\cellcolor{CoolGray!50}&4. \textit{Hot-Swapping} \& \textit{Auto-Detection:} Supports hot-plugging and automatic detection of new, swapped, or removed input/output channels (sensors, actuators, interfaces, communication), updating its operation and configuration with no need for manual operator intervention.\\
		\cellcolor{CoolGray!50}&\multicolumn{1}{>{\centering\arraybackslash\columncolor{Platinum!60}}m{\CCWidth}}{C. Commissioning, Flexibility \& Self-Adaptation}\\
		\cellcolor{CoolGray!50}&5. \textit{Guided} \& \textit{Autonomous Commissioning:} Features intuitive commissioning workflows, including automated calibration and adaptation procedures, providing feedback on missing/ expected/ proposed in-/outputs or adjustments, enabling non-expert users to deploy and onboard the model with minimal manual setup.\\
		\cellcolor{CoolGray!50}&6. \textit{Flexible Input/Output Adaptation:} Automatically configures itself to accept changing sensor modalities, command interfaces, and action outputs, functioning consistently even if components are substituted, or temporarily degraded/ missing – requiring no manual software changes.\\
		\cellcolor{CoolGray!50}&7. \textit{Autonomous Fault Handling} \& \textit{Feedback:} Autonomously detects and diagnoses sensor, actuator or integration faults and missing interfaces – provides clear, actionable feedback or explanations to users for resolution and continued safe operation.\\
		\cellcolor{CoolGray!50}&\multicolumn{1}{>{\centering\arraybackslash\columncolor{Platinum!60}}m{\CCWidth}}{D. Documentation \& Integration Performance}\\
		\cellcolor{CoolGray!50}&8. \textit{Comprehensive Integration Documentation:} Includes detailed, accessible documentation for all supported platforms, integration methods, autonomous adaptation and recognition mechanisms, commissioning and troubleshooting steps, supplemented with example configurations and practical best practices.\\
		\cellcolor{CoolGray!50}&9. \textit{Integration Performance Monitoring} \& \textit{Learning:} Continually logs integration performance, adaptation strategies/ speed, and incident recoveries – using historical deployment data to inform, accelerate, and improve future onboarding and commissioning processes.\\
		\cellcolor{CoolGray!50}&10. \textit{Cost-Reduction Memory} \& \textit{Adaptive Optimisation:} Systematically records operational experiences that relate to resource and cost efficiency – such as energy-conserving behaviours, internal module optimisations, and hardware usage patterns – and leverages this memory to autonomously propose or implement more cost-effective action plans, configurations, and adaptation strategies in future deployments.\\
		\cellcolor{CoolGray!50}&\cellcolor{CoolGray!50}\\
		\cellcolor{CoolGray!50}\textbf{I8:}&\multicolumn{1}{>{\centering\arraybackslash\columncolor{Platinum!60}}m{\CCWidth}}{A. Capability, Limitation \& Reasoning Transparency}\\
		\cellcolor{CoolGray!50}\shortstack{\textbf{Explainability}\\ \& \\ \textbf{Trust}\\~\\~}&1. \textit{Task Feasibility} \& \textit{Capability Explanation:} Supports clear, user-facing explanations of what tasks it can or cannot solve – including limitations, current capabilities, and the feasibility of the currently instructed task – both before and during execution, with accessible rationale for any constraint.\\
		\cellcolor{CoolGray!50}&2. \textit{Reason Logging} \& \textit{Accessible Justification:} Autonomously generates, logs, and exposes the reasoning behind its critical decisions, action choices, adaptations, and detected failures, ensuring each is transparently documented and retrievable for audit, debugging, self-improvement and user review.\\
		\cellcolor{CoolGray!50}&3. \textit{Explains Influences} \& \textit{Disturbances:} Provides explanations regarding environmental factors (e.g. lighting, scene setup, object/material properties, distractors, detected disturbances) and how these influence its reasoning, behaviour, or performance, along with advice or solutions for mitigation.\\
		\cellcolor{CoolGray!50}&\multicolumn{1}{>{\centering\arraybackslash\columncolor{Platinum!60}}m{\CCWidth}}{B. Self-Assessment, Learning, \& Feedback Integration}\\
		\cellcolor{CoolGray!50}&4. \textit{Continuous Self-Assessment with Explainability:} Consistently self-evaluates capabilities, performance, and limitations and their impacts in the current context (e.g. hardware, sensor setup, noise conditions, distribution shifts), providing the evaluation alongside an explanation and leveraging these insights for ongoing online learning, self-improvement and user-insight.\\
		\cellcolor{CoolGray!50}&5. \textit{User Feedback} \& \textit{Adaptive Improvement:} Supports intuitive interfaces for user feedback and corrections during and after execution (e.g. thumbs up/down, follow-up queries, correction inputs), acknowledges understanding, adapts behaviour accordingly, and provides explanations of resulting changes; uses feedback to refine future reasoning.\\
		\cellcolor{CoolGray!50}&6. \textit{Monitors} \& \textit{Logs Explainability for Learning:} Keeps persistent records of all generated explainability outputs (reasoning, assessment, error reports), using these logs to continuously improve future behaviours, onboarding, and commissioning processes.\\
		\cellcolor{CoolGray!50}&\multicolumn{1}{>{\centering\arraybackslash\columncolor{Platinum!60}}m{\CCWidth}}{C. Explainability Methods, Interfaces \& Outputs}\\
		\cellcolor{CoolGray!50}&7. \textit{Flexible, Intuitive Explainability Interfaces:} Offers explanation outputs in multiple modalities (text, speech, visualisation, etc.), supporting both non-expert operators and technical users, with methods to visualise/explain impact of user commands, input/output importance, intermediate steps, and internal processes at runtime.\\
		\cellcolor{CoolGray!50}&8. \textit{Confidence Score Explanation:} All confidence scores (for actions, predictions, or decisions) are supplied with user-facing explanations – detailing what influenced high or low values, such as data loss, fallback mode, sensor unreliability, or ambiguous scene attributes.\\
		\cellcolor{CoolGray!50}&9. \textit{Explainability Across System Components:} Integrates explainability tools and interfaces across all modules, enabling explainable actions, outputs, and internal reasoning during setup, operation, debugging, and benchmarking; all methods are thoroughly documented with guidance for triggering and interpreting results.\\
		\cellcolor{CoolGray!50}&\multicolumn{1}{>{\centering\arraybackslash\columncolor{Platinum!60}}m{\CCWidth}}{D. Security, Compliance, \& Benchmarking Explainability}\\
		\cellcolor{CoolGray!50}&10. \textit{Security} \& \textit{Authorisation Feedback:} Detects and explains potential security issues, adversarial attacks, or authorisation events – providing specifics on why the event was flagged, input used, type of threat, and how to strengthen defences or permissions for future operations.\\
		\cellcolor{CoolGray!50}&11. \textit{Rule} \& \textit{Norm Compliance with Explainability:} Evaluates and explains compliance with operational guidelines, safety norms, and domain-specific rules (both upon being instructed and during/after execution), visualising/documenting adherence, deviations, and model reasoning behind its decision.\\
		\cellcolor{CoolGray!50}&12. \textit{Performance Benchmarking} \& \textit{OOD Feedback:} Integration of explainability methods in benchmarking, evaluation, and reporting – e.g. visualising focus, highlighting difficulties in \gls{ood} situations, or explaining how improvements, data changes, or model modifications affect learned skills or generalisation, etc.\\
		\cellcolor{CoolGray!50}&\multicolumn{1}{>{\centering\arraybackslash\columncolor{Platinum!60}}m{\CCWidth}}{E. Trust \& Transparency during Runtime}\\
		\cellcolor{CoolGray!50}&13. \textit{Behaviour Consistency} \& \textit{Pre-Adaptation Notification:} Maintains consistent, transparent explanations for identical inputs and conditions, avoids “super-hero syndrome” (thinking to be capable to solve every task), and informs the user before adapting behaviour due to error, environment change, or new context.\\
		\cellcolor{CoolGray!50}&14. \textit{Error, Anomaly} \& \textit{Safeguard Explanations:} Immediately triggers explanation processes for detected errors, anomalies, or failed safeguards – detailing what happened, why, and the recovery strategies attempted or proposed. Explanation is automatically provided to relevant system components/ logs/ etc.\\
		\cellcolor{CoolGray!50}&\cellcolor{CoolGray!50}\\
		\cellcolor{CoolGray!50}\textbf{I9:}&\multicolumn{1}{>{\centering\arraybackslash\columncolor{Platinum!60}}m{\CCWidth}}{A. Multimodal Acceptance \& Fusion}\\
		\cellcolor{CoolGray!50}\shortstack{\textbf{Sensor Fusion}\\ \& \\ \textbf{Perception}\\~\\~\\~}&1. \textit{Multimodal Sensor Acceptance} \& \textit{Fusion:} Accepts diverse sensors and supports their real-time fusion and processing across modalities (visual, tactile, audio, etc.), including redundant interchangeable sensors, with the ability to interpret and combine inputs from multiple simultaneous sources for robust perception.\\
		\cellcolor{CoolGray!50}&2. \textit{Flexible Input Combination} \& \textit{Goal Specification:} Accepts task instructions, goals, and feedback through multiple modalities (speech/text, gesture, image, etc.), and can leverage several modalities simultaneously for improved perception, understanding, and command robustness.\\
		\cellcolor{CoolGray!50}&3. \textit{Security in Sensor Operations:} Utilises various sensor sources/ modalities to detect adversarial attacks, facilitates secure authorisation, and ensures safe detection of humans/workers.\\
		\cellcolor{CoolGray!50}&\multicolumn{1}{>{\centering\arraybackslash\columncolor{Platinum!60}}m{\CCWidth}}{B. Input Monitoring, Fault Tolerance \& Adaptation}\\
		\cellcolor{CoolGray!50}&4. \textit{Continuous Input Monitoring} \& \textit{Feedback:} Continuously checks sensor inputs for noise, faults, degradation, absence, or sub-optimal orientation, providing real-time feedback during integration, commissioning and operation, reporting missing or degraded inputs as well as recommendations for improvement.\\
		\cellcolor{CoolGray!50}&5. \textit{Robust Fault Tolerance} \& \textit{Degradation Handling:} Detects and actively compensates for sensor failures, degradation, or external disturbances by dynamically shifting input weighting, switching perception strategies, or prioritising the most reliable sensor(s) for the current scene, environment, and task context.\\
		\cellcolor{CoolGray!50}&6. \textit{Autonomous Disturbance Recognition} \& \textit{Countermeasures:} Recognises environmental disturbances that critically affect perception (e.g. poor lighting, moisture, obstructed view) and initiates context-sensitive countermeasures (like halting, alerting users, removing obstacles, or activating cleaning routines), and adapts perception strategies online.\\
		\cellcolor{CoolGray!50}&\multicolumn{1}{>{\centering\arraybackslash\columncolor{Platinum!60}}m{\CCWidth}}{C. Dynamic Input Prioritisation \& Adaptation}\\
		\cellcolor{CoolGray!50}&7. \textit{Dynamic Sensor Prioritisation:} Automatically ranks and prioritises active sensors and modalities based on current instruction, scene conditions, quality, relevance, historical context and reliability, adapting input strategy in real-time to environmental changes and operational demands.\\
		\cellcolor{CoolGray!50}&8. \textit{Online Learning for Perception Skills:} Continuously learns and adapts perception strategy (e.g. how to interpret or counteract noisy or degraded sensor values, shifting focus to non-degraded sensors, …) using historical data and reasoning explanations to improve future sensor fusion and perception capabilities.\\
		\cellcolor{CoolGray!50}&\multicolumn{1}{>{\centering\arraybackslash\columncolor{Platinum!60}}m{\CCWidth}}{D. Scene Understanding, Interpretation \& Output}\\
		\cellcolor{CoolGray!50}&9. \textit{Comprehensive Scene Understanding via Fusion:} Recognises objects, materials, spatial arrangements and environmental constraints by integrating information from multiple, complementary modalities – supplies rich scene interpretation, spatial awareness, and accurate environmental representations for downstream robot control.\\
		\cellcolor{CoolGray!50}&10. \textit{Perceptual Confidence} \& \textit{Explanation Output:} Provides a perceptual confidence score, summarising the certainty of scene/object perception and input quality. Complemented by interpreting input importance (which sensors/modalities influenced results) and supply of user-friendly explanations visualising/ describing reasoning behind understanding, degraded input handling, and current limitations.\\
		\cellcolor{CoolGray!50}&11. \textit{Perceptual Feature Exposure} \& \textit{Modularity:} Exposes interpretable intermediate representations and perceptual features to system components or external interfaces, supporting modularity, human supervision, and transparent debugging of perception skills.\\
		\cellcolor{CoolGray!50}&\multicolumn{1}{>{\centering\arraybackslash\columncolor{Platinum!60}}m{\CCWidth}}{E. Evaluation, Documentation \& Knowledge Exchange}\\
		\cellcolor{CoolGray!50}&12. \textit{Modular Integration} \& \textit{Documentation:} Enables flexible, exchangeable or adaptive observation spaces (autonomous/manual adaptation; fine-tuning optional) with support for hot-swapping sensor setups, providing clear feedback during integration and maintaining up-to-date documentation of all supported sensor types, modality usage, and perception features.\\
		\cellcolor{CoolGray!50}&13. \textit{Perception Evaluation} \& \textit{Benchmarking:} Benchmarks and documents perception skills (input importance ranking, noise adaptation, consistency, and reproducibility), and supplies results for model evaluation, external audit, and improvement.\\
		\cellcolor{CoolGray!50}&14. \textit{Peer-to-Peer Perception Knowledge Exchange:} Can exchange sensor/perception information and interpretations with other systems, enabling collective/global perception enhancement and collaborative scene/environmental understanding.\\
		\cellcolor{CoolGray!50}&\cellcolor{CoolGray!50}\\
		\cellcolor{CoolGray!50}\textbf{I10:}&\multicolumn{1}{>{\centering\arraybackslash\columncolor{Platinum!60}}m{\CCWidth}}{A. Coverage of Tasks, Domains, \& Environments}\\
		\cellcolor{CoolGray!50}\shortstack{\textbf{Standardised} \\ \textbf{Benchmarking}\\ \& \\ \textbf{Evaluation}}&1.	\textit{Benchmark Diversity} \& \textit{Scope:} Evaluated using standardised benchmarks that span a wide range of tasks, domains, environments, object types, and material properties – including known, partially known, and unknown scenarios – explicitly testing performance in worst-case distribution shift or \gls{ood} situations.\\
		\cellcolor{CoolGray!50}&2. \textit{Hardware Platform Benchmarking:} Evaluates model on multiple hardware platforms/ uses a hardware-aware benchmark, documenting resource utilisation, energy consumption, real-time performance, and cost-effectiveness across devices and sensor/actuator setups.\\
		\cellcolor{CoolGray!50}&\multicolumn{1}{>{\centering\arraybackslash\columncolor{Platinum!60}}m{\CCWidth}}{B. Robustness, Generalisation \& Transparency}\\
		\cellcolor{CoolGray!50}&3. \textit{Repeatability} \& \textit{Reproducibility:} Demonstrates repeatable, reproducible model/system behaviour in benchmark runs, validating consistency over multiple trials, hardware settings, and environmental variations.\\
		\cellcolor{CoolGray!50}&4. \textit{Robustness} \& \textit{Fault Tolerance Evaluation:} Systematically benchmarks performance against faults, disturbances, distractor objects, sensor degradation, and intentional adversarial attacks; measures recovery strategies, fallback/safeguard execution, and online adaptation when exposed to (un)seen disturbances.\\
		\cellcolor{CoolGray!50}&5. \textit{Generalisation} \& \textit{Adaptation Assessment:} Evaluates the model’s generalisation and online adaptation capabilities – including response to distribution shifts, novel commands, unknown inputs, and task or environmental changes – using both synthetic and real-world scenarios.\\
		\cellcolor{CoolGray!50}&6. \textit{Transparent Metrics} \& \textit{Reporting:} Publishes clear, traceable benchmark metrics – including performance, accuracy (\gls{vpa}), safety, robustness, fault tolerance, energy usage, real-time performance and compliance – with full documentation for all test scenarios.\\
		\cellcolor{CoolGray!50}&\multicolumn{1}{>{\centering\arraybackslash\columncolor{Platinum!60}}m{\CCWidth}}{C. Instruction, Output \& Integration Assessment}\\
		\cellcolor{CoolGray!50}&7. \textit{Instruction, Command} \& \textit{Modality Coverage:} Benchmarks instructability and goal specification for all supported input modalities (e.g. speech, text, images, gestures), focusing on interpretation accuracy, and solution strategy adherence.\\
		\cellcolor{CoolGray!50}&8. \textit{Integration} \& \textit{Commissioning Evaluation:} Assesses ease and effectiveness of integration into different systems and workflows, including input/output adaptation, fine-tuning procedures, and usability before, during, and after commissioning.\\
		\cellcolor{CoolGray!50}&\multicolumn{1}{>{\centering\arraybackslash\columncolor{Platinum!60}}m{\CCWidth}}{D. Specific Benchmarking}\\
		\cellcolor{CoolGray!50}&9. \textit{Module} \& \textit{Skill-Specific Benchmarking:} Separate evaluation of (a) independent core modules, learned skills, tool selection quality, etc. and (b) whole-system-level performance/ metrics.\\
		\cellcolor{CoolGray!50}&10. \textit{Safety, Security} \& \textit{Norm Compliance Validation:} Explicitly evaluates implemented authorisation, safety features, resilience against adversarial inputs, and system compliance with relevant industrial standards, rules, and domain-specific guidelines.\\
		\cellcolor{CoolGray!50}&11. \textit{Explainability, Reasoning} \& \textit{Feedback Benchmarks:} Evaluation of built-in explainability methods, overall reasoning transparency, feedback usage/ usability, and correction/ user guidance understanding – demonstrating black-box mitigation and feedback consistency.\\
		\cellcolor{CoolGray!50}&12. \textit{Data Requirement} \& \textit{Autonomous Runtime Assessment:} Verifies the model’s data requirements for pretraining, fine-tuning, and online adaptation. Benchmarks autonomous self-evaluation during runtime learning, adaptation, and self-improvement.\\
		\cellcolor{CoolGray!50}&13. \textit{Human Interaction} \& \textit{Object Recognition Evaluation:} Evaluates human-robot interaction and cooperation capabilities. Benchmarks HRC/HRI-related object/human recognition for known and unknown cases, considering detection rates, explanation quality, and adaptation of resulting behaviour.\\
		\cellcolor{CoolGray!50}&\cellcolor{CoolGray!50}\\
		\cellcolor{CoolGray!50}\textbf{I11:}&\multicolumn{1}{>{\centering\arraybackslash\columncolor{Platinum!60}}m{\CCWidth}}{A. Data Efficiency \& Finetuning}\\
		\cellcolor{CoolGray!50}\shortstack{\textbf{Data} \\ \textbf{Requirements}\\ \& \\ \textbf{Usage}}&1. \textit{Low-Data Adaptation} \& \textit{Few-Shot Requirements:} Needs only few or zero new examples for effective adaptation to novel tasks, environments, hardware, or operators, using suitable techniques (e.g. in-context learning, online self-adaptation, modular fine-tuning) with minimal resource utilisation.\\
		\cellcolor{CoolGray!50}&2. \textit{Efficient Task, Environment} \& \textit{Modality Adaptation:} Supports adaptation of observation/action spaces or input modalities without needing retraining or extensive new data and supports independent module adaptation to minimise both overall and domain-specific data requirements.\\
		\cellcolor{CoolGray!50}&\multicolumn{1}{>{\centering\arraybackslash\columncolor{Platinum!60}}m{\CCWidth}}{B. Pretraining Coverage \& Robustness}\\
		\cellcolor{CoolGray!50}&3. \textit{Comprehensive} \& \textit{Diverse Pretraining Dataset:} Pretraining covers broad task types, environments, domains, object/material variations, instruction modalities, distractors, noise, faults, adversarial attacks, redundant/fallback routines, and safety/data privacy scenarios, using both synthetic and real-world data for robust generalisation and OOD handling.\\
		\cellcolor{CoolGray!50}&4. \textit{Documented Data Composition} \& \textit{Simulation-Real Transfer:} Maintains accessible documentation on training/finetuning data composition, variation, and procedures. Including evaluation of transfer between simulation and real-world settings and tracks performance on transfer degradation.\\
		\cellcolor{CoolGray!50}&\multicolumn{1}{>{\centering\arraybackslash\columncolor{Platinum!60}}m{\CCWidth}}{C. Data Sufficiency, Self-Assessment \& Explainability}\\
		\cellcolor{CoolGray!50}&5. \textit{Automatic Data Sufficiency Assessment} \& \textit{Feedback:} Continuously analyses historic and incoming data for sufficiency, coverage gaps, redundancy, and quality issues during adaptation or operation. Provides alerts and actionable recommendations for missing, underrepresented, or low-quality data (e.g. based on task, scene understanding, …).\\
		\cellcolor{CoolGray!50}&6. \textit{Explainability on Data Usage} \& \textit{Improvement:} Supplies explainable analyses of data composition/ identified gaps, adaptation impact, improvement, and focus shift during training/finetuning. Recommends targeted data enrichment or refinement to address detected blind spots or weaknesses.\\
		\cellcolor{CoolGray!50}&\multicolumn{1}{>{\centering\arraybackslash\columncolor{Platinum!60}}m{\CCWidth}}{D. Data Flexibility, Interface \& Compliance}\\
		\cellcolor{CoolGray!50}&7. \textit{Flexible Data Type Acceptance:} Accepts and learns (on- and offline) from diverse data formats and modalities (e.g. trajectory, demonstration, egocentric, multi-modal sensory).\\
		\cellcolor{CoolGray!50}&8. \textit{Compliance with Privacy, Ethics} \& \textit{Law:} Ensures all data usage can comply with provided legal, ethical, and corporate policies – leveraging privacy safeguards (e.g. data anonymisation, operator-specific data isolation, encrypted logging, …) where needed.\\
		\cellcolor{CoolGray!50}&\multicolumn{1}{>{\centering\arraybackslash\columncolor{Platinum!60}}m{\CCWidth}}{E. Ongoing Monitoring, Confidence \& Robustness}\\
		\cellcolor{CoolGray!50}&9. \textit{Ongoing Adaptation/Finetuning Confidence Scores:} Outputs initial and ongoing confidence scores after (fine)tuning and during online adaptation, estimating both data sufficiency/appropriateness and the model's expected operational robustness and readiness under changing environments, hardware, or tasks.\\
		\cellcolor{CoolGray!50}&10. \textit{Resilience Through Redundancy} \& \textit{Fault Tolerance:} Incorporates redundancy, noise, faults, and fallback examples in pretraining and fine-tuning to support robust adaptation, autonomous countermeasure activation, and continued reliable operation under adverse data scenarios.\\
		\cellcolor{CoolGray!50}&11. \textit{Self-Assessment to Reduce Input} \& \textit{Modalities:} Evaluates and documents own data and input usage, minimising dependence on unnecessary modalities and optimising resource efficiency during all operational phases.\\
\end{longtable}

\normalsize
\twocolumn
\clearpage
\section{RFM-Overview}\label{app:rfms}
\onecolumn

\renewcommand{\arraystretch}{1.35}
\small

\begin{longtable}{@{}
		>{\centering\arraybackslash}m{0.9cm}
		>{\centering\arraybackslash}m{1.1cm}
		>{\centering\arraybackslash}m{1.1cm}
		*{11}{>{\centering\arraybackslash}m{0.62cm}}
		@{}}
	\caption{Overview of all $324$ identified RFMs that include manipulation capabilities (for control \& integrated; see Section~\ref{ssec:methods-crfm} for composition details).
		\textit{Reporting their proportional industrial applicability ratings for each individual implication, and a total score over all implications. Per-criterion ratings are created utilising the publication's context, by leveraging our LLM-based pipeline as described in Section~\ref{ssec:survey-proc}. Values are rounded to two decimal places.}}
	\label{tab:rfms-all}\\
	
	\rowcolor{CoolGray!50}
	\cellcolor{Platinum!60}\textbf{\rule{0pt}{2.5ex}Ref} &
	\textbf{\rule{0pt}{2.5ex}Model} &
	\textbf{\rule{0pt}{2.5ex}Total~$\downarrow$} &
	\textbf{\rule{0pt}{2.5ex}I1} &
	\textbf{\rule{0pt}{2.5ex}I2} &
	\textbf{\rule{0pt}{2.5ex}I3} &
	\textbf{\rule{0pt}{2.5ex}I4} &
	\textbf{\rule{0pt}{2.5ex}I5} &
	\textbf{\rule{0pt}{2.5ex}I6} &
	\textbf{\rule{0pt}{2.5ex}I7} &
	\textbf{\rule{0pt}{2.5ex}I8} &
	\textbf{\rule{0pt}{2.5ex}I9} &
	\textbf{\rule{0pt}{2.5ex}I10} &
	\textbf{\rule{0pt}{2.5ex}I11} \\
	\endfirsthead
	
	\rowcolor{CoolGray!50}
	\cellcolor{Platinum!60}\textbf{\rule{0pt}{2.5ex}Ref} &
	\textbf{\rule{0pt}{2.5ex}Model} &
	\textbf{\rule{0pt}{2.5ex}Total} &
	\textbf{\rule{0pt}{2.5ex}I1} &
	\textbf{\rule{0pt}{2.5ex}I2} &
	\textbf{\rule{0pt}{2.5ex}I3} &
	\textbf{\rule{0pt}{2.5ex}I4} &
	\textbf{\rule{0pt}{2.5ex}I5} &
	\textbf{\rule{0pt}{2.5ex}I6} &
	\textbf{\rule{0pt}{2.5ex}I7} &
	\textbf{\rule{0pt}{2.5ex}I8} &
	\textbf{\rule{0pt}{2.5ex}I9} &
	\textbf{\rule{0pt}{2.5ex}I10} &
	\textbf{\rule{0pt}{2.5ex}I11} \\
	\endhead
	
	\multicolumn{14}{r}{\footnotesize\textit{continued on next page}}\\
	\endfoot
	
	\endlastfoot

	\cellcolor{CoolGray!50}\cite{teamGeminiRobotics152025} & Gemini Robotics 1.5 & 0.12 & 0.44 & 0.06 & 0.07 & 0.07 & 0.00 & 0.00 & 0.00 & 0.07 & 0.14 & 0.23 & 0.09\\
	\cellcolor{CoolGray!50}\cite{sunCollabVLASelfReflectiveVisionLanguageAction2025} & Collab VLA & 0.11 & 0.28 & 0.00 & 0.20 & 0.07 & 0.08 & 0.11 & 0.00 & 0.07 & 0.21 & 0.08 & 0.09\\
	\cellcolor{CoolGray!50}\cite{liInformationTheoreticGraphFusion2025} & GF-VLA & 0.11 & 0.28 & 0.00 & 0.00 & 0.14 & 0.08 & 0.00 & 0.00 & 0.07 & 0.14 & 0.23 & 0.27\\
	\cellcolor{CoolGray!50}\cite{linOneTwoVLAUnifiedVisionLanguageAction2025} & OneTwo VLA & 0.11 & 0.33 & 0.00 & 0.20 & 0.14 & 0.00 & 0.11 & 0.10 & 0.07 & 0.14 & 0.00 & 0.09\\
	\cellcolor{CoolGray!50}\cite{liSelfCorrectingVisionLanguageActionModel2025} & SC-VLA & 0.11 & 0.33 & 0.06 & 0.00 & 0.21 & 0.23 & 0.11 & 0.00 & 0.00 & 0.00 & 0.15 & 0.09\\
	\cellcolor{CoolGray!50}\cite{wuHiBerNACHierarchicalBrainemulated2025} & HiBer NAC & 0.10 & 0.22 & 0.11 & 0.00 & 0.00 & 0.08 & 0.11 & 0.10 & 0.00 & 0.21 & 0.15 & 0.09\\
	\cellcolor{CoolGray!50}\cite{liuMultiTaskInteractiveRobot2024} & Sirius-Fleet & 0.10 & 0.39 & 0.22 & 0.00 & 0.14 & 0.00 & 0.00 & 0.00 & 0.00 & 0.00 & 0.08 & 0.09\\
	\cellcolor{CoolGray!50}\cite{huangTactileVLAUnlockingVisionLanguageAction2025} & Tactile-VLA & 0.10 & 0.28 & 0.00 & 0.00 & 0.07 & 0.15 & 0.00 & 0.00 & 0.14 & 0.07 & 0.15 & 0.18\\
	\cellcolor{CoolGray!50}\cite{zhongDexGraspVLAVisionLanguageActionFramework2025} & DexGrasp VLA & 0.09 & 0.28 & 0.00 & 0.00 & 0.07 & 0.08 & 0.00 & 0.00 & 0.07 & 0.00 & 0.31 & 0.18\\
	\cellcolor{CoolGray!50}\cite{teamGeminiRoboticsBringing2025} & Gemini Robotics & 0.09 & 0.22 & 0.00 & 0.00 & 0.07 & 0.00 & 0.00 & 0.00 & 0.14 & 0.14 & 0.23 & 0.18\\
	\cellcolor{CoolGray!50}\cite{fangRobixUnifiedModel2025} & Robix & 0.09 & 0.22 & 0.11 & 0.20 & 0.00 & 0.00 & 0.00 & 0.00 & 0.00 & 0.00 & 0.31 & 0.09\\
	\cellcolor{CoolGray!50}\cite{liuVITAENaturalEmbodied2025} & VITA-E & 0.09 & 0.28 & 0.11 & 0.20 & 0.00 & 0.00 & 0.11 & 0.00 & 0.07 & 0.00 & 0.08 & 0.09\\
	\cellcolor{CoolGray!50}\cite{driessPaLMEEmbodiedMultimodal2023} & PaLM-E & 0.09 & 0.33 & 0.00 & 0.00 & 0.29 & 0.00 & 0.00 & 0.00 & 0.00 & 0.07 & 0.00 & 0.18\\
	\cellcolor{CoolGray!50}\cite{daiRACERRichLanguageGuided2025} & RACER & 0.09 & 0.28 & 0.00 & 0.00 & 0.07 & 0.00 & 0.00 & 0.00 & 0.14 & 0.00 & 0.15 & 0.27\\
	\cellcolor{CoolGray!50}\cite{yuanEmbodiedR1ReinforcedEmbodied2025} & Embodied-R1 & 0.09 & 0.28 & 0.00 & 0.00 & 0.00 & 0.08 & 0.00 & 0.00 & 0.00 & 0.14 & 0.23 & 0.18\\
	\cellcolor{CoolGray!50}\cite{chenVLMimicVisionLanguage2024} & VLMimic & 0.08 & 0.33 & 0.00 & 0.00 & 0.14 & 0.00 & 0.00 & 0.00 & 0.07 & 0.00 & 0.15 & 0.09\\
	\cellcolor{CoolGray!50}\cite{chenInternVLAM1SpatiallyGuided2025} & Intern VLA-M1 & 0.08 & 0.28 & 0.00 & 0.07 & 0.07 & 0.00 & 0.00 & 0.00 & 0.00 & 0.07 & 0.15 & 0.18\\
	\cellcolor{CoolGray!50}\cite{yangFPCVLAVisionLanguageActionFramework2025} & FPC-VLA & 0.08 & 0.28 & 0.06 & 0.00 & 0.14 & 0.08 & 0.00 & 0.00 & 0.00 & 0.00 & 0.23 & 0.00\\
	\cellcolor{CoolGray!50}\cite{liHAMSTERHierarchicalAction2025} & HAMSTER & 0.07 & 0.11 & 0.00 & 0.00 & 0.00 & 0.00 & 0.00 & 0.10 & 0.00 & 0.07 & 0.23 & 0.36\\
	\cellcolor{CoolGray!50}\cite{liCoAVLAImprovingVisionLanguageAction2025} & CoA-VLA & 0.07 & 0.11 & 0.00 & 0.00 & 0.00 & 0.08 & 0.00 & 0.10 & 0.07 & 0.14 & 0.23 & 0.09\\
	\cellcolor{CoolGray!50}\cite{huangThinkActVisionLanguageActionReasoning2025} & ThinkAct & 0.07 & 0.28 & 0.00 & 0.00 & 0.14 & 0.00 & 0.00 & 0.00 & 0.00 & 0.00 & 0.15 & 0.18\\
	\cellcolor{CoolGray!50}\cite{leeMolmoActActionReasoning2025} & MolmoAct & 0.07 & 0.17 & 0.00 & 0.07 & 0.00 & 0.08 & 0.00 & 0.00 & 0.07 & 0.07 & 0.15 & 0.18\\
	\cellcolor{CoolGray!50}\cite{belkhaleRTHActionHierarchies2024} & RT-H & 0.07 & 0.22 & 0.00 & 0.13 & 0.00 & 0.08 & 0.00 & 0.00 & 0.07 & 0.07 & 0.00 & 0.18\\
	\cellcolor{CoolGray!50}\cite{kimOpenVLAOpenSourceVisionLanguageAction2024} & OpenVLA & 0.07 & 0.22 & 0.00 & 0.00 & 0.07 & 0.00 & 0.00 & 0.00 & 0.00 & 0.00 & 0.38 & 0.09\\
	\cellcolor{CoolGray!50}\cite{buSynergisticGeneralizedEfficient2025} & RoboDual & 0.07 & 0.17 & 0.00 & 0.00 & 0.00 & 0.00 & 0.11 & 0.00 & 0.00 & 0.07 & 0.23 & 0.27\\
	\cellcolor{CoolGray!50}\cite{huangVoxPoserComposable3D2023} & VoxPoser & 0.07 & 0.17 & 0.00 & 0.13 & 0.07 & 0.15 & 0.00 & 0.00 & 0.00 & 0.00 & 0.15 & 0.09\\
	\cellcolor{CoolGray!50}\cite{buUniVLALearningAct2025} & UniVLA & 0.07 & 0.11 & 0.00 & 0.00 & 0.00 & 0.08 & 0.00 & 0.00 & 0.07 & 0.00 & 0.23 & 0.36\\
	\cellcolor{CoolGray!50}\cite{buClosedLoopVisuomotorControl2024} & CLOVER & 0.07 & 0.28 & 0.00 & 0.00 & 0.07 & 0.00 & 0.00 & 0.00 & 0.07 & 0.00 & 0.23 & 0.00\\
	\cellcolor{CoolGray!50}\cite{chenNanoVLARoutingDecoupled2025} & NanoVLA & 0.07 & 0.11 & 0.00 & 0.00 & 0.07 & 0.08 & 0.00 & 0.20 & 0.00 & 0.00 & 0.31 & 0.00\\
	\cellcolor{CoolGray!50}\cite{kangCLIPRTLearningLanguageConditioned2025} & CLIP-RT & 0.07 & 0.17 & 0.00 & 0.13 & 0.07 & 0.00 & 0.00 & 0.00 & 0.07 & 0.00 & 0.08 & 0.18\\
	\cellcolor{CoolGray!50}\cite{lynchLanguageConditionedImitation2021} & LCIL & 0.07 & 0.11 & 0.00 & 0.27 & 0.00 & 0.00 & 0.11 & 0.00 & 0.00 & 0.00 & 0.08 & 0.18\\
	\cellcolor{CoolGray!50}\cite{liSwitchVLAExecutionAwareTask2025} & Switch VLA & 0.06 & 0.22 & 0.00 & 0.07 & 0.00 & 0.00 & 0.11 & 0.00 & 0.00 & 0.00 & 0.08 & 0.18\\
	\cellcolor{CoolGray!50}\cite{liUnifiedVideoAction2025} & UVA & 0.06 & 0.11 & 0.00 & 0.00 & 0.07 & 0.08 & 0.00 & 0.00 & 0.00 & 0.00 & 0.31 & 0.09\\
	\cellcolor{CoolGray!50}\cite{liuIDAGCAdaptiveGeneralized2025} & IDAGC & 0.06 & 0.11 & 0.00 & 0.13 & 0.00 & 0.00 & 0.00 & 0.00 & 0.00 & 0.14 & 0.08 & 0.18\\
	\cellcolor{CoolGray!50}\cite{xuEfficientAlignmentUnconditioned2025} & A2 & 0.06 & 0.06 & 0.00 & 0.00 & 0.00 & 0.15 & 0.00 & 0.00 & 0.07 & 0.07 & 0.15 & 0.18\\
	\cellcolor{CoolGray!50}\cite{blackZeroShotRoboticManipulation2023} & SuSIE & 0.06 & 0.11 & 0.00 & 0.00 & 0.00 & 0.08 & 0.00 & 0.00 & 0.00 & 0.00 & 0.23 & 0.27\\
	\cellcolor{CoolGray!50}\cite{panOmniManipGeneralRobotic2025} & Omni Manip & 0.06 & 0.28 & 0.00 & 0.00 & 0.07 & 0.00 & 0.00 & 0.00 & 0.00 & 0.00 & 0.15 & 0.09\\
	\cellcolor{CoolGray!50}\cite{liuFoAMForesightAugmentedMultiTask2025} & FoAM & 0.06 & 0.17 & 0.00 & 0.00 & 0.07 & 0.08 & 0.00 & 0.00 & 0.00 & 0.00 & 0.08 & 0.27\\
	\cellcolor{CoolGray!50}\cite{zhangVTLAVisionTactileLanguageActionModel2025} & VTLA & 0.06 & 0.06 & 0.00 & 0.00 & 0.14 & 0.08 & 0.00 & 0.00 & 0.00 & 0.07 & 0.15 & 0.18\\
	\cellcolor{CoolGray!50}\cite{zeGNFactorMultiTaskReal2024} & GNFactor & 0.06 & 0.17 & 0.00 & 0.00 & 0.00 & 0.00 & 0.00 & 0.00 & 0.07 & 0.07 & 0.15 & 0.18\\
	\cellcolor{CoolGray!50}\cite{wenDiffusionVLAGeneralizableInterpretable2025} & Diffusion-VLA & 0.06 & 0.22 & 0.00 & 0.00 & 0.07 & 0.00 & 0.00 & 0.00 & 0.07 & 0.07 & 0.00 & 0.18\\
	\cellcolor{CoolGray!50}\cite{liGRMGLeveragingPartially2024} & GR-MG & 0.06 & 0.22 & 0.00 & 0.00 & 0.00 & 0.00 & 0.00 & 0.00 & 0.00 & 0.00 & 0.23 & 0.18\\
	\cellcolor{CoolGray!50}\cite{nasirianyRTAffordanceAffordancesAre2024} & RT-Affordance & 0.06 & 0.17 & 0.00 & 0.00 & 0.00 & 0.00 & 0.00 & 0.00 & 0.07 & 0.07 & 0.08 & 0.27\\
	\cellcolor{CoolGray!50}\cite{yangInstructVLAVisionLanguageActionInstruction2025} & Instruct VLA & 0.06 & 0.17 & 0.00 & 0.00 & 0.00 & 0.00 & 0.00 & 0.00 & 0.07 & 0.07 & 0.23 & 0.09\\
	\cellcolor{CoolGray!50}\cite{bharadhwajRoboAgentGeneralizationEfficiency2024} & Robo Agent & 0.05 & 0.17 & 0.00 & 0.00 & 0.14 & 0.00 & 0.00 & 0.00 & 0.00 & 0.00 & 0.15 & 0.09\\
	\cellcolor{CoolGray!50}\cite{liPointVLAInjecting3D2025} & PointVLA & 0.05 & 0.06 & 0.00 & 0.00 & 0.00 & 0.08 & 0.00 & 0.00 & 0.00 & 0.00 & 0.23 & 0.27\\
	\cellcolor{CoolGray!50}\cite{quEO1InterleavedVisionTextAction2025} & EO-1 & 0.05 & 0.11 & 0.00 & 0.00 & 0.00 & 0.00 & 0.11 & 0.00 & 0.00 & 0.07 & 0.15 & 0.18\\
	\cellcolor{CoolGray!50}\cite{tangIncentivizingMultimodalReasoning2025} & Reason Manip & 0.05 & 0.22 & 0.00 & 0.00 & 0.00 & 0.00 & 0.00 & 0.00 & 0.00 & 0.14 & 0.08 & 0.09\\
	\cellcolor{CoolGray!50}\cite{zhaiIgnitingVLMsEmbodied2025} & WALL-OSS & 0.05 & 0.17 & 0.00 & 0.07 & 0.00 & 0.00 & 0.11 & 0.00 & 0.00 & 0.00 & 0.08 & 0.18\\
	\cellcolor{CoolGray!50}\cite{fengVidarEmbodiedVideo2025} & Vidar & 0.05 & 0.11 & 0.00 & 0.00 & 0.00 & 0.00 & 0.00 & 0.00 & 0.07 & 0.00 & 0.15 & 0.27\\
	\cellcolor{CoolGray!50}\cite{liCronusVLAEfficientRobust2025} & Cronus VLA & 0.05 & 0.06 & 0.00 & 0.00 & 0.14 & 0.00 & 0.00 & 0.00 & 0.07 & 0.00 & 0.23 & 0.09\\
	\cellcolor{CoolGray!50}\cite{singhOGVLAOrthographicImage2025} & OG-VLA & 0.05 & 0.11 & 0.00 & 0.00 & 0.07 & 0.00 & 0.00 & 0.00 & 0.00 & 0.00 & 0.31 & 0.09\\
	\cellcolor{CoolGray!50}\cite{huangOTTERVisionLanguageActionModel2025} & OTTER & 0.05 & 0.17 & 0.00 & 0.00 & 0.00 & 0.00 & 0.00 & 0.00 & 0.07 & 0.00 & 0.23 & 0.09\\
	\cellcolor{CoolGray!50}\cite{teamGigaBrain0WorldModelPowered2025} & Giga Brain-0 & 0.05 & 0.06 & 0.00 & 0.00 & 0.00 & 0.00 & 0.00 & 0.10 & 0.00 & 0.14 & 0.08 & 0.27\\
	\cellcolor{CoolGray!50}\cite{sunEmmaXEmbodiedMultimodal2024} & Emma-X & 0.05 & 0.11 & 0.00 & 0.00 & 0.00 & 0.00 & 0.00 & 0.00 & 0.07 & 0.07 & 0.23 & 0.09\\
	\cellcolor{CoolGray!50}\cite{xueLeVERBHumanoidWholeBody2025} & LeVERB & 0.05 & 0.17 & 0.00 & 0.00 & 0.00 & 0.00 & 0.00 & 0.00 & 0.00 & 0.00 & 0.23 & 0.18\\
	\cellcolor{CoolGray!50}\cite{yanManiFlowGeneralRobot2025} & ManiFlow & 0.05 & 0.11 & 0.00 & 0.00 & 0.07 & 0.00 & 0.00 & 0.00 & 0.00 & 0.00 & 0.23 & 0.18\\
	\cellcolor{CoolGray!50}\cite{chiDiffusionPolicyVisuomotor2023} & Diffusion Policy & 0.05 & 0.11 & 0.06 & 0.00 & 0.14 & 0.08 & 0.11 & 0.00 & 0.00 & 0.00 & 0.08 & 0.00\\
	\cellcolor{CoolGray!50}\cite{stoneOpenWorldObjectManipulation2023} & MOO & 0.05 & 0.22 & 0.00 & 0.00 & 0.00 & 0.00 & 0.00 & 0.00 & 0.00 & 0.07 & 0.00 & 0.27\\
	\cellcolor{CoolGray!50}\cite{liuMLAMultisensoryLanguageAction2025} & MLA & 0.05 & 0.11 & 0.00 & 0.00 & 0.00 & 0.00 & 0.00 & 0.00 & 0.07 & 0.14 & 0.15 & 0.09\\
	\cellcolor{CoolGray!50}\cite{garciaGeneralizableVisionLanguageRobotic2025} & GemBench & 0.05 & 0.17 & 0.00 & 0.00 & 0.00 & 0.00 & 0.00 & 0.00 & 0.07 & 0.00 & 0.31 & 0.00\\
	\cellcolor{CoolGray!50}\cite{guhurInstructiondrivenHistoryawarePolicies2022} & IDHAP & 0.05 & 0.17 & 0.00 & 0.00 & 0.07 & 0.00 & 0.00 & 0.00 & 0.07 & 0.00 & 0.08 & 0.18\\
	\cellcolor{CoolGray!50}\cite{liBridgeVLAInputOutputAlignment2025} & Bridge VLA & 0.05 & 0.17 & 0.00 & 0.00 & 0.00 & 0.08 & 0.00 & 0.00 & 0.00 & 0.00 & 0.23 & 0.09\\
	\cellcolor{CoolGray!50}\cite{wangProgrammaticallyGroundedCompositionally2023} & PRO-GRAM PORT & 0.05 & 0.11 & 0.00 & 0.00 & 0.00 & 0.00 & 0.00 & 0.00 & 0.07 & 0.07 & 0.15 & 0.18\\
	\cellcolor{CoolGray!50}\cite{kuangRAMRetrievalBasedAffordance2024} & RAM & 0.05 & 0.22 & 0.00 & 0.00 & 0.00 & 0.08 & 0.00 & 0.00 & 0.00 & 0.00 & 0.08 & 0.18\\
	\cellcolor{CoolGray!50}\cite{luoVisualEmbodiedBrain2025} & VeBrain & 0.05 & 0.28 & 0.00 & 0.00 & 0.00 & 0.08 & 0.00 & 0.00 & 0.00 & 0.00 & 0.08 & 0.09\\
	\cellcolor{CoolGray!50}\cite{zhengUniversalActionsEnhanced2025} & UniAct & 0.05 & 0.11 & 0.00 & 0.00 & 0.00 & 0.08 & 0.00 & 0.00 & 0.00 & 0.00 & 0.23 & 0.18\\
	\cellcolor{CoolGray!50}\cite{maActraOptimizedTransformer2024} & Actra & 0.05 & 0.17 & 0.00 & 0.00 & 0.07 & 0.00 & 0.00 & 0.00 & 0.07 & 0.00 & 0.08 & 0.09\\
	\cellcolor{CoolGray!50}\cite{goyalRVT2LearningPrecise2024} & RVT-2 & 0.05 & 0.17 & 0.00 & 0.00 & 0.07 & 0.00 & 0.00 & 0.00 & 0.00 & 0.00 & 0.15 & 0.09\\
	\cellcolor{CoolGray!50}\cite{niuTimeUnifiedDiffusionPolicy2025} & TUDP & 0.05 & 0.00 & 0.00 & 0.00 & 0.00 & 0.08 & 0.00 & 0.10 & 0.00 & 0.07 & 0.15 & 0.18\\
	\cellcolor{CoolGray!50}\cite{zhangUniCoDEnhancingRobot2025} & UniCoD & 0.05 & 0.17 & 0.00 & 0.00 & 0.00 & 0.08 & 0.00 & 0.00 & 0.00 & 0.00 & 0.15 & 0.09\\
	\cellcolor{CoolGray!50}\cite{guRTTrajectoryRoboticTask2023} & RT-Trajectory & 0.05 & 0.11 & 0.00 & 0.00 & 0.00 & 0.00 & 0.00 & 0.00 & 0.07 & 0.00 & 0.15 & 0.18\\
	\cellcolor{CoolGray!50}\cite{zhangInSpireVisionLanguageActionModels2025} & InSpire & 0.05 & 0.22 & 0.00 & 0.00 & 0.00 & 0.00 & 0.00 & 0.00 & 0.07 & 0.00 & 0.15 & 0.00\\
	\cellcolor{CoolGray!50}\cite{baiLearningSeeAct2025} & TVVE & 0.05 & 0.11 & 0.00 & 0.00 & 0.00 & 0.08 & 0.00 & 0.00 & 0.00 & 0.00 & 0.31 & 0.00\\
	\cellcolor{CoolGray!50}\cite{wangScalingProprioceptiveVisualLearning2024} & HPT & 0.05 & 0.06 & 0.00 & 0.00 & 0.07 & 0.00 & 0.00 & 0.00 & 0.00 & 0.00 & 0.15 & 0.27\\
	\cellcolor{CoolGray!50}\cite{songHumeIntroducingSystem22025} & Hume & 0.05 & 0.22 & 0.00 & 0.00 & 0.07 & 0.00 & 0.00 & 0.00 & 0.00 & 0.00 & 0.15 & 0.00\\
	\cellcolor{CoolGray!50}\cite{stepputtisLanguageConditionedImitationLearning2020} & LCIL2 & 0.05 & 0.11 & 0.00 & 0.00 & 0.07 & 0.00 & 0.00 & 0.00 & 0.07 & 0.07 & 0.08 & 0.09\\
	\cellcolor{CoolGray!50}\cite{yeLatentActionPretraining2025} & LAP-VLA & 0.05 & 0.06 & 0.00 & 0.00 & 0.00 & 0.00 & 0.00 & 0.00 & 0.00 & 0.00 & 0.23 & 0.27\\
	\cellcolor{CoolGray!50}\cite{brohanRT1RoboticsTransformer2023} & RT-1 & 0.05 & 0.17 & 0.00 & 0.00 & 0.00 & 0.00 & 0.00 & 0.00 & 0.07 & 0.00 & 0.15 & 0.09\\
	\cellcolor{CoolGray!50}\cite{jonesSightFinetuningGeneralist2025} & FuSe & 0.05 & 0.06 & 0.00 & 0.00 & 0.00 & 0.00 & 0.00 & 0.00 & 0.00 & 0.21 & 0.00 & 0.27\\
	\cellcolor{CoolGray!50}\cite{fangSAM2ActIntegratingVisual2025} & SAM2 Act & 0.05 & 0.11 & 0.00 & 0.00 & 0.00 & 0.08 & 0.00 & 0.00 & 0.00 & 0.00 & 0.31 & 0.00\\
	\cellcolor{CoolGray!50}\cite{huVideoPredictionPolicy2025} & Video Prediction Policy & 0.05 & 0.06 & 0.00 & 0.00 & 0.00 & 0.00 & 0.00 & 0.00 & 0.00 & 0.07 & 0.15 & 0.27\\
	\cellcolor{CoolGray!50}\cite{liLanguageGuidedObjectCentricDiffusion2025} & Lan-o3dp & 0.05 & 0.17 & 0.06 & 0.00 & 0.07 & 0.00 & 0.00 & 0.00 & 0.00 & 0.00 & 0.08 & 0.09\\
	\cellcolor{CoolGray!50}\cite{fanInterleaveVLAEnhancingRobot2025} & Interleave VLA & 0.05 & 0.17 & 0.00 & 0.00 & 0.00 & 0.00 & 0.00 & 0.00 & 0.07 & 0.00 & 0.08 & 0.18\\
	\cellcolor{CoolGray!50}\cite{teamCarefulExaminationLarge2025} & LBMfDM & 0.05 & 0.06 & 0.00 & 0.00 & 0.00 & 0.00 & 0.00 & 0.00 & 0.00 & 0.00 & 0.23 & 0.27\\
	\cellcolor{CoolGray!50}\cite{saxenaMResTMultiResolutionSensing2024} & MResT & 0.05 & 0.11 & 0.00 & 0.00 & 0.00 & 0.08 & 0.11 & 0.00 & 0.00 & 0.07 & 0.08 & 0.09\\
	\cellcolor{CoolGray!50}\cite{huangGraphCoTVLA3DSpatialAware2025} & GraphCoT VLA & 0.05 & 0.06 & 0.00 & 0.00 & 0.00 & 0.00 & 0.00 & 0.00 & 0.21 & 0.14 & 0.00 & 0.09\\
	\cellcolor{CoolGray!50}\cite{niuPretrainingAutoregressiveRobotic2025} & ARM4R & 0.05 & 0.17 & 0.00 & 0.00 & 0.07 & 0.08 & 0.00 & 0.00 & 0.00 & 0.00 & 0.08 & 0.09\\
	\cellcolor{CoolGray!50}\cite{xuSPECISkillPrompts2025} & SPECI & 0.04 & 0.11 & 0.00 & 0.00 & 0.00 & 0.00 & 0.00 & 0.00 & 0.00 & 0.00 & 0.15 & 0.18\\
	\cellcolor{CoolGray!50}\cite{reedGeneralistAgent2022} & Gato & 0.04 & 0.00 & 0.00 & 0.00 & 0.00 & 0.00 & 0.00 & 0.00 & 0.07 & 0.00 & 0.15 & 0.27\\
	\cellcolor{CoolGray!50}\cite{linEvo0VisionLanguageActionModel2025} & Evo-0 & 0.04 & 0.11 & 0.00 & 0.00 & 0.07 & 0.08 & 0.00 & 0.00 & 0.00 & 0.00 & 0.15 & 0.00\\
	\cellcolor{CoolGray!50}\cite{huangMoTVLAVisionLanguageActionModel2025} & MoT VLA & 0.04 & 0.11 & 0.00 & 0.00 & 0.07 & 0.00 & 0.00 & 0.00 & 0.00 & 0.00 & 0.15 & 0.09\\
	\cellcolor{CoolGray!50}\cite{zhangGEVRMGoalExpressiveVideo2025} & GEVRM & 0.04 & 0.11 & 0.00 & 0.00 & 0.07 & 0.00 & 0.00 & 0.00 & 0.00 & 0.00 & 0.15 & 0.09\\
	\cellcolor{CoolGray!50}\cite{wenTinyVLAFastDataEfficient2025} & Tiny VLA & 0.04 & 0.17 & 0.00 & 0.00 & 0.07 & 0.00 & 0.00 & 0.00 & 0.00 & 0.00 & 0.15 & 0.00\\
	\cellcolor{CoolGray!50}\cite{yangLoHoVLAUnifiedVisionLanguageAction2025} & LoHo VLA & 0.04 & 0.22 & 0.00 & 0.00 & 0.07 & 0.00 & 0.00 & 0.00 & 0.00 & 0.00 & 0.08 & 0.00\\
	\cellcolor{CoolGray!50}\cite{shiMemoryVLAPerceptualCognitiveMemory2025} & Memory VLA & 0.04 & 0.17 & 0.00 & 0.00 & 0.00 & 0.00 & 0.00 & 0.00 & 0.00 & 0.00 & 0.23 & 0.00\\
	\cellcolor{CoolGray!50}\cite{liuRDT1BDiffusionFoundation2025} & RDT-1B & 0.04 & 0.11 & 0.00 & 0.00 & 0.00 & 0.08 & 0.11 & 0.00 & 0.00 & 0.00 & 0.00 & 0.18\\
	\cellcolor{CoolGray!50}\cite{nvidiaGR00TN1Open2025} & GR00T N1 & 0.04 & 0.11 & 0.00 & 0.00 & 0.00 & 0.00 & 0.00 & 0.00 & 0.00 & 0.00 & 0.08 & 0.27\\
	\cellcolor{CoolGray!50}\cite{wuUnleashingLargeScaleVideo2023} & GR-1 & 0.04 & 0.17 & 0.00 & 0.00 & 0.07 & 0.00 & 0.00 & 0.00 & 0.00 & 0.00 & 0.00 & 0.18\\
	\cellcolor{CoolGray!50}\cite{jiangVIMAGeneralRobot2023} & VIMA & 0.04 & 0.17 & 0.00 & 0.00 & 0.07 & 0.00 & 0.00 & 0.00 & 0.00 & 0.00 & 0.00 & 0.18\\
	\cellcolor{CoolGray!50}\cite{jangBCZZeroShotTask2022} & BC-Z & 0.04 & 0.11 & 0.00 & 0.07 & 0.00 & 0.00 & 0.00 & 0.00 & 0.00 & 0.00 & 0.08 & 0.18\\
	\cellcolor{CoolGray!50}\cite{chenIntentionVLAGeneralizableEfficient2025} & Intention VLA & 0.04 & 0.17 & 0.00 & 0.00 & 0.00 & 0.00 & 0.11 & 0.00 & 0.00 & 0.07 & 0.08 & 0.00\\
	\cellcolor{CoolGray!50}\cite{shridharCLIPortWhatWhere2021} & CLIPort & 0.04 & 0.11 & 0.00 & 0.00 & 0.00 & 0.00 & 0.00 & 0.00 & 0.00 & 0.00 & 0.23 & 0.09\\
	\cellcolor{CoolGray!50}\cite{chenFastinSlowDualSystemFoundation2025} & Fast-in-Slow & 0.04 & 0.17 & 0.00 & 0.00 & 0.00 & 0.08 & 0.00 & 0.00 & 0.00 & 0.00 & 0.08 & 0.09\\
	\cellcolor{CoolGray!50}\cite{xiaCAGECausalAttention2024} & CAGE & 0.04 & 0.11 & 0.00 & 0.00 & 0.00 & 0.00 & 0.00 & 0.00 & 0.00 & 0.00 & 0.23 & 0.09\\
	\cellcolor{CoolGray!50}\cite{wangSKILSemanticKeypoint2025} & SKIL & 0.04 & 0.06 & 0.00 & 0.00 & 0.00 & 0.08 & 0.00 & 0.00 & 0.00 & 0.00 & 0.15 & 0.18\\
	\cellcolor{CoolGray!50}\cite{yangARVRMImitatingHuman2025} & AR-VRM & 0.04 & 0.11 & 0.00 & 0.00 & 0.00 & 0.00 & 0.00 & 0.00 & 0.07 & 0.00 & 0.08 & 0.18\\
	\cellcolor{CoolGray!50}\cite{sunGeoVLAEmpowering3D2025} & GeoVLA & 0.04 & 0.06 & 0.00 & 0.00 & 0.00 & 0.00 & 0.00 & 0.00 & 0.00 & 0.00 & 0.31 & 0.09\\
	\cellcolor{CoolGray!50}\cite{liangSkillDiffuserInterpretableHierarchical2024} & Skill Diffuser & 0.04 & 0.11 & 0.00 & 0.00 & 0.00 & 0.00 & 0.00 & 0.00 & 0.07 & 0.07 & 0.15 & 0.00\\
	\cellcolor{CoolGray!50}\cite{chenGoalVLAImageGenerativeVLMs2025} & Goal-VLA & 0.04 & 0.11 & 0.00 & 0.00 & 0.00 & 0.08 & 0.00 & 0.00 & 0.00 & 0.00 & 0.15 & 0.09\\
	\cellcolor{CoolGray!50}\cite{linVOTEVisionLanguageActionOptimization2025} & VOTE & 0.04 & 0.11 & 0.00 & 0.00 & 0.00 & 0.08 & 0.00 & 0.10 & 0.00 & 0.00 & 0.15 & 0.00\\
	\cellcolor{CoolGray!50}\cite{wuAffordDPGeneralizableDiffusion2025} & AffordDP & 0.04 & 0.11 & 0.00 & 0.00 & 0.00 & 0.00 & 0.00 & 0.00 & 0.00 & 0.00 & 0.08 & 0.27\\
	\cellcolor{CoolGray!50}\cite{ranasinghePixelMotionUniversal2025} & LangToMo & 0.04 & 0.06 & 0.00 & 0.00 & 0.00 & 0.00 & 0.00 & 0.00 & 0.07 & 0.07 & 0.15 & 0.09\\
	\cellcolor{CoolGray!50}\cite{cheangGR2GenerativeVideoLanguageAction2024} & GR-2 & 0.04 & 0.17 & 0.00 & 0.00 & 0.00 & 0.00 & 0.00 & 0.00 & 0.00 & 0.00 & 0.15 & 0.09\\
	\cellcolor{CoolGray!50}\cite{aminP06VLAThat2025} & $\pi_{0*.6}$ & 0.04 & 0.00 & 0.00 & 0.00 & 0.07 & 0.15 & 0.00 & 0.00 & 0.07 & 0.00 & 0.08 & 0.09\\
	\cellcolor{CoolGray!50}\cite{reussFLOWERDemocratizingGeneralist2025} & FLOWER & 0.04 & 0.22 & 0.00 & 0.00 & 0.00 & 0.00 & 0.00 & 0.00 & 0.00 & 0.00 & 0.15 & 0.00\\
	\cellcolor{CoolGray!50}\cite{chiMinDLearningDualSystem2025} & MinD & 0.04 & 0.00 & 0.00 & 0.00 & 0.14 & 0.00 & 0.00 & 0.00 & 0.07 & 0.07 & 0.15 & 0.00\\
	\cellcolor{CoolGray!50}\cite{chenMotoLatentMotion2025} & Moto & 0.04 & 0.11 & 0.00 & 0.00 & 0.00 & 0.00 & 0.00 & 0.00 & 0.07 & 0.00 & 0.08 & 0.18\\
	\cellcolor{CoolGray!50}\cite{mandiCACTIFrameworkScalable2023} & CACTI & 0.04 & 0.22 & 0.00 & 0.00 & 0.00 & 0.00 & 0.00 & 0.00 & 0.00 & 0.00 & 0.08 & 0.09\\
	\cellcolor{CoolGray!50}\cite{kangTaskContextAwareDiffusionPolicy2025} & TCADP-LG & 0.04 & 0.06 & 0.00 & 0.00 & 0.00 & 0.08 & 0.00 & 0.10 & 0.00 & 0.07 & 0.08 & 0.09\\
	\cellcolor{CoolGray!50}\cite{shentuLLMsActionsLatent2025} & LCB & 0.04 & 0.11 & 0.00 & 0.07 & 0.00 & 0.00 & 0.00 & 0.10 & 0.07 & 0.07 & 0.00 & 0.00\\
	\cellcolor{CoolGray!50}\cite{shukorSmolVLAVisionLanguageActionModel2025} & Smol VLA & 0.04 & 0.06 & 0.00 & 0.00 & 0.07 & 0.00 & 0.00 & 0.20 & 0.00 & 0.00 & 0.15 & 0.00\\
	\cellcolor{CoolGray!50}\cite{argusCVLAEfficientCameraSpace2025} & cVLA & 0.03 & 0.06 & 0.00 & 0.00 & 0.00 & 0.00 & 0.00 & 0.00 & 0.00 & 0.00 & 0.08 & 0.27\\
	\cellcolor{CoolGray!50}\cite{cheangGR3TechnicalReport2025} & GR-3 & 0.03 & 0.11 & 0.00 & 0.00 & 0.00 & 0.00 & 0.00 & 0.00 & 0.00 & 0.00 & 0.08 & 0.18\\
	\cellcolor{CoolGray!50}\cite{xiongHyperVLAEfficientInference2025} & Hyper VLA & 0.03 & 0.06 & 0.00 & 0.00 & 0.00 & 0.08 & 0.00 & 0.00 & 0.00 & 0.00 & 0.15 & 0.09\\
	\cellcolor{CoolGray!50}\cite{chenEmbodimentEquivariantVisionLanguageAction2025} & E2VLA & 0.03 & 0.06 & 0.00 & 0.00 & 0.07 & 0.08 & 0.00 & 0.00 & 0.07 & 0.00 & 0.00 & 0.09\\
	\cellcolor{CoolGray!50}\cite{jiRoboBrainUnifiedBrain2025} & Robo Brain & 0.03 & 0.06 & 0.00 & 0.00 & 0.00 & 0.00 & 0.00 & 0.00 & 0.14 & 0.00 & 0.08 & 0.09\\
	\cellcolor{CoolGray!50}\cite{maoOmniDGeneralizableRobot2025} & OmniD & 0.03 & 0.00 & 0.00 & 0.00 & 0.00 & 0.08 & 0.00 & 0.00 & 0.00 & 0.00 & 0.23 & 0.09\\
	\cellcolor{CoolGray!50}\cite{yuForceVLAEnhancingVLA2025} & Force VLA & 0.03 & 0.11 & 0.00 & 0.00 & 0.14 & 0.00 & 0.00 & 0.00 & 0.00 & 0.00 & 0.00 & 0.09\\
	\cellcolor{CoolGray!50}\cite{chenVisionLanguageModelPredictive2025} & VLMPC & 0.03 & 0.11 & 0.00 & 0.00 & 0.00 & 0.08 & 0.00 & 0.00 & 0.00 & 0.00 & 0.15 & 0.00\\
	\cellcolor{CoolGray!50}\cite{collinsAMPLIFYActionlessMotion2025} & AMPLIFY & 0.03 & 0.06 & 0.00 & 0.00 & 0.00 & 0.00 & 0.00 & 0.00 & 0.00 & 0.00 & 0.15 & 0.18\\
	\cellcolor{CoolGray!50}\cite{voReFineVLAReasoningAwareTeacherGuided2025} & ReFine VLA & 0.03 & 0.11 & 0.00 & 0.00 & 0.00 & 0.00 & 0.00 & 0.00 & 0.14 & 0.07 & 0.00 & 0.00\\
	\cellcolor{CoolGray!50}\cite{liCogACTFoundationalVisionLanguageAction2024} & CogACT & 0.03 & 0.11 & 0.00 & 0.00 & 0.00 & 0.00 & 0.00 & 0.00 & 0.00 & 0.00 & 0.23 & 0.00\\
	\cellcolor{CoolGray!50}\cite{zhangGRAPEGeneralizingRobot2025} & GRAPE & 0.03 & 0.11 & 0.00 & 0.00 & 0.00 & 0.08 & 0.00 & 0.00 & 0.00 & 0.00 & 0.15 & 0.00\\
	\cellcolor{CoolGray!50}\cite{yanDNActDiffusionGuided2024} & DNAct & 0.03 & 0.11 & 0.00 & 0.00 & 0.00 & 0.00 & 0.00 & 0.00 & 0.00 & 0.07 & 0.15 & 0.00\\
	\cellcolor{CoolGray!50}\cite{yangMagmaFoundationModel2025} & Magma & 0.03 & 0.11 & 0.00 & 0.00 & 0.00 & 0.00 & 0.00 & 0.00 & 0.00 & 0.00 & 0.00 & 0.27\\
	\cellcolor{CoolGray!50}\cite{caiGROOT2WeaklySupervised2024} & GROOT-2 & 0.03 & 0.11 & 0.00 & 0.00 & 0.00 & 0.00 & 0.00 & 0.00 & 0.07 & 0.00 & 0.08 & 0.09\\
	\cellcolor{CoolGray!50}\cite{liuTriVLATripleSystemBasedUnified2025} & TriVLA & 0.03 & 0.11 & 0.00 & 0.00 & 0.00 & 0.00 & 0.00 & 0.00 & 0.00 & 0.00 & 0.15 & 0.09\\
	\cellcolor{CoolGray!50}\cite{wangThisThatLanguageGestureControlled2025} & This\&That & 0.03 & 0.06 & 0.00 & 0.00 & 0.00 & 0.00 & 0.00 & 0.00 & 0.00 & 0.07 & 0.15 & 0.09\\
	\cellcolor{CoolGray!50}\cite{huTARADTaskAwareRobot2025} & TARAD & 0.03 & 0.17 & 0.00 & 0.00 & 0.00 & 0.00 & 0.00 & 0.00 & 0.00 & 0.00 & 0.00 & 0.18\\
	\cellcolor{CoolGray!50}\cite{yuUsingBothDemonstrations2023} & DeL-TaCo & 0.03 & 0.06 & 0.00 & 0.00 & 0.00 & 0.00 & 0.00 & 0.00 & 0.00 & 0.00 & 0.15 & 0.18\\
	\cellcolor{CoolGray!50}\cite{shaoConcept2RobotLearningManipulation2020} & Concept2 Robot & 0.03 & 0.06 & 0.00 & 0.00 & 0.00 & 0.08 & 0.00 & 0.00 & 0.00 & 0.00 & 0.15 & 0.09\\
	\cellcolor{CoolGray!50}\cite{jiaPointMapPolicyStructuredPoint2025} & Point Map Policy & 0.03 & 0.00 & 0.00 & 0.00 & 0.00 & 0.00 & 0.00 & 0.00 & 0.07 & 0.07 & 0.15 & 0.09\\
	\cellcolor{CoolGray!50}\cite{dalalLocalPoliciesEnable2025} & LP-ZS-LH & 0.03 & 0.17 & 0.00 & 0.00 & 0.00 & 0.00 & 0.00 & 0.00 & 0.00 & 0.00 & 0.08 & 0.09\\
	\cellcolor{CoolGray!50}\cite{chenVidBotLearningGeneralizable2025} & VidBot & 0.03 & 0.06 & 0.06 & 0.00 & 0.00 & 0.00 & 0.11 & 0.00 & 0.00 & 0.00 & 0.08 & 0.09\\
	\cellcolor{CoolGray!50}\cite{ahnCanNotSay2022} & SayCan & 0.03 & 0.11 & 0.00 & 0.00 & 0.00 & 0.00 & 0.00 & 0.00 & 0.07 & 0.07 & 0.08 & 0.00\\
	\cellcolor{CoolGray!50}\cite{chenACDiTAdaptiveCoordination2025} & AC-DiT & 0.03 & 0.00 & 0.00 & 0.00 & 0.00 & 0.00 & 0.00 & 0.00 & 0.07 & 0.07 & 0.15 & 0.09\\
	\cellcolor{CoolGray!50}\cite{zhaoVLASVisionLanguageActionModel2025} & VLAS & 0.03 & 0.06 & 0.00 & 0.00 & 0.00 & 0.00 & 0.00 & 0.00 & 0.00 & 0.07 & 0.15 & 0.09\\
	\cellcolor{CoolGray!50}\cite{myersPolicyAdaptationLanguage2024} & PALO & 0.03 & 0.11 & 0.00 & 0.00 & 0.00 & 0.00 & 0.00 & 0.00 & 0.00 & 0.00 & 0.15 & 0.09\\
	\cellcolor{CoolGray!50}\cite{zhengTraceVLAVisualTrace2025} & Trace VLA & 0.03 & 0.17 & 0.00 & 0.00 & 0.00 & 0.00 & 0.00 & 0.00 & 0.00 & 0.00 & 0.15 & 0.00\\
	\cellcolor{CoolGray!50}\cite{chenVillaXEnhancingLatent2025} & Villa-X & 0.03 & 0.00 & 0.00 & 0.00 & 0.00 & 0.00 & 0.00 & 0.00 & 0.00 & 0.07 & 0.23 & 0.09\\
	\cellcolor{CoolGray!50}\cite{wangSparseDiffusionPolicy2024} & Sparse Diffusion Policy & 0.03 & 0.11 & 0.00 & 0.00 & 0.00 & 0.00 & 0.00 & 0.00 & 0.07 & 0.07 & 0.00 & 0.09\\
	\cellcolor{CoolGray!50}\cite{spiridonovGeneralistRobotManipulation2025} & GRM-BALD & 0.03 & 0.06 & 0.00 & 0.00 & 0.00 & 0.00 & 0.00 & 0.00 & 0.00 & 0.00 & 0.15 & 0.18\\
	\cellcolor{CoolGray!50}\cite{bharadhwajGen2ActHumanVideo2024} & Gen2Act & 0.03 & 0.11 & 0.00 & 0.00 & 0.00 & 0.00 & 0.00 & 0.00 & 0.07 & 0.00 & 0.00 & 0.18\\
	\cellcolor{CoolGray!50}\cite{yanRoboTronManiAllinOneMultimodal2025} & Robo Tron-Mani & 0.03 & 0.06 & 0.00 & 0.00 & 0.00 & 0.00 & 0.00 & 0.00 & 0.00 & 0.07 & 0.15 & 0.09\\
	\cellcolor{CoolGray!50}\cite{zhangSAMELeveragingVisual2024} & SAM-E & 0.03 & 0.06 & 0.00 & 0.00 & 0.00 & 0.00 & 0.00 & 0.00 & 0.07 & 0.07 & 0.08 & 0.09\\
	\cellcolor{CoolGray!50}\cite{liuHybridVLACollaborativeDiffusion2025} & Hybrid VLA & 0.03 & 0.11 & 0.06 & 0.00 & 0.00 & 0.00 & 0.00 & 0.00 & 0.00 & 0.00 & 0.15 & 0.00\\
	\cellcolor{CoolGray!50}\cite{buckerLATTELAnguageTrajectory2023} & LATTE & 0.03 & 0.11 & 0.00 & 0.00 & 0.00 & 0.08 & 0.00 & 0.00 & 0.00 & 0.00 & 0.08 & 0.09\\
	\cellcolor{CoolGray!50}\cite{yangCoMoLearningContinuous2025} & CoMo & 0.03 & 0.06 & 0.00 & 0.00 & 0.00 & 0.00 & 0.00 & 0.00 & 0.00 & 0.00 & 0.15 & 0.18\\
	\cellcolor{CoolGray!50}\cite{zhangHiRTEnhancingRobotic2025} & HiRT & 0.03 & 0.17 & 0.00 & 0.00 & 0.00 & 0.00 & 0.00 & 0.00 & 0.00 & 0.00 & 0.15 & 0.00\\
	\cellcolor{CoolGray!50}\cite{zhengFLARERobotLearning2025} & FLARE & 0.03 & 0.06 & 0.00 & 0.00 & 0.00 & 0.00 & 0.00 & 0.00 & 0.00 & 0.07 & 0.00 & 0.18\\
	\cellcolor{CoolGray!50}\cite{tianPredictiveInverseDynamics2024} & Seer & 0.03 & 0.11 & 0.00 & 0.00 & 0.00 & 0.00 & 0.00 & 0.00 & 0.00 & 0.00 & 0.08 & 0.09\\
	\cellcolor{CoolGray!50}\cite{wenDVLADiffusionVisionLanguageAction2025} & dVLA & 0.03 & 0.06 & 0.00 & 0.00 & 0.00 & 0.00 & 0.00 & 0.00 & 0.07 & 0.07 & 0.00 & 0.09\\
	\cellcolor{CoolGray!50}\cite{yangTransferringFoundationModels2025} & TPM & 0.03 & 0.06 & 0.00 & 0.00 & 0.07 & 0.00 & 0.00 & 0.00 & 0.00 & 0.00 & 0.08 & 0.09\\
	\cellcolor{CoolGray!50}\cite{houDitaScalingDiffusion2025} & Dita & 0.03 & 0.11 & 0.00 & 0.00 & 0.00 & 0.00 & 0.00 & 0.00 & 0.00 & 0.00 & 0.08 & 0.09\\
	\cellcolor{CoolGray!50}\cite{yaoThinkSmallAct2025} & Think Small-PPL & 0.03 & 0.06 & 0.00 & 0.00 & 0.00 & 0.00 & 0.00 & 0.00 & 0.00 & 0.07 & 0.08 & 0.09\\
	\cellcolor{CoolGray!50}\cite{karamchetiLanguageDrivenRepresentationLearning2023} & LDRL & 0.03 & 0.11 & 0.00 & 0.00 & 0.07 & 0.00 & 0.00 & 0.00 & 0.00 & 0.00 & 0.08 & 0.00\\
	\cellcolor{CoolGray!50}\cite{jiangRynnVLA001UsingHuman2025} & Rynn VLA-001 & 0.03 & 0.06 & 0.00 & 0.00 & 0.00 & 0.00 & 0.00 & 0.10 & 0.00 & 0.00 & 0.08 & 0.09\\
	\cellcolor{CoolGray!50}\cite{lanBFABestFeatureAwareFusion2025} & BFA & 0.03 & 0.00 & 0.00 & 0.00 & 0.00 & 0.00 & 0.00 & 0.10 & 0.07 & 0.07 & 0.08 & 0.00\\
	\cellcolor{CoolGray!50}\cite{reussMultimodalDiffusionTransformer2024} & MM Diff Tran & 0.03 & 0.06 & 0.00 & 0.00 & 0.00 & 0.00 & 0.00 & 0.00 & 0.00 & 0.00 & 0.08 & 0.18\\
	\cellcolor{CoolGray!50}\cite{mengPreservingCombiningKnowledge2024} & Robo LL-RL & 0.03 & 0.06 & 0.00 & 0.00 & 0.00 & 0.08 & 0.00 & 0.00 & 0.00 & 0.00 & 0.15 & 0.00\\
	\cellcolor{CoolGray!50}\cite{koLEARNINGACTACTIONLESS2024} & LAA VTDC & 0.03 & 0.11 & 0.00 & 0.00 & 0.00 & 0.00 & 0.00 & 0.00 & 0.00 & 0.00 & 0.08 & 0.09\\
	\cellcolor{CoolGray!50}\cite{shafiullahBringingRobotsHome2023} & Dobb·E & 0.03 & 0.00 & 0.00 & 0.00 & 0.00 & 0.00 & 0.00 & 0.10 & 0.00 & 0.00 & 0.15 & 0.09\\
	\cellcolor{CoolGray!50}\cite{liVisionLanguageFoundationModels2024} & Robo Flamingo & 0.03 & 0.06 & 0.00 & 0.00 & 0.00 & 0.00 & 0.00 & 0.10 & 0.00 & 0.00 & 0.08 & 0.09\\
	\cellcolor{CoolGray!50}\cite{quSpatialVLAExploringSpatial2025} & Spatial VLA & 0.03 & 0.11 & 0.00 & 0.00 & 0.07 & 0.00 & 0.00 & 0.00 & 0.00 & 0.00 & 0.08 & 0.00\\
	\cellcolor{CoolGray!50}\cite{black$p_05$VisionLanguageActionModel2025} & $\pi_{0.5}$ & 0.03 & 0.06 & 0.00 & 0.00 & 0.00 & 0.00 & 0.00 & 0.00 & 0.00 & 0.00 & 0.15 & 0.09\\
	\cellcolor{CoolGray!50}\cite{dengGraspVLAGraspingFoundation2025} & Grasp VLA & 0.03 & 0.06 & 0.00 & 0.00 & 0.00 & 0.00 & 0.00 & 0.00 & 0.00 & 0.00 & 0.08 & 0.18\\
	\cellcolor{CoolGray!50}\cite{yangFP33DFoundation2025} & FP3 & 0.03 & 0.11 & 0.00 & 0.00 & 0.07 & 0.00 & 0.00 & 0.00 & 0.00 & 0.00 & 0.00 & 0.09\\
	\cellcolor{CoolGray!50}\cite{liTaskReconstructionExtrapolation2025} & $\pi_{0}$-Text Latent & 0.03 & 0.06 & 0.00 & 0.00 & 0.00 & 0.00 & 0.00 & 0.00 & 0.07 & 0.00 & 0.15 & 0.00\\
	\cellcolor{CoolGray!50}\cite{fuICRTInContextImitation2025} & ICRT & 0.03 & 0.11 & 0.00 & 0.00 & 0.00 & 0.00 & 0.00 & 0.00 & 0.00 & 0.00 & 0.00 & 0.18\\
	\cellcolor{CoolGray!50}\cite{bhat3DCAVLALeveraging2025} & 3D-CAVLA & 0.03 & 0.06 & 0.00 & 0.00 & 0.00 & 0.00 & 0.00 & 0.00 & 0.00 & 0.07 & 0.08 & 0.09\\
	\cellcolor{CoolGray!50}\cite{zhaoUnveilingPotentialVisionLanguageAction2025} & OE-VLA & 0.03 & 0.11 & 0.00 & 0.00 & 0.00 & 0.00 & 0.00 & 0.00 & 0.00 & 0.07 & 0.00 & 0.09\\
	\cellcolor{CoolGray!50}\cite{huangEmbodiedGeneralistAgent2024} & LEO & 0.03 & 0.06 & 0.00 & 0.00 & 0.00 & 0.00 & 0.00 & 0.00 & 0.00 & 0.07 & 0.08 & 0.09\\
	\cellcolor{CoolGray!50}\cite{liCogVLACognitionAlignedVisionLanguageAction2025} & Cog VLA & 0.03 & 0.06 & 0.00 & 0.00 & 0.00 & 0.08 & 0.00 & 0.00 & 0.07 & 0.00 & 0.08 & 0.00\\
	\cellcolor{CoolGray!50}\cite{songReconVLAReconstructiveVisionLanguageAction2025} & Recon VLA & 0.03 & 0.11 & 0.00 & 0.00 & 0.00 & 0.00 & 0.00 & 0.00 & 0.07 & 0.00 & 0.08 & 0.00\\
	\cellcolor{CoolGray!50}\cite{radosavovicRealWorldRobotLearning2022} & Real World-RL-MVP & 0.03 & 0.06 & 0.00 & 0.00 & 0.00 & 0.00 & 0.00 & 0.00 & 0.00 & 0.00 & 0.08 & 0.18\\
	\cellcolor{CoolGray!50}\cite{wenDexVLAVisionLanguageModel2025} & DexVLA & 0.03 & 0.06 & 0.00 & 0.00 & 0.00 & 0.00 & 0.00 & 0.00 & 0.00 & 0.00 & 0.15 & 0.09\\
	\cellcolor{CoolGray!50}\cite{meesGroundingLanguageVisual2023} & GLVA & 0.03 & 0.11 & 0.00 & 0.00 & 0.00 & 0.00 & 0.00 & 0.00 & 0.00 & 0.00 & 0.15 & 0.00\\
	\cellcolor{CoolGray!50}\cite{ze3DDiffusionPolicy2024} & 3D Diffusion Policy & 0.03 & 0.11 & 0.00 & 0.00 & 0.00 & 0.00 & 0.00 & 0.00 & 0.00 & 0.00 & 0.08 & 0.09\\
	\cellcolor{CoolGray!50}\cite{luoBeingH0VisionLanguageActionPretraining2025} & Being-H0 & 0.03 & 0.06 & 0.00 & 0.00 & 0.00 & 0.00 & 0.00 & 0.00 & 0.00 & 0.00 & 0.08 & 0.18\\
	\cellcolor{CoolGray!50}\cite{fangAirExo2ScalingGeneralizable2025} & AirExo-2 & 0.03 & 0.00 & 0.00 & 0.00 & 0.00 & 0.00 & 0.00 & 0.10 & 0.00 & 0.07 & 0.15 & 0.00\\
	\cellcolor{CoolGray!50}\cite{brohanRT2VisionLanguageActionModels2023} & RT-2 & 0.03 & 0.06 & 0.00 & 0.00 & 0.00 & 0.00 & 0.00 & 0.00 & 0.00 & 0.00 & 0.15 & 0.09\\
	\cellcolor{CoolGray!50}\cite{shridharPerceiverActorMultiTaskTransformer2022} & Perceiver-Actor & 0.03 & 0.06 & 0.00 & 0.00 & 0.00 & 0.00 & 0.00 & 0.00 & 0.00 & 0.00 & 0.08 & 0.18\\
	\cellcolor{CoolGray!50}\cite{biHRDTHumanManipulation2025} & H-RDT & 0.03 & 0.06 & 0.00 & 0.00 & 0.00 & 0.00 & 0.00 & 0.00 & 0.00 & 0.00 & 0.08 & 0.18\\
	\cellcolor{CoolGray!50}\cite{yangTraMoELearningTrajectory2025} & Tra-MoE & 0.03 & 0.00 & 0.00 & 0.00 & 0.00 & 0.08 & 0.00 & 0.00 & 0.00 & 0.00 & 0.15 & 0.09\\
	\cellcolor{CoolGray!50}\cite{zhen3DVLA3DVisionLanguageAction2024} & 3D-VLA & 0.03 & 0.06 & 0.00 & 0.00 & 0.00 & 0.00 & 0.00 & 0.00 & 0.00 & 0.07 & 0.08 & 0.09\\
	\cellcolor{CoolGray!50}\cite{itoIntegratedLearningRobot2022} & ILRMS & 0.03 & 0.00 & 0.00 & 0.00 & 0.07 & 0.00 & 0.00 & 0.00 & 0.07 & 0.00 & 0.00 & 0.18\\
	\cellcolor{CoolGray!50}\cite{wangVLAAdapterEffectiveParadigm2025} & VLA-Adapter & 0.03 & 0.06 & 0.00 & 0.00 & 0.00 & 0.00 & 0.00 & 0.00 & 0.00 & 0.00 & 0.23 & 0.00\\
	\cellcolor{CoolGray!50}\cite{meesWhatMattersLanguage2022} & Lang Cond-RoboIL & 0.03 & 0.06 & 0.00 & 0.00 & 0.00 & 0.00 & 0.00 & 0.00 & 0.00 & 0.00 & 0.15 & 0.09\\
	\cellcolor{CoolGray!50}\cite{kobayashiBiVLABilateralControlBased2025} & Bi-VLA & 0.03 & 0.11 & 0.00 & 0.00 & 0.00 & 0.00 & 0.00 & 0.00 & 0.00 & 0.00 & 0.00 & 0.18\\
	\cellcolor{CoolGray!50}\cite{lyuDyWADynamicsadaptiveWorld2025} & DyWA & 0.03 & 0.06 & 0.00 & 0.00 & 0.00 & 0.08 & 0.00 & 0.00 & 0.00 & 0.00 & 0.15 & 0.00\\
	\cellcolor{CoolGray!50}\cite{zhangDreamVLAVisionLanguageActionModel2025} & Dream VLA & 0.03 & 0.06 & 0.00 & 0.00 & 0.00 & 0.00 & 0.00 & 0.00 & 0.00 & 0.07 & 0.08 & 0.09\\
	\cellcolor{CoolGray!50}\cite{houDiffusionTransformerPolicy2025} & Diffusion Transformer Policy & 0.03 & 0.11 & 0.00 & 0.00 & 0.00 & 0.00 & 0.00 & 0.00 & 0.07 & 0.00 & 0.00 & 0.09\\
	\cellcolor{CoolGray!50}\cite{black$p_0$VisionLanguageActionFlow2024} & $\pi_{0}$ & 0.02 & 0.06 & 0.00 & 0.00 & 0.00 & 0.00 & 0.00 & 0.00 & 0.00 & 0.00 & 0.08 & 0.09\\
	\cellcolor{CoolGray!50}\cite{staroverovFineTuningMultimodalTransformer2023} & Rozum Former & 0.02 & 0.00 & 0.00 & 0.00 & 0.00 & 0.00 & 0.00 & 0.00 & 0.00 & 0.00 & 0.00 & 0.27\\
	\cellcolor{CoolGray!50}\cite{bousmalisRoboCatSelfImprovingGeneralist2023} & RoboCat & 0.02 & 0.00 & 0.00 & 0.00 & 0.00 & 0.00 & 0.00 & 0.00 & 0.00 & 0.00 & 0.15 & 0.09\\
	\cellcolor{CoolGray!50}\cite{chenIGORImageGOalRepresentations2024} & IGOR & 0.02 & 0.06 & 0.00 & 0.00 & 0.00 & 0.00 & 0.00 & 0.00 & 0.00 & 0.00 & 0.00 & 0.18\\
	\cellcolor{CoolGray!50}\cite{zhaoLearningFineGrainedBimanual2023} & ACT & 0.02 & 0.00 & 0.00 & 0.00 & 0.00 & 0.00 & 0.00 & 0.10 & 0.00 & 0.00 & 0.08 & 0.09\\
	\cellcolor{CoolGray!50}\cite{dongImitDiffTransferringFoundationModel2025} & ImitDiff & 0.02 & 0.11 & 0.00 & 0.00 & 0.00 & 0.00 & 0.00 & 0.00 & 0.00 & 0.00 & 0.08 & 0.00\\
	\cellcolor{CoolGray!50}\cite{renSurferProgressiveReasoning2024} & Surfer & 0.02 & 0.11 & 0.00 & 0.00 & 0.00 & 0.00 & 0.00 & 0.00 & 0.00 & 0.00 & 0.08 & 0.00\\
	\cellcolor{CoolGray!50}\cite{mazzagliaHybridTrainingVisionLanguageAction2025} & HyT-VLA & 0.02 & 0.00 & 0.00 & 0.00 & 0.00 & 0.00 & 0.00 & 0.00 & 0.00 & 0.07 & 0.15 & 0.00\\
	\cellcolor{CoolGray!50}\cite{kimFineTuningVisionLanguageActionModels2025} & Open VLA OFT & 0.02 & 0.00 & 0.00 & 0.00 & 0.00 & 0.00 & 0.00 & 0.00 & 0.00 & 0.00 & 0.15 & 0.09\\
	\cellcolor{CoolGray!50}\cite{huangVLMTDPVLMguidedTrajectoryconditioned2025} & VLM-TDP & 0.02 & 0.06 & 0.00 & 0.00 & 0.00 & 0.00 & 0.00 & 0.00 & 0.00 & 0.00 & 0.15 & 0.00\\
	\cellcolor{CoolGray!50}\cite{myersGoalRepresentationsInstruction2023} & GRIF & 0.02 & 0.00 & 0.00 & 0.00 & 0.00 & 0.00 & 0.00 & 0.00 & 0.00 & 0.00 & 0.08 & 0.18\\
	\cellcolor{CoolGray!50}\cite{goyalRVTRoboticView2023} & RVT & 0.02 & 0.00 & 0.00 & 0.00 & 0.00 & 0.08 & 0.00 & 0.00 & 0.00 & 0.00 & 0.15 & 0.00\\
	\cellcolor{CoolGray!50}\cite{liuInstructionFollowingAgentsMultimodal2023} & Instruct RL & 0.02 & 0.06 & 0.00 & 0.00 & 0.00 & 0.00 & 0.00 & 0.00 & 0.00 & 0.00 & 0.00 & 0.18\\
	\cellcolor{CoolGray!50}\cite{maCDPRobustAutoregressive2025} & CDP & 0.02 & 0.06 & 0.00 & 0.00 & 0.00 & 0.00 & 0.00 & 0.00 & 0.00 & 0.00 & 0.08 & 0.09\\
	\cellcolor{CoolGray!50}\cite{fanDiffusionTrajectoryGuidedPolicy2025} & DTP & 0.02 & 0.06 & 0.00 & 0.00 & 0.00 & 0.00 & 0.00 & 0.00 & 0.00 & 0.00 & 0.08 & 0.09\\
	\cellcolor{CoolGray!50}\cite{zhongFlowVLAVisualChain2025} & Flow VLA & 0.02 & 0.06 & 0.00 & 0.00 & 0.00 & 0.00 & 0.00 & 0.00 & 0.00 & 0.00 & 0.08 & 0.09\\
	\cellcolor{CoolGray!50}\cite{tanThinkTwiceAct2025} & FLASH VLA & 0.02 & 0.00 & 0.00 & 0.00 & 0.00 & 0.00 & 0.00 & 0.10 & 0.00 & 0.00 & 0.08 & 0.09\\
	\cellcolor{CoolGray!50}\cite{liRoboMUTUALRoboticMultimodal2024} & Robo-MUTUAL & 0.02 & 0.00 & 0.00 & 0.00 & 0.00 & 0.00 & 0.00 & 0.00 & 0.00 & 0.00 & 0.15 & 0.09\\
	\cellcolor{CoolGray!50}\cite{gaoFLIPFlowCentricGenerative2025} & FLIP & 0.02 & 0.06 & 0.00 & 0.00 & 0.00 & 0.00 & 0.00 & 0.00 & 0.07 & 0.00 & 0.08 & 0.00\\
	\cellcolor{CoolGray!50}\cite{dengLearningLanguageConditionedDeformable2024} & LCDOM & 0.02 & 0.06 & 0.00 & 0.00 & 0.00 & 0.00 & 0.00 & 0.00 & 0.00 & 0.00 & 0.00 & 0.18\\
	\cellcolor{CoolGray!50}\cite{songAcceleratingVisionLanguageActionModel2025} & PD-VLA & 0.02 & 0.06 & 0.00 & 0.00 & 0.07 & 0.00 & 0.00 & 0.00 & 0.00 & 0.00 & 0.08 & 0.00\\
	\cellcolor{CoolGray!50}\cite{yangPushingLimitsCrossEmbodiment2024} & Cross EmbL & 0.02 & 0.06 & 0.00 & 0.00 & 0.00 & 0.00 & 0.00 & 0.00 & 0.00 & 0.00 & 0.08 & 0.09\\
	\cellcolor{CoolGray!50}\cite{doshiScalingCrossEmbodiedLearning2024} & Cross Former & 0.02 & 0.11 & 0.00 & 0.00 & 0.00 & 0.00 & 0.00 & 0.00 & 0.00 & 0.00 & 0.00 & 0.09\\
	\cellcolor{CoolGray!50}\cite{zhangReinboTAmplifyingRobot2025} & ReinboT & 0.02 & 0.00 & 0.00 & 0.00 & 0.00 & 0.00 & 0.00 & 0.00 & 0.00 & 0.00 & 0.08 & 0.18\\
	\cellcolor{CoolGray!50}\cite{shenExpertiseNeedNot2025} & AdaMoE & 0.02 & 0.06 & 0.00 & 0.00 & 0.00 & 0.00 & 0.00 & 0.00 & 0.00 & 0.00 & 0.15 & 0.00\\
	\cellcolor{CoolGray!50}\cite{griebenowOpenVocabularyRoboticObject2025} & OWL×T5 & 0.02 & 0.06 & 0.00 & 0.00 & 0.00 & 0.00 & 0.00 & 0.00 & 0.00 & 0.00 & 0.00 & 0.18\\
	\cellcolor{CoolGray!50}\cite{maContrastiveImitationLearning2024} & $\sum$-agent & 0.02 & 0.06 & 0.00 & 0.00 & 0.00 & 0.00 & 0.00 & 0.00 & 0.07 & 0.00 & 0.08 & 0.00\\
	\cellcolor{CoolGray!50}\cite{renSurferWorldModelBased2025} & Surfer & 0.02 & 0.11 & 0.00 & 0.00 & 0.00 & 0.00 & 0.00 & 0.00 & 0.00 & 0.00 & 0.08 & 0.00\\
	\cellcolor{CoolGray!50}\cite{heLearningActionableDiscrete2024} & ADDP-AVP & 0.02 & 0.00 & 0.00 & 0.00 & 0.00 & 0.00 & 0.00 & 0.00 & 0.00 & 0.00 & 0.08 & 0.18\\
	\cellcolor{CoolGray!50}\cite{wenObjectCentricInstructionAugmentation2024} & OCI & 0.02 & 0.00 & 0.00 & 0.00 & 0.00 & 0.00 & 0.00 & 0.00 & 0.00 & 0.00 & 0.15 & 0.09\\
	\cellcolor{CoolGray!50}\cite{thomasPLEXMakingMost2023} & PLEX & 0.02 & 0.06 & 0.00 & 0.00 & 0.00 & 0.00 & 0.00 & 0.00 & 0.00 & 0.00 & 0.00 & 0.18\\
	\cellcolor{CoolGray!50}\cite{huangEnerVerseEnvisioningEmbodied2025} & EnerVerse & 0.02 & 0.00 & 0.00 & 0.00 & 0.00 & 0.00 & 0.00 & 0.00 & 0.07 & 0.00 & 0.08 & 0.09\\
	\cellcolor{CoolGray!50}\cite{zhangAutoregressiveActionSequence2025} & ARP & 0.02 & 0.06 & 0.00 & 0.00 & 0.00 & 0.00 & 0.00 & 0.00 & 0.07 & 0.00 & 0.08 & 0.00\\
	\cellcolor{CoolGray!50}\cite{huangRoboGroundRoboticManipulation2025} & Robo Ground & 0.02 & 0.06 & 0.00 & 0.00 & 0.00 & 0.00 & 0.00 & 0.00 & 0.00 & 0.07 & 0.08 & 0.00\\
	\cellcolor{CoolGray!50}\cite{leiDynamicMixtureProgressive2025} & DMPEL & 0.02 & 0.06 & 0.00 & 0.00 & 0.00 & 0.00 & 0.00 & 0.00 & 0.00 & 0.00 & 0.08 & 0.09\\
	\cellcolor{CoolGray!50}\cite{zhuObjectVLAEndtoEndOpenWorld2025} & Object VLA & 0.02 & 0.00 & 0.00 & 0.00 & 0.00 & 0.00 & 0.00 & 0.00 & 0.00 & 0.00 & 0.08 & 0.18\\
	\cellcolor{CoolGray!50}\cite{liuRoboUniViewVisualLanguageModel2024} & Robo Uni View & 0.02 & 0.06 & 0.00 & 0.00 & 0.00 & 0.08 & 0.00 & 0.00 & 0.00 & 0.00 & 0.08 & 0.00\\
	\cellcolor{CoolGray!50}\cite{liangDiscreteDiffusionVLA2025} & Discrete Diffusion VLA & 0.02 & 0.11 & 0.00 & 0.00 & 0.00 & 0.00 & 0.00 & 0.00 & 0.00 & 0.00 & 0.08 & 0.00\\
	\cellcolor{CoolGray!50}\cite{wenROSAHarnessingRobot2025} & ROSA & 0.02 & 0.11 & 0.00 & 0.00 & 0.00 & 0.00 & 0.00 & 0.00 & 0.00 & 0.00 & 0.08 & 0.00\\
	\cellcolor{CoolGray!50}\cite{zhaoCoTVLAVisualChainofThought2025} & CoT-VLA & 0.02 & 0.00 & 0.00 & 0.00 & 0.00 & 0.00 & 0.00 & 0.00 & 0.00 & 0.07 & 0.08 & 0.09\\
	\cellcolor{CoolGray!50}\cite{tianPDFactorLearningTriPerspective2025} & PDFactor & 0.02 & 0.00 & 0.00 & 0.00 & 0.00 & 0.08 & 0.00 & 0.00 & 0.00 & 0.00 & 0.08 & 0.09\\
	\cellcolor{CoolGray!50}\cite{liuVoxActBVoxelBasedActing2024} & VoxAct-B & 0.02 & 0.06 & 0.00 & 0.00 & 0.00 & 0.00 & 0.00 & 0.10 & 0.00 & 0.00 & 0.08 & 0.00\\
	\cellcolor{CoolGray!50}\cite{fanLongVLAUnleashingLongHorizon2025} & Long-VLA & 0.02 & 0.11 & 0.00 & 0.00 & 0.00 & 0.00 & 0.00 & 0.00 & 0.00 & 0.00 & 0.00 & 0.09\\
	\cellcolor{CoolGray!50}\cite{ghoshOctoOpenSourceGeneralist2024} & Octo & 0.02 & 0.06 & 0.00 & 0.00 & 0.00 & 0.00 & 0.00 & 0.00 & 0.00 & 0.00 & 0.00 & 0.18\\
	\cellcolor{CoolGray!50}\cite{caesarEnablingActionCrossmodality2024} & CrossT5 & 0.01 & 0.00 & 0.00 & 0.00 & 0.00 & 0.00 & 0.00 & 0.00 & 0.00 & 0.00 & 0.00 & 0.18\\
	\cellcolor{CoolGray!50}\cite{gePolicyAdaptationFoundation2023} & PAFMF & 0.01 & 0.00 & 0.00 & 0.00 & 0.00 & 0.00 & 0.00 & 0.00 & 0.00 & 0.00 & 0.15 & 0.00\\
	\cellcolor{CoolGray!50}\cite{dingHumanoidVLAUniversalHumanoid2025} & Humanoid VLA & 0.01 & 0.06 & 0.00 & 0.00 & 0.00 & 0.00 & 0.00 & 0.00 & 0.00 & 0.00 & 0.00 & 0.09\\
	\cellcolor{CoolGray!50}\cite{wangRoboBERTEndtoendMultimodal2025} & Robo BERT & 0.01 & 0.06 & 0.00 & 0.00 & 0.00 & 0.00 & 0.00 & 0.00 & 0.00 & 0.00 & 0.08 & 0.00\\
	\cellcolor{CoolGray!50}\cite{zhuEfficientContinualAdaptation2025} & OMLA & 0.01 & 0.00 & 0.00 & 0.00 & 0.00 & 0.00 & 0.00 & 0.00 & 0.00 & 0.00 & 0.08 & 0.09\\
	\cellcolor{CoolGray!50}\cite{liangDreamitateRealWorldVisuomotor2024} & Dreamitate & 0.01 & 0.06 & 0.00 & 0.00 & 0.00 & 0.00 & 0.00 & 0.00 & 0.00 & 0.07 & 0.00 & 0.00\\
	\cellcolor{CoolGray!50}\cite{yangEgoVLALearningVisionLanguageAction2025} & EgoVLA & 0.01 & 0.06 & 0.00 & 0.00 & 0.00 & 0.00 & 0.00 & 0.00 & 0.00 & 0.00 & 0.08 & 0.00\\
	\cellcolor{CoolGray!50}\cite{silvaLanConLearnLearningLanguage2022} & LanCon-Learn & 0.01 & 0.06 & 0.00 & 0.00 & 0.00 & 0.00 & 0.00 & 0.00 & 0.00 & 0.00 & 0.00 & 0.09\\
	\cellcolor{CoolGray!50}\cite{jamesCoarsetoFineQattentionEfficient2022} & CtF Q-attention & 0.01 & 0.00 & 0.00 & 0.00 & 0.00 & 0.00 & 0.00 & 0.00 & 0.00 & 0.00 & 0.08 & 0.09\\
	\cellcolor{CoolGray!50}\cite{nguyenPixelMotionDiffusion2025} & DAWN & 0.01 & 0.00 & 0.00 & 0.00 & 0.00 & 0.00 & 0.00 & 0.00 & 0.00 & 0.07 & 0.08 & 0.00\\
	\cellcolor{CoolGray!50}\cite{liuStaMoUnsupervisedLearning2025} & StaMo & 0.01 & 0.00 & 0.00 & 0.00 & 0.00 & 0.00 & 0.00 & 0.00 & 0.00 & 0.00 & 0.08 & 0.09\\
	\cellcolor{CoolGray!50}\cite{daviesTenmaRobustCrossEmbodiment2025} & Tenma & 0.01 & 0.06 & 0.00 & 0.00 & 0.00 & 0.00 & 0.00 & 0.00 & 0.00 & 0.00 & 0.00 & 0.09\\
	\cellcolor{CoolGray!50}\cite{wangVQVLAImprovingVisionLanguageAction2025} & VQ-VLA & 0.01 & 0.00 & 0.00 & 0.00 & 0.00 & 0.00 & 0.00 & 0.00 & 0.00 & 0.00 & 0.08 & 0.09\\
	\cellcolor{CoolGray!50}\cite{luResearchTaskDecomposition2024} & VLA-Arm-TDMTO & 0.01 & 0.00 & 0.00 & 0.00 & 0.00 & 0.00 & 0.00 & 0.00 & 0.00 & 0.00 & 0.15 & 0.00\\
	\cellcolor{CoolGray!50}\cite{yangVlaserVisionLanguageActionModel2025} & Vlaser & 0.01 & 0.00 & 0.00 & 0.00 & 0.00 & 0.00 & 0.00 & 0.00 & 0.00 & 0.00 & 0.08 & 0.09\\
	\cellcolor{CoolGray!50}\cite{weiImitationAllYou2023} & DualMind & 0.01 & 0.00 & 0.00 & 0.00 & 0.00 & 0.00 & 0.00 & 0.00 & 0.07 & 0.00 & 0.00 & 0.09\\
	\cellcolor{CoolGray!50}\cite{royM2DistillMultiModalDistillation2025} & M2Distill & 0.01 & 0.00 & 0.00 & 0.00 & 0.00 & 0.00 & 0.00 & 0.00 & 0.00 & 0.00 & 0.08 & 0.09\\
	\cellcolor{CoolGray!50}\cite{wenVidManExploitingImplicit2024} & VidMan & 0.01 & 0.06 & 0.00 & 0.00 & 0.00 & 0.00 & 0.00 & 0.00 & 0.00 & 0.00 & 0.08 & 0.00\\
	\cellcolor{CoolGray!50}\cite{hanDualProcessVLA2024} & DP-VLA & 0.01 & 0.00 & 0.00 & 0.00 & 0.00 & 0.00 & 0.11 & 0.00 & 0.00 & 0.00 & 0.08 & 0.00\\
	\cellcolor{CoolGray!50}\cite{lvF1VisionLanguageActionModel2025} & F1 & 0.01 & 0.00 & 0.00 & 0.00 & 0.00 & 0.00 & 0.00 & 0.00 & 0.07 & 0.00 & 0.08 & 0.00\\
	\cellcolor{CoolGray!50}\cite{kooHAMLETSwitchYour2025} & HAMLET & 0.01 & 0.00 & 0.00 & 0.00 & 0.00 & 0.00 & 0.00 & 0.00 & 0.07 & 0.00 & 0.08 & 0.00\\
	\cellcolor{CoolGray!50}\cite{zhang4DVLASpatiotemporalVisionLanguageAction2025} & 4D-VLA & 0.01 & 0.06 & 0.00 & 0.00 & 0.00 & 0.00 & 0.00 & 0.00 & 0.00 & 0.00 & 0.08 & 0.00\\
	\cellcolor{CoolGray!50}\cite{gervetAct3D3DFeature2023} & Act3D & 0.01 & 0.06 & 0.00 & 0.00 & 0.00 & 0.00 & 0.00 & 0.00 & 0.00 & 0.00 & 0.00 & 0.09\\
	\cellcolor{CoolGray!50}\cite{zhangUPVLAUnifiedUnderstanding2025} & UP-VLA & 0.01 & 0.06 & 0.00 & 0.00 & 0.00 & 0.00 & 0.00 & 0.00 & 0.00 & 0.00 & 0.08 & 0.00\\
	\cellcolor{CoolGray!50}\cite{zhuoABMAttentionManipulation2024} & ABM & 0.01 & 0.06 & 0.00 & 0.00 & 0.00 & 0.00 & 0.00 & 0.00 & 0.00 & 0.00 & 0.00 & 0.09\\
	\cellcolor{CoolGray!50}\cite{cenWorldVLAAutoregressiveAction2025} & World VLA & 0.01 & 0.00 & 0.00 & 0.00 & 0.00 & 0.00 & 0.00 & 0.00 & 0.00 & 0.00 & 0.08 & 0.09\\
	\cellcolor{CoolGray!50}\cite{wangUnifiedVisionLanguageActionModel2025} & UniVLA & 0.01 & 0.00 & 0.00 & 0.00 & 0.00 & 0.00 & 0.00 & 0.00 & 0.00 & 0.00 & 0.00 & 0.18\\
	\cellcolor{CoolGray!50}\cite{wangBitVLA1bitVisionLanguageAction2025} & BitVLA & 0.01 & 0.00 & 0.00 & 0.00 & 0.00 & 0.00 & 0.00 & 0.10 & 0.00 & 0.00 & 0.00 & 0.09\\
	\cellcolor{CoolGray!50}\cite{zhouChatVLA2VisionLanguageActionModel2025} & Chat VLA-2 & 0.01 & 0.06 & 0.00 & 0.00 & 0.00 & 0.00 & 0.00 & 0.00 & 0.00 & 0.00 & 0.08 & 0.00\\
	\cellcolor{CoolGray!50}\cite{yueDeeRVLADynamicInference2024} & DeeR-VLA & 0.01 & 0.06 & 0.00 & 0.00 & 0.00 & 0.00 & 0.00 & 0.10 & 0.00 & 0.00 & 0.00 & 0.00\\
	\cellcolor{CoolGray!50}\cite{patratskiySpatialTracesEnhancing2025} & Spatial Traces & 0.01 & 0.00 & 0.00 & 0.00 & 0.00 & 0.00 & 0.00 & 0.00 & 0.00 & 0.00 & 0.08 & 0.09\\
	\cellcolor{CoolGray!50}\cite{deyReVLARevertingVisual2025} & ReVLA & 0.01 & 0.06 & 0.00 & 0.00 & 0.00 & 0.00 & 0.00 & 0.00 & 0.00 & 0.00 & 0.08 & 0.00\\
	\cellcolor{CoolGray!50}\cite{jinAlphaBlockEmbodiedFinetuning2023} & Alpha Block & 0.01 & 0.06 & 0.00 & 0.07 & 0.00 & 0.00 & 0.00 & 0.00 & 0.00 & 0.00 & 0.00 & 0.00\\
	\cellcolor{CoolGray!50}\cite{bonattiPACTPerceptionActionCausal2023} & PACT & 0.01 & 0.00 & 0.00 & 0.00 & 0.00 & 0.00 & 0.00 & 0.00 & 0.00 & 0.00 & 0.00 & 0.18\\
	\cellcolor{CoolGray!50}\cite{xuNaturalVLMLeveragingFinegrained2024} & Natural VLM & 0.01 & 0.06 & 0.00 & 0.00 & 0.00 & 0.00 & 0.00 & 0.00 & 0.00 & 0.00 & 0.00 & 0.09\\
	\cellcolor{CoolGray!50}\cite{chengOmniVTLAVisionTactileLanguageActionModel2025} & Omni VTLA & 0.01 & 0.00 & 0.00 & 0.00 & 0.00 & 0.00 & 0.00 & 0.00 & 0.07 & 0.00 & 0.00 & 0.09\\
	\cellcolor{CoolGray!50}\cite{driessKnowledgeInsulatingVisionLanguageAction2025} & KIVLA & 0.01 & 0.06 & 0.00 & 0.00 & 0.00 & 0.00 & 0.00 & 0.00 & 0.00 & 0.00 & 0.00 & 0.09\\
	\cellcolor{CoolGray!50}\cite{fangKeypointAbstractionUsing2024} & KALM & 0.01 & 0.00 & 0.00 & 0.00 & 0.00 & 0.00 & 0.00 & 0.00 & 0.00 & 0.00 & 0.08 & 0.09\\
	\cellcolor{CoolGray!50}\cite{jiangBetterYouLearn2025} & Light VLA & 0.01 & 0.00 & 0.00 & 0.00 & 0.00 & 0.00 & 0.00 & 0.10 & 0.07 & 0.00 & 0.00 & 0.00\\
	\cellcolor{CoolGray!50}\cite{suMotionActionDiffusing2025} & MBA & 0.01 & 0.00 & 0.00 & 0.00 & 0.00 & 0.08 & 0.00 & 0.00 & 0.00 & 0.00 & 0.08 & 0.00\\
	\cellcolor{CoolGray!50}\cite{songAviActionVolumetric2025} & Avi & 0.01 & 0.06 & 0.00 & 0.00 & 0.00 & 0.00 & 0.00 & 0.00 & 0.00 & 0.07 & 0.00 & 0.00\\
	\cellcolor{CoolGray!50}\cite{zhouChatVLAUnifiedMultimodal2025} & Chat VLA & 0.01 & 0.06 & 0.00 & 0.00 & 0.00 & 0.00 & 0.00 & 0.00 & 0.00 & 0.00 & 0.00 & 0.09\\
	\cellcolor{CoolGray!50}\cite{liMasteringRobotManipulation2024} & MIDAS & 0.01 & 0.06 & 0.00 & 0.00 & 0.00 & 0.00 & 0.00 & 0.00 & 0.00 & 0.00 & 0.00 & 0.09\\
	\cellcolor{CoolGray!50}\cite{bingMetaReinforcementLearningLanguage2022} & Meta RL-LI & 0.01 & 0.06 & 0.00 & 0.00 & 0.00 & 0.00 & 0.00 & 0.00 & 0.00 & 0.00 & 0.08 & 0.00\\
	\cellcolor{CoolGray!50}\cite{tanBLM$_1$BoundlessLarge2025} & BLM1 & 0.01 & 0.06 & 0.00 & 0.00 & 0.00 & 0.00 & 0.00 & 0.00 & 0.00 & 0.00 & 0.00 & 0.09\\
	\cellcolor{CoolGray!50}\cite{huMiniDiffuserFast2025} & Mini Diffuser & 0.01 & 0.06 & 0.00 & 0.00 & 0.00 & 0.00 & 0.00 & 0.00 & 0.00 & 0.00 & 0.00 & 0.09\\
	\cellcolor{CoolGray!50}\cite{wuDiscretePolicyLearning2025} & Discrete Policy & 0.01 & 0.00 & 0.00 & 0.00 & 0.00 & 0.00 & 0.00 & 0.00 & 0.00 & 0.00 & 0.08 & 0.09\\
	\cellcolor{CoolGray!50}\cite{PI06_model_card} & $\pi_0.6$ (model card) & 0.01 & 0.06 & 0.00 & 0.00 & 0.00 & 0.00 & 0.00 & 0.00 & 0.00 & 0.00 & 0.00 & 0.09\\
	\cellcolor{CoolGray!50}\cite{hungNORASmallOpenSourced2025} & NORA & 0.01 & 0.06 & 0.00 & 0.00 & 0.00 & 0.00 & 0.00 & 0.00 & 0.00 & 0.00 & 0.00 & 0.00\\
	\cellcolor{CoolGray!50}\cite{liVIRTVisionInstructed2024} & VIRT & 0.01 & 0.00 & 0.00 & 0.00 & 0.00 & 0.00 & 0.00 & 0.00 & 0.00 & 0.00 & 0.08 & 0.00\\
	\cellcolor{CoolGray!50}\cite{chenPolarNet3DPoint2023} & PolarNet & 0.01 & 0.00 & 0.00 & 0.00 & 0.00 & 0.00 & 0.00 & 0.00 & 0.00 & 0.00 & 0.08 & 0.00\\
	\cellcolor{CoolGray!50}\cite{zhangMoLeVLADynamicLayerskipping2025} & MoLe-VLA & 0.01 & 0.00 & 0.00 & 0.00 & 0.00 & 0.00 & 0.00 & 0.10 & 0.00 & 0.00 & 0.00 & 0.00\\
	\cellcolor{CoolGray!50}\cite{agibot-world-contributorsAgiBotWorldColosseo2025} & AgiBot World Colosseo & 0.01 & 0.00 & 0.00 & 0.00 & 0.00 & 0.00 & 0.00 & 0.00 & 0.00 & 0.00 & 0.00 & 0.09\\
	\cellcolor{CoolGray!50}\cite{chebotarQTransformerScalableOffline2023} & Q-Trans-former & 0.01 & 0.00 & 0.00 & 0.00 & 0.00 & 0.00 & 0.00 & 0.00 & 0.00 & 0.00 & 0.08 & 0.00\\
	\cellcolor{CoolGray!50}\cite{wenLLaDAVLAVisionLanguage2025} & LLaDA-VLA & 0.01 & 0.06 & 0.00 & 0.00 & 0.00 & 0.00 & 0.00 & 0.00 & 0.00 & 0.00 & 0.00 & 0.00\\
	\cellcolor{CoolGray!50}\cite{liangNeverEndingBehaviorCloningAgent2025} & NBAgent & 0.01 & 0.06 & 0.00 & 0.00 & 0.00 & 0.00 & 0.00 & 0.00 & 0.00 & 0.00 & 0.00 & 0.00\\
	\cellcolor{CoolGray!50}\cite{liFastMemorybasedDataEfficient2025} & LiteVLP & 0.01 & 0.00 & 0.00 & 0.00 & 0.00 & 0.00 & 0.00 & 0.00 & 0.00 & 0.00 & 0.00 & 0.09\\
	\cellcolor{CoolGray!50}\cite{zhengInterpretableRoboticManipulation2024} & Ex-PERACT & 0.01 & 0.00 & 0.00 & 0.00 & 0.00 & 0.00 & 0.00 & 0.00 & 0.00 & 0.00 & 0.00 & 0.09\\
	\cellcolor{CoolGray!50}\cite{liuRoboMambaEfficientVisionLanguageAction2024} & Robo Mamba & 0.01 & 0.00 & 0.00 & 0.00 & 0.00 & 0.00 & 0.00 & 0.00 & 0.00 & 0.00 & 0.08 & 0.00\\
	\cellcolor{CoolGray!50}\cite{jangContextVLAVisionLanguageActionModel2025} & Context VLA & 0.01 & 0.00 & 0.00 & 0.00 & 0.00 & 0.00 & 0.00 & 0.00 & 0.00 & 0.00 & 0.08 & 0.00\\
	\cellcolor{CoolGray!50}\cite{kooRetoVLAReusingRegister2025} & RetoVLA & 0.01 & 0.00 & 0.00 & 0.00 & 0.00 & 0.00 & 0.00 & 0.00 & 0.00 & 0.00 & 0.08 & 0.00\\
	\cellcolor{CoolGray!50}\cite{zhangGroundingActionsCamera2025} & OC-VLA & 0.01 & 0.00 & 0.00 & 0.00 & 0.00 & 0.00 & 0.00 & 0.00 & 0.00 & 0.00 & 0.00 & 0.09\\
	\cellcolor{CoolGray!50}\cite{liControlVLAFewshotObjectcentric2025} & Control VLA & 0.01 & 0.00 & 0.00 & 0.00 & 0.00 & 0.00 & 0.00 & 0.00 & 0.00 & 0.00 & 0.00 & 0.09\\
	\cellcolor{CoolGray!50}\cite{dasariIngredientsRoboticDiffusion2024} & RoboDiT & 0.01 & 0.00 & 0.00 & 0.00 & 0.00 & 0.00 & 0.00 & 0.00 & 0.00 & 0.00 & 0.00 & 0.09\\
	\cellcolor{CoolGray!50}\cite{chenECFlowEnablingVersatile2025} & EC-Flow & 0.01 & 0.00 & 0.00 & 0.00 & 0.00 & 0.00 & 0.00 & 0.00 & 0.00 & 0.00 & 0.00 & 0.09\\
	\cellcolor{CoolGray!50}\cite{xianChainedDiffuserUnifyingTrajectory2023} & Chained Diffuser & 0.01 & 0.00 & 0.00 & 0.00 & 0.00 & 0.00 & 0.00 & 0.00 & 0.00 & 0.00 & 0.00 & 0.09\\
	\cellcolor{CoolGray!50}\cite{liObjectFocusActorDataefficient2025} & OFA & 0.01 & 0.00 & 0.00 & 0.00 & 0.00 & 0.00 & 0.00 & 0.00 & 0.00 & 0.00 & 0.00 & 0.09\\
	\cellcolor{CoolGray!50}\cite{sejnovaBridgingLanguageVision2024} & VAE-Rob Manip & 0.01 & 0.00 & 0.00 & 0.00 & 0.00 & 0.00 & 0.00 & 0.00 & 0.00 & 0.00 & 0.00 & 0.09\\
	\cellcolor{CoolGray!50}\cite{abouzeidGeoAwareVLAImplicitGeometry2025} & Geo Aware-VLA & 0.00 & 0.00 & 0.00 & 0.00 & 0.00 & 0.00 & 0.00 & 0.00 & 0.00 & 0.00 & 0.00 & 0.00\\
	\cellcolor{CoolGray!50}\cite{songCEEDVLAConsistencyVisionLanguageAction2025} & CEED-VLA & 0.00 & 0.00 & 0.00 & 0.00 & 0.00 & 0.00 & 0.00 & 0.00 & 0.00 & 0.00 & 0.00 & 0.00\\
	\cellcolor{CoolGray!50}\cite{wangSpecVLASpeculativeDecoding2025} & Spec-VLA & 0.00 & 0.00 & 0.00 & 0.00 & 0.00 & 0.00 & 0.00 & 0.00 & 0.00 & 0.00 & 0.00 & 0.00\\
	\cellcolor{CoolGray!50}\cite{liRoboFlamingoPlusFusionDepth2025} & Robo Flamingo-Plus & 0.00 & 0.00 & 0.00 & 0.00 & 0.00 & 0.00 & 0.00 & 0.00 & 0.00 & 0.00 & 0.00 & 0.00\\
	\cellcolor{CoolGray!50}\cite{nguyenRoboticCLIPFinetuningCLIP2024} & Robotic-CLIP & 0.00 & 0.00 & 0.00 & 0.00 & 0.00 & 0.00 & 0.00 & 0.00 & 0.00 & 0.00 & 0.00 & 0.00\\
	\cellcolor{CoolGray!50}\cite{budzianowskiEdgeVLAEfficientVisionLanguageAction2025} & EdgeVLA & 0.00 & 0.00 & 0.00 & 0.00 & 0.00 & 0.00 & 0.00 & 0.00 & 0.00 & 0.00 & 0.00 & 0.00\\
	\cellcolor{CoolGray!50}\cite{niuLLARVAVisionActionInstruction2024} & LLARVA & 0.00 & 0.00 & 0.00 & 0.00 & 0.00 & 0.00 & 0.00 & 0.00 & 0.00 & 0.00 & 0.00 & 0.00\\
	\cellcolor{CoolGray!50}\cite{geVGGTDPGeneralizableRobot2025} & VGGT-DP & 0.00 & 0.00 & 0.00 & 0.00 & 0.00 & 0.00 & 0.00 & 0.00 & 0.00 & 0.00 & 0.00 & 0.00\\
	\cellcolor{CoolGray!50}\cite{zhuScalingDiffusionPolicy2025} & ScaleDP & 0.00 & 0.00 & 0.00 & 0.00 & 0.00 & 0.00 & 0.00 & 0.00 & 0.00 & 0.00 & 0.00 & 0.00\\
\end{longtable}
\normalsize
\twocolumn
\begin{table}[t]
	\centering
	\renewcommand{\arraystretch}{1.35}
	\small
	\caption{Criterion fulfilment for I1 -- Adaptability \& Flexibility. 
		}
	\label{tab:i1-criteria-summary}
	
	\begin{tabular}{@{}
			>{\raggedright\arraybackslash}m{4.75cm}
			>{\centering\arraybackslash}m{0.5cm}
			>{\centering\arraybackslash}m{1.3cm}
			@{}}
		\rowcolor{CoolGray!50}
		\cellcolor{Platinum!60}\textbf{\rule{0pt}{2.5ex}Criterion} &
		\textbf{\rule{0pt}{2.5ex}$\sum\downarrow$} &
		\textbf{\rule{0pt}{2.5ex}\% fulfilled} \\
		\cellcolor{CoolGray!50}Multi-Task Generalisation & 160 & 49.38\% \\
		\cellcolor{CoolGray!50}Generalisation to Environmental Variation & 101 & 31.17\% \\
		\cellcolor{CoolGray!50}Autonomous Handling of Distractors/Obstacles & 87 & 26.85\% \\
		\cellcolor{CoolGray!50}Online/Mid-Execution Adaptation & 49 & 15.12\% \\
		\cellcolor{CoolGray!50}Autonomous Incorporation of Skills & 22 & 6.79\% \\
		\cellcolor{CoolGray!50}Online Environment Adaptation and Response & 21 & 6.48\% \\
		\cellcolor{CoolGray!50}Self-Evaluation \& Command Adherence & 19 & 5.86\% \\
		\cellcolor{CoolGray!50}Multi-Hardware Support & 16 & 4.94\% \\
		\cellcolor{CoolGray!50}Mid-Execution Feedback Responsiveness & 14 & 4.32\% \\
		\cellcolor{CoolGray!50}Explained Adaptation Decisions & 10 & 3.09\% \\
		\cellcolor{CoolGray!50}Multimodal Command Interface & 7 & 2.16\% \\
		\cellcolor{CoolGray!50}Memory of Previous Faults/Distractors & 4 & 1.23\% \\
		\cellcolor{CoolGray!50}Restricted Autonomous Task Switching & 3 & 0.93\% \\
		\cellcolor{CoolGray!50}Autonomous Fallbacks & 2 & 0.62\% \\
		\cellcolor{CoolGray!50}Energy/System-Aware Adaptation & 1 & 0.31\% \\
		\cellcolor{CoolGray!50}Operational Adaptation Confidence & 1 & 0.31\% \\
		\cellcolor{CoolGray!50}Strategy-Goal Alignment & 1 & 0.31\% \\
		\cellcolor{CoolGray!50}Hot-Swappable \& Dimensionality Adaptation & 0 & 0.00\% \\
	\end{tabular}
\end{table}
\begin{table}[t]
	\centering
	\renewcommand{\arraystretch}{1.35}
	\small
	\caption{Criterion fulfilment for I2 -- Safety \& Compliance.
	}
	\label{tab:i2-criteria-summary}
	\begin{tabular}{@{}
			>{\raggedright\arraybackslash}m{4.75cm}
			>{\centering\arraybackslash}m{0.5cm}
			>{\centering\arraybackslash}m{1.3cm}
			@{}}
		\rowcolor{CoolGray!50}
		\cellcolor{Platinum!60}\textbf{\rule{0pt}{2.5ex}Criterion} &
		\textbf{\rule{0pt}{2.5ex}$\sum\downarrow$} &
		\textbf{\rule{0pt}{2.5ex}\% fulfilled} \\
		\cellcolor{CoolGray!50}Redundant, Independent Safety Mechanisms & 5 & 1.54\% \\
		\cellcolor{CoolGray!50}Command Compliance with Online-Safety & 4 & 1.23\% \\
		\cellcolor{CoolGray!50}Proactive, Plan-Level Safety & 3 & 0.93\% \\
		\cellcolor{CoolGray!50}Learning from Safety Events & 2 & 0.62\% \\
		\cellcolor{CoolGray!50}Decoupled, Real-Time Safety Control & 1 & 0.31\% \\
		\cellcolor{CoolGray!50}Reactive, Real-Time Safety & 1 & 0.31\% \\
		\cellcolor{CoolGray!50}Rule-Adherence \& Autonomous Rejection & 1 & 0.31\% \\
		\cellcolor{CoolGray!50}Authorised Operator Detection \& Override & 0 & 0.00\% \\
		\cellcolor{CoolGray!50}External Safety Integration & 0 & 0.00\% \\
		\cellcolor{CoolGray!50}Fault Detection \& Sensing & 0 & 0.00\% \\
		\cellcolor{CoolGray!50}Integrable Safety Systems & 0 & 0.00\% \\
		\cellcolor{CoolGray!50}Persistence of Safety Capabilities in Adaptation & 0 & 0.00\% \\
		\cellcolor{CoolGray!50}Persistent Safety Logs & 0 & 0.00\% \\
		\cellcolor{CoolGray!50}Safe, Autonomous Recovery & 0 & 0.00\% \\
		\cellcolor{CoolGray!50}Safety Confidence Output & 0 & 0.00\% \\
		\cellcolor{CoolGray!50}Safety Documentation \& Traceability & 0 & 0.00\% \\
		\cellcolor{CoolGray!50}Standards-Compliant & 0 & 0.00\% \\
		\cellcolor{CoolGray!50}Transparent Safety Reasoning & 0 & 0.00\% \\	
	\end{tabular}
\end{table}
\begin{table}[t]
	\centering
	\renewcommand{\arraystretch}{1.35}
	\small
	\caption{Criterion fulfilment for I3 -- HRI \& HRC. 
	}
	\label{tab:i3-criteria-summary}
	\begin{tabular}{@{}
			>{\raggedright\arraybackslash}m{4.75cm}
			>{\centering\arraybackslash}m{0.5cm}
			>{\centering\arraybackslash}m{1.3cm}
			@{}}
		\rowcolor{CoolGray!50}
		\cellcolor{Platinum!60}\textbf{\rule{0pt}{2.5ex}Criterion} &
		\textbf{\rule{0pt}{2.5ex}$\sum\downarrow$} &
		\textbf{\rule{0pt}{2.5ex}\% fulfilled} \\
		\cellcolor{CoolGray!50}Intuitive Feedback Acceptance & 13 & 4.01\% \\
		\cellcolor{CoolGray!50}Collaborative Adaptation & 8 & 2.47\% \\
		\cellcolor{CoolGray!50}Transparent Reasoning Explanation & 5 & 1.54\% \\
		\cellcolor{CoolGray!50}Real-time Interruptibility & 4 & 1.23\% \\
		\cellcolor{CoolGray!50}Continuous Responsiveness & 2 & 0.62\% \\
		\cellcolor{CoolGray!50}Active Operator Exclusivity & 0 & 0.00\% \\
		\cellcolor{CoolGray!50}Exception Handling \& Safe State Transition & 0 & 0.00\% \\
		\cellcolor{CoolGray!50}Human Presence Awareness in Workspace & 0 & 0.00\% \\
		\cellcolor{CoolGray!50}Human-Centric Adaptation & 0 & 0.00\% \\
		\cellcolor{CoolGray!50}Industrial HRI/HRC Standards Compliance & 0 & 0.00\% \\
		\cellcolor{CoolGray!50}Instruction Confidence & 0 & 0.00\% \\
		\cellcolor{CoolGray!50}Learned Operator Preferences & 0 & 0.00\% \\
		\cellcolor{CoolGray!50}Multimodal Instruction \& Output & 0 & 0.00\% \\
		\cellcolor{CoolGray!50}Operator Data Privacy & 0 & 0.00\% \\
		\cellcolor{CoolGray!50}Operator Identification \& Authorisation & 0 & 0.00\% \\
	\end{tabular}
\end{table}
\begin{table}[t]
	\centering
	\renewcommand{\arraystretch}{1.35}
	\small
	\caption{Criterion fulfilment for I4 -- Robustness \& Reliability. 
	}
	\label{tab:i4-criteria-summary}
	\begin{tabular}{@{}
			>{\raggedright\arraybackslash}m{5cm}
			>{\centering\arraybackslash}m{0.5cm}
			>{\centering\arraybackslash}m{1.3cm}
			@{}}
		\rowcolor{CoolGray!50}
		\cellcolor{Platinum!60}\textbf{\rule{0pt}{2.5ex}Criterion} &
		\textbf{\rule{0pt}{2.5ex}$\sum\downarrow$} &
		\textbf{\rule{0pt}{2.5ex}\% fulfilled} \\
		\cellcolor{CoolGray!50}Disturbance Resilience \& Continuous Operation & 44 & 13.58\% \\
		\cellcolor{CoolGray!50}Environmental Robustness \& Adaptation & 14 & 4.32\% \\
		\cellcolor{CoolGray!50}Graceful Degradation \& Strategy Adaptation & 6 & 1.85\% \\
		\cellcolor{CoolGray!50}Reproducible \& Benchmarked Fault Diagnosis & 5 & 1.54\% \\
		\cellcolor{CoolGray!50}Adaptive Error Learning \& Memorisation & 3 & 0.93\% \\
		\cellcolor{CoolGray!50}Logging, Explanation \& Operator Guidance & 1 & 0.31\% \\
		\cellcolor{CoolGray!50}Automated Recovery \& Safe-State Transition & 0 & 0.00\% \\
		\cellcolor{CoolGray!50}Autonomous Fault Categorisation \& Severity Response & 0 & 0.00\% \\
		\cellcolor{CoolGray!50}Consistent \& Documented Fault Handling & 0 & 0.00\% \\
		\cellcolor{CoolGray!50}Multi-Level Fault \& Anomaly Detection & 0 & 0.00\% \\
		\cellcolor{CoolGray!50}Online Confidence \& Reliability Assessment & 0 & 0.00\% \\
		\cellcolor{CoolGray!50}Robust Redundancy \& Modality Management & 0 & 0.00\% \\
		\cellcolor{CoolGray!50}Supervisor Oversight \& Fault Review & 0 & 0.00\% \\
		\cellcolor{CoolGray!50}Threat \& Attack Detection & 0 & 0.00\% \\
	\end{tabular}
\end{table}
\begin{table}[t]
	\centering
	\renewcommand{\arraystretch}{1.35}
	\small
	\caption{Criterion fulfilment for I5 -- Precision \& Accuracy. 
	}
	\label{tab:i5-criteria-summary}
	\begin{tabular}{@{}
			>{\raggedright\arraybackslash}m{5cm}
			>{\centering\arraybackslash}m{0.5cm}
			>{\centering\arraybackslash}m{1.3cm}
			@{}}
		\rowcolor{CoolGray!50}
		\cellcolor{Platinum!60}\textbf{\rule{0pt}{2.5ex}Criterion} &
		\textbf{\rule{0pt}{2.5ex}$\sum\downarrow$} &
		\textbf{\rule{0pt}{2.5ex}\% fulfilled} \\
		\cellcolor{CoolGray!50}Consistency Over Time \& Contexts & 38 & 11.73\% \\
		\cellcolor{CoolGray!50}Skill Integration with Known \gls{vpa} & 9 & 2.78\% \\
		\cellcolor{CoolGray!50}\gls{vpa} Learning from Experience \& Feedback & 6 & 1.85\% \\
		\cellcolor{CoolGray!50}HRI/HRC Integration & 3 & 0.93\% \\
		\cellcolor{CoolGray!50}User-Defined Trade-offs & 2 & 0.62\% \\
		\cellcolor{CoolGray!50}Disturbance \& Uncertainty Management & 1 & 0.31\% \\
		\cellcolor{CoolGray!50}\gls{vpa} Instructability & 1 & 0.31\% \\
		\cellcolor{CoolGray!50}Achievable \gls{vpa}-Proposals & 0 & 0.00\% \\
		\cellcolor{CoolGray!50}Autonomous \gls{vpa} Adaptation & 0 & 0.00\% \\
		\cellcolor{CoolGray!50}Outcome Logging \& Failure Reasoning & 0 & 0.00\% \\
		\cellcolor{CoolGray!50}Supervision \& Verification Module & 0 & 0.00\% \\
		\cellcolor{CoolGray!50}Transparent Documentation of \gls{vpa} Capabilities & 0 & 0.00\% \\
		\cellcolor{CoolGray!50}\gls{vpa}-Confidence Score \& Explanation & 0 & 0.00\% \\
	\end{tabular}
\end{table}
\begin{table}[t]
	\centering
	\renewcommand{\arraystretch}{1.35}
	\small
	\caption{Criterion fulfilment for I6 -- Real-time Performance. 
	}
	\label{tab:i6-criteria-summary}
	\begin{tabular}{@{}
			>{\raggedright\arraybackslash}m{5cm}
			>{\centering\arraybackslash}m{0.5cm}
			>{\centering\arraybackslash}m{1.3cm}
			@{}}
		\rowcolor{CoolGray!50}
		\cellcolor{Platinum!60}\textbf{\rule{0pt}{2.5ex}Criterion} &
		\textbf{\rule{0pt}{2.5ex}$\sum\downarrow$} &
		\textbf{\rule{0pt}{2.5ex}\% fulfilled} \\
		\cellcolor{CoolGray!50}Goal-Responsive Solution Strategies & 12 & 3.70\% \\
		\cellcolor{CoolGray!50}Historical Data Utilisation & 2 & 0.62\% \\
		\cellcolor{CoolGray!50}System-wide Real-time Compliance & 1 & 0.31\% \\
		\cellcolor{CoolGray!50}Task, Input \& Environment Independence & 1 & 0.31\% \\
		\cellcolor{CoolGray!50}Autonomous Adaptation \& Safety Response & 0 & 0.00\% \\
		\cellcolor{CoolGray!50}Dynamic Resource Management \& Determinism & 0 & 0.00\% \\
		\cellcolor{CoolGray!50}Live Metrics, Logging \& Explainability & 0 & 0.00\% \\
		\cellcolor{CoolGray!50}Real-Time Confidence/Alerting & 0 & 0.00\% \\
		\cellcolor{CoolGray!50}Transparent Documentation \& Compliance & 0 & 0.00\% \\
	\end{tabular}
\end{table}
\begin{table}[t]
	\centering
	\renewcommand{\arraystretch}{1.35}
	\small
	\caption{Criterion fulfilment for I7 -- Cost-effectiveness \& Integration Capabilities. 
	}
	\label{tab:i7-criteria-summary}
	\begin{tabular}{@{}
			>{\raggedright\arraybackslash}m{5cm}
			>{\centering\arraybackslash}m{0.5cm}
			>{\centering\arraybackslash}m{1.3cm}
			@{}}
		\rowcolor{CoolGray!50}
		\cellcolor{Platinum!60}\textbf{\rule{0pt}{2.5ex}Criterion} &
		\textbf{\rule{0pt}{2.5ex}$\sum\downarrow$} &
		\textbf{\rule{0pt}{2.5ex}\% fulfilled} \\
		\cellcolor{CoolGray!50}Dynamic Resource Allocation & 16 & 4.94\% \\
		\cellcolor{CoolGray!50}Low-Cost Hardware Readiness & 8 & 2.47\% \\
		\cellcolor{CoolGray!50}Comprehensive Integration Documentation & 1 & 0.31\% \\
		\cellcolor{CoolGray!50}Autonomous Fault Handling \& Feedback & 0 & 0.00\% \\
		\cellcolor{CoolGray!50}Cost-Reduction Memory \& Adaptive Optimisation & 0 & 0.00\% \\
		\cellcolor{CoolGray!50}Flexible Input/Output Adaptation & 0 & 0.00\% \\
		\cellcolor{CoolGray!50}Guided \& Autonomous Commissioning & 0 & 0.00\% \\
		\cellcolor{CoolGray!50}Hot-Swapping \& Auto-Detection & 0 & 0.00\% \\
		\cellcolor{CoolGray!50}Integration Performance Monitoring \& Learning & 0 & 0.00\% \\
		\cellcolor{CoolGray!50}Universal Integration without Retraining & 0 & 0.00\% \\
	\end{tabular}
\end{table}
\begin{table}[t]
	\centering
	\renewcommand{\arraystretch}{1.35}
	\small
	\caption{Criterion fulfilment for I8 -- Explainability \& Trust. 
	}
	\label{tab:i8-criteria-summary}
	\begin{tabular}{@{}
			>{\raggedright\arraybackslash}m{5cm}
			>{\centering\arraybackslash}m{0.5cm}
			>{\centering\arraybackslash}m{1.3cm}
			@{}}
		\rowcolor{CoolGray!50}
		\cellcolor{Platinum!60}\textbf{\rule{0pt}{2.5ex}Criterion} &
		\textbf{\rule{0pt}{2.5ex}$\sum\downarrow$} &
		\textbf{\rule{0pt}{2.5ex}\% fulfilled} \\
		\cellcolor{CoolGray!50}Performance Benchmarking \& OOD Feedback & 63 & 19.44\% \\
		\cellcolor{CoolGray!50}Explains Influences \& Disturbances & 4 & 1.23\% \\
		\cellcolor{CoolGray!50}Reason Logging \& Accessible Justification & 3 & 0.93\% \\
		\cellcolor{CoolGray!50}Task Feasibility \& Capability Explanation & 3 & 0.93\% \\
		\cellcolor{CoolGray!50}Error, Anomaly \& Safeguard Explanations & 2 & 0.62\% \\
		\cellcolor{CoolGray!50}User Feedback \& Adaptive Improvement & 2 & 0.62\% \\
		\cellcolor{CoolGray!50}Behaviour Consistency \& Pre-Adaptation Notification & 0 & 0.00\% \\
		\cellcolor{CoolGray!50}Confidence Score Explanation & 0 & 0.00\% \\
		\cellcolor{CoolGray!50}Continuous Self-Assessment with Explainability & 0 & 0.00\% \\
		\cellcolor{CoolGray!50}Explainability Across System Components & 0 & 0.00\% \\
		\cellcolor{CoolGray!50}Flexible, Intuitive Explainability Interfaces & 0 & 0.00\% \\
		\cellcolor{CoolGray!50}Monitors \& Logs Explainability for Learning & 0 & 0.00\% \\
		\cellcolor{CoolGray!50}Rule \& Norm Compliance with Explainability & 0 & 0.00\% \\
		\cellcolor{CoolGray!50}Security \& Authorisation Feedback & 0 & 0.00\% \\
	\end{tabular}
\end{table}
\begin{table}[t]
	\centering
	\renewcommand{\arraystretch}{1.35}
	\small
	\caption{Criterion fulfilment for I9 -- Sensor Fusion \& Perception. 
	}
	\label{tab:i9-criteria-summary}
	\begin{tabular}{@{}
			>{\raggedright\arraybackslash}m{5cm}
			>{\centering\arraybackslash}m{0.5cm}
			>{\centering\arraybackslash}m{1.3cm}
			@{}}
		\rowcolor{CoolGray!50}
		\cellcolor{Platinum!60}\textbf{\rule{0pt}{2.5ex}Criterion} &
		\textbf{\rule{0pt}{2.5ex}$\sum\downarrow$} &
		\textbf{\rule{0pt}{2.5ex}\% fulfilled} \\
		\cellcolor{CoolGray!50}Perceptual Feature Exposure \& Modularity & 32 & 9.88\% \\
		\cellcolor{CoolGray!50}Comprehensive Scene Understanding via Fusion & 29 & 8.95\% \\
		\cellcolor{CoolGray!50}Multimodal Sensor Acceptance \& Fusion & 11 & 3.40\% \\
		\cellcolor{CoolGray!50}Flexible Input Combination \& Goal Specification & 10 & 3.09\% \\
		\cellcolor{CoolGray!50}Dynamic Sensor Prioritisation & 2 & 0.62\% \\
		\cellcolor{CoolGray!50}Peer-to-Peer Perception Knowledge Exchange & 1 & 0.31\% \\
		\cellcolor{CoolGray!50}Autonomous Disturbance Recognition \& Countermeasures & 0 & 0.00\% \\
		\cellcolor{CoolGray!50}Continuous Input Monitoring \& Feedback & 0 & 0.00\% \\
		\cellcolor{CoolGray!50}Modular Integration \& Documentation & 0 & 0.00\% \\
		\cellcolor{CoolGray!50}Online Learning for Perception Skills & 0 & 0.00\% \\
		\cellcolor{CoolGray!50}Perception Evaluation \& Benchmarking & 0 & 0.00\% \\
		\cellcolor{CoolGray!50}Perceptual Confidence \& Explanation Output & 0 & 0.00\% \\
		\cellcolor{CoolGray!50}Robust Fault Tolerance \& Degradation Handling & 0 & 0.00\% \\
		\cellcolor{CoolGray!50}Security in Sensor Operations & 0 & 0.00\% \\
	\end{tabular}
\end{table}
\begin{table}[t]
	\centering
	\renewcommand{\arraystretch}{1.35}
	\small
	\caption{Criterion fulfilment for I10 -- Standardised Benchmarking \& Evaluation. 
	}
	\label{tab:i10-criteria-summary}
	\begin{tabular}{@{}
			>{\raggedright\arraybackslash}m{5cm}
			>{\centering\arraybackslash}m{0.5cm}
			>{\centering\arraybackslash}m{1.3cm}
			@{}}
		\rowcolor{CoolGray!50}
		\cellcolor{Platinum!60}\textbf{\rule{0pt}{2.5ex}Criterion} &
		\textbf{\rule{0pt}{2.5ex}$\sum\downarrow$} &
		\textbf{\rule{0pt}{2.5ex}\% fulfilled} \\
	\cellcolor{CoolGray!50}Repeatability \& Reproducibility & 145 & 44.75\% \\
	\cellcolor{CoolGray!50}Module \& Skill-Specific Benchmarking & 129 & 39.81\% \\
	\cellcolor{CoolGray!50}Generalisation \& Adaptation Assessment & 90 & 27.78\% \\
	\cellcolor{CoolGray!50}Benchmark Diversity \& Scope & 47 & 14.51\% \\
	\cellcolor{CoolGray!50}Instruction, Command, \& Modality Coverage & 3 & 0.93\% \\
	\cellcolor{CoolGray!50}Explainability, Reasoning, \& Feedback Benchmarks & 1 & 0.31\% \\
	\cellcolor{CoolGray!50}Integration \& Commissioning Evaluation & 1 & 0.31\% \\
	\cellcolor{CoolGray!50}Data Requirement \& Autonomous Runtime Assessment & 0 & 0.00\% \\
	\cellcolor{CoolGray!50}Hardware Platform Benchmarking & 0 & 0.00\% \\
	\cellcolor{CoolGray!50}Human Interaction \& Object Recognition Evaluation & 0 & 0.00\% \\
	\cellcolor{CoolGray!50}Robustness \& Fault Tolerance Evaluation & 0 & 0.00\% \\
	\cellcolor{CoolGray!50}Safety, Security, \& Norm Compliance Validation & 0 & 0.00\% \\
	\cellcolor{CoolGray!50}Transparent Metrics \& Reporting & 0 & 0.00\% \\
	\end{tabular}
\end{table}
\begin{table}[t]
	\centering
	\renewcommand{\arraystretch}{1.35}
	\small
	\caption{Criterion fulfilment for I11 -- Data Requirements \& Usage. 
	}
	\label{tab:i11-criteria-summary}
	\begin{tabular}{@{}
			>{\raggedright\arraybackslash}m{5cm}
			>{\centering\arraybackslash}m{0.5cm}
			>{\centering\arraybackslash}m{1.3cm}
			@{}}
		\rowcolor{CoolGray!50}
		\cellcolor{Platinum!60}\textbf{\rule{0pt}{2.5ex}Criterion} &
		\textbf{\rule{0pt}{2.5ex}$\sum\downarrow$} &
		\textbf{\rule{0pt}{2.5ex}\% fulfilled} \\
		\cellcolor{CoolGray!50}Flexible Data Type Acceptance & 158 & 48.77\% \\
		\cellcolor{CoolGray!50}Low-Data Adaptation \& Few-Shot Requirements & 136 & 41.98\% \\
		\cellcolor{CoolGray!50}Documented Data Composition \& Simulation-Real Transfer & 33 & 10.19\% \\
		\cellcolor{CoolGray!50}Efficient Task, Environment, \& Modality Adaptation & 14 & 4.32\% \\
		\cellcolor{CoolGray!50}Resilience Through Redundancy \& Fault Tolerance & 2 & 0.62\% \\
		\cellcolor{CoolGray!50}Self-Assessment to Reduce Input \& Modalities & 1 & 0.31\% \\
		\cellcolor{CoolGray!50}Automatic Data Sufficiency Assessment \& Feedback & 0 & 0.00\% \\
		\cellcolor{CoolGray!50}Compliance with Privacy, Ethics, \& Law & 0 & 0.00\% \\
		\cellcolor{CoolGray!50}Comprehensive, Diverse Pretraining Dataset & 0 & 0.00\% \\
		\cellcolor{CoolGray!50}Explainability on Data Usage \& Improvement & 0 & 0.00\% \\
		\cellcolor{CoolGray!50}Ongoing Adaptation/Finetuning Confidence Scores & 0 & 0.00\% \\
	\end{tabular}
\end{table}
\end{appendices}

\end{document}